%% file: main.tex
\documentclass[letterpaper,12pt,titlepage]{report}
\input{macros}

\begin{document}

\pagenumbering{roman}
\input{src/00-titlepage}
\input{src/01-abstract}
\input{src/02-0-acknowledgements}
\input{src/02-2-dedication}

\tableofcontents
\listoffigures
\listoftables
\listofalgorithms
\clearpage

\pagenumbering{arabic}
\input{src/10-intro}
\input{src/20-background}
\part{Learning to Parse through Cross-Modal Grounding}
\label{part:syntax-vision}
\input{src/30-vgnsl}
\input{src/40-avnsl}
\input{src/50-structiou}

\part{Learning to Parse through Program Execution}
\label{part:semantics-execution}
\input{src/60-g2l2}
\input{src/70-mbrexec}
\part{Learning to Parse through Cross-Lingual Grounding}
\label{part:cross-lingual}
\input{src/80-mlpalign}
\input{src/90-subdp}
\input{src/discussion}
\hypersetup{linkcolor=cyan}
\setcitestyle{numbers}
\bibliographystyle{bib/acl_natbib}
\bibliography{bib/fspubs, bib/refs}

\end{document}

%% file: macros.tex
\usepackage[left=1in, right=1in, top=1.2in, bottom=1.4in, footskip=0.5in]{geometry}
\usepackage[utf8]{inputenc}
\usepackage[dvipsnames]{xcolor}
\usepackage[ruled,vlined,noresetcount]{algorithm2e}
\usepackage{algpseudocode}
\usepackage{amsmath}
\usepackage{amsfonts}
\usepackage{amssymb}
\usepackage{amsthm}
\usepackage{arydshln}
\usepackage{bbold}
\usepackage{booktabs}
\usepackage{bbm}
\usepackage{bm}
\usepackage{ccg}
\usepackage{cjhebrew}
\usepackage{enumitem}
\usepackage{float}
\usepackage[T1]{fontenc}
\usepackage{footnote}
\usepackage[linguistics]{forest}
\usepackage{graphicx}
\usepackage[hidelinks,backref=page,colorlinks=true,linkcolor=black,citecolor=black,urlcolor=black]{hyperref}
\usepackage{listings}
\usepackage[bb=boondox]{mathalfa}
\usepackage{mathpazo}
\usepackage{mathtools}
\usepackage{microtype}
\usepackage{multirow}
\usepackage[numbers,authoryear]{natbib}
\usepackage{subcaption}
\usepackage{tablefootnote}
\usepackage{titlesec}
\usepackage{tikz}
\usepackage{tikz-dependency}
\usepackage{url}
\usepackage{xspace}
\usepackage[capitalize,noabbrev]{cleveref}
\crefname{chapter}{Chapter}{Chapters}
\crefname{section}{\S}{\S\S}
\Crefname{section}{\S}{\S\S}
\crefname{table}{Table}{Tables}
\crefname{figure}{Figure}{Figures}
\crefname{algorithm}{Algorithm}{}
\crefname{equation}{Eq.}{}
\crefname{appendix}{Appendix}{}
\crefformat{section}{\S#2#1#3}
\crefname{lemma}{Lemma}{Lemmas}
\crefname{proposition}{Proposition}{Propositions}
\crefname{definition}{Definition}{Definitions}
\crefname{corollary}{Corollary}{Corollaries}
\crefname{example}{Example}{Examples}
\crefname{problem}{Problem}{Problems}
\crefname{remark}{Remark}{Remarks}

\setlist[itemize]{itemsep=1pt, topsep=1pt}
\setlist[enumerate]{itemsep=1pt, topsep=1pt}

\DeclareMathOperator*{\argmax}{arg\,max}

\newcommand{\stderr}{$\pm$}

\newcommand{\interalia}[1]{\cite[][\textit{inter alia}]{#1}}
\makeatletter
\def\adl@drawiv#1#2#3{%
        \hskip.5\tabcolsep
        \xleaders#3{#2.5\@tempdimb #1{1}#2.5\@tempdimb}%
                #2\z@ plus1fil minus1fil\relax
        \hskip.5\tabcolsep}
\newcommand{\cdashlinelr}[1]{%
  \noalign{\vskip\aboverulesep
           \global\let\@dashdrawstore\adl@draw
           \global\let\adl@draw\adl@drawiv}
  \cdashline{#1}
  \noalign{\global\let\adl@draw\@dashdrawstore
           \vskip\belowrulesep}}
\makeatother

\titlespacing*{\paragraph}{0pt}{1ex plus 0.5ex minus 0.2ex}{0.5ex}

\newcommand{\simalign}{SimAlign\xspace}
\newcommand{\vgnsl}{\textsc{VG-NSL}\xspace}
\newcommand{\vgnslai}{\textsc{VG-NSL}+\textsc{ai}\xspace}
\newcommand{\avnsl}{\textsc{AV-NSL}\xspace}

\newcommand{\subdp}{\textsc{SubDP}\xspace}

\newcommand{\gtlt}{\textsc{G2L2}\xspace}
\newcommand{\ckyee}{\textsc{CKY-E$^2$}\xspace}
\newcommand{\mbrexec}{\textsc{MBR-exec}\xspace}
\newcommand{\structiou}{\textsc{Struct-IoU}\xspace}
\newcommand{\parseval}{\textsc{ParsEval}\xspace}

\usetikzlibrary{decorations.pathreplacing}

\newtheorem{theorem}{Theorem}[chapter]
\theoremstyle{definition}
\newtheorem{corollary}[theorem]{Corollary}
\newtheorem{example}[theorem]{Example}
\newtheorem{proposition}[theorem]{Proposition}
\newtheorem{definition}[theorem]{Definition}
\newtheorem{problem}[theorem]{Problem}
\DeclareMathOperator*{\suchthat}{\textit{s.t.}}

\newcommand{\iou}{\textsc{IoU}\xspace}
\DeclareMathOperator*{\mathiou}{\iou}
\newcommand{\cI}{\mathcal{I}}
\newcommand{\cT}{\mathcal{T}}

\newcommand{\cU}{\mathcal{U}}

\newcommand{\bd}{{\bm{d}}}

\newcommand{\bn}{{\bm{n}}}
\newcommand{\bp}{{\bm{p}}}
\newcommand{\bq}{{\bm{q}}}
\newcommand{\bA}{{\bm{A}}}
\newcommand{\br}{{\bm{r}}}
\newcommand{\bz}{{\bm{z}}}
\tikzset{
  structiounode/.style={fill,circle,inner sep=1pt,outer sep=0pt}
}

\newcommand{\mbrbleu}{\textsc{MBR-bleu}\xspace}
\newcommand{\maxlikelihood}{ML\xspace}
\newcommand{\maxavglikelihood}{\textsc{MaLL}\xspace}

\newcommand{\mycell}[1]{\begin{tabular}[t]{@{}l@{}l}#1\end{tabular}}
\newcommand{\sigmadot}[2]{\sigma\left(\langle#1, e_{\textbf{#2}}\rangle\right)}

%% file: src/00-titlepage.tex
\begin{titlepage}
    \begin{center}
        \vspace*{\fill}
        \thispagestyle{empty}
        {\LARGE Learning Language Structures through Grounding} \\
        \vspace*{\baselineskip}
        \large
        by \\
        Haoyue Freda Shi \\
        \vspace*{4\baselineskip}
        A thesis submitted
        \\ in partial fulfillment of the requirements for \\
        the degree of\\
        \vspace*{\baselineskip}
        Doctor of Philosophy in Computer Science \\
        \vspace{\baselineskip}
        at the \\
        TOYOTA TECHNOLOGICAL INSTITUTE AT CHICAGO \\
        Chicago, Illinois \\
        \vspace*{\baselineskip}
        June 2024 \\
        \vspace*{2\baselineskip}

        \textit{Thesis Committee}: \\
        Kevin Gimpel (Thesis Advisor) \\
        Karen Livescu (Thesis Advisor) \\
        Roger P. Levy \\
        Luke Zettlemoyer \\
        \vspace*{\fill}
        \clearpage

        \vspace*{\fill}
        \thispagestyle{empty}
        Copyright \copyright\ 2024 by Haoyue Freda Shi. \\
        All rights reserved.
        \vspace*{\fill}
    \end{center}
\end{titlepage}

%% file: src/01-abstract.tex
\begin{abstract}
Language is highly structured, with syntactic and semantic structures, to some extent, agreed upon by speakers of the same language.
With implicit or explicit awareness of such structures, humans can learn and use language efficiently and generalize to sentences that contain unseen words.
Motivated by human language learning, in this dissertation, we consider a family of machine learning tasks that aim to learn language structures through \textit{grounding}.
We seek distant supervision from other data sources (i.e., grounds), including but not limited to other modalities (e.g., vision), execution results of programs, and other languages.

We demonstrate the potential of this task formulation and advocate for its adoption through three schemes, each shown in a separate part of this dissertation.
In Part I, we consider learning syntactic parses through visual grounding.
We propose the task of visually grounded grammar induction, which aims at learning to predict the constituency parse tree of a sentence by reading the sentence and looking at the corresponding image.
We present the first models to induce syntactic structures from visually grounded text and speech, and find that the visual grounding signals can help improve the parsing quality over language-only models.
As a side contribution, we propose a novel evaluation metric that enables the evaluation of speech parsing without text or automatic speech recognition systems involved.
In Part II, we propose two execution-aware methods to map sentences into corresponding semantic structures (i.e., programs).
One of them enables nearly perfect compositional generalization to unseen sentences with mild assumptions on domain knowledge, and the other significantly improves the performance of few-shot semantic parsing by leveraging the execution results of programs as a source of grounding signals.
In Part III, we propose methods that learn language structures from annotations in other languages.
Specifically, we propose a method that sets a new state-of-the-art performance on cross-lingual word alignment, without using any annotated parallel data.
We then leverage the learned word alignments to improve the performance of zero-shot cross-lingual dependency parsing, by proposing a novel substructure-based projection method that preserves structural knowledge learned from the source language.
\end{abstract}

%% file: src/02-0-acknowledgements.tex
\chapter*{Acknowledgements}
I am incredibly fortunate to have Professors Kevin Gimpel and Karen Livescu as my Ph.D. advisors.
Over the past years, Karen and Kevin have offered me the highest level of freedom (that I can imagine) on research topics while reminding me of my dissertation outline and have been incredibly supportive in everything.
I usually filled our research meetings with scattered, vague, and probably crazy ideas, many of which did not make much sense, and many of them have not been realized yet.
No matter how senseless the ideas were, my advisors always listened to me patiently and provided constructive feedback that helped make the ideas concrete and realistic.
They answered all my beginner questions such as \textit{``what is prosody''} and \textit{``what is bitext''}, and responded with an absolute yes to almost all my unreasonable requests, from editing papers on holiday nights to having them as my secondary reviewer---as the advisor to my future students, I wish myself could be (even half) as kind and patient as them.
I promise to keep the spirit of scientific rigor, curiosity, and kindness that they have shown to me, and I hope to be able to pass it on to my future students.

I have learned a lot from my mentors through internships and cross-institutional collaborations.
I am grateful to Roger Levy and Luke Zettlemoyer for the training, guidance, and support they offered me, and for their feedback on this dissertation as committee members.
I am fortunate to have been mentored twice by Sida Wang, the first author of \href{https://aclanthology.org/P16-1224.pdf}{my favorite ACL paper} so far, and have learned a lot of concrete research skills from him.
I am grateful to Denny Zhou for hosting me in the final internship and for his support on all my projects.
I am also in debt to my academic mentors before coming to TTIC.
Thanks to Lei Li and Hao Zhou for their guidance and continued support---I worked with them at the ByteDance AI lab, and now it is great to see they have both returned to academia and have started fostering the next generation of computer scientists.
Thanks to Sam Bowman for the initial guidance when I entered the fantastic world of neural NLP and, perhaps more importantly, for introducing TTIC to me.

Jiayuan Mao deserves special thanks for being my closest collaborator and friend over the past years, for brainstorming with me, for providing fruitful comments on almost all my papers, for sharing research or non-research stuff that amused me a lot, and for teaching me random knowledge in robotics, music, and photography.

I am grateful to Michael Bowling, Nidhi Hegde, and Dale Schuurmans---the extended definition of \textit{grounding} in this dissertation has become much clearer and more articulable after the discussion with them at a wonderful dinner in Edmonton.
Thanks to David McAllester for his insightful opinions on grounding---he is an anti-grounding person,\footnote{Although I believe the definition of grounding (\cref{sec:bg-what-is-grounding}) in this dissertation is different from the one that David holds his opinion against.} but his opinions have significantly helped me clarify my thoughts on grounding.
Thanks to Yoav Artzi, Noriyuki Kojima, and Wentao Wang, as well as many anonymous conference reviewers, for their feedback on the work presented in this dissertation.

I am grateful to my coauthors, mentors and peers who have taught me a lot: thanks to Armen Aghajanyan, Xinyun Chen, Ed Chi, Dipanjan Das, David Dohan, Daniel Fried, Yoon Kim, Lingyu Gao, David Harwath, Jim Glass, Jessy Lin, Marjan Ghazvininejad, Vikram Gupta, Jeff Lai, Mike Lewis, Kanishka Misra, Puyuan Peng, Mrinmaya Sachan, Nathan Scales, Nathanael Sch\"{a}rli, Bowen Shi, Mirac Suzgun, Josh Tenenbaum, Shubham Toshniwal, Eric Wallace, Xuezhi Wang, Jason Wei, Jiajun Wu, Scott Yih, Ruiqi Zhong, and folks who organized and participated in the NL-Augmenter project for the wonderful collaboration; thanks to Allyson Ettinger, Sanghee Kim, Jiangtian Li, Jiaxuan Li, Zi Lin, Peng Qian, Weiwei Sun and Yuhan Zhang for inspiring conversations in linguistics and cognitive sciences; thanks to Hongyuan Mei and Lili Mou for discussions and encouragement around research ideas and career plans; thanks to the members of the TTIC speech and language reading group and the later TTIC-UChicago joint NLP reading group, especially Mingda Chen, Chung-Ming Chien, Ju-Chieh Chou, Kartik Goyal, Yushi Hu, Ruotian Luo, Richard Pang, Ankita Pasad, Shane Settle, Karl Stratos, Chenhao Tan, Qingming Tang, Chih-chan Tien, Lifu Tu, Haochen Wang, Sam Wiseman, Davis Yoshida, David Yunis, and Jiawei Zhou for all the delightful conversations.

TTIC is a fantastic place to do research, and I am grateful to the faculty and staff members for their efforts to make TTIC even better.
I have benefitted a lot from the awesome courses and/or conversations with Avrim Blum, Julia Chuzhoy, Greg Shakhnarovich, Madhur Tulsiani, Matthew Turk, and Matthew Walter, as well as the (probably globally best) administrative support from Adam Bohlander, Rose Bradford, Erica Cocom, Chrissy Coleman, Jessica Jacobson, Deree Kobets, Mary Marre, Alicia McClarin, and Amy Minick.

I gratefully acknowledge that I have been supported by a Google Ph.D. fellowship since 2021.
The fellowship is an honor, a financial endowment that provides me research freedom and, more importantly, great mental support during the darkest time caused by COVID-19 and the subsequent disasters on the faraway land I most care about.

I would like to express my gratitude to my friends and extended family for their support and encouragement.
Thanks to Qian Li, Jingye Tian, and Yuemei Zhang for sharing news and thoughts with me and being super responsive whenever I pinged them for a random chat.
Thanks to Xinyuan Zhang for all her cute arts.
Thanks to Yvonne Han for always sharing her excellent cooking and brilliant tiny items, and for showing me that the world is much more fantastic than what I have already explored.
Thanks to Xiao Han for being my amazing cousin and sharing his research, life stories, and thoughts with me.
Thanks to Hexiang Hu and Tete Xiao for keeping me posted on all kinds of news and for discussing the industrial NLP breakthroughs with me.
Thanks to Jiading Fang and Han Shao for their friendship and for keeping me more informed.
Thanks to Hanqing Zhao, whom I consider the best poet I have ever had the pleasure to meet in person, for sharing her poems and thoughts with me, and for shaping me into (what I proudly call) a 0th-generation Hanqingist poetry critic.
Thanks to friends in Canada for making my life there colorful (misspelled intentionally :)---we will have a longer journey together!

I never met the people listed below or was not even born when some of them passed away.
However, I am grateful for their impact on my life: thanks to John Stith Pemberton for inventing Coca-Cola---I suspect that I would not have been able to complete this dissertation without Coke.
Thanks to Adonis, Elizabeth Bishop, Luis Cernuda, and Wis\l{}awa Szymborska for their great poetry that accompanied me during sleepless nights.

I have considered substituting the following paragraph with a banal one but eventually opted to retain it in its current form.
I would indeed prefer not to be born if I had the option, but family is never a reason for my pessimism.
I could not be more blessed to have my family---Mom, Dad, Yudong, and the furry family members Wis\l{}awa, Ludwig, and Heinrich: thanks much to you all for always standing with me so that I gain enough courage to face the ridiculously meaningless life.

Fortunately, this dissertation is not solely about meaning, and we stand on the ground.

%% file: src/02-2-dedication.tex
\begin{center}
{\LARGE \textbf{DEDICATION}}
\hspace{0pt}
\vfill
\textit{
To Mingxin Liu (1994--2019): \\
Yet another piece of evidence that we miss you.
}
\vfill
\hspace{0pt}
\end{center}

%% file: src/10-intro.tex
\chapter{Introduction}
Language is highly structured.
Most natural languages naturally appear with sequential structures: sentences usually consist of a sequence of words; phrase structures are considered fundamental to natural languages for the ability of humans to handle nonadjacent and hierarchical dependencies between words \interalia{chomsky1957syntactic};
over the past decades, multiple other syntactic and semantic structure formalisms have been proposed and studied \interalia{tesniere1959elements,joshi-etal-1975-tree,steedman-2000-syntactic}.

On the other hand, humans learn and use these language structures naturally and implicitly---through communication and interaction with others, humans develop the ability to understand and produce grammatical sentences to describe objects and scenarios in the world.
Explicitly annotated structures, however, are almost never shown to humans during the learning process.

In this dissertation, we consider the following types of language structures as our representative targets of learning:
\begin{itemize}
    \item \textbf{Phrase structures}.
          Phrase structures are fundamental to natural languages for the human ability to handle nonadjacent and hierarchical dependencies between words \interalia{chomsky1957syntactic}. In this dissertation, we specifically consider the problem of learning phrase structures from parallel vision.
    \item \textbf{Dependency structures}.
          Dependency structures \citep{tesniere1959elements} are another syntactic structure widely used in NLP.
          Unlike phrase structures, dependency structures allow more flexibility by directly modeling word relationships.
          In this dissertation, we consider learning dependency structures through cross-lingual grounding signals, where we start with a well-trained dependency parser in a high-resource language and use bitext to guide the learning process in low-resource languages.
    \item \textbf{Combinatory categorial grammar (CCG)}.
          Beyond syntax, CCG provides a joint syntactic and semantic formalism for natural language.
          As a phrase-structure grammar,\footnote{
              We acknowledge that there are different conventions in defining phrase-structure grammar. For example, \citet{jurafsky-martin-2009-speech} use the term \textit{phrase-structure grammar} interchangeably with \textit{context-free grammar}, which excludes CCG.
              Here, we use the term \textit{phrase-structure grammar} to refer to a grammar that models the hierarchical phrase structures, in contrast to dependency grammar.
          } CCG additionally provides a type-driven compositionality mechanism for semantic representation and composition.
          We consider the problem of learning CCG from parallel vision and execution results of the induced programs.
    \item \textbf{Executable programs}.
          We consider executable programs as semantic structures of corresponding natural language utterances, where executing them with optionally world knowledge as the input grounds the abstract sentences into the real world.
          To this end, we investigate the problem of execution-informed semantic parsing, which converts natural language to executable programs without explicit supervision.
\end{itemize}

Enabling machine learning of language structures resembling human-like learning holds both theoretical and practical potential.
From a theoretical perspective, learning language structures without explicit supervision can benefit the study of syntax acquisition by providing evidence supporting or challenging the poverty of the stimulus hypothesis \citep{piattelli1980language}.
From a practical perspective, language structures learned without explicit supervision can affordably enhance the compositional generalization ability of text processing systems to out-of-distribution data \interalia{havrylov2019cooperative,mao2021grammar}, which provides an alternative to the traditional supervised learning paradigm that requires large-scale human annotations.
Automatic learning and prediction of language structures can also benefit a wide range of natural language processing (NLP) applications, such as linguistics research, second language instruction, and program synthesis from natural language commands.

This dissertation addresses the challenges of learning language structures through grounding, from syntax to semantics.
We are interested in exploring the potential of grounding signals that are naturally parallel to natural languages as a source of indirect supervision.
We focus on using grounding signals in the real world, which typically require less human effort than those needed for supervised learning.
In addition to providing richer information about surrounding environments, these grounding signals enable more significant potential in real-world applications by offering a more comprehensive model of the environments where humans live.
Specifically, we investigate several representative types of grounding signals, including:
\begin{itemize}
    \item \textbf{Parallel vision}.
          Captions are often paired with images that share the same or similar meanings in a different modality.
          Thus, we can consider using visual information to supervise the learning of language structures.
    \item \textbf{Execution results of programs}.
          For natural language sentences with associated executable programs, we may execute the programs and use the execution results to guide the learning process of language structures. Given an optional input, a program can be executed with appropriate interpreters.
          The output, i.e., the execution results, can be considered a source of grounding signal.
    \item \textbf{Parallel sentences in other languages (bitext)}.
          There are around 7,000 languages all over the world.
          Sentences in different languages often have parallel ones with similar meanings in other languages.
          Once we have a reliable understanding of sentences in one language, we can use parallel sentences as grounding signals to improve or even enable understanding of other languages.
\end{itemize}

\section{Dissertation Structure and Contribution Summary}
This dissertation consists of three parts, exploring the potential of grounding in learning both syntactic and semantic structures of natural language.
Before delving into the details of each part, we first provide a brief overview of the background by introducing the involved language structures and evaluation metrics for the learning systems and offering an extended definition of grounding (\cref{chapter:background}).
Work in this dissertation can be formulated as learning language structures through grounding, where all work involved in this dissertation is based on the assumption that grounding signals provide indirect but useful supervision for learning language structures.
We provide a different learning scheme to instantiate the general framework in each part.

In \cref{part:syntax-vision}, we discuss learning syntactic structures through visual grounding.
We introduce the task of visually grounded grammar induction, which we use as a testbed to investigate the potential of grounding signals in learning syntactic structures.
We present \vgnsl(\cref{chapter:vgnsl}), a neural model that learns to parse sentences into its constituency parse structures by grounding them to corresponding visual scenes.
Experiments show that \vgnsl can induce phrase structures from parallel vision, outperforming existing text-only grammar induction methods, and can be extended to multiple languages.
We extend \vgnsl to induce syntactic structures from visually grounded speech (\cref{chapter:avnsl}), and show that the model can induce meaningful phrase structures by breaking down an utterance into a few constituent-like segments.
Along this line, we propose a new evaluation metric for speech constituency parsing (\cref{chapter:structiou}), to enable the evaluation of speech parsing models in the absence of ground-truth transcriptions or an automatic speech recognition system.
This evaluation metric can be naturally extended to evaluate text constituency parsing, providing an additional perspective to the existing evaluation metrics, such as \parseval \citep{black-etal-1991-procedure}.

In \cref{part:semantics-execution}, we consider learning semantic structures through grounding with program execution.
We first extend the task of syntax induction (\cref{chapter:vgnsl,chapter:avnsl}) to joint syntax and semantics induction through joint visual grounding and program execution (\cref{chapter:g2l2}).
We show that the theoretically motivated model enables nearly perfect compositional generalization to unseen sentences and scenes, with mild assumptions on inductive biases and domain knowledge.
Considering general-purpose semantic parsing that converts natural language utterances to executable programs, we ground the pre-trained language models to the real world by executing the induced programs and propose a new decoding method to improve the execution accuracy of output programs (i.e., semantic parses of sentences; \cref{chapter:mbrexec}).
For the first time, we show that few-shot (e.g., fifteen examples) translation from natural language to executable programs can achieve the performance of supervised methods, which require thousands of annotated training examples.

In \cref{part:cross-lingual}, we investigate the potential of cross-lingual grounding signals in learning dependency parsing structures, another representative formalism of syntactic parsing.
We first propose a lightweight method to extract cross-lingual word alignment, which may also be considered a type of language structure, from pre-trained contextualized language models (\cref{chapter:mlpalign}).
Our system achieves state-of-the-art performance across languages without accessing parallel sentences usually used in previous work.
We then propose a method to learn zero-shot cross-lingual dependency syntax by grounding it to a supervised parser trained on another language (\cref{chapter:subdp}).
We demonstrate that our method efficiently preserves and transfers structural knowledge learned in the source-language (i.e., ground) parsing system to the target language, achieving a new state-of-the-art performance in zero-shot cross-lingual dependency parsing.

Finally, we conclude the dissertation by discussing the contributions of each part and the potential future directions beyond learning language structures through grounding (\cref{chapter:discussion}).

\section{A Quick Guide to Readers}
This dissertation may be particularly interesting to readers interested in one or more of the following topics in natural language processing and computational linguistics: machine language acquisition and grounded language learning, syntactic and semantic parsing, program synthesis, lexical semantics, and cross-lingual NLP.
Additionally, the evaluation metric proposed in \cref{chapter:structiou} may be interesting to readers looking for a problem that can be solved in polynomial time with tree-based dynamic programming algorithms.
We list the following topics and the relevant chapters in which they are discussed:
\begin{itemize}
    \item Grounding: \cref{sec:bg-what-is-grounding}.
    \item Modeling language acquisition: \cref{chapter:vgnsl,chapter:avnsl,chapter:g2l2}.
    \item Algorithms for structured prediction: \cref{chapter:structiou,chapter:g2l2,chapter:subdp}.
    \item Syntactic parsing: \cref{chapter:vgnsl,chapter:avnsl,chapter:structiou,chapter:structiou,chapter:g2l2,chapter:subdp}.
    \item Semantic parsing: \cref{chapter:g2l2,chapter:mbrexec}.
    \item Cross-modal NLP: \cref{chapter:vgnsl,chapter:avnsl,chapter:g2l2}.
    \item Cross-lingual NLP: \cref{chapter:mlpalign,chapter:subdp}.
\end{itemize}

We encourage readers familiar with each piece of work to read the corresponding discussions, i.e., the last section of each chapter.
In these discussions, we provide a more comprehensive analysis, placing the work within the broader context of the field as of the year 2024 and suggesting potential directions for future research.

%% file: src/20-background.tex
\chapter{Background}
\label{chapter:background}
This chapter provides background information on the topics relevant to the work presented in this dissertation, including different types of language structures and grounding signals.
In \cref{sec:bg-language-structures}, we provide background knowledge on syntax (\cref{sec:bg-constituency-syntax,sec:bg-dependency-syntax}), semantics (\cref{sec:bg-executable-programs}), and cross-lingual word alignment (\cref{sec:bg-cross-lingual-alignment}) as language structures.
For more details, we recommend readers refer to the textbooks on NLP, e.g., \cite{jurafsky-martin-2009-speech}.
In \cref{sec:bg-what-is-grounding}, we discuss the concept of \textit{grounding} in this dissertation.
Our definition slightly extends the one given by \citet{harnad-1990-symbol}, and arguably covers the notion of \textit{grounding} in natural language processing, (computational) linguistics, robotics, and cognitive sciences.
In this dissertation, vision, program execution results, and cross-lingual supervision are considered representative types of grounding signals.

\section{Language Structures}
\label{sec:bg-language-structures}
\subsection{Phrase Structures and Phrase-Structure Grammars}
\label{sec:bg-constituency-syntax}
A group of words is considered as a \textit{constituent},\footnote{For simplicity in this dissertation, we only consider continuous constituents formed by consecutive words in a sentence---while this case covers most of the constituents in English declarative sentences, it is worth noting that there exist discontinuous constituents, which typically appear in English interrogative sentences, or even declarative sentences in other languages with more flexible word orders (e.g., German and Czech). } or a \textit{phrase}, if it behaves as a single unit in a sentence.
A \textit{constituency parse tree} represents the hierarchical phrase structure of a sentence by recursively breaking down the sentence into constituents.
For example, the sentence ``\textit{The cat sat on the mat}'' can be parsed into a constituency parse tree in \cref{fig:bg-constituency-syntax-example}.
In this tree, the sentence (represented by the symbol S) is divided into a noun phrase (NP) and a verb phrase (VP), and the phrases are further divided into smaller constituents until the leaves are reached.
The symbols in the tree can be classified into two categories: non-terminal symbols (e.g., S, NP, VP, and PP), which usually represent multi-word phrases in an NLP context, and pre-terminal symbols (e.g., DT, NN, V, and IN), which usually represents word categories (or more specifically, part-of-speech tags).

\begin{figure}[t!]
    \centering
    \begin{subfigure}{0.45\textwidth}
        \begin{forest}
            [S
                    [NP
                            [DT [\it The]]
                            [NN [\it cat]]
                    ]
                    [VP
                            [V [\it sat]]
                            [PP
                                    [IN [\it on]]
                                    [NP
                                            [DT [\it the]]
                                            [NN [\it mat]]
                                    ]
                            ]
                    ]
            ]
        \end{forest}
    \end{subfigure}
    \begin{subfigure}{0.45\textwidth}
        \begin{tabular}{ll}
            \toprule
            \bf Symbol & \bf Description                      \\
            \midrule
            \multicolumn{2}{l}{\textit{Non-terminal symbols}} \\
            \midrule
            S          & Sentence                             \\
            NP         & Noun Phrase                          \\
            VP         & Verb Phrase                          \\
            PP         & Prepositional Phrase                 \\
            \midrule
            \multicolumn{2}{l}{\textit{Pre-terminal symbols}} \\
            \midrule
            DT         & Determiner                           \\
            NN         & Noun                                 \\
            V          & Verb                                 \\
            IN         & Preposition                          \\
            \bottomrule
        \end{tabular}
    \end{subfigure}
    \caption{The constituency parse tree of the sentence ``\textit{The cat sat on the mat}''.}
    \label{fig:bg-constituency-syntax-example}
\end{figure}

Constituency parse trees can be generated with \textit{phrase-structure grammars}, which specifies a set of rules that describe how non-terminal symbols can be rewritten into sequences of terminal and non-terminal symbols.
A representative formalism for phrase-structure grammar is \textit{context-free grammar} (CFG), which consists of a set of production rules that specify how non-terminal symbols can be rewritten into sequences of terminal and non-terminal symbols, and the rewriting process of one non-terminal symbol is independent of the context.
For example, the CFG that generates the constituency parse tree in \cref{fig:bg-constituency-syntax-example} can be defined as follows:
\begin{align*}
    \text{S}  & \rightarrow \text{NP VP} \\
    \text{NP} & \rightarrow \text{DT NN} \\
    \text{VP} & \rightarrow \text{V PP}  \\
    \text{PP} & \rightarrow \text{IN NP}
\end{align*}
with additional lexical rules that rewrite the pre-terminal symbols into terminal symbols (i.e., words in this case).

In the past few decades, the task of supervised constituency parsing has been widely studied in NLP, and various methods have been proposed to predict constituency parse trees from sentences \interalia{collins-koo-2005-discriminative,charniak-johnson-2005-coarse,mcclosky-etal-2006-effective}.
At the training stage, the model is trained to predict the manually annotated constituency parse tree, e.g., the Penn Treebank \citep[PTB; ][]{marcus-etal-1993-building}, given a sentence and optionally the ground-truth part-of-speech tags.
For unsupervised constituency parsing or phrase-structure induction, the task is often formulated as predicting the constituency parse trees, given from a set of sentences without parse tree annotations \interalia{klein2002generative}---the model will induce the phrase structures from statistical patterns in the sentences and the inductive biases specified for the model.

\begin{figure}[t!]
    \centering
    \begin{subfigure}{0.45\textwidth}
        \centering
        \begin{forest}
            [S
                    [NP
                            [DT [\it The]]
                            [NN [\it cat]]
                    ]
                    [VP
                            [V [\it sat]]
                            [PP
                                    [IN [\it on]]
                                    [NP
                                            [DT [\it the]]
                                            [NN [\it mat]]
                                    ]
                            ]
                    ]
            ]
        \end{forest}

        \begin{tabular}{l}
            \toprule
            \bf Brackets   \\
            \midrule
            \bf S: [1, 6]  \\
            \bf NP: [1, 2] \\
            \bf VP: [3, 6] \\
            PP: [4, 6]     \\
            NP: [5, 6]     \\
            \bottomrule
        \end{tabular}
    \end{subfigure}
    \begin{subfigure}{0.45\textwidth}
        \centering
        \begin{forest}
            [S
                    [NP
                            [DT [\it The]]
                            [NN [\it cat]]
                    ]
                    [VP
                            [V [\it sat]]
                            [NP
                                    [DT [\it on]]
                                    [NN [\it the]]
                                    [NN [\it mat]]
                            ]
                    ]
            ]
        \end{forest}
        \vspace{30pt}

        \begin{tabular}{l}
            \toprule
            \bf Brackets   \\
            \midrule
            \bf S: [1, 6]  \\
            \bf NP: [1, 2] \\
            \bf VP: [3, 6] \\
            NP: [4, 6]     \\
            \bottomrule
        \end{tabular}
    \end{subfigure}
    \caption[Illustration of the bracket-based $F_1$ score.]{Illustration of the bracket-based $F_1$ score. The predicted tree (right) is compared with the gold tree (left) to compute the precision, recall, and $F_1$ score over brackets. The brackets that exist in both trees are in boldface.}
    \label{fig:bg-constituency-syntax-example-eval}
\end{figure}
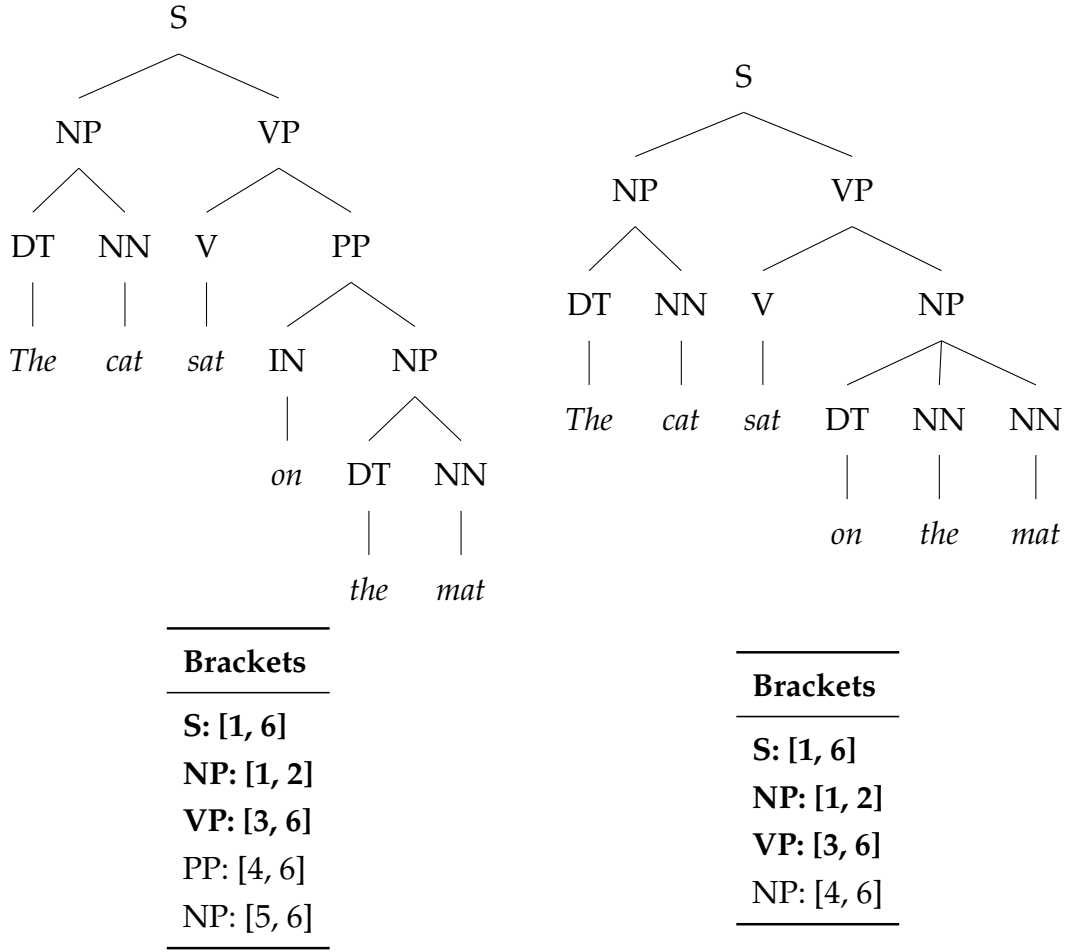

Standard metrics exist for evaluating the quality of constituency parse trees by comparing the predicted parse tree with the gold parse tree.
Among them, the most widely used metric is the $F_1$ score over brackets \citep{black-etal-1991-procedure,sekine-collins-1997-evalb}, which is computed based on the precision and recall of constituents (represented by the brackets).
Consider the trees presented in \cref{fig:bg-constituency-syntax-example-eval} as an example.
The predicted tree is compared with the gold tree to compute the precision, recall, and $F_1$ score over brackets.
In the evaluation process, the constituency parse trees are represented by brackets, where each bracket corresponds to the left and right boundaries (inclusive) of a constituent.
In the example, there are five brackets in the gold tree and four brackets in the predicted tree.
Among them, three brackets are in common, and the precision, recall, and $F_1$ score, which is the harmonic mean between the precision and recall, are computed as follows:
\begin{align*}
    \textit{Precision} & = \frac{3}{4},                                                                      \\
    \textit{Recall}    & = \frac{3}{5},                                                                      \\
    F_1                & = \frac{2}{\frac{1}{\textit{Precision}} + \frac{1}{\textit{Recall}}} = \frac{2}{3}.
\end{align*}

The task of phrase-structure grammar induction (also known as unsupervised constituency parsing) aims to induce phrase-structure grammar from a set of sentences without any explicit supervision.
The task is often formulated as finding the best parse tree for each sentence in the training set and then inducing the grammar that generates the parse trees.
Conventionally, the evaluation of phrase-structure grammar induction has been focused on comparing the unlabeled constituency parse trees, which ignore the specific labels of the non-terminal symbols; that is, all non-terminal symbols are treated as the same.
Work on grammar induction presented in this dissertation (\cref{chapter:vgnsl,chapter:avnsl}) also follows this convention, and the evaluation metrics are based on the unlabeled parse trees.

\subsection{Dependency Syntax}
\label{sec:bg-dependency-syntax}
\begin{figure}[t]
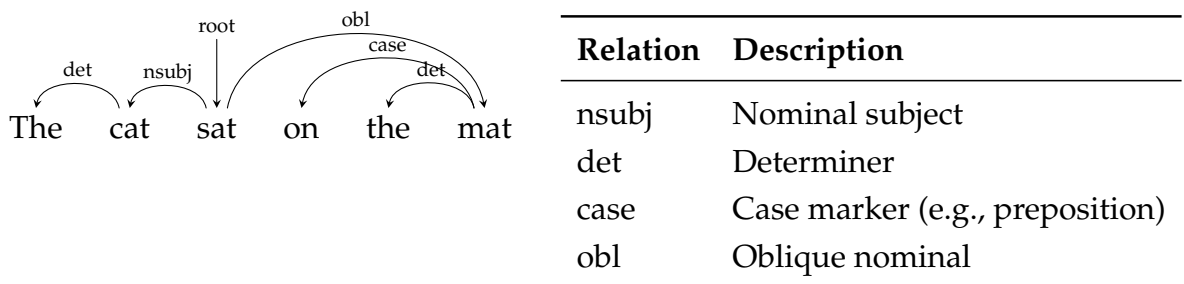

    \centering
    \begin{subfigure}{0.45\textwidth}
        \begin{dependency}[theme = simple]
            \begin{deptext}[column sep=1em]
                The \& cat \& sat \& on \& the \& mat \\
            \end{deptext}
            \depedge{2}{1}{\small det}
            \depedge{3}{2}{\small nsubj}
            \deproot[edge unit distance=1.5ex]{3}{\small root}
            \depedge{6}{4}{\small case}
            \depedge{6}{5}{\small det}
            \depedge{3}{6}{\small obl}
        \end{dependency}
    \end{subfigure}
    \begin{subfigure}{0.5\textwidth}
        \begin{tabular}{ll}
            \toprule
            \bf Relation & \bf Description                 \\
            \midrule
            nsubj        & Nominal subject                 \\
            det          & Determiner                      \\
            case         & Case marker (e.g., preposition) \\
            obl          & Oblique nominal                 \\
            \bottomrule
        \end{tabular}
    \end{subfigure}
    \caption{The dependency parse tree of the sentence ``\textit{The cat sat on the mat}'', annotated following the Universal Dependencies \citep{nivre-etal-2020-universal} scheme.}
    \label{fig:bg-dependency-syntax-example}
\end{figure}

In contrast to phrase-structure grammar, dependency grammar represents syntax by binary asymmetric relations between words in a sentence.
A dependency parse tree represents the dependency relations between words in a sentence, where each word is a node in the tree, and the edges between the nodes represent the dependency relations \citep{tesniere1959elements}.
Each edge is directed from the \textit{head} to the \textit{dependent}, and the head of an entire sentence is denoted a special root node.
According to the definition of a tree, each node has exactly one incoming edge (connecting itself as the dependent to its head) except for the root node, and there is a unique path from the root node to any other node in the tree.
As an example, the dependency parse tree of the sentence ``\textit{The cat sat on the mat}'' is shown in \cref{fig:bg-dependency-syntax-example}.

In supervised dependency parsing \interalia{mcdonald-2006-discriminative,nivre-2004-incrementality,nivre-2008-algorithms}, the task is to predict the dependency parse tree of a sentence given the sentence and, optionally, the ground-truth part-of-speech tags.
The predicted parse tree is evaluated by comparing it with the ground-truth parse tree, and the evaluation metrics include the labeled attachment score (LAS) and the unlabeled attachment score (UAS).
The LAS is the percentage of words in the sentence assigned with both the correct head and correct dependency relation, and the UAS is the percentage of words in the sentence assigned the correct head regardless of the dependency relation.
Consider the trees presented in \cref{fig:bg-dependency-syntax-example} as an example.
The predicted tree is compared with the gold tree to compute the LAS and UAS.
There are six words (and, therefore, six edges).
The predicted tree has four correct edges, one edge with the correct head and dependent but an incorrect label (\textit{sat} to \textit{mat}), and one edge with incorrect head (\textit{the} to \textit{on}), resulting in the LAS of $\frac{4}{6}$ and the UAS of $\frac{5}{6}$.

\begin{figure}[t]
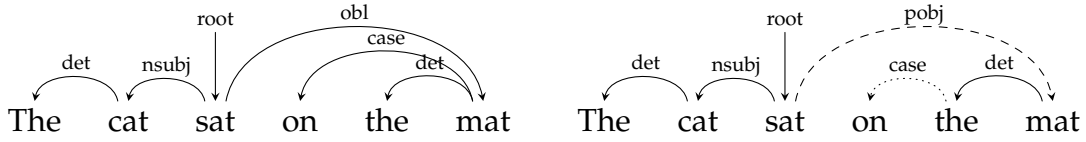

    \centering
    \begin{subfigure}{0.45\textwidth}
        \centering
        \begin{dependency}[theme = simple]
            \begin{deptext}[column sep=1em]
                The \& cat \& sat \& on \& the \& mat \\
            \end{deptext}
            \depedge{2}{1}{\small det}
            \depedge{3}{2}{\small nsubj}
            \deproot[edge unit distance=1.5ex]{3}{\small root}
            \depedge{6}{4}{\small case}
            \depedge{6}{5}{\small det}
            \depedge{3}{6}{\small obl}
        \end{dependency}
    \end{subfigure}
    \begin{subfigure}{0.45\textwidth}
        \centering
        \begin{dependency}[theme = simple]
            \begin{deptext}[column sep=1em]
                The \& cat \& sat \& on \& the \& mat \\
            \end{deptext}
            \depedge{2}{1}{\small det}
            \depedge{3}{2}{\small nsubj}
            \deproot[edge unit distance=1.5ex]{3}{\small root}
            \depedge[dotted]{5}{4}{\small case}
            \depedge{6}{5}{\small det}
            \depedge[dashed]{3}{6}{\small pobj}
        \end{dependency}
    \end{subfigure}
    \caption[Illustration of how LAS and UAS work as dependency parsing metrics.]{
        Illustration of how LAS and UAS work as dependency parsing metrics.
        The predicted tree (right) is compared with the gold tree (left) to compute the LAS and UAS.
        The edges in the predicted tree that LAS considers mismatched are dashed, whereas those by UAS (and, therefore, also LAS) are dotted.
    }
\end{figure}

\subsection{Executable Programs as Semantics}
\label{sec:bg-executable-programs}
\subsubsection{Programs}
In this dissertation, we follow existing work in NLP \interalia{zelle-mooney-1996-learning,zettlemoyer-collins-2005-learning,liang-etal-2013-learning} to consider the task of semantic parsing as translating natural language utterances into logical forms.
In addition, an executable program (e.g., a Python program) can be considered as a logical form in a relaxed sense, and we can therefore use the execution results to evaluate program semantics.
For example, let $u_i$ denote a natural-language utterance and $p_i$ denote its paired ground-truth Python program.
Suppose we have a test set of $N$ pairs of utterances and ground-truth programs $\{(u_1, p_1), \ldots, (u_N, p_N)\}$.
The execution accuracy of predicted programs $P' = \{p'_i\}$ is defined as the percentage of the programs that are executed correctly, i.e., the outputs of the program match the expected ones:
\begin{align*}
    \textit{Execution Accuracy}(P') = \frac{1}{N} \sum_{i=1}^N \mathbbm{1}\left[\textit{Execute}(p_i') = \textit{Execute}(p_i)\right],
\end{align*}
where $\mathbbm{1}[\cdot]$ is the indicator function that returns 1 if the condition is true and 0 otherwise, and the function $\textit{Execute}(\cdot)$ maps a program to its execution results.
For simplicity, we assume that the execution results are deterministic and can be represented as a string to support the comparison.
Practically, some programs require input arguments to execute, and we synthesize a few input cases so that we can approximate the program semantics by comparing the execution results on specific input cases (see more discussions in \cref{chapter:mbrexec}).
In addition, since some programs may not terminate, we empirically set a timeout threshold for the execution of each program and consider programs that exceed the threshold to have a different execution result from any other programs.

\subsubsection{Combinatory Categorial Grammars}
Linguists have presented joint formalism of syntax and semantics.
This dissertation considers the combinatory categorial grammars \citep[CCGs;][]{steedman-2000-syntactic} as a representative formalism that combines syntax and semantics.
Each word token in a sentence is associated with a syntactic category and a semantic program---the syntactic category here can be viewed as the function signature in a programming language, and the semantic program corresponds to the function implementation.\footnote{Generally, different word tokens with the same word type may be associated with different syntactic categories and semantic programs. A representative example is the word \textit{word}, which can be a noun or a verb in different contexts.}

For illustrative purposes, we provide the CCG derivation of the sentence ``\textit{A cat drinks milk}'' in \cref{fig:bg-ccg-example}.
Each lexical entry is associated with a syntactic category (e.g., \textit{NP} and \textit{(S\bs NP)/NP}) and a semantic program (represented by lambda calculus), and the derivation process combines the categories and programs of the words in the sentence to derive the final semantic program of the sentence.
At each step, the derivation process applies a rule that combines the categories and programs of two adjacent words to derive a new category and program.
The operators / and \bs\xspace denote the forward and backward application, respectively---taking forward application as an example, the category $X/Y$ expects an argument of category $Y$ at the right side, and the category $Y$ is combined with the category $X/Y$ to derive a new category $X$.
Note that the syntactic categories are not necessarily the same as the part-of-speech tags, and the semantic programs can be instantiated in executable programs.
As a comprehensive example, we offer a detailed introduction of the instantiation of CCG we used in this dissertation in \cref{chapter:g2l2}.
In this work, we consider the execution accuracy of the finally derived function (in the \cref{fig:bg-ccg-example} example, \textit{drinks}'(\textit{cat}', \textit{milk}')) as the primary evaluation metric.

\begin{figure}
    \centering
    \deriv{4}{
        \rm A & \rm cat &\rm drinks &\rm milk \\
        \uline{2}&\uline{1}&\uline{1}\\
        & \it NP &\it (S{\bs}NP)/NP &\it NP\\
        & \textit{cat}' & \lambda{} x.\lambda y. \textit{drinks}'(y, x) & \textit{milk}' \\
        & & \fapply{2} \\
        & & \mc{2}{\it S\bs NP}\\
        & & \mc{2}{{\lambda{} y. \textit{drinks}'(y, \textit{milk}')}} \\
        \bapply{4} \\
        \mc{4}{\it S}\\
        \mc{4}{\textit{drinks}'(\textit{cat}', \textit{milk}')} \\
    }
    \caption[An example of a CCG derivation for the sentence ``\textit{A cat drinks milk}.'']{An example of a CCG derivation for the sentence ``\textit{A cat drinks milk.}'' $>$ and $<$ denote forward and backward application, respectively.
        \label{fig:bg-ccg-example}
    }
\end{figure}

\subsection{Cross-Lingual Word Alignment}
\label{sec:bg-cross-lingual-alignment}
\begin{figure}[t]
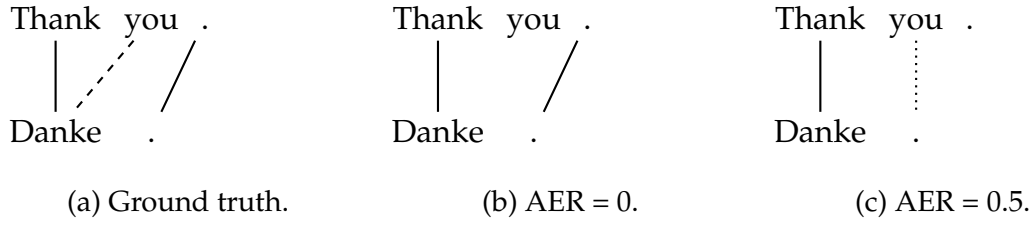

    \centering
    \begin{subfigure}{0.3\textwidth}
        \begin{dependency}[arc edge, arc angle=70, text only label, label style={above}]
            \begin{deptext}[column sep=.1cm]
                Thank \& you \& .\\[30pt]
                Danke \& .\\
            \end{deptext}
            \draw [-, thick, solid] (\wordref{1}{1}) -- (\wordref{2}{1});
            \draw [-, thick, dashed] (\wordref{1}{2}) -- (\wordref{2}{1});
            \draw [-, thick, solid] (\wordref{1}{3}) -- (\wordref{2}{2});
        \end{dependency}
        \caption{Ground truth.}
        \label{fig:bg-cross-lingual-alignment-example-a}
    \end{subfigure}
    \begin{subfigure}{0.3\textwidth}
        \begin{dependency}[arc edge, arc angle=70, text only label, label style={above}]
            \begin{deptext}[column sep=.1cm]
                Thank \& you \& . \\[30pt]
                Danke \& .\\
            \end{deptext}
            \draw [-, thick, solid] (\wordref{1}{1}) -- (\wordref{2}{1});
            \draw [-, thick, solid] (\wordref{1}{3}) -- (\wordref{2}{2});
        \end{dependency}
        \caption{AER = 0.}
    \end{subfigure}
    \begin{subfigure}{0.3\textwidth}
        \begin{dependency}[arc edge, arc angle=70, text only label, label style={above}]
            \begin{deptext}[column sep=.1cm]
                Thank \& you \& . \\[30pt]
                Danke \& .\\
            \end{deptext}
            \draw [-, thick, solid] (\wordref{1}{1}) -- (\wordref{2}{1});
            \draw [-, thick, dotted] (\wordref{1}{2}) -- (\wordref{2}{2});
        \end{dependency}
        \caption{AER = 0.5.}
    \end{subfigure}
    \caption[Example of cross-lingual word alignment between English and German.]{
        Example of cross-lingual word alignment between English and German.
        Solid line: required alignment; dashed line: optional alignment; dotted line: predicted alignment that does not exist in the ground truth.
    }
    \label{fig:bg-cross-lingual-alignment-example}
\end{figure}

The final language structure considered in this dissertation is cross-lingual word alignment, which aims to align words in a sentence in one language to words in a sentence in another language.
The alignment is typically represented as a set of word pairs, where each pair consists of a word in the source language and a word in the target language.
Such word alignment is often used as a preprocessing step for various cross-lingual tasks, especially for statistical machine translation \interalia{berger-etal-1994-candide}.
In addition to being a type of language structure, cross-lingual word alignment plays an important role in transferring knowledge from one language to another, and can therefore facilitate cross-lingual grounding (which we will discuss later in \cref{chapter:subdp}).
As an example, \cref{fig:bg-cross-lingual-alignment-example-a} shows the word alignment between the English sentence ``\textit{Thank you}'' and German sentence ``\textit{Danke}.''

The annotations of word alignment pairs can be categorized into two types: required alignment and optional alignment.
The required alignment is the word pairs that a word alignment model must align ---the model will be penalized if it fails to align these pairs.
In contrast, the optional alignment is the word pairs that the model may align, but the model is not penalized whether it aligns this pair or not.
In addition, the model will also be penalized if it aligns the word pairs that are not in the ground-truth annotation as either a required or an optional alignment.

The evaluation metric for cross-lingual word alignment is the alignment error rate, which is computed as follows.
Let $P = \{(s_i, t_i)\}$ denote the set of predicted alignment pairs, $R$ denote the required alignment pairs (solid lines in \cref{fig:bg-cross-lingual-alignment-example}), and $A$ denote the union of the required and optional alignments (both solid and dashed lines in \cref{fig:bg-cross-lingual-alignment-example}).
The alignment error rate (AER) is defined by
$$
    \textit{AER} = 1 - \frac{|P \cap A| + |P\cap R|}{|P| + |R|}.
$$

\cref{tab:bg-language-structures} summarizes the language structures considered in this dissertation, along with the corresponding evaluation metrics.

\begin{table}[t]
    \centering
    \begin{tabular}{p{0.12\textwidth}p{0.5\textwidth}p{0.25\textwidth}}
        \toprule
                                      & \bf Language Structure                & \bf Evaluation Metric                   \\
        \midrule
        \it (\cref{chapter:vgnsl})    & Constituency syntax                   & Bracket $F_1$                           \\
        \it (\cref{chapter:avnsl})    & Constituency syntax over spoken words & Bracket $F_1$, \cref{chapter:structiou} \\
        \it (\cref{chapter:g2l2})     & Combinatory categorial grammars       & Execution accuracy                      \\
        \it (\cref{chapter:mbrexec})  & Python programs                       & Execution accuracy                      \\
        \it (\cref{chapter:mlpalign}) & Word alignment                        & Alignment error rate                    \\
        \it (\cref{chapter:subdp})    & Dependency syntax                     & LAS, UAS                                \\
        \bottomrule
    \end{tabular}
    \caption{Language structures considered in this dissertation and the corresponding primary evaluation metrics.}
    \label{tab:bg-language-structures}
\end{table}

\section{Grounding}
\label{sec:bg-what-is-grounding}
The symbol grounding problem \citep{harnad-1990-symbol} presents a fundamental issue in the areas of artificial intelligence and cognitive sciences, which concerns how \textit{``the semantic interpretation of a formal symbol system be made intrinsic to the system, rather than just parasitic on the meanings in our heads,''} and  how \textit{``the meanings of the meaningless symbol tokens, manipulated solely on the basis of their (arbitrary) shapes, be grounded in anything but other meaningless symbols.''}
That is, the symbols in the system should be linked to meanings in the real world.\footnote{
    We acknowledge that \textit{meanings in the real world} is a vague description, and the definition of \textit{meaning} can vary drastically depending on the context.
}
However, there are discrepancies in the definition of \textit{grounding} under different contexts \citep{chai-etal-2018-language,mollo-milliere-2023-vector}.
In addition to the high-level concept of interpreting symbols in the real world, grounding can be interpreted differently in various scenarios, including but not necessarily limited to the following NLP-centric ones,\footnote{
    \citet{mollo-milliere-2023-vector} have offered a comprehensive discussion from a more philosophical perspective.
    Roughly speaking, our Scenario 1 corresponds to their \textit{referential grounding} and \textit{sensorimotor grounding}, Scenario 2 corresponds to their \textit{relational grounding} and \textit{epistemic grounding}, and Scenario 3 corresponds to their
    \textit{communicative grounding}.
} where some of them are indeed instantiations of the high-level definition by \citet{harnad-1990-symbol}:
\begin{itemize}
    \item \textbf{Scenario 1} (NLP and robotics): The process that links symbols (e.g., written words) and sequences of symbols (e.g., written sentences) to real-world entities and scenarios, represented by sensory signals, such as image \interalia{plummer-etal-2015-flickr30k} and audio \interalia{settle-etal-2019-acoustically}.
          While this definition follows the high-level concept of grounding, it is debatable whether all sensory signals represent meanings in an acceptable sense---for example, the spoken language is often considered a symbol system that lacks meanings; therefore, acoustically grounded word embeddings \citep{settle-etal-2019-acoustically} do not ground the symbol system represented by words to their meanings.
          Similarly, it depends on the specific visual content to determine whether the visual signals represent meanings---for example, it is arguable that pictures of written text do not represent meanings in the same way as pictures of objects do.
    \item \textbf{Scenario 2} (NLP and computational linguistics): The process of linking entities in natural language to an existing knowledge base, which is also referred to as the entity linking task \interalia{bunescu-pasca-2006-using}.
          This definition can be considered as an instantiation or extension of the symbol grounding problem in the context of NLP, with additional supervision (e.g., annotated links) being considered to train the machine processing systems.
    \item \textbf{Scenario 3} (NLP, pragmatics, and cognitive sciences): The common ground shared by the speaker and listener in a conversation in order to collaboratively communicate \interalia{grice-1975-logic,levinson-1983-pragmatics,clark-brennan-1991-grounding}.
          In NLP, this definition is often considered in the domain of dialogue systems \interalia{zhou-etal-2018-dataset}---the system should be able to share a common ground with the users to generate appropriate natural language utterances in contexts.
          This definition arguably contradicts the one offered by \citet{harnad-1990-symbol} in a narrow and literal sense, as it does not necessarily require the symbols to be linked to meanings in the real world; instead, the interpretation of the utterances are left to the speaker and listener.
\end{itemize}
The discussion on the above scenarios is extended from the taxonomy offered by \citet{chai-etal-2018-language}, where they consider the term grounding in two senses: \textit{semantic grounding} (Scenario 2, and part of Scenario 1) and \textit{communicative grounding} (Scenario 3).
For a detailed discussion covering more recent related work on language grounding, we recommend readers refer to the survey by \citet{bisk-etal-2020-experience}.
For a comprehensive discussion of recent work in NLP from a common-ground-centric view (i.e., Scenario 3 above), we recommend the survey by \citet{chandu-etal-2021-grounding}.

In this dissertation, we downplay the philosophical aspects of grounding and its connection with meanings, and give an extended definition of grounding from an arguably more natural perspective in machine learning.
We consider grounding as a process that connects data from one source $\mathcal{X}$ (analogous to language, as a symbol system in the definition by \citeauthor{harnad-1990-symbol}) to another source $\mathcal{Y}$ (i.e., the ground, analogous to the real world), where we require:
\begin{enumerate}
    \item $\mathcal{X}$ is the primary source of data, and $\mathcal{Y}$ provides information that can be helpful on the task associated with $\mathcal{X}$.
          At the inference phase, $\mathcal{X}$ is the required input data to the trained model, and whether to involve $\mathcal{Y}$ as an additional data source during inference can be optional.
    \item $\mathcal{X}$ and $\mathcal{Y}$ have shared information (so that we can find meaningful correspondence between them).
\end{enumerate}
In the following discussion, we will call $\mathcal{X}$ the \textit{primary data source} and $\mathcal{Y}$ the \textit{ground}.
We will refer to the specific data used to represent $\mathcal{Y}$ as the \textit{grounding signals}.

We assume that the data from $\mathcal{X}$ and $\mathcal{Y}$ are sampled in a paired manner from the underlying joint distribution $p(x, y)$.
In information theoretic terms, Condition 2 above can be expressed as
\begin{align*}
    I(X; Y) = \mathbb{E}_{x, y\sim p(x, y)} \log \left(\frac{p(x, y)}{p(x)p(y)}\right) > 0,
\end{align*}
where $I(X; Y)$ denotes the mutual information between $X$ and $Y$, which can be derived from the joint distribution $p(x, y)$.

In real-world applications, different data sources usually carry non-identical information and noise; therefore, the conditional entropy $H(Y \mid X) = \mathbb{E}_{x, y\sim p(x, y)} \log p(x,y) - \log p(x)$ is almost always positive; that is, the ground $\mathcal{Y}$ almost always contains additional information compared to what can be inferred from the primary data source $\mathcal{X}$.
However, we do not require the conditional entropy $H(Y \mid X)$ to be strictly positive in this dissertation.
Consider the following synthetic data, where the primary data source gives written words \textit{triangle} and \textit{square}, and the ground gives only one corresponding shape of the same size for each word in visual signals.
Such a setting arguably does not tell us much about the rules for recognizing triangles and squares---since there is only one visual shape presented for one word, there is no way to enable generalizability of the recognition without any built-in inductive bias---and the conditional entropy $H(Y \mid X)$ is zero, but having $\mathcal{Y}$ may still be better than nothing for recognizing the visual shape from the written words.
In such cases, inductive biases associated with models, such as the localization bias introduced by the convolutional neural networks \citep{lecun-etal-1998-gradient}, can be combined with the ground to help the specific task associated with the primary data source $\mathcal{X}$.

The core difference between the definition of grounding in this dissertation and most existing literature is whether to emphasize the roles of forms and meanings.
In most existing literature, the term grounding is either explicitly or implicitly connected with linking \textit{symbolic forms} to their \textit{meanings}, where the symbolic forms considered are usually language.
While our definition is compatible with this view, we do not require the ground to be meanings---in this dissertation, any additional data source that can be used for a specific task with primary data source $\mathcal{X}$ can be considered as the ground, regardless of whether it is related to meanings or not.
Similarly, the form of the source data $\mathcal{X}$ may be extended beyond symbol systems---as an example, performing image segmentation with supervision solely from parallel text \interalia{xu2022groupvit} can be considered as a grounding process, where the primary data source is visual data and the textual data serve as the ground.

The term grounding also differs from a few related machine learning terms.
Compared to multi-view machine learning, such as canonical correlation analysis \citep[CCA;][]{hotelling-1936-relations}, the grounding process distinguishes the primary data source and the ground rather than treating them as two equally important views of the same data.
The transfer learning \citep{bozinovski-fulgosi-1976-influence} schema can also be considered as a special case of grounding, where the ground is the source domain and the primary data source is the target domain---work presented in \cref{chapter:subdp} can also be considered as a transfer learning technique.
However, transfer learning uses the same model architecture for the source and target domains, which is not required by grounding.

\cref{tab:bg-grounding-examples} analyzes the primary data source and the ground in the grounding settings considered in the above scenarios and individual chapters of this dissertation.
Since this dissertation focuses on learning language structures, the primary data source is natural language in different forms, including text and speech.
In contrast, we consider various data sources as the ground, including vision, program execution results, and other languages.

\begin{table}[t]
    \centering
    \begin{tabular}{p{0.13\textwidth}p{0.3\textwidth}p{0.49\textwidth}}
        \toprule
                                             & \bf Data Source $\mathcal{X}$ & \bf Ground $\mathcal{Y}$                    \\
        \midrule
        \it ~\citeauthor{harnad-1990-symbol} & Symbol system                 & Real-world meanings                         \\
        \it Scenario 1                       & Natural language (text)       & Sensory data                                \\
        \it Scenario 2                       & Natural language (text)       & Knowledge base                              \\
        \it Scenario 3                       & Natural language              & Shared common ground                        \\
        \midrule
        \it \cref{chapter:vgnsl}             & Natural language (text)       & Vision                                      \\
        \it \cref{chapter:avnsl}             & Natural language (speech)     & Vision                                      \\
        \it \cref{chapter:g2l2}              & Natural language (text)       & Vision and/or program execution results     \\
        \it \cref{chapter:mbrexec}           & Natural language (text)       & Program execution results                   \\
        \it \cref{chapter:mlpalign}          & Natural language (text)       & Natural language (text in another language) \\
        \it \cref{chapter:subdp}             & Natural language (text)       & Natural language (text in another language) \\
        \bottomrule
    \end{tabular}
    \caption[The primary data source and ground in grounding settings in various scenarios.]{The primary data source and ground in various scenarios. The top section discusses the scenarios mentioned above in this section (\cref{sec:bg-what-is-grounding}), and the bottom section discusses the grounding settings considered in individual chapters of this dissertation.}
    \label{tab:bg-grounding-examples}
\end{table}

%% file: src/30-vgnsl.tex
\chapter{Syntax Acquisition from Visually Grounded Text}
\label{chapter:vgnsl}
\textit{Content in this chapter has been published as a conference paper at ACL 2019 \citep{shi2019visually}. Jiayuan Mao has contributed significantly to this work.
}

\input{figures/301-vgnsl-intro.tex}
Consider images paired with descriptive texts (i.e., captions) in English (\cref{fig:vgnsl-intro}).
Given sufficiently many such pairs and no prior knowledge of English, one can infer the correspondence between certain words and visual attributes (e.g., recognizing that ``\textit{a cat}'' refers to the objects in the blue boxes).
Additionally, one can assume that visually concrete spans of words should be processed as a whole and thus form constituents in the sentence.
Such a process can happen for noun phrases, verb phrases, and prepositional phrases.
This intuition motivates using image-text pairs to facilitate automated language learning, including syntax and semantics.
Specifically, in this chapter, we focus on learning syntactic structures, where we propose the Visually Grounded Neural Syntax Learner (\vgnsl; \cref{fig:vgnsl-model}).

\vgnsl acquires syntax in the form of phrase structures by looking at images and reading captions.
The \vgnsl model consists of two modules: a textual module for inferring structures and representations for captions, and a visual-semantic module for matching constituents with images.
At a high level, \vgnsl builds latent constituency trees of word sequences and recursively composes representations for constituents.
Next, it matches the visual and textual representations.
The training procedure is built on the hypothesis that a better syntactic structure contributes to a better representation of constituents, leading to better alignment between vision and language.
We use no human-labeled constituency trees or other syntactic labeling (such as part-of-speech tags).
Instead, we define a \textit{concreteness} score of constituents based on their matching with images and use it to guide the parsing of sentences.
At the test time, no images paired with the text are needed.

In our experiments, we compare \vgnsl with prior approaches to unsupervised constituency parsing, most of which do not use visual grounding.
Our main findings are listed as follows:
\begin{enumerate}
    \item \vgnsl improves over the best previous text-only approaches to unsupervised constituency parsing in terms of $F_1$ scores with gold parse trees.
    \item While many existing approaches are quite unstable to the choice of random initialization, \vgnsl exhibits consistent parsing results across multiple training runs.
    \item Through analysis of the performance of different models on different types of constituents, we find that \vgnsl shows substantial improvement in noun phrases and prepositional phrases, which are common in image captions.
    \item \vgnsl is more data-efficient than prior text-only grammar induction models and achieves comparable performance using only 20\% of the training captions.
    \item The \textit{concreteness} score, which emerges during the matching between constituents and images, correlates well with a similar measure defined by linguists.
\end{enumerate}
In addition, \vgnsl can be easily extended to multiple languages, which we evaluate on the Multi30K dataset \citep{elliott2016multi30k,elliott2017findings} consisting of German and French image captions.

\section{Related Work}
\label{sec:vgnsl-related-work}
\paragraph{Phrase-structure grammar induction from text.}
Recent work has proposed several approaches for inducing latent phrase structures \interalia{choi2018learning,yogatama2017learning,maillard-clark-2018-latent} from the distant supervision of downstream tasks.
However, many of the methods are not able to produce linguistically sound structures, or even consistent ones with fixed data and hyperparameters but different random initialization \citep{williams2018latent}.

A related line of research is to induce latent syntactic structure via language modeling.
This approach has achieved remarkable performance on unsupervised constituency parsing \citep{shen2018neural, shen2019ordered}, especially in identifying the boundaries of higher-level (i.e., larger) constituents.
To our knowledge, the Parsing-Reading-Predict Network \citep[PRPN;][]{shen2018neural} and the Ordered Neuron LSTM
\citep[ON-LSTM;][]{shen2019ordered} currently produce the best fully unsupervised constituency parsing results.
One issue with PRPN, however, is that it tends to produce meaningless parses for lower-level (smaller) constituents \citep{htut2018grammar}.

Over the last two decades, there has been extensive study targeting unsupervised constituency parsing \citep{klein2002generative,klein2004corpus,klein2005natural,bod2006all,ponvert2011simple} and dependency parsing \citep{klein2004corpus,smith2006annealing,spitkovsky2010from,han2017dependency}. However, all of these approaches are based on linguistic annotations. Specifically, they operate on the part-of-speech tags of words instead of word tokens.
One exception is \citet{spitkovsky2011unsupervised}, which produces dependency parse trees based on automatically induced pseudo tags.

\paragraph{Grounded language acquisition.}
Grounded language acquisition has been studied for image-caption data \citep{christie2016resolving}, video-caption data \citep{siddharth2014seeing,yu2015compositional}, and visual reasoning \citep{mao2019neurosymbolic}. However, existing approaches rely on human labels or rules for classifying visual attributes or actions. Instead, our model induces syntax structures with no human-defined labels or rules.

Meanwhile, encoding representations into a joint visual-semantic embedding space \citep{ngiam2011multimodal} is a widely studied approach, and has achieved remarkable results on image-caption retrieval \citep{kiros2014unifying,faghri2017vse++,shi2018learning}, image caption generation \citep{kiros2014unifying,karpathy2015deep,ma2015multimodal}, and visual question answering \citep{malinowski2015ask}.
This work uses this idea to match visual and textual representations.

\paragraph{Concreteness estimation.}
\citet{turney2011literal} define concrete words as those referring to things, events, and properties that we can perceive directly with our senses.
Subsequent work has studied word-level concreteness estimation based on text \citep{turney2011literal,hill2013concreteness}, human judgments \citep{silberer2012grounded,hill2014concreteness,brysbaert2014concreteness}, and multi-modal data \citep{hill2014learning,hill2014multi,kiela2014improving,young2014image,hessel2018quantifying,silberer2017visually}.
As with \citet{hessel2018quantifying} and \citet{kiela2014improving}, our model uses multi-modal data to estimate concreteness.
Compared with them, we define concreteness for spans instead of words and use it to induce linguistic structures.

\section{The Visually Grounded Neural Syntax Learner}
\label{sec:vgnsl-model}
\input{figures/302-vgnsl-model.tex}
Given a set of paired images and captions, our goal is to learn representations and structures for words and constituents.
Toward this goal, we propose the Visually Grounded Neural Syntax Learner (\vgnsl), an approach for the grounded acquisition of syntax of natural language.
\vgnsl is motivated by the idea of semantic bootstrapping \citep{pinker1984language}, which suggests that children acquire syntax by first understanding the meaning of words and phrases and linking them with the syntax of words.

At a high level (\cref{fig:vgnsl-model}), \vgnsl consists of two modules.
First, given an input caption (i.e., a sentence or a phrase), as a sequence of tokens, \vgnsl builds a constituency parse tree, and recursively composes representations for every constituent.
Next, it matches textual constituent representations with visual inputs.
Both modules are jointly optimized with natural supervision: the model acquires constituency structures, composes textual representations, and links them with visual scenes, by looking at images and reading paired captions.

\subsection{Textual Representations and Structures}
\vgnsl starts by composing a binary constituency structure of the text, using an easy-first bottom-up parser.
The composition of the tree from a caption of length $n$ consists of $n - 1$ steps.
Let $\mathbf{X}^{(t)} = \left(\mathbf{x}^{(t)}_1, \mathbf{x}^{(t)}_2, \cdots, \mathbf{x}^{(t)}_k\right)$ denote the textual representations of a sequence of constituents after step $t$, where $k = n - t$.
For simplicity, we use $\mathbf{X}^{(0)}$ to denote the \textit{word embeddings} for all tokens (the initial representations).

At step $t$, a score function $\textit{score}(\cdot; \Theta)$, parameterized by $\Theta$, is evaluated on all pairs of consecutive constituents,
resulting in a vector $\textit{\textbf{score}}(\mathbf{X}^{(t-1)}; \Theta)$ of length $n - t - 1$:
\begin{align*}
\textit{score}&\left(\mathbf{X}^{(t-1)};\Theta\right)_j
\triangleq \textit{score}\left(\left[\mathbf{x}^{(t-1)}_j,\mathbf{x}^{(t-1)}_{j + 1}\right];\Theta\right).
\end{align*}
We implement $\textit{score}(\cdot; \Theta)$ as a two-layer ReLU-activated feed-forward network.

A pair of constituents $\left(\mathbf{x}^{(t-1)}_{j^*}, \mathbf{x}^{(t-1)}_{j^* + 1}\right)$ is sampled from all pairs of consecutive constituents, following the distribution produced by a {\tt softmax} operator over the scores:\footnote{~At test time, we take the {\tt argmax}.}
\begin{align*}
    p_\Theta\left(j^*\right) &
    = \frac{\exp\left(\textit{score}\left(\mathbf{X}^{(t-1)}; \Theta\right)_{j^*}\right)}{
    \sum_{j} \exp\left(\textit{score}\left(\mathbf{X}^{(t-1)}; \Theta\right)_j\right)
    } .
\end{align*}
The selected pair is combined to form a single new constituent, whereas the rest of the constituents are directly copied into the next step.
Thus, after step $t$, the number of constituents is decreased by 1.
The textual representation for the new constituent is defined as the L2-normalized sum of the component constituents:
\begin{align*}
    \textit{combine}\left(\mathbf{x}^{(t-1)}_{j^*}, \mathbf{x}^{(t-1)}_{j^*+1}\right) \triangleq \frac{\mathbf{x}^{(t-1)}_{j^*} + \mathbf{x}^{(t-1)}_{j^*+1}}{\left\|\mathbf{x}^{(t-1)}_{j^*} + \mathbf{x}^{(t-1)}_{j^*+1}\right\|_2}.
\end{align*}
We find that using a more complex encoder for constituents, such as GRUs \citep{cho2014learning}, will cause the representations to be highly biased towards a few salient words in the sentence \citep[e.g., the encoder encodes only the word ``cat'' while ignoring the rest part of the caption;][]{shi2018learning,wu2019unified}.
This significantly degrades the performance of linguistic structure induction.

We repeat the above score-sample-combine process for $n - 1$ steps until all words in the input text have been combined into a single constituent (\cref{figure:vgnsl-parsing-module}), which denotes the ending of the inference process of the constituency parse tree.
Since we combine two consecutive constituents at each time step, the derived tree $\mathbf{t}$ contains $2n - 1$ constituents (including $n$ individual words as terminals and $n-1$ nonterminals).
\input{figures/303-vgnsl-parser.tex}

\subsection{Visual-Semantic Embeddings}
We follow an approach similar to that of \citet{kiros2014unifying} to define the visual-semantic embedding (VSE) space for paired images and text constituents.
Let $\mathbf{v}^{(i)}$ denote the vector representation of an image $i$, and $\mathbf{c}^{(i)}_{j}$ denote the vector representation of the $j^\textit{th}$ constituent of its corresponding text caption---during the matching with images, we ignore the tree structure and index the constituents as a list.
A function $m_\Phi(\cdot, \cdot)$ is defined as the matching score between images and texts:
\begin{align*}
    m_\Phi\left(\mathbf{v}, \mathbf{c}\right) &\triangleq \cos(\Phi \mathbf{v}, \mathbf{c}),
\end{align*}
where the parameter vector $\Phi$ aligns the visual and textual representations into a joint space.

\subsection{Training}
We optimize the visual-semantic representations (affected by parameters $\Phi$ and the word embeddings) and constituency structures (affected by parameters $\Theta$) iteratively.
In each iteration, given constituency parsing results of the caption, $\Phi$ is optimized for matching the visual and the textual representations.
Next, given the visual grounding of constituents, $\Theta$ is optimized for producing constituents that can be better matched with images.
Specifically, we optimize textual representations and the visual-semantic embedding space using a hinge-based triplet ranking loss:
\begin{align*}
    \mathcal{L}(\mathcal{C}, \Phi; \mathcal{V}) =  & \sum_{\begin{subarray}{c}
    i, k \neq i,
    j, \ell
\end{subarray}} \left[m_\Phi(\mathbf{c}^{(k)}_{\ell}, \mathbf{v}^{(i)}) - m_\Phi(\mathbf{c}^{(i)}_{j}, \mathbf{v}^{(i)}) + \delta\right]_+ \\
    & + \sum_{\begin{subarray}{c}
    i, k \neq i,
    j
\end{subarray}}\left[m_\Phi(\mathbf{c}^{(i)}_{j}, \mathbf{v}^{(k)}) - m_\Phi(\mathbf{c}^{(i)}_{j}, \mathbf{v}^{(i)}) + \delta\right]_+,
\end{align*}
where $i$ and $k$ index over all image-caption pairs in the dataset, while $j$ and $\ell$ enumerate over all constituents of a specific caption ($c^{(i)}$ and $c^{(k)}$, respectively), $\mathcal{V} = \left\{\mathbf{v}^{(i)}\right\}$ is the set of corresponding image representations, $\mathcal{C} = \left\{\mathbf{c}^{(i)}_{j}\right\}$ is the set of textual representations of all constituents, $\delta$ is a hyperparameter denoting a constant margin, and $[\cdot]_+$ denotes $\max(0, \cdot)$.
The loss $\mathcal{L}$ extends the loss for image-caption retrieval of \citet{kiros2014unifying}, by introducing the alignments between entire images and sub-sentence constituents.
For simplicity, we use frozen image representation from a pre-trained image encoder and only optimize the textual representations $\mathcal{C}$ and the linear transformation $\Phi$ that projects the image representations into the joint space.

We also optimize textual structures for a better alignment between the derived constituents and the images.
Intuitively, we would like adjectives to be associated (combined) with the corresponding nouns, and verbs and prepositions to be associated (combined) with the corresponding subjects and objects, respectively.
Specifically, we use REINFORCE \citep{williams1992simple} as the gradient estimator for $\Theta$.
Consider the parsing process of a specific caption $c^{(i)}$, and denote the corresponding image embedding $\mathbf{v}^{(i)}$.
For a constituent $\mathbf{c}^{(i)}_j$ of $c^{(i)}$, we define its (visual) concreteness $\textit{concrete}\left(\mathbf{c}^{(i)}_j; \mathcal{V},\mathcal{C}\right)$ as:
\begin{align}
    \textit{concrete}\left(\mathbf{c}^{(i)}_j; \mathcal{V},\mathcal{C}\right) \nonumber = &\sum_{k\neq i, p} \left[m(\mathbf{c}^{(i)}_j, \mathbf{v}^{(i)}) - m(\mathbf{c}^{(k)}_{p}, \mathbf{v}^{(i)}) - \delta'\right]_+ \nonumber \\
    & +\sum_{k\neq i} \left[m(\mathbf{c}^{(i)}_j, \mathbf{v}^{(i)}) - m(\mathbf{c}^{(i)}_j, \mathbf{v}^{(k)}) - \delta'\right]_+,
    \label{eq:vgnsl-concreteness}
\end{align}
where $\delta'$ is a fixed margin. At step $t$, we define the reward function for a combination of a pair of constituents $\left(\mathbf{x}^{(t-1)}_q, \mathbf{x}^{(t-1)}_{q + 1}\right)$ as:
\begin{align}
    r\left(\mathbf{x}^{(t-1)}_q, \mathbf{x}^{(t-1)}_{q + 1}\right) = \textit{concrete}\left(\mathbf{z}, \mathbf{v}^{(i)}\right),
\label{eq:vgnsl-reward}
\end{align}
where $\mathbf{z} \triangleq \textit{combine}\left(\mathbf{x}^{(t-1)}_q, \mathbf{x}^{(t-1)}_{q + 1}\right)$. In plain words, at each step, we encourage the model to compose a constituent that maximizes the alignment between the new constituent and the corresponding image. During training, we sample constituency parse trees of captions and reinforce each composition step using \cref{eq:vgnsl-reward}.
Concretely, we update $\Theta$ using the following gradient estimator:
\begin{align*}
    \textcolor{black}{\Theta} \leftarrow \textcolor{black}{\Theta} + \eta\cdot\nabla_{\textcolor{black}{\Theta}} \sum_{i, j} p_\Theta(c_j^{(i)}) {\textit{concrete}\left(\mathbf{c}_j^{(i)}; \mathcal{V}, \mathcal{C} \right)},
\end{align*}
where $\Theta$ denotes the parameters of the score function in the parser that account for the sampled constituency parse trees, $\eta$ is the learning rate, and $p_\Theta(c_j^{(i)})$ is the probability of selecting the constituent $\mathbf{c}_j^{(i)}$ at the parsing step.
At the inference stage, since we do not need to estimate the concreteness scores, no paired images of text are needed.

\subsection{The Abstract-Initial Inductive Bias}
\label{sec:vgnsl-abstract-initial-bias}
English and many other Indo-European languages are usually head-initial \citep{baker2001atoms}.
For example, in verb phrases or prepositional phrases, the verb (or the preposition) precedes the complements (e.g., the object of the verb).
Consider the simple noun-phrase caption \textit{a white cat on the lawn}.
While the association of the adjective (\textit{white}) can be induced from the visual grounding of phrases, whether the preposition (\textit{on}) should be associated with \textit{a white cat} or \textit{the lawn} is more challenging to induce.
Given an empirical observation that prepositions are less visually concrete than nouns, we impose the abstract-initial inductive bias to guide the learner to correctly associate prepositions with their complements, determiners with corresponding noun phrases, and complementizers with the corresponding relative clauses.
Specifically, we discourage abstract constituents (i.e., constituents that cannot be grounded in the image) from being combined with a preceding constituent, by modifying the original reward definition (\cref{eq:vgnsl-reward}) as:
\begin{equation}
    \begin{aligned}
    r'\left(\mathbf{x}^{(t-1)}_j , \mathbf{x}^{(t-1)}_{j + 1}\right) = \frac{r\left(\mathbf{x}^{(t-1)}_j, \mathbf{x}^{(t-1)}_{j + 1}\right)}{\lambda \cdot \textit{abstract}\left(\mathbf{x}^{(t-1)}_{j + 1};\mathcal{V},\mathcal{C}\right) + 1} \ ,
\end{aligned}
\label{eq:vgnsl-inductive-bias}
\end{equation}
where $\lambda$ is a scalar hyperparameter, $\mathbf{v}^{(i)}$ is the image embedding corresponding to the caption being parsed, and $abstract$ denotes the {\it abstractness} of the span, defined analogously to concreteness (\cref{eq:vgnsl-concreteness}):
\begin{align*}
    \textit{abstract}\left(\mathbf{c}^{(i)}_j; \mathcal{V},\mathcal{C}\right) = &
    \sum_{k\neq i, p} \left[m(\mathbf{c}^{(k)}_{p}, \mathbf{v}^{(i)}) - m(\mathbf{c}^{(i)}_j, \mathbf{v}^{(i)}) + \delta'\right]_+ \\
    & +\sum_{k\neq i} \left[m(\mathbf{c}^{(i)}_j, \mathbf{v}^{(k)}) - m(\mathbf{c}^{(i)}_j, \mathbf{v}^{(i)}) + \delta'\right]_+,
\end{align*}

The intuition here is that the initial heads for prepositional phrases (e.g., {\it on}) and relative clauses (e.g., {\it which, where}) are usually abstract words, especially in the domain of image captions.
During training, we encourage the model to associate these abstract words with the succeeding constituents instead of the preceding ones.
It is worth noting that such an inductive bias is language-specific, and cannot be applied to head-final languages such as Japanese \citep{baker2001atoms}.
We leave the design of head-directionality inductive biases for other languages for future work.

\section{Experiments}
\label{sec:vgnsl-experiments}
We evaluate \vgnsl for unsupervised parsing in a few ways: $F_1$ score with gold trees, self-agreement across different choices of random initialization, performance on different types of constituents, and data efficiency. In addition, we find that the \textit{concreteness} score acquired by \vgnsl is consistent with a similar measure defined by linguists.
We focus on English for the main experiments but also extend to German and French.

\subsection{Datasets and Metrics}
We use the standard split of the MSCOCO dataset \citep{lin2014microsoft}, following \citet{karpathy2015deep}.
It contains 82,783 images for training, 1,000 for development, and another 1,000 for testing.
Each image is associated with five captions.

For the evaluation of constituency parsing, the Penn Treebank \citep[PTB;][]{marcus-etal-1993-building} is a widely used, manually annotated dataset.
However, PTB consists of sentences from abstract domains, e.g., the \textit{Wall Street Journal} (WSJ), which are not visually grounded and whose linguistic structures can hardly be induced by \vgnsl.
Here, we evaluate models on the MSCOCO test set, which is well-matched to the training domain; we leave the extension of our work to more abstract domains for future work.
We apply Benepar \citep{kitaev-klein-2018-constituency},\footnote{~\url{https://pypi.org/project/benepar}} an off-the-shelf constituency parser with state-of-the-art performance (95.52 $F_1$ score) on the WSJ test set,\footnote{~We also manually label the constituency parse trees for 50 captions randomly sampled from the MSCOCO test split, where Benepar has an $F_1$ score of 95.65 with the manual labels.
Details can be found at \url{https://home.ttic.edu/~freda/thesis_release/benepar_coco}.} to parse the captions in the MSCOCO test set as gold constituency parse trees.
We evaluate all the investigated models using the $F_1$ score compared to these gold parse trees.\footnote{~Following convention \citep{black-etal-1991-procedure,sekine-collins-1997-evalb}, we report the $F_1$ score across all constituents in the dataset, instead of the average of sentence-level $F_1$ scores. Without further note, all parsing $F_1$ scores reported in this dissertation are calculated in this way.}

\subsection{Baselines}
\label{sec:baseline}
We compare \vgnsl with various baselines for unsupervised tree structure modeling of texts, where we categorize the baselines by their training objective or supervision.

\paragraph{Trivial tree structures.}
Similarly to recent work on latent tree structures \citep{williams2018latent,htut2018grammar,shi2018tree}, we include three types of \textit{trivial} baselines without linguistic information: random binary trees, left-branching binary trees, and right-branching binary trees.

\paragraph{Syntax acquisition by language modeling and statistics.} \citet{shen2018neural} propose the Parsing-Reading-Predict Network (PRPN), which predicts syntactic distances \citep{shen2018straight} between adjacent words, and composes a binary tree based on the syntactic distances to improve language modeling.
The learned distances can be mapped into a binary constituency parse tree, by recursively splitting the sentence between the two consecutive words with the largest syntactic distance.

Ordered neurons \citep[ON-LSTM;][]{shen2019ordered} is a recurrent unit based on the LSTM cell \citep{hochreiter1997long}  that explicitly regularizes different neurons in a cell to represent short-term or long-term information.
After being trained on the language modeling task, \citet{shen2019ordered} suggest that the gate values in ON-LSTM cells can be viewed as syntactic distances \citep{shen2018straight} between adjacent words to induce latent tree structures.
We train both PRPN and ON-LSTM on all captions in the MSCOCO training set and use a fixed version of the syntactic distance method (\cref{algo:vgnsl-fixed-syntactic-distance}) to compose constituency parse trees.\footnote{~As pointed out by \citet{dyer2019critical}, the original syntactic distance--based tree composition method \citep{shen2018neural} is biased towards right-branching trees. We fix this issue by treating the distances in all positions equally. }
\input{algorithms/301-syndist}

Motivated by the syntactic distance approaches \citep{shen2018neural,shen2019ordered}, we also introduce another baseline, PMI, which uses negative pointwise mutual information between adjacent words as the syntactic distance.
We compose constituency parse trees based on the distances in the same way as PRPN and ON-LSTM.

\paragraph{Syntax acquisition from downstream tasks.}
\citet{choi2018learning} propose to compose binary constituency parse trees directly from downstream tasks using the Gumbel softmax trick \citet{jang2016categorical}.
We integrate a Gumbel tree-based caption encoder into the visual semantic embedding approach \citep{kiros2014unifying}, where the model is trained on the downstream task of image-caption retrieval with a hinge-based triplet ranking loss.

\paragraph{Syntax acquisition from concreteness estimation.}
Since we apply concreteness information to train \vgnsl, it is worth comparing against unsupervised constituency parsing based on previous approaches for predicting word concreteness.
This set of baselines includes semi-supervised concreteness estimation \citep{turney2011literal}, crowdsourced labeling \citep{brysbaert2014concreteness}, and multimodal estimation \citep{hessel2018quantifying}.
Note that none of these approaches has been applied to unsupervised constituency parsing.

Based on the concreteness score of words, we introduce another baseline similar to \vgnsl (\cref{algo:vgnsl-concreteness-tree}).
Specifically, at each step, we combine two consecutive constituents with the largest average concreteness and use the average concreteness as the score for the composed constituent.
The algorithm generates binary constituency parse trees of captions.
For a fair comparison, we implement a variant of this algorithm that also adopts the abstract-initial inductive bias (\cref{sec:vgnsl-abstract-initial-bias}).
\input{algorithms/302-concreteness.tex}

\subsection{Implementation Details}
Across all experiments and all models (including baselines such as PRPN, ON-LSTM, and Gumbel), the embedding dimension for words and constituents is 512.
For \vgnsl, we use a pre-trained ResNet-152 \citep{he2016deep}, trained on ImageNet \citep{russakovsky2015imagenet}, to extract vector embeddings for images. Thus, $\Phi$ is a mapping from a 2048-D image embedding space to a 512-D visual-semantic embedding space.
As for the $\textit{score}$ function in constituency parsing, we use a hidden dimension of 128 and ReLU activation.
All \vgnsl models are trained for 30 epochs. We use an Adam optimizer \citep{kingma2015adam} with an initial learning rate $5\times 10^{-4}$ to train \vgnsl.
The learning rate is re-initialized to $5 \times 10^{-5}$ after 15 epochs.
We tune other hyperparameters of \vgnsl on the development set using the self-agreement $F_1$ score \citep{williams2018latent} over five runs with different choices of random initialization.

\input{tables/301-vgnsl-main-results.tex}
\subsection{Results: Unsupervised Constituency Parsing}
We evaluate the induced constituency parse trees via the overall $F_1$ score, as well as the recall of four types of constituents: noun phrases (NP), verb phrases (VP), prepositional phrases (PP), and adjective phrases (ADJP) (\cref{table:vgnsl-main-result}).
We also evaluate the robustness of models trained with fixed data and hyperparameters, but different random initialization, in two ways: via the standard deviation of performance across multiple runs, and via the self-agreement $F_1$ score \citep{williams2018latent}, which is the average $F_1$ taken over pairs of different runs.

Among all of the models that do not require extra labels, \vgnsl with the abstract-initial inductive bias (\vgnslai) achieves the best $F_1$ score.
PRPN \citep{shen2018neural} and a concreteness estimation-based baseline \citep{hessel2018quantifying} both produce competitive results.
It is worth noting that the PRPN baseline reaches this performance without any information from images.
However, the performance of PRPN is less stable than that of \vgnsl across random initialization.
In contrast to its state-of-the-art performance on the WSJ full set \citep{shen2019ordered}, we observe that ON-LSTM does not perform well on the MSCOCO caption dataset. However, it remains the best model for adjective phrases, which is consistent with the result reported by \citet{shen2019ordered}.

In addition to the best overall $F_1$ scores, \vgnslai achieves competitive scores across most phrase types (NP, VP, and PP).
Our models (\vgnsl and \vgnslai) perform the best on NP and PP, which are the most common visually grounded phrases in the MSCOCO dataset.
In addition, our models produce much higher self $F_1$ than the baselines \citep{shen2018neural,shen2019ordered,choi2018learning}, showing that they reliably produce reasonable constituency parse trees with different initialization.

We also test the effectiveness of using pre-trained word embeddings.
Specifically, for \vgnslai + FastText, we use the pre-trained FastText embeddings (300-D; \citealp{joulin2016fasttext}), concatenated with 212-D trainable embeddings, as the word embeddings.
Using pre-trained word embeddings further improves performance to an average $F_1$ of 54.4\% while keeping a comparable self $F_1$.

\subsection{Results: Data Efficiency}
\input{figures/304-vgnsl-data-efficiency.tex}
We compare the data efficiency for PRPN (the strongest baseline method), ON-LSTM, \vgnsl, and \vgnslai.
We train the models using 1\%, 2\%, 5\%, 10\%, 20\%, 50\% and 100\% of the MSCOCO training set, and report the overall $F_1$ and self $F_1$ scores on the test set (\cref{fig:vgnsl-data-efficiency}).

Compared to PRPN trained on the full training set, \vgnsl and \vgnslai reach comparable performance using only 20\% of the data (i.e., 8K images with 40K captions).
\vgnsl tends to quickly become more stable (in terms of the self $F_1$ score) as the amount of data increases, while PRPN and ON-LSTM remain less stable.

\subsection{Analysis: Consistency with Linguistic Concreteness}
\input{tables/302-vgnsl-concreteness.tex}
During training, \vgnsl acquires concreteness estimates for constituents via \cref{eq:vgnsl-concreteness}. Here, we evaluate the consistency between word-level concreteness estimates induced by \vgnsl and those produced by other methods \citep{turney2011literal,brysbaert2014concreteness,hessel2018quantifying}.
Specifically, we measure the correlation between the concreteness estimated by \vgnsl on the MSCOCO test set and existing linguistic concreteness definitions (\cref{tab:vgnsl-concreteness}).
For any word type $z$, we estimate its concreteness by taking the average of $\textit{concrete}(\mathbf{z}; \mathcal{V},\mathcal{C})$, across all word tokens of $z$ in the dataset---the calculation of $\textit{concrete}(\mathbf{z}; \mathcal{V},\mathcal{C})$ is based on the visual-semantic embeddings of constituents, and requires the corresponding image to the caption where the word token appears.
The high correlation between \vgnsl and the concreteness scores produced by \citet{turney2011literal} and \citet{brysbaert2014concreteness} supports the argument that the linguistic concept of concreteness can be acquired in an unsupervised way.
Our model also achieves a high correlation with \citet{hessel2018quantifying}, which also estimates word concreteness based on visual information.

\subsection{Analysis: Self Agreement as Model Selection Criterion}
\input{tables/303-vgnsl-self-agreement.tex}
We introduce a novel hyperparameter tuning and model selection method based on the self-agreement $F_1$ score.

Let $\mathcal{M}_\mathcal{H}^{(i, j)}$ denote the j$^\textit{th}$ checkpoint of the i-th model trained with hyperparameters $\mathcal{H}$, where $\mathcal{M}_\mathcal{H}^{(i_1, \cdot)}$ and  $\mathcal{M}_\mathcal{H}^{(i_2, \cdot)}$ differ in their random initialization. The hyperparameters $\mathcal{H}$ are tuned to maximize:
\begin{align*}
    \sum_{1\leq i < k \leq N} & \max_{|j_i - j_k| < \delta} F_1\left(\mathcal{M}_\mathcal{H}^{(i, j_i)}, \mathcal{M}_\mathcal{H}^{(k, j_k)}\right),
\end{align*}
where $F_1(\cdot, \cdot)$ denotes the $F_1$ score between the trees generated by two models, $N$ the number of different runs, and $\delta$ the margin to ensure only nearby checkpoints are compared.\footnote{~In all of our experiments, $N=5, \delta=2$.}

After finding the best hyperparameters $\mathcal{H}_0$, we train the model for another $N$ times with different random initialization, and select the best models by
\begin{align*}
    \argmax_{\{j_\ell \}_{\ell=1}^N} \sum_{1\leq i < k \leq N} F_1\left(\mathcal{M}_{\mathcal{H}_0}^{(i, j_i)}, \mathcal{M}_{\mathcal{H}_0}^{(k, j_k)}\right).
\end{align*}

We compare the performance of \vgnsl selected by the self $F_1$ score and that selected by recall at 1 in image-to-text retrieval \citep[R@1 in \cref{tab:vgnsl-self-agreement-F1};][]{kiros2014unifying}.
As a model selection criterion, self $F_1$ consistently outperforms R@1 (avg. $F_1$: 50.4 vs. 47.7 and 53.3 vs. 53.1 for \vgnsl and \vgnslai, respectively).
Meanwhile, it is worth noting that even if we select \vgnsl by R@1, it shows better stability compared with PRPN and ON-LSTM (\cref{table:vgnsl-main-result}), in terms of the score variance across different random initialization and self $F_1$.
Specifically, the variance of avg. $F_1$ is always less than 0.6 while the self $F_1$ is greater than 80.

Note that the PRPN and ON-LSTM models are not tuned using self $F_1$, since these models are usually trained for hundreds or thousands of epochs and, thus, it is computationally expensive to evaluate self $F_1$.
We leave the efficient tuning of these baselines by self $F_1$ for future work.

\subsection{Extension to Multiple Languages}
\input{tables/304-vgnsl-other-languages.tex}
We extend our experiments to the Multi30K dataset, which is built on the Flickr30K dataset \citep{young2014image} and consists of English, German \citep{elliott2016multi30k}, and French \citep{elliott2017findings} captions.
For Multi30K, there are 29,000 images in the training set, 1,014 in the development set, and 1,000 in the test set. Each image is associated with one caption in each covered language.

We compare our models to PRPN and ON-LSTM in terms of overall $F_1$ scores (\cref{table:vgnsl-other-languages}).
\vgnsl with the abstract-initial inductive bias consistently performs the best across the three languages, all of which are highly head-initial \citep{baker2001atoms}---this set of results supports our intuition to implement the abstract-initial inductive bias.
Note that the $F_1$ scores here are not comparable to those in \cref{table:vgnsl-main-result}, since Multi30K (English) has 13 times fewer captions than MSCOCO.

\section{Conclusion and Discussion}
We have proposed a simple but effective model, the Visually Grounded Neural Syntax Learner, for visually grounded language structure acquisition.
\vgnsl jointly learns parse trees and visually grounded textual representations.
In our experiments, we find that this approach to grounded language learning produces parsing models that are both accurate and stable.
In addition, the learning process of \vgnsl is much more data-efficient than prior text-only approaches.
Along the way, the model acquires estimates of word concreteness.

The results suggest multiple future research directions, which we discuss as follows:
\begin{enumerate}
\item
\vgnsl matches text embeddings directly with embeddings of entire images.
Its performance may be boosted by considering structured representations of both images (e.g., \citealp{lu2016visual,wu2019unified}) and texts.
Evidence from recent work, such as \citet{hong2021vlgrammar} and \citet{wan2022unsupervised}, suggests that the awareness of image structures can improve the performance of text syntactic analysis.
In a reversed direction, where vision is the primary modality to be processed, \citet{xu2022groupvit} have also shown that text, as a secondary modality that provides distant supervision, can improve the performance of image segmentation, which is a crucial step towards a hierarchical understanding of images.

On the other hand, while different modalities---to some extent---share the structures, they also possess their unique features and information.
We look forward to future work that integrates the structures of multiple modalities into a unified model while also considering the unique features of each modality.
\item
Thus far, we have used a shared vector representation for both syntax and semantics, but it may be useful to disentangle their representations.
In \cref{chapter:g2l2}, we discuss the induction of combinatory categorial grammar \citep[CCG;][] {steedman-2000-syntactic} from visually grounded text, which is a meaningful step towards this direction.
In addition, the disentanglement of syntax and semantics may enable more interpretable and controllable study relevant to syntactic \citep{brown1957linguistic} and semantic \citep{pinker1984language} bootstrapping.
\item
Our best parsing model is based on the abstract-initial inductive bias, which is designed based on insights drawn from the structure of head-initial languages.
However, automatically acquiring such effective inductive biases from data remains challenging \citep{kemp2006learning,gauthier2018word}, and we suggest another future direction to learn inductive biases from data with plausibly minimal human intervention.
\item It may be possible to extend our approach to other linguistic tasks such as dependency parsing \citep{christie2016resolving}, coreference resolution \citep{kottur2018visual}, and learning pragmatics beyond semantics \citep{andreas2016reasoning}.
Towards this line, we recommend readers to check out \citet{su-etal-2021-dependency}.
\end{enumerate}
There are also limitations to the idea of grounded language acquisition. In particular, the current approach has thus far been applied to understanding grounded texts in a single domain (static visual scenes for \vgnsl), which fundamentally lacks the dynamics of real-world interactions.
Its applicability may be extended by learning shared representations across multiple modalities \citep{castrejon2016learning} or integrating with pure text-domain models (e.g., integrating with probabilistic context-free grammars; \citealp{zhao-titov-2020-visually}).

%% file: figures/301-vgnsl-intro.tex
\begin{figure}[t]
    \centering
    \vspace{-40pt}
    \includegraphics[width=0.75\textwidth,page=2]{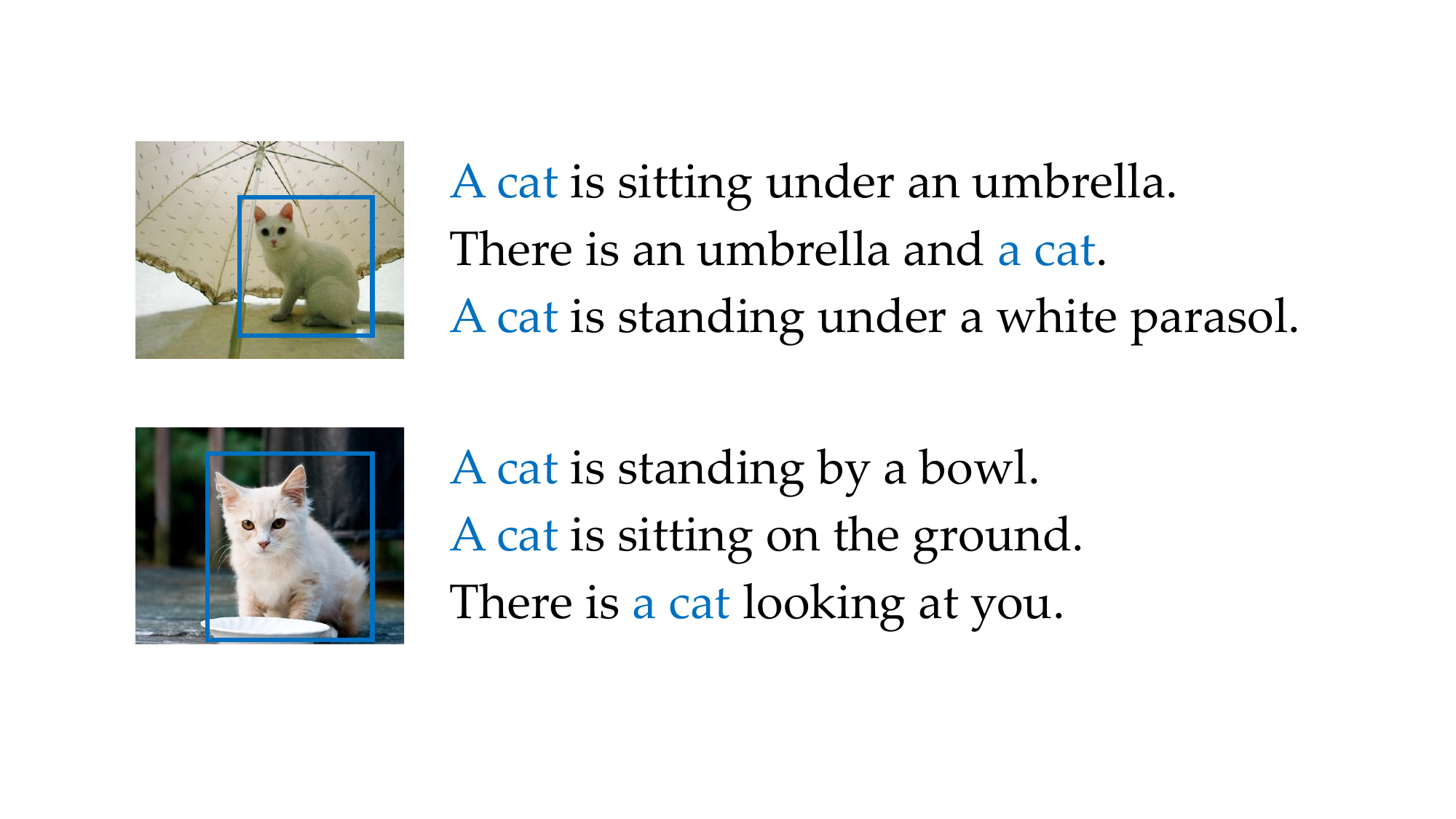}
    \vspace{-40pt}
    \caption[Illustration of visual correspondence of phrases.]{
        \label{fig:vgnsl-intro} Illustration of visual correspondence of phrases.
        We propose to use image-caption pairs to extract constituents from text based on the assumption that similar textual spans (in blue) should be matched to similar visual objects (in blue boxes), and these concrete spans form constituents.
        Best viewed in color.
    }
\end{figure}

%% file: figures/302-vgnsl-model.tex
\begin{figure}[t]
    \centering
    \includegraphics[width=\textwidth]{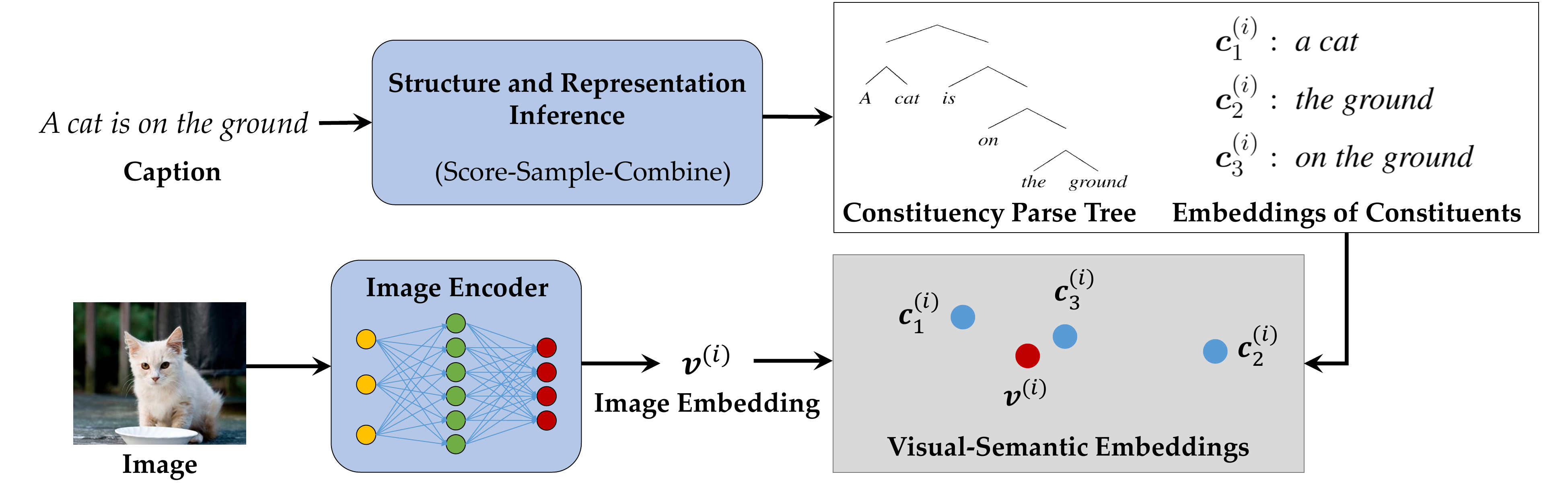}
    \caption[Illustration of the \vgnsl model.]{
        Illustration of the \vgnsl model. 
        \vgnsl consists of two modules: a textual module for inferring structures and representations for captions, and a visual-semantic module for matching constituents with images. 
        \vgnsl induces constituency parse trees of captions by looking at images and reading paired captions.
    }
    \label{fig:vgnsl-model}
\end{figure}

%% file: figures/303-vgnsl-parser.tex
\begin{figure}[t]
    \centering
    \includegraphics[width=0.65\textwidth,page=1]{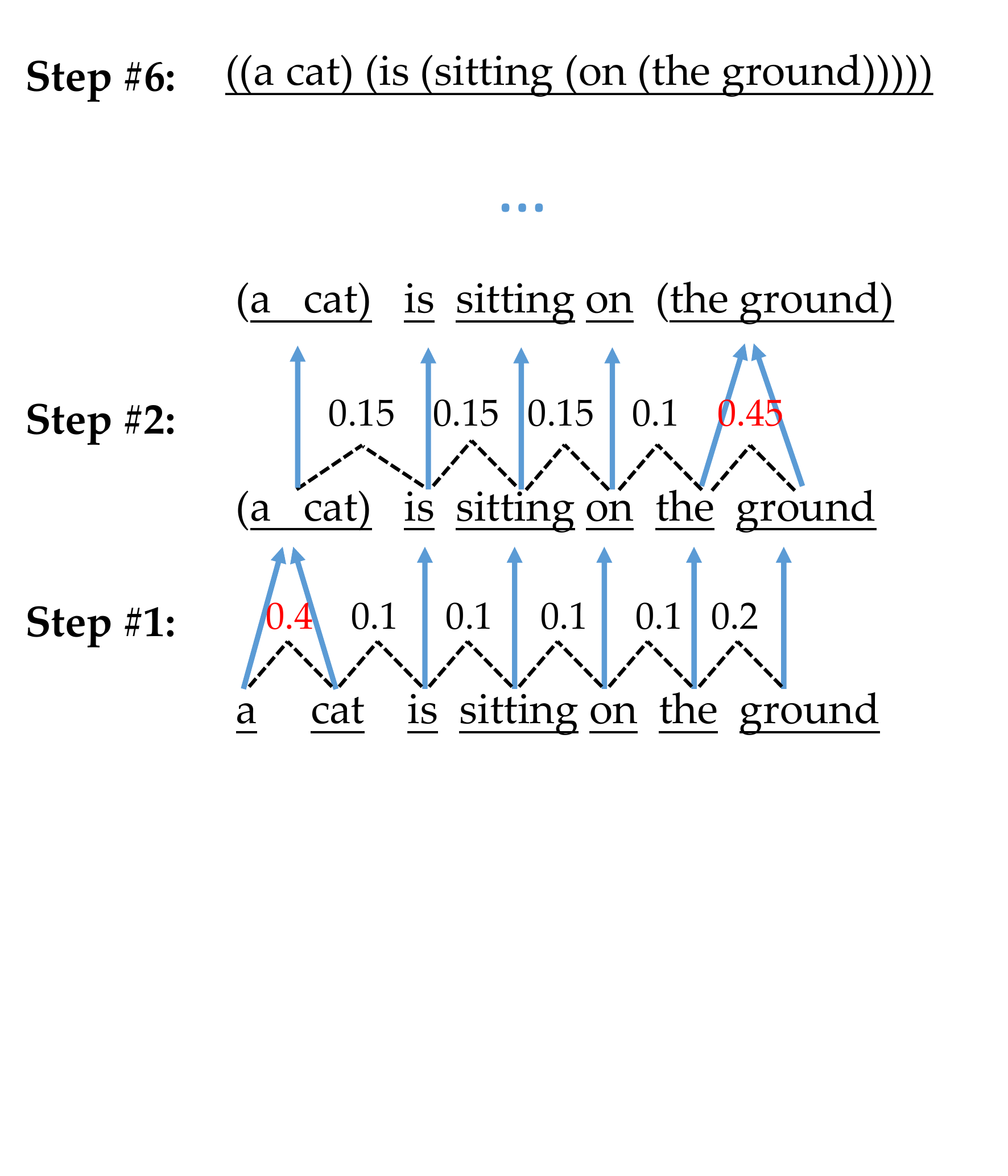}
    \vspace{-120pt}
    \caption[An illustration of how \vgnsl composes a constituency parse tree.]{
        An illustration of how \vgnsl composes a constituency parse tree.
        At each step, the score function \textit{score} is evaluated on all pairs of consecutive constituents (dashed lines).
        Next, a pair of constituents is sampled from all pairs, following a distribution computed by the \texttt{softmax} of all predicted scores.
        The selected two constituents are combined into a larger one, while the other constituents remain unchanged (solid lines).
        For the 7-word sentence shown in the figure, it requires 6 steps to compose the final parse tree.
    }
    \label{figure:vgnsl-parsing-module}
\end{figure}

%% file: algorithms/301-syndist.tex
\begin{algorithm}[t!]
    \SetAlgoLined
    \SetKwInOut{Input}{Output}
    \SetKwFunction{parse}{parse}
    \KwIn{text length $m$, list of syntactic distances $\boldsymbol{d} = (d_1, d_2, \ldots, d_{m-1})$}
    \KwOut{Boundaries of constituents $B = \{(L_i, R_i)\}_{i=1, \ldots, m-1}$}
    $B$ = \parse{$\boldsymbol{d}$, $1$, $m$} \\~\\
    \textbf{Function} \parse{$\boldsymbol{d}$, $\textit{left}$, $\textit{right}$} { \\
        \If{$\textit{left} = \textit{right}$}{
            \KwRet \FuncSty{$\emptyset$}
        }
        $p = \argmax_{j \in [\textit{left}, \textit{right-1}]} d_j$ \\
        $\textit{boundaries} = $ \FuncSty{Union}(\{($\textit{left}$, $\textit{right}$)\}, \parse($\boldsymbol{d}$, \textit{left}, $p$), \parse($\boldsymbol{d}$, $p+1$, \textit{right})) \\

        \KwRet \textit{boundaries}
    }
    \caption{\label{algo:vgnsl-fixed-syntactic-distance} Constituency parsing based on given syntactic distances.}
\end{algorithm}

%% file: algorithms/302-concreteness.tex
\begin{algorithm}[t!]
    \SetAlgoLined
    \SetKwInOut{Input}{Output}
    \KwIn{ list of normalized word concreteness scores $\boldsymbol{a} = (a_1, a_2, \ldots, a_m)$, hyperparameter $\tau$ controlling the level of abstract-initiality}
    \KwOut{Boundaries of constituents $B = \{(L_i, R_i)\}_{i=1, \ldots, m-1}$}
    \For{$j=1$ to $m$} {
        $\textit{left}_j = j$ \\
        $\textit{right}_j = j$
    }
     \While{$\textit{len}(\boldsymbol{a}) > 1$}{
      $p = \argmax_{j} \left( a_{j} + \tau a_{j+1} \right)$ \\
      add $(\textit{left}_p, \textit{right}_{p+1})$ to $B$ \\
      $\boldsymbol{a} = \boldsymbol{a}_{< p}$ + ($\frac{a_p + a_{p+1}}{2}$) + $\boldsymbol{a}_{>p+1}$ \\
      $\textbf{\textit{left}} = \textbf{\textit{left}}_{< p}$ + ($\textit{left}_p$) + $\textbf{\textit{left}}_{>p+1}$\\
      $\textbf{\textit{right}} = \textbf{\textit{right}}_{< p}$ + ($\textit{right}_{p+1}$) + $\textbf{\textit{right}}_{>p+1}$
     }
     \caption{\label{algo:vgnsl-concreteness-tree} Constituency parsing based on concreteness estimation.}
\end{algorithm}

%% file: tables/301-vgnsl-main-results.tex
\begin{table}[!t]
    \centering
    \small
    \setlength{\tabcolsep}{0.5pt}
    \begin{tabular}{lrlrlrlrlrlr}
        \toprule
        \textbf{Model} & \multicolumn{2}{c}{\textbf{NP}} & \multicolumn{2}{c}{\textbf{VP}} & \multicolumn{2}{c}{\textbf{PP}} & \multicolumn{2}{c}{\textbf{ADJP}} & \multicolumn{2}{c}{\textbf{Avg. $\textbf{F}_1$}} & \multicolumn{1}{c}{\textbf{Self $\textbf{F}_1$}} \\
        \midrule
        Random  & 47.3&$_{\pm 0.3}$ & 10.5&$_{\pm 0.4}$ & 17.3&$_{\pm 0.7}$ & 33.5&$_{\pm 0.8}$ & 27.1&$_{\pm 0.2}$ & 32.4 \\
        Left    & 51.4&& 1.8 && 0.2 && 16.0 && $23.3$ && N/A\\
        Right   & 32.2&& 23.4&& 18.7 && 14.4 && $22.9$ && N/A\\
        PMI     & 54.2&& 16.0 && 14.3 && 39.2 && $30.5$ && N/A\\
        PRPN \citep{shen2018neural}    & ~~~72.8&$_{\pm 9.7}$~~~ & ~~~33.0&$_{\pm 9.1}$ ~~~& ~~~61.6&$_{\pm 9.9}$ ~~~& ~~~35.4&$_{\pm 4.3}$ ~~~& ~~~52.5&$_{\pm 2.6}$ ~~~& 60.3 \\
        ON-LSTM \citep{shen2019ordered}     & 74.4&$_{\pm 7.1}$ & 11.8&$_{\pm 5.6}$ & 41.3&$_{\pm 16.4}$ & \textbf{44.0}&$_{\pm 14.0}$ &  45.5&$_{\pm 3.3}$ & 69.3 \\
        Gumbel \citep{choi2018learning}$^\dagger$ & 50.4&$_{\pm 0.3}$ & 8.7&$_{\pm 0.3}$ & 15.5&$_{\pm 0.0}$ & 34.8&$_{\pm 1.6}$ & 27.9&$_{\pm 0.2}$ & 40.1 \\
        \midrule
        \vgnsl (ours)$^\dagger$ & \textbf{79.6}&$_{\pm 0.4}$ & 26.2&$_{\pm 0.4}$ & 42.0&$_{\pm 0.6}$ & 22.0&$_{\pm 0.4}$ & 50.4&$_{\pm 0.3}$ & 87.1\\
        \vgnslai (ours)$^\dagger$ & 74.6&$_{\pm 0.5}$ & 32.5&$_{\pm 1.5}$ & \textbf{66.5}&$_{\pm 1.2}$ & 21.7&$_{\pm 1.1}$ & 53.3&$_{\pm 0.2}$ & \textbf{90.2} \\
        \vgnslai + FastText (ours)*$^\dagger$ & 78.8&$_{\pm 0.5}$ & 24.4&$_{\pm 0.9}$ & 65.6&$_{\pm 1.1}$ & 22.0&$_{\pm 0.7}$ & \textbf{54.4}&$_{\pm 0.4}$ & 89.8 \\
        \toprule
        \multicolumn{12}{c}{\normalsize \textit{Concreteness estimation--based models}} \\
        \midrule
        \citet{turney2011literal}* &  65.5 && 30.8 && 35.3 && 30.4 && 42.5 && N/A\\
        \citet{turney2011literal}+HI*  & 74.5 && 26.2  && 47.6 && 25.6 && 48.9 && N/A \\
        \citet{brysbaert2014concreteness}* & 54.1 && 27.8 && 27.0 && 33.1 &&	34.1 && N/A \\
        \citet{brysbaert2014concreteness}+HI* & 73.4 && 23.9 && 50.0 && 26.1 && 47.9 && N/A \\
        \citet{hessel2018quantifying}$^\dagger$ & 50.9 && 21.7 &&	32.8 && 27.5 && 33.2  && N/A \\
        \citet{hessel2018quantifying}+HI$^\dagger$ &	72.5 && \textbf{34.4} &&	65.8 && 26.2 &&	52.9 && N/A \\
        \bottomrule
    \end{tabular}
    \caption[Recall of specific typed phrases and overall $\text{F}_1$ score on MSCOCO.]{
        Recall of specific typed phrases and overall $\text{F}_1$ score, evaluated on the MSCOCO test set, averaged over 5 runs with different random initializations.
        We also include self-agreement $\text{F}_1$ score \citep{williams2018latent} across the 5 runs.
        \stderr denotes standard deviation.
        * denotes models requiring extra labels and/or corpus, and $\dagger$ denotes models requiring a pre-trained visual feature extractor.
        We highlight the best number in each column.
        The Left, Right, and PMI baselines, as well as concreteness estimation--based models have no standard deviation or self $\text{F}_1$ (shown as N/A) as they are deterministic given the training and/or testing data.

    }
    \label{table:vgnsl-main-result}
\end{table}

%% file: figures/304-vgnsl-data-efficiency.tex
\begin{figure}[t]
    \centering
    \begin{subfigure}[t]{0.48\textwidth}
    \centering
    \includegraphics[width=\textwidth]{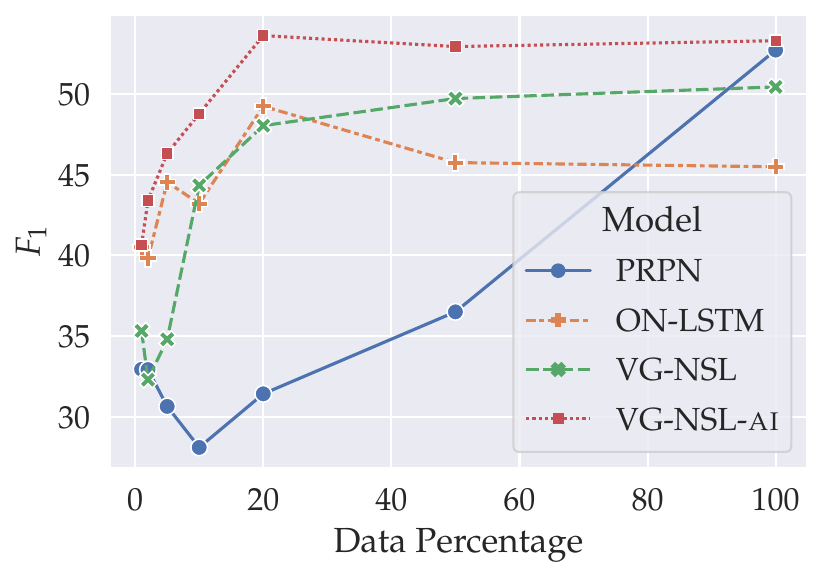}\vspace{-0.3em}
    \caption{The data percentage--$F_1$ curves.}
    \end{subfigure}
    \begin{subfigure}[t]{0.48\textwidth}
    \centering
    \includegraphics[width=\textwidth]{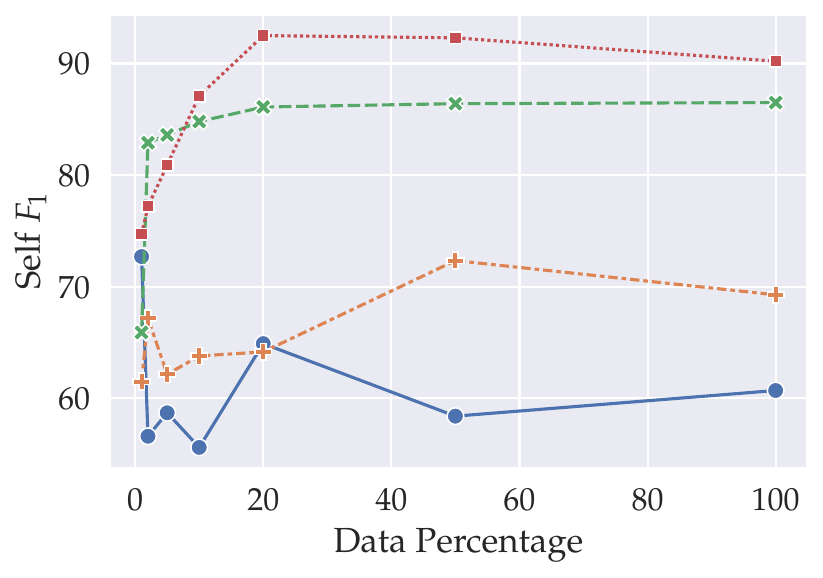}\vspace{-0.3em}
    \caption{The data percentage--self $F_1$ curves.}
    \end{subfigure}
    \caption[$F_1$ score and self $F_1$ score with respect to the amount of training data.]{$F_1$ score and self $F_1$ score with respect to the amount of training data. All numbers are averaged over 5 runs with different random initialization.
    }
    \label{fig:vgnsl-data-efficiency}
\end{figure}

%% file: tables/302-vgnsl-concreteness.tex
\begin{table}[t]
    \centering
    \begin{tabular}{lcc}
    \toprule
        \textbf{Model/Method} & \vgnsl & (+\textsc{ai}) \\
    \midrule
        \citet{turney2011literal} & 0.74 & 0.72 \\
        \citet{brysbaert2014concreteness} & 0.71 & 0.71 \\
        \citet{hessel2018quantifying} & 0.84 & 0.85 \\
    \bottomrule
    \end{tabular}
    \caption[Agreement between \vgnsl concreteness and existing concreteness estimation methods.]{
        Agreement between our concreteness estimates and existing models or labels, evaluated via the Pearson correlation coefficient computed over the most frequent 100 words in the MSCOCO test set, averaged over 5 runs with different random initialization.
    }
    \label{tab:vgnsl-concreteness}
\end{table}

%% file: tables/303-vgnsl-self-agreement.tex
\begin{table}[t]
    \centering
    \begin{tabular}{lccr}
    \toprule
    \bf Model & \bf Criterion & \bf Avg. $\textbf{\textit{F}}_1$ & \bf Self $\textbf{\textit{F}}_1$ \\
    \midrule
    \vgnsl & Self $F_1$ & \textbf{50.4} $_{\pm0.3}$ & \textbf{87.1} \\
    \vgnsl & R@1 & 47.7 $_{\pm0.6}$ & 83.4 \\
    \midrule
    \vgnslai & Self $F_1$ & \textbf{53.3} $_{\pm0.2}$ & \textbf{90.2} \\
    \vgnslai& R@1& 53.1 $_{\pm0.2}$ & 88.7 \\
    \bottomrule
    \end{tabular}
    \caption[$F_1$ and self $F_1$ scores of \vgnsl and \vgnslai with different model selection methods.]{
        Average $F_1$ scores and self $F_1$ scores of \vgnsl and \vgnslai with different model selection methods.
        R@1 denotes using recall at 1 \citep{kiros2014unifying} as the model selection criterion. All hyperparameters are tuned with the self-agreement $F_1$ score.
        The numbers are comparable to those in \cref{table:vgnsl-main-result}.
    }
    \label{tab:vgnsl-self-agreement-F1}
\end{table}

%% file: tables/304-vgnsl-other-languages.tex
\begin{table}[t]
    \centering
    \setlength{\tabcolsep}{1pt}
    \begin{tabular}{lrlrlrl}
         \toprule
         \textbf{Model} & \multicolumn{2}{c}{\textbf{EN}} & \multicolumn{2}{c}{\textbf{DE}} & \multicolumn{2}{c}{\textbf{FR}} \\
         \midrule
          PRPN & ~~30.8&$_{\pm 17.9}$~ & ~~31.5&$_{\pm 8.9}$~ & ~~27.5&$_{\pm 7.0}$~ \\
          ON-LSTM   & ~~\textbf{38.7}&$_{\pm 12.7}$  &  ~~34.9&$_{\pm 12.3}$ & ~~27.7&$_{\pm 5.6}$~ \\
          \vgnsl & 33.5&$_{\pm 0.2}$ & 36.3&$_{\pm 0.2}$ & 34.3&$_{\pm 0.6}$ \\
          \vgnslai & \textbf{38.7}&$_{\pm 0.2}$ & \textbf{38.3}&$_{\pm 0.2}$ & \textbf{38.1}&$_{\pm 0.6}$ \\
         \bottomrule
    \end{tabular}
    \caption[$F_1$ scores on the Multi30K.]{
        $F_1$ scores on the Multi30K test split \citep{young2014image,elliott2016multi30k,elliott2017findings}, averaged over 5 runs with different random initialization. $\pm$ denotes the standard deviation.
    }
    \label{table:vgnsl-other-languages}
\end{table}

%% file: src/40-avnsl.tex
\chapter{Syntax Acquisition from Visually Grounded Speech}
\label{chapter:avnsl}
\textit{Content in this chapter has been published as a workshop paper at ASRU 2023 \citep{lai2023audio}. Cheng-I Jeff Lai and Puyuan Peng have contributed significantly to this work. }

Given the results presented in \cref{chapter:vgnsl} on parsing visually grounded text, we now focus on the problem of learning to parse visually grounded speech without explicit syntactic supervision.

\input{figures/401-avnsl-teaser.tex}
In this chapter, we present a pipeline approach: the core idea is to first segment the speech waveform into sequences of word segments, and subsequently induce phrase structure using the inferred segment-level continuous representations.
We introduce the Audio-Visual Neural Syntax Learner (\avnsl; \cref{fig:avnsl-teaser}) that learns phrase structures by listening to audio and looking at images, without ever being exposed to text.
The speech utterances are represented by sequences of continuous speech segment representations derived from a pre-trained model that simultaneously discovers word-like units and learns segment representations \citep{peng2022word}.
Adapting the model structures of \vgnsl (\cref{chapter:vgnsl}), \avnsl (1) learns to map the representations of speech segments and images into a shared embedding space, resulting in higher similarity scores for segments and images that convey similar meanings, (2) estimates the visual \textit{concreteness} of speech segments using the learned embedding space, and (3) outputs speech segments with higher concreteness as the constituents.
By training on paired images and spoken captions, \avnsl can infer meaningful phrase structures comparable to those derived by naturally supervised text parsers, for both English and German.
Our findings extend prior work in unsupervised language acquisition from speech and grounded grammar induction and present one approach to bridge the gap between the two topics.
As a by-product, we also improve over the previous state of the art in unsupervised word segmentation.

\section{Related Work}
\label{chapter:avnsl-related}
\paragraph{Grounded grammar induction.}
Since the proposal of the visually grounded grammar induction task \citep{shi2019visually}, there has been subsequent research on the topic \interalia{zhao-titov-2020-visually,zhang2021video,wan2022unsupervised}.
To the best of our knowledge, existing work on grammar induction from distant supervision has been based almost exclusively on text input.
The most relevant work to ours is \citet{zhang2021video}, where speech features are treated as an auxiliary input for video-text grammar induction; that is, the model proposed by \citet{zhang2021video} still requires text data and an off-the-shelf automatic speech recognizer.
In contrast to existing approaches, \avnsl employs raw speech data and bypasses text to induce constituency parse trees, utilizing distant supervision from parallel audio-visual data.

\paragraph{Spoken word discovery.}
Following the pioneering work in spoken term discovery \citep{park2007unsupervised}, a line of work has been done to discover repetitive patterns or keywords from unannotated speech \interalia{jansen2011efficient, mcinnes2011unsupervised, zhang2013unsupervised}.
Other related work has considered tasks such as unsupervised word segmentation and spoken term discovery \interalia{lee2015unsupervised,kamper2017segmental,chorowski2021aligned, bhati2021segmental}, and the ZeroSpeech challenges \citep{dunbar2022self} have been a major driving force in the field.
A recent line of work \citep{harwath2017learning,harwath2018jointly,harwath2019towards,harwarth2020learning} has shown that word-like and phone-like units can be acquired from speech by analyzing audio-visual retrieval models.
\citet{peng2022word} shows that word discovery naturally emerges from a visually grounded, self-supervised speech model, by analyzing the model's self-attention heads.
In contrast, \avnsl attempts to induce phrase structure, in the form of constituency parsing on top of unsupervised word segments.

\begin{figure}[t]
    \begin{center}
        \hspace{50pt}
        \begin{minipage}{0.45\textwidth}
            \begin{forest}
                [\textcolor{blue}{1-10}
                    [\textcolor{teal}{1-5}
                        [\textcolor{cyan}{1-3}]
                        [\textcolor{orange}{3-4}]
                    ]
                    [\textcolor{magenta}{5-10}
                        [\textcolor{olive}{5-8}]
                        [\textcolor{violet}{8-10}]
                    ]
                ]
            \end{forest}
        \end{minipage}
        \begin{minipage}{0.4\textwidth}
            \begin{forest}
                [\textcolor{blue}{1-10}
                    [\textcolor{teal}{1-2}]
                    [\textcolor{magenta}{2-10}
                        [\textcolor{cyan}{2-3}]
                        [\textcolor{olive}{3-10}
                            [\textcolor{orange}{3-4}]
                            [\textcolor{violet}{4-10}]
                        ]
                    ]
                ]
            \end{forest}
        \end{minipage}
    \end{center}
    \caption[An illustrative example of the evaluation metric used in \citet{tseng-etal-2023-cascading}.]{An illustrative example of the evaluation metric used in \citet{tseng-etal-2023-cascading}.
        Node labels represent the time spans of speech segments, measured in a specific unit (e.g., seconds).
        These two trees will receive a perfect $F_1$ score under their metric.
        Best viewed in color: nodes in the same color are aligned by the max-sum-duration-based bipartite matching algorithm, and each pair of aligned nodes contributes $1$ to the numerator of both precision and recall.
    }
    \label{fig:tseng-etal-eval}
\end{figure}
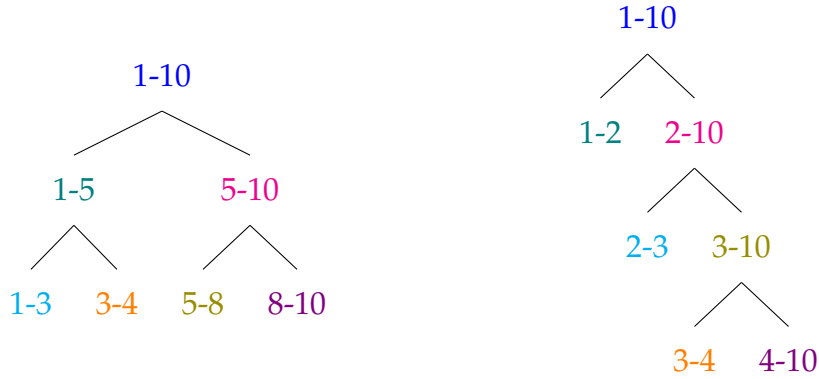

\paragraph{Speech parsing and its applications.}
Early work on speech parsing can be traced back to SParseval \citep{roark-etal-2006-sparseval}, a toolkit that evaluates text parsers given potentially errorful automatic speech recognition (ASR) results.
In the past, syntax has also been studied in the context of speech prosody \citep{wagner2010experimental,kohn2018empirical}, and another line of work \citep{tran-etal-2018-parsing, tran2020role,tran-ostendorf-2021-assessing} has incorporated acoustic-prosodic features for text parsing with auxiliary speech input.
\citet{lou2019neural} trains a text parser \citep{kitaev-klein-2018-constituency} to detect speech disfluencies, and \citet{pupier2022end} trains a text dependency parser from speech jointly with an ASR model.
Concurrent work \citep{tseng-etal-2023-cascading} pipelines an unsupervised ASR system with DIORA \citep{drozdov-etal-2019-unsupervised} to perform unsupervised speech parsing.
While the predicted parse trees are, in principle, comparable to the results in this work, the evaluation metric they adopted may introduce extra difficulty in interpreting the results---concretely, they use a bipartite matching algorithm to align the constituent spans in both trees by maximizing the sum of overlap time durations between aligned spans, treat the aligned spans as precise/recalled constituents, and report the $F_1$ score as in \parseval \citep{black-etal-1991-procedure}.
In an extreme case (\cref{fig:tseng-etal-eval}), such a metric assigns a perfect $F_1$ score to two trees that are not similar in structural terms, which further motivates us to develop a new evaluation metric, \structiou, in \cref{chapter:structiou}.

On the application side, syntactic parses of texts have been applied to prosody modeling in end-to-end text-to-speech \citep{guo2019exploiting, tyagi2020dynamic, kaiki2021using}.
While this work builds upon pre-existing text parsing algorithms, we focus on phrase structure induction in the absence of text.

\section{Method}
\label{sec:avnsl-method}
Given paired spoken captions and images, \avnsl infers phrase structures from speech utterances without relying on text.
The basis of \avnsl is the Visually-Grounded Neural Syntax Learner (\vgnsl; \cref{chapter:vgnsl}), which learns to induce constituency parse trees by guiding a sequential sampling process with text-image matching.
We refer the readers to \cref{chapter:vgnsl} for a detailed description of \vgnsl, and here we focus on the modifications and extensions to \vgnsl that enable \avnsl to parse speech.
To implement \avnsl, we obtain sequences of word segments and extract segment-level self-supervised representations.
The extracted segment representations are then used to induce phrase structures, analogously to the randomly initialized word representations in \vgnsl.
In this section, we describe in detail the unique components of \avnsl.

\subsection{Audio-Visual Word Segmentation and Representation}
\input{figures/402-avnsl-vghubert-insertion-demo.tex}
We use VG-HuBERT \citep{peng2022word}, a model trained to associate spoken captions with natural images via retrieval, for word segmentation.
After training, spoken word segmentation emerges via magnitude thresholding of the self-attention heads of the audio encoder: at layer $\ell$, we (1) sort the attention weights from the \texttt{[CLS]} token to other tokens in descending order, and (2) apply a threshold $p$ to retain the top $p\%$ of the overall attention magnitude (\cref{fig:avnsl-vghubert-insertion-demo-a}).

Empirically, however, the VG-HuBERT word segmentation model tends to ignore function words such as \textit{a} and \textit{of}.
This is natural because function words are often less visually grounded and may not be as salient in the attention maps.
Therefore, we devise a simple heuristic to pick up function word segments by inserting a short word segment wherever there is a gap of more than $s$ seconds that VG-HuBERT fails to place a segment (\cref{fig:avnsl-vghubert-insertion-demo-b}).
We additionally apply unsupervised voice activity detection \citep{tan2020rvad} to restrict segment insertion to only voiced regions.
The length of the insertion gap $s$, the VG-HuBERT segmentation layer $l$, attention magnitude threshold $p\%$, and model training snapshots across random seeds and training steps are all chosen in an unsupervised fashion using minimal Bayes risk decoding.

We use the word segments output by VG-HuBERT to calculate the representations.
Let $R =\{r_j\}_{j=1}^T$ denote the frame-level representation sequence, where $T$ is the speech sequence length.
Audio-visual word segmentation returns an alignment $A(i) = r_{p:q}$ that maps the $i^\textit{th}$ word segment to the $p^\textit{th}$ to $q^\textit{th}$ acoustic frames.
The segment-level continuous representation for the $i^\textit{th}$ word is
$w_i^0 = \sum_{t\in A(i)}a_{i,t}r_{i,t}$,
where $a_{i,t}$ is the attention weights over the segments specified by $A(i)$.
In \avnsl, $R$ is the layer representation from a pre-trained speech model (e.g., VG-HuBERT), and $a_{i, t}$ is the \texttt{[CLS]} token attention weights over frames within each segment.

\subsection{Self-Training}
\label{subsec:avnsl-self-training}
\citet{shi2020role} have shown that self-training can usually improve parsing performance: the approach involves training an additional parser to fit the output generated by a pre-existing learned parser.
Concretely, they use Benepar \cite{kitaev-klein-2018-constituency}, a supervised neural constituency parser, as the base model for self-training, where it (1) takes a sentence as the input, (2) maps it to word representations, and (3) predicts a score for all spans of being in the constituency parse tree.
The model evaluates all possible tree structures for inference and outputs the highest-scoring one.

Following this idea, we apply self-training to improve \avnsl.
We extend Benepar to the speech domain and introduce s-Benepar, which takes segment-level continuous mean-pooling HuBERT representations instead of discrete words as the input and outputs the constituency parse trees.

\subsection{Unsupervised Decoding}
\label{subsec:avnsl-unsupervised-decoding}
Following the decoding design of \vgnsl, we extend consistency-based decoding to \avnsl, which is also similar in spirit to minimum Bayes risk (MBR) decoding (\cref{chapter:mbrexec}), for both spoken word segmentation and phrase-structure induction.
Given a loss function $\ell_\textit{MBR}(O_1, O_2)$ between two outputs $O_1$ and $O_2$, and a set of $k$ outputs $\mathcal{O} = \{O_1, \ldots, O_k\}$, we select the optimal output
$$
    \hat{O} = \arg\min_{O'\in \mathcal{O}} \sum_{O''\in \mathcal{O}} \ell_\textit{MBR}(O', O'').
$$
The output candidate set $\mathcal{O}$ is obtained by sampling from the output distribution.
We defer more detailed discussions regarding MBR decoding to \cref{chapter:mbrexec}.

For word segmentation, we define the loss between two segmentation proposals $\mathcal{S}_1$ and $\mathcal{S}_2$ as
$\ell_\textit{MBR}(\mathcal{S}_1, \mathcal{S}_2) = -\textsc{mIoU}(\mathcal{S}_1, \mathcal{S}_2),$ where $\textsc{mIoU}(\cdot,\cdot)$ denotes the mean intersection over union ratio across all matched pairs of predicted word spans.
We match the predicted word spans using the maximum weight-matching algorithm \citep{galil1986efficient}, where word spans correspond to vertices, and we define edge weights by the temporal overlap between the corresponding spans.

For phrase structure induction, the loss function between two parse trees $\mathcal{T}_1$ and $\mathcal{T}_2$ is
$\ell_\textit{MBR}(\mathcal{T}_1, \mathcal{T}_2) = 1 - F_1(\mathcal{T}_1, \mathcal{T}_2),$ where $F_1(\cdot, \cdot)$ denotes the $F_1$ score between the two trees.

\section{Experiments}
\label{sec:avnsl-experiments}
\subsection{Setup}
\label{subsec:avnsl-exp-setting}
\paragraph{Datasets.}
We first evaluate models on SpokenCOCO \citep{hsu2020text}, the spoken version of MSCOCO \citep{lin2014microsoft}, where humans verbally read the text captions in English.
It contains 83k/5k/5k images for training, validation, and testing.
Each image has five corresponding captions.

We also extend our experiments to German, where we synthesize German speech from the Multi30K captions \citep{elliott2016multi30k}.\footnote{~Synthesized with the pre-trained German Tacotron2 model (\url{https://github.com/thorstenMueller/Thorsten-Voice}).}
It contains 29k/1k/1k images for training, validation, and testing.
Each image has one corresponding caption.
Following \vgnsl (\cref{chapter:vgnsl}), we use pre-trained English Benepar \citep{kitaev-klein-2018-constituency} to generate the oracle parse trees for captions.

\paragraph{Preprocessing.}
We use the Montreal Forced Aligner \citep{mcauliffe-etal-2017-montreal} trained on the specific language (i.e., English or German) to obtain the oracle word segmentation.
We remove utterances that have mismatches between ASR transcripts and text captions.

\subsection{Baselines and Toplines}
\label{subsec:avnsl-baselines}
We consider the following baselines and modeling alternatives to examine each component of \avnsl:

\begin{enumerate}
    \item \textbf{Trivial tree structures.}
          Following \vgnsl (\cref{chapter:vgnsl}), we include baselines without linguistic information: random binary trees, left-branching binary trees, and right-branching binary trees.

    \item \textbf{AV-cPCFG.}
          We train compound probabilistic context-free grammars \citep[cPCFG;][]{kim-etal-2019-compound} on word-level discrete speech tokens given by VG-HuBERT.
          Unlike in \avnsl, the segment representations are discretized via k-means to obtain word-level indices.
          AV-cPCFG uses visual cues only for segmentation and segment representations, not for phrase structure induction.

    \item \textbf{DPDP-cPCFG.}
          In contrast to AV-cPCFG, DPDP-cPCFG does not rely on any visual grounding throughout.
          We use DPDP \citep{kamper2022word} and pre-trained HuBERT \citep{hsu2021hubert} followed by k-means to obtain discrete word indices.\footnote{~We sweep the number of word clusters over $\{1\text{k}, 2\text{k}, 4\text{k}, 8\text{k}, 12\text{k}, 16\text{k}\}.$}

    \item \textbf{Oracle \avnsl} (topline).
          To remove the uncertainty of unsupervised word segmentation, we directly train \avnsl on top of the oracle word segmentation via forced alignment.
          Due to the absence of VG-HuBERT, the frame-level representations $R$ are obtained from pre-trained HuBERT while the attention weights $a_{i,t}$ are parameterized by a 1-layer MLP, jointly trained with the tree sampling module instead.
\end{enumerate}

\subsection{Evaluation Metrics}
\label{subsec:avnsl-eval-metrics}

\paragraph{Word segmentation.}
We use the standard word boundary prediction metrics (precision, recall, and the $F_1$ score), calculated by comparing the temporal position between inferred and forced-aligned word boundaries.
An inferred boundary located within $\pm 20$\textit{ms} of a forced aligned boundary is considered a successful prediction.

\paragraph{Parsing.}
For parsing with the oracle word segmentations, we use \parseval \citep{black-etal-1991-procedure} to calculate the $F_1$ score between the predicted and reference parse trees.
For parsing with inferred word segmentation, due to the mismatch in the number of nodes between the predicted and reference parse trees, we use the structured average intersection-over-union ratio (\structiou) as an additional metric---this is a new metric developed in this dissertation, which will be presented in details in \cref{chapter:structiou}.

\structiou considers both word segmentation quality and temporal overlap between induced constituents.
Concretely, the input is two constituency parse trees over the same speech utterance, $\mathcal{T}_1=\{a_i\}_{i=1}^n$ and $\mathcal{T}_2=\{b_j\}_{j=1}^m$, where $a_i$ and $b_j$ are time spans.
Suppose $a_i$ from $\mathcal{T}_1$ is aligned to $b_j$ from $\mathcal{T}_2$.  In a valid alignment, the following conditions must be satisfied:
\begin{enumerate}
    \item Any descendant of $a_i$ may either align to a descendant of $b_j$ or be left unaligned;
    \item Any ancestor of $a_i$ may either align to an ancestor of $b_j$ or be left unaligned;
    \item Any descendant of $b_j$, may either align to a descendant of $a_i$ or be left unaligned;
    \item Any ancestor of $b_j$, may either align to an ancestor of $a_i$ or be left unaligned.
\end{enumerate}
Given a Boolean matrix $\bm{A}$, where $A_{i,j}=1$ denotes that $a_i$ aligns to $b_j$, we compute the structured average \textsc{IoU} between $\mathcal{T}_1$ and $\mathcal{T}_2$ over $\bm{A}$ by
$$\structiou(\mathcal{T}_1, \mathcal{T}_2; \bm{A}) = \frac{2}{n+m} \left(\sum_{i=1}^{n_1} \sum_{j=1}^{n_2} A_{i,j} \textsc{IoU}(a_i, b_j)\right),$$
and the final evaluation result is obtained by maximizing the \structiou score across all valid alignments.
Dynamic programming allows us to calculate the optimal \structiou score within $\mathcal{O}(n^2m^2)$ time.
We will describe the detailed problem formulation and algorithm of \structiou in \cref{chapter:structiou}.

\subsection{Results: Unsupervised Word Segmentation}
\label{subsec:avnsl-word-segmentation}
\input{tables/402-avnsl-word-segmentation.tex}
We validate the effectiveness of our unsupervised word segmentation approach.
We first compare our improved VG-HuBERT with segment insertion to the original VG-HuBERT \citep{peng2022word} and DPDP \citep{kamper2022word}, a speech-only word segmentation method (\cref{tab:avnsl-word-segmentation}).
We find that segment insertion improves the recall and hurts the precision while achieving the highest $F_1$ score.

Next, we compare MBR-based and supervised decoding.
For practice efficiency, we implement MBR-based decoding as follows: we first run a pilot hyperparameter selection, performing word segmentation on all candidates in the SpokenCOCO validation set, and subsequently choose the $10$ most selected sets of hyperparameters to perform another round of MBR selection on the training set.

For German word segmentation, we employ identical models and settings as those used for English, as \citet{peng2023syllable} have shown that the word segmentation capability of English VG-HuBERT demonstrates cross-lingual generalization without any adaptation.
On German Multi30K, our method achieves an $F_1$ score of $37.46$ with MBR, which outperforms supervised hyperparameter tuning ($36.45$).

\subsection{Results: Unsupervised Phrase Structure Induction}
\label{subsec:avnsl-phrase-structure-induction}
\input{tables/401-avnsl-main-results.tex}
We quantitatively show that \avnsl learns meaningful phrase structure given word segments (\cref{tab:avnsl-main-results}).
The best performing \avnsl is based on our improved VG-HuBERT with MBR top 10 selection for word segmentation, VG-HuBERT layers as the segment representations, and another MBR decoding over phrase structure induction hyperparameters, including training checkpoints and segment representation layers.
Comparing \avnsl against AV-cPCFG and AV-cPCFG against DPDP-cPCFG, we empirically show the necessity of training \avnsl on \textit{continuous} segment representation instead of discretized speech tokens, and the effectiveness of visual grounding in our overall model design.

\input{tables/403-avnsl-self-training.tex}
Next, we compare the performance of \avnsl with and without self-training (\cref{tab:avnsl-self-training}) and find that self-training with an s-Benepar backbone improves the best \avnsl performance from 0.521 (\cref{tab:avnsl-main-results}) to 0.538.

\input{tables/404-avnsl-oracle-word-segmentation.tex}
Thirdly, \cref{tab:avnsl-oracle-word-segmentation-results} isolates phrase structure induction from word segmentation quality with oracle \avnsl.
Unlike in \cref{tab:avnsl-main-results}, we can adopt the \parseval $F_1$ score \citep{black-etal-1991-procedure} for evaluation since there is no mismatch in the number of tree nodes.
Unsupervised oracle \avnsl matches or outperforms text-based \vgnsl with proper segment-level representations.
Similarly to \cref{tab:avnsl-self-training}, self-training with s-Benepar on oracle \avnsl trees further improves the syntax induction results, almost matching that of right-branching trees.

Perhaps surprisingly, right-branching trees (RBT) with oracle and VG-HuBERT word segmentation reach the best English \structiou and $F_1$ scores on SpokenCOCO, respectively.
We note that the RBTs highly align with the head initiality of English \citep{baker2001atoms}, especially in our setting where all punctuation marks were removed.
In contrast, our experiments on German show that \avnsl outperforms both RBTs and left-branching trees in terms of \structiou (\cref{tab:avnsl-german-results}).\footnote{For German grammar induction with oracle segmentation, oracle \avnsl attains 33.94 $F_1$ while LBT/RBT attain 26.70/25.30 $F_1$ respectively.}
\input{tables/405-avnsl-german-results.tex}

\subsection{Analysis}
\label{subsec:avnsl-analysis}

\textbf{Recall of each type of constituent.}
Similarly to \cref{chapter:vgnsl}, we present the recall of specific types of constituents (\cref{tab:avnsl-recall-analysis}).
While \vgnsl benefits from the abstract-initial bias, where abstract words are encouraged to appear at the beginning of a constituent, \avnsl outperforms all variations of \vgnsl on all constituent categories except NP.
We hypothesize that the spoken utterances are noisier than the written text, and it is, therefore, more challenging for \avnsl to capture the noun phrases; however, the spoken utterances contain arguably more information, such as prosody, that can be used to infer other types of constituents.
We leave the investigation of these hypotheses to future work.
\input{tables/406-avnsl-recall-analysis.tex}

\paragraph{Ablation study.}
We introduce three ablations to evaluate the efficacy of high-quality word segmentation, visual representation, and speech representation (\cref{tab:avnsl-ablation-study}).
Concretely, we train \avnsl with the following modifications:
\begin{enumerate}
    \item Given the number of words $n$, we divide the speech utterances uniformly into $n$ chunks to get the word segmentation and use the same visual representations as \avnsl.
    \item We replace visual representations with random vectors, where each pixel is independently sampled from a uniform distribution, and use the oracle word segmentation.
    \item We replace the self-supervised speech representations (i.e., HuBERT) with log-Mel spectrograms.
\end{enumerate}
We observe significant performance drops in all settings, compared to \avnsl with oracle word segmentations.
This set of results complements \cref{tab:avnsl-main-results}, stressing that precise word segmentation and both high-quality visual and speech representations are all necessary for phrase structure induction from speech.
\input{tables/407-avnsl-ablation.tex}

\section{Conclusion and Discussion}
\label{sec:avnsl-conclusion}
Previous research has achieved notable progress in zero-resource speech processing and grammar induction by employing multi-modal techniques.
In our study, we propose an approach to model human language acquisition that leverages the visual modality as the source of grounding signals.
Our approach, \avnsl, encompasses extracting word-level representations from speech and deriving syntactic structures from those representations, thereby eliminating the reliance on text.
Through quantitative and qualitative analyses, we demonstrate in both English and German experiments that our proposed model infers meaningful constituency parse trees based on continuous word segment representations.
Our work represents the initial step in grammar induction within textless settings, paving the way for future research endeavors, which include but are not limited to (1) building end-to-end models that take spoken utterances and produce their syntactic analysis, (2) understanding the contribution of various grounding signals to grammar induction, and (3) modeling human language acquisition in grounded environments.

As a side contribution, we find that an unsupervised word segmentation model, VG-HuBERT, and heuristics-based segment insertion outperform the previous state-of-the-art method in unsupervised word segmentation.
This result suggests that we may combine the advantages of audio-visual \interalia{peng2022word} and speech-only \interalia{kamper2022word} word segmentation models to improve the quality of word segmentation further: intuitively, the former better extracts word boundaries of more visually concrete words (see more discussions in \cref{chapter:vgnsl}), whereas the latter provides a general option for a broader range of words.

Lastly, we note that the work presented in this chapter should be considered a preliminary exploration of unsupervised speech parsing, together with \citet{tseng-etal-2023-cascading}.
While it is clear that visually grounded text parsing models can significantly outperform right-branching baselines (\cref{chapter:vgnsl}), our best-performing visually grounded speech parsing model still lags behind the simple right-branching trees in English.
As discussed in the prior content, we note that the right-branching trees highly align with the head initiality of English, and the results presented in this chapter are nontrivial as we do not incorporate any linguistic priors.
On the other hand, this set of results also implies significant room for improvement in the quality of visually grounded speech parsing.
In particular, an end-to-end approach that jointly learns word segmentation and phrase structure induction may be a promising direction for future research.

%% file: figures/401-avnsl-teaser.tex
\begin{figure}[t]
    \centering
    \includegraphics[width=\textwidth,page=2]{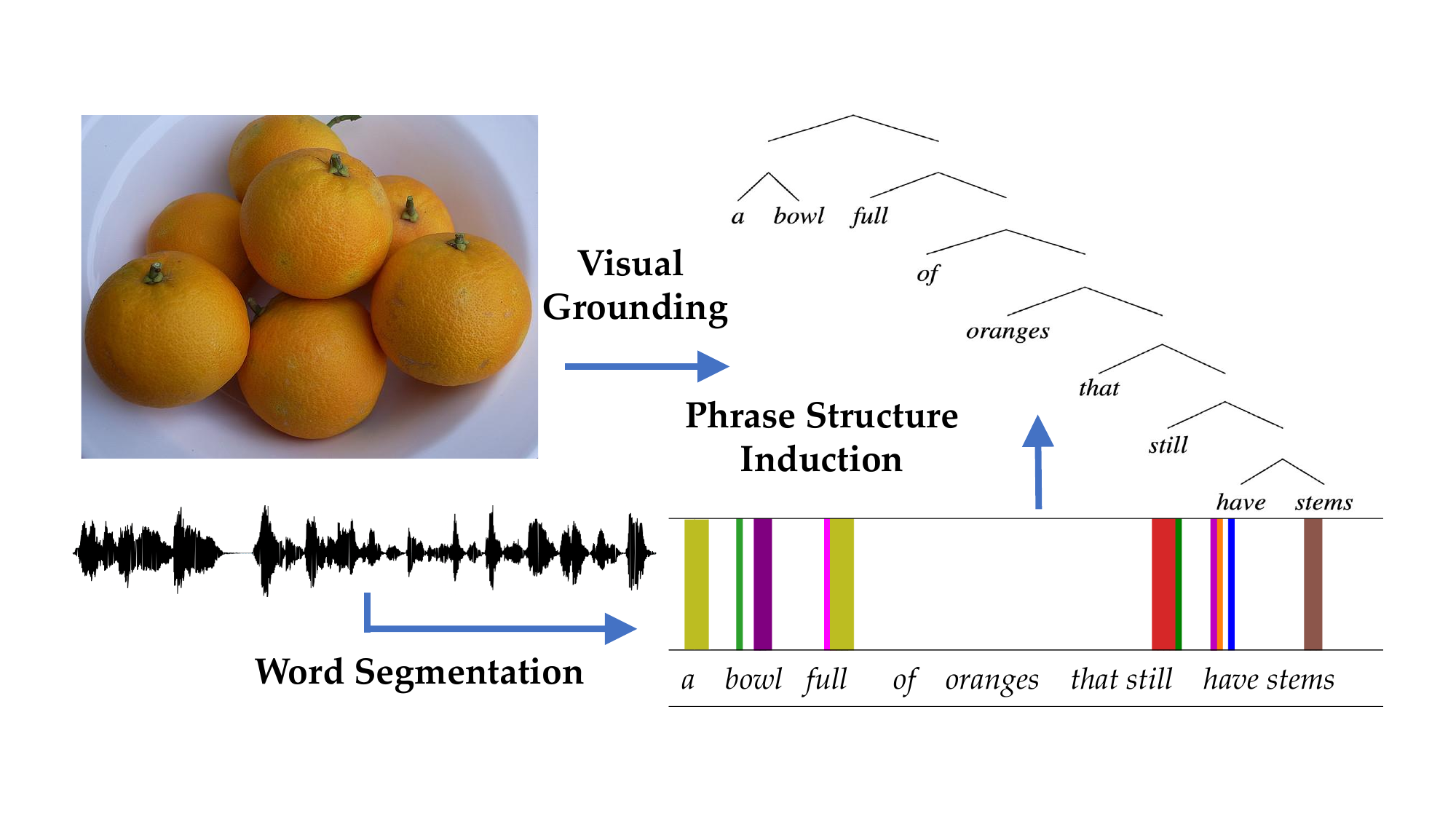}
    \vspace{-40pt}
    \caption[Illustration of the \avnsl model.]{Illustration of the \avnsl model. We adopt the \vgnsl model architecture to perform phrase structure induction from visually grounded speech.}
    \label{fig:avnsl-teaser}
\end{figure}

%% file: figures/402-avnsl-vghubert-insertion-demo.tex
\begin{figure}[t]
    \centering
    \begin{subfigure}[t]{\textwidth}
        \centering
        \includegraphics[width=0.8\textwidth,page=3]{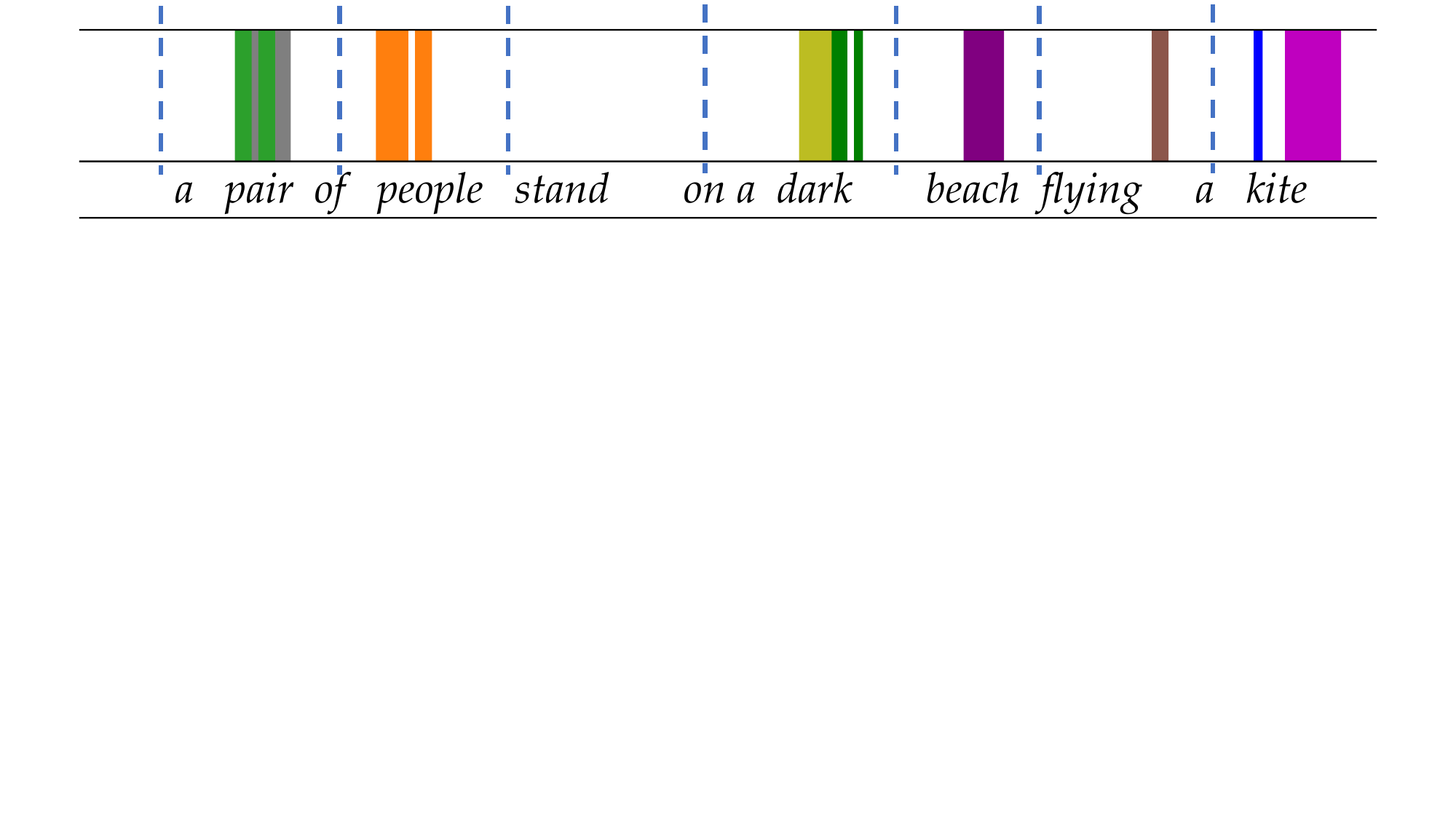}
        \vspace{-150pt}
        \caption{
            Example of VG-HuBERT word segmentation (top): different colors denote different attention heads, and color transparency represents the magnitude of the attention weights.
            Adjacent attention boundaries (vertical dashed lines) are used as the word boundaries.
            \label{fig:avnsl-vghubert-insertion-demo-a}
        }
    \end{subfigure}

    \vspace{-80pt}

    \begin{subfigure}[t]{\textwidth}
        \centering
        \includegraphics[width=0.8\textwidth,page=4]{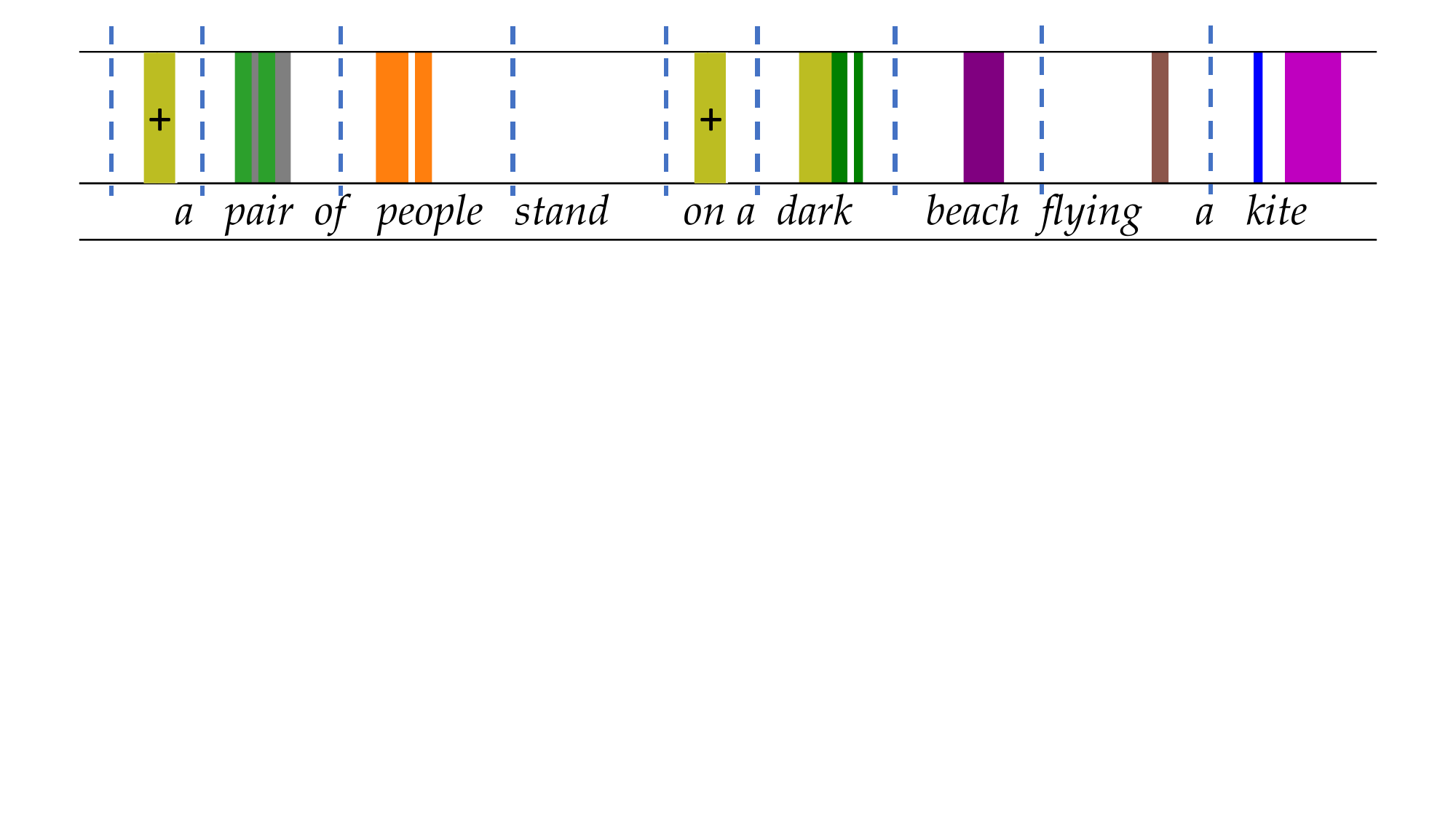}
        \vspace{-140pt}
        \caption{
            \textit{Segment insertion} (bottom): short segments (marked with ``+'') are placed in long enough gaps between existing segments to recover function words.
            \label{fig:avnsl-vghubert-insertion-demo-b}
        }
    \end{subfigure}

    \vspace{-100pt}
    \caption[Examples of VG-HuBERT word segmentation and segment insertion.]{Examples of VG-HuBERT word segmentation and segment insertion (this work). Best viewed in color.}
    \label{fig:avnsl-vghubert-insertion-demo}
\end{figure}

%% file: tables/402-avnsl-word-segmentation.tex
\begin{table}[t]
    \centering
    \begin{tabular}{cccccc}
        \toprule
        \bf Method & \bf Decoding   & \bf Precision & \bf Recall & \bf $F_1$ \\
        \midrule
        DPDP \citep{kamper2022word} &supervised & 17.37&9.00&11.85 \\
        VG-HuBERT \citep{peng2022word} & supervised &\bf 36.19&27.22&31.07 \\
        VG-HuBERT &  supervised &34.34 &29.85 &31.94 \\
        w/ seg. ins. (ours) & MBR     & 33.31 &\bf 34.90 &\bf 34.09 \\
        \bottomrule
    \end{tabular}
    \caption[English word segmentation results on the SpokenCOCO validation set.]{
    English word segmentation results on the SpokenCOCO validation set.
    Supervised decoding methods require an annotated development set to choose the best hyperparameters.
    The best number in each column is in boldface.
    VG-HuBERT with segment insertion and MBR decoding achieves the best boundary $F_1$.
    }
    \label{tab:avnsl-word-segmentation}
\end{table}

%% file: tables/401-avnsl-main-results.tex
\begin{table}[t]
    \centering
    \begin{tabular}{lcccc}
        \toprule
        \multicolumn{3}{c}{\bf Model} & {\bf Output} & \multirow{2}{*}{\bf \structiou} \\
        \cmidrule(lr){1-3}
        {\bf Syntax Induction} & {\bf Segmentation} & {\bf Seg. Repr.} & {\bf Selection} & {} \\
        \midrule
        Right-Branching & VG-HuBERT &  & & \textbf{0.546} \\
        Right-Branching & DPDP & & & 0.478 \\
        \midrule
        AV-cPCFG & VG-HuBERT & discrete & supervised & 0.499 \\
        AV-cPCFG & VG-HuBERT & discrete & supervised & 0.481 \\
        DPDP-cPCFG & DPDP & discrete & supervised & 0.465 \\
        \midrule
        AV-NSL & VG-HuBERT & continuous & MBR & 0.516 \\
        AV-NSL & VG-HuBERT & continuous & MBR & 0.521 \\
        \bottomrule
    \end{tabular}
    \caption[English phrase structure induction results on SpokenCOCO.]{
        English phrase structure induction results on SpokenCOCO.
        Subscripts denote layer number, e.g., HuBERT$_{10}$ denotes the 10\textsuperscript{th} layer representation from HuBERT.
        We list the best-performing hyperparameters for each modeling choice.
    }
    \label{tab:avnsl-main-results}
\end{table}

%% file: tables/403-avnsl-self-training.tex
\begin{table}[t]
    \centering
    \begin{tabular}{lcc}
    \toprule
    \bf Segment Representation & \bf Output Selection & \bf \structiou \\
    \midrule
    HuBERT & last ckpt. & \textbf{0.538} \\
    {HuBERT$_{2,4,6,8,10,12}$} & MBR & 0.536 \\
    \bottomrule
    \end{tabular}
    \caption[Results of self-training with s-Benepar.]{Results of self-training with s-Benepar, trained on outputs from the best AV-NSL model (\structiou = 0.521) from \cref{tab:avnsl-main-results}.
    Inputs to s-Benepar are segment-level HuBERT representations instead of VG-HuBERT representations.
    }
    \label{tab:avnsl-self-training}
\end{table}

%% file: tables/404-avnsl-oracle-word-segmentation.tex
\begin{table}[t]
    \centering
    \begin{tabular}{llcc}
        \toprule
        \multicolumn{2}{c}{\bf Model} & {\bf Output} & \multirow{2}{*}{$F_1$} \\
        \cmidrule(lr){1-2}
        {\bf Syntax Induction} & {\bf Seg. Representation} & {\bf Selection} & {} \\
        \midrule
        Right-Branching & N/A & N/A & \textbf{57.39} \\
        \midrule
        VG-NSL & word embeddings & Supervised & 53.11 \\
        oracle AV-NSL & HuBERT$_{2}$ & Supervised & 55.51 \\
        \midrule
        oracle AV-NSL $\rightarrow$ s-Benepar & HuBERT$_{2}$ & MBR & 57.24 \\
        \bottomrule
    \end{tabular}
    \caption[\parseval $F_1$ scores given oracle segmentation.]{\parseval $F_1$ scores given oracle segmentation. The best number is in boldface.
    }
    \label{tab:avnsl-oracle-word-segmentation-results}
\end{table}

%% file: tables/405-avnsl-german-results.tex
\begin{table}[t]
    \centering
    \begin{tabular}{llcc}
        \toprule
        \multicolumn{2}{c}{\bf Model} & {\bf Output} & \multirow{2}{*}{\bf \textsc{SAIoU}} \\
        \cmidrule(lr){1-2}
        {\bf Induction} & {\bf Segmentation} & {\bf Selection} & {} \\
        \midrule
        Right-Branching & VG-HuBERT+MBR$_{10}$ & N/A & 0.456 \\
        Left-Branching & VG-HuBERT+MBR$_{10}$ & N/A & 0.461 \\
        \midrule
        AV-NSL & VG-HuBERT+MBR$_{10}$ &  MBR & \textbf{0.487} \\
        \bottomrule
    \end{tabular}
    \caption[Phrase structure induction results on (spoken) German Multi30K.]{Phrase structure induction results on the (spoken) German Multi30K test set.
    The best number is in boldface. }
    \label{tab:avnsl-german-results}
\end{table}

%% file: tables/406-avnsl-recall-analysis.tex
\begin{table}[t]
    \centering
    \begin{tabular}{lccccc}
            \toprule
            \multirow{2}{*}{\bf Model} & \multirow{2}{*}{$F_1$} & \multicolumn{4}{c}{\bf Constituent Recall} \\ \cmidrule(lr){3-6}
            & & \bf NP & \bf VP & \bf PP & \bf ADJP \\
            \midrule
            \vgnsl \citep{shi2019visually} & 50.4 & \textbf{79.6} & 26.2 & 42.0 & 22.0 \\
            \vgnslai & 53.3 & 74.6 & 32.5 & 66.5 & 21.7 \\
            \vgnslai + FastText & 54.4 & 78.8 & 24.4 & 65.6 & 22.0 \\
            \midrule
            oracle \avnsl & \textbf{55.5} & 55.5 & \textbf{68.1} & \textbf{66.6} & \textbf{22.1} \\
            \bottomrule
    \end{tabular}
    \caption[Recall for specific typed phrases on SpokenCOCO.]{
        Recall of specific typed phrases, including noun phrases (NP), verb phrases (VP), prepositional phrases (PP), and adjective phrases (ADJP), as well as the overall $F_1$ score, evaluated on SpokenCOCO test set.
        \vgnsl numbers are the same as those in \cref{chapter:vgnsl}.
    }
    \label{tab:avnsl-recall-analysis}
\end{table}

%% file: tables/407-avnsl-ablation.tex
\begin{table}[t]
    \centering
    \begin{tabular}{llcc}
        \toprule
        \multicolumn{2}{c}{\bf Model} & \multirow{2}{*}{\bf Visual} & \multirow{2}{*}{$F_1$} \\
        \cmidrule(lr){1-2}
        {\bf Word Segmentation} & {\bf Seg. Repre.} & {} & \\
        \midrule
        MFA & HuBERT$_2$ & ResNet 101 & 55.51\\
        \midrule
        Uniform & HuBERT$_{2}$ & ResNet 101 & 48.97 \\
        MFA & HuBERT$_2$ & random & 31.23 \\
        MFA & logMel spec & ResNet 101 & 42.01 \\
        \bottomrule
    \end{tabular}
    \caption{\parseval $F_1$ scores for ablations over word segmentation, visual representation, and speech representation.}
    \label{tab:avnsl-ablation-study}
\end{table}

%% file: src/50-structiou.tex
\chapter{\structiou: A Structured Alignment--Based Evaluation Metric for Constituency Parsing}
\label{chapter:structiou}
Models for \textit{textless} constituency parsing have been proposed in \cref{chapter:avnsl} \citep{lai2023audio} and other concurrent work \citep{tseng-etal-2023-cascading}.
In contrast to earlier work that parses manually labeled \interalia{charniak-johnson-2001-edit} or automatic \interalia{kahn-ostendorf-2012-joint} speech transcriptions, these models construct constituency parse trees over automatically recognized spoken word boundaries, where each word is represented with a time range of the spoken utterance, without using any form of text.
To evaluate these textless models, we need a metric that compares the predicted tree (over spoken word boundaries) with the manually labeled ground-truth tree (over written words) and faithfully reflects the parsing quality.
Since the automatically recognized word boundaries may be imperfect, the metric should also reflect the changes in parsing quality due to word boundary errors.
To the best of our knowledge, none of the existing metrics meets these requirements, as they are all designed to compare parse trees over discrete word sequences instead of continuous time ranges.

Motivated by the need for textless speech parsing evaluation \citep{lai2023audio,tseng-etal-2023-cascading}, in this work, we introduce the structured average intersection-over-union ratio (\structiou; \cref{fig:structiou-example-speech-parse-tree}), a metric that compares two parse trees over time ranges.
We relax the definition of segment trees \citep{bentley1977solutions} to represent speech constituency parse trees, where each node is associated with an interval that represents the time range of the corresponding spoken word or constituent.
To obtain the ``ground-truth'' speech parse trees, we use the forced alignment algorithm \citep{mcauliffe-etal-2017-montreal}, a supervised and highly accurate method that aligns written words to time ranges of the corresponding spoken utterance, to project the ground-truth text parses onto the time domain.
\structiou is calculated by aligning the same-label nodes in the predicted and ground-truth parse trees, following structured constraints that preserve parent-child relations.
The calculation of \structiou can be formulated as an optimization problem (\cref{sec:structiou-problem-formulation}) with a polynomial-time solution (\cref{sec:structiou-solution}) in terms of the number of tree nodes.

\input{figures/501-structiou-teaser.tex}

Although \structiou is designed to evaluate speech parsing, it is also applicable to text parsing evaluation.
We analyze \structiou for both purposes: in speech parsing evaluation, \structiou robustly takes into account both the structure information and word boundaries; in text parsing evaluation, while maintaining a high correlation with the \parseval $F_1$ score, \structiou shows a higher tolerance to potential syntactic ambiguity.

\section{Related Work}
\label{sec:structiou-related-work}
\paragraph{Text constituency parsing and evaluation.}
In the past decades, there has been much effort in building and improving constituency parsing models \interalia{collins-koo-2005-discriminative,charniak-johnson-2005-coarse,mcclosky-etal-2006-effective,durrett-klein-2015-neural,cross-huang-2016-span,dyer-etal-2016-recurrent,choe-charniak-2016-parsing,stern-etal-2017-minimal,kitaev-klein-2018-constituency}.
\parseval \citep{black-etal-1991-procedure} has been the standard evaluation metric for constituency parsing in most scenarios, which takes the ground truth and predicted trees and calculates the harmonic mean of precision and recall of labeled spans.
For morphologically rich languages, \textsc{TedEval} \citep{tsarfaty-etal-2012-joint} extends \parseval to accept multiple morphological analyses over sequences of words.
All of these metrics are designed to evaluate parses over discrete word sequences and cannot be easily extended to evaluate speech parses over continuous time ranges.
Although our metric, \structiou, is designed to evaluate speech constituency parsing, it can be easily extended for text parsing evaluation, reflecting a different aspect from existing metrics (\cref{sec:structiou-text-constituency-parsing-evaluation-english}).

\paragraph{Speech constituency parsing and its evaluation.}
Work on conversational speech parsing has focused on addressing the unique challenges posed by speech, including speech recognition errors \citep{kahn-ostendorf-2012-joint,marin-ostendorf-2014-domain}, unclear sentence boundaries \citep{kahn-etal-2004-parsing}, disfluencies \citep{jamshid-lou-johnson-2020-improving,kahn-etal-2005-effective,lease-johnson-2006-early}, as well as integrating prosodic features into the parsing systems \citep{tran-etal-2018-parsing,tran-ostendorf-2021-assessing}.
On the evaluation side, the closest work to ours is \textsc{SParsEval} \citep{roark-etal-2006-sparseval}, which extends \parseval to account for speech recognition errors by allowing for word-level insertion, deletion, and substitution.
In contrast, our metric \structiou applies to the cases where no speech recognizer is applied or available.

\paragraph{Other structured evaluation metrics for parsing.}
There have been evaluation metrics of abstract meaning representations \citep[AMRs;][]{cai-knight-2013-smatch}, where two AMR graphs are matched by solving an NP-complete integer linear programming problem.
While our work shares the spirit with theirs, we focus on the evaluation of speech constituency parsing over continuous word boundaries.
A polynomial-time exact solution exists for our optimization problem.

\section{Preliminaries}
\label{sec:structiou-preliminaries}
For simplicity, we use real-valued open intervals to represent speech spans, although most of the following definitions and conclusions can be easily extended to closed and half-open intervals.
We start with basic operations over open intervals (\cref{sec:structiou-open-interval-operations}) and then introduce relaxed segment trees (\cref{sec:structiou-extended-segment-tree}), which are used to represent parse trees.
\subsection{Open Intereval Operations}
\label{sec:structiou-open-interval-operations}

\begin{definition}
	\label{def:open-interval-length}
	The \textbf{length} of a real-valued open interval $I=(a, b)$, where $a < b$, is $|I| = b-a$.
\end{definition}

\begin{definition}
	\label{def:structiou-intersection-size}
	The \textbf{intersection size} of open intervals $I_1$ and $I_2$ is
	\begin{align*}
		\cI(I_1, I_2) = \left\{
		\begin{aligned}
			 & 0             &  & \text{if } I_1 \cap I_2 = \emptyset \\
			 & |I_1\cap I_2| &  & \text{otherwise.}
		\end{aligned}
		\right.
	\end{align*}
\end{definition}

\begin{definition}
	\label{def:union-size}
	The \textbf{union size} of open intervals $I_1$ and $I_2$ is
	$\cU(I_1, I_2) = |I_1| + |I_2| - \cI(I_1, I_2)$.
\end{definition}

\begin{definition}
	\label{def:structiou-intersection-over-union}
	The \textbf{intersection over union} (\iou) ratio between open intervals $I_1$ and $I_2$ is
	\begin{align*}
		\iou(I_1, I_2) = \frac{\cI(I_1, I_2)}{\cU(I_1, I_2)}.
	\end{align*}
\end{definition}
We will use \iou as the similarity metric between two intervals.

\subsection{Relaxed Segment Trees}
\label{sec:structiou-extended-segment-tree}
To represent parse trees, we relax the definition of a segment tree \citep{bentley1977solutions} as follows.

\begin{definition}
	\label{def:structiou-segment-tree-node}
	A \textbf{node} $\bn$ of a relaxed segment tree is a triple
	$\bn = \langle
		I_{\bn},
		C_{\bn},
		\ell_{\bn}
		\rangle$, where
	\begin{enumerate}
		\item $I_{\bn} = (s_{\bn}, e_{\bn})$ is an open interval (i.e., segment) associated with the node $\bn$, where $s_{\bn} < e_{\bn}$;
		\item $C_{\bn}$ is a finite set of disjoint children nodes of $\bn$: for any $\bp, \bq \in C_\bn (\bp \neq \bq)$, $I_\bp \cap I_\bq = \emptyset$.
		      $C_{\bn} = \emptyset$ if and only if $\bn$ is a terminal node;
		\item For a nonterminal node $\bn$, $s_{\bn} = \min_{\bp\in C_{\bn}} s_{\bp},$ and $e_{\bn} = \max_{\bp\in C_{\bn}} e_{\bp}$.
	\end{enumerate}
\end{definition}
\begin{corollary}
	\label{corollary:structiou-segment-tree-parent-cover}
	For nodes $\bp, \bn$, if $\bp \in C_\bn$, then $I_\bp \subseteq I_\bn$.
\end{corollary}
\begin{proof}
	According to the definition of open intervals and \cref{def:structiou-segment-tree-node} (3),
	\begin{align*}
		            & a_\bn \leq a_\bp < b_\bp \leq b_\bn                        \\
		\Rightarrow & I_\bp = (a_\bp, b_\bp) \subseteq (a_\bn, b_\bn) = I_{\bn}.
	\end{align*}
\end{proof}

\begin{definition}
	\label{def:structiou-ancestor}
	Node $\bp$ is an \textbf{ancestor} of node $\bq$ if there exists a sequence of nodes $\bn_0,$ $\bn_1,$ $\ldots,$ $\bn_k (k\geq 1)$ such that (i.) $\bn_0 = \bp$, (ii.) $\bn_k = \bq$, and (iii.) for any $i \in [k]$,\footnote{$[k]=\{1, 2, \ldots, k\}$, where $k \in \mathbb{N}$.} $\bn_{i} \in C_{\bn_{i-1}}$.
\end{definition}

\begin{corollary}
	\label{corollary:structiou-ancestor-cover}
	If node $\bp$ is an ancestor of node $\bq$, then $I_\bp \supseteq I_\bq$.
\end{corollary}
\begin{proof}
	According to Definition~\ref{def:structiou-ancestor}, there exists a sequence of nodes $\bn_0, \bn_1, \ldots, \bn_k (k\geq 1)$ such that (1) $\bn_0 = \bp$, (2) $\bn_k = \bq$ and (3) for any $i \in [k]$, $\bn_{i} \in \mathbf{C}_{\bn_{i-1}}$.

	Corollary~\ref{corollary:structiou-segment-tree-parent-cover} implies that for any $i \in [k]$, $I_{\bn_{i-1}} \supseteq I_{\bn_i} \Rightarrow I_{\bn_0}\supseteq I_{\bn_k}\Rightarrow I_\bp \supseteq I_\bq$.
\end{proof}

\begin{definition}
	\label{def:structiou-descendant}
	Node $\bp$ is a \textbf{descendant} of node $\bq$ if $\bq$ is an ancestor of $\bp$.
\end{definition}

\begin{definition}
	\label{def:structiou-ex-segment-tree}
	A \textbf{relaxed segment tree} $\cT = \langle\br_\cT, N_\cT\rangle$ is a tuple, where
	\begin{enumerate}
		\item $\br_\cT$ is the root node of $\cT$;
		\item $N_\cT = \{\br_\cT\} \cup \{\bn: \bn \text{ is a descendant of } \br_\cT\}$ is a finite set of all nodes in $\cT$.
	\end{enumerate}
\end{definition}
\begin{example}
	A relaxed segment tree can represent a constituency parse tree over spoken word time ranges (\cref{fig:structiou-example-speech-parse-tree-1}).
\end{example}

\begin{corollary}
	\label{corollary:structiou-segment-tree-characterization}
	A relaxed segment tree can be uniquely characterized by its root node.
\end{corollary}
\begin{proof}
	$(\Rightarrow)$ Definition~\ref{def:structiou-ex-segment-tree} implies that each relaxed segment tree has one root node.

	$(\Leftarrow)$ Given a specific node $\bn$, we have the unique set \\
	$\mathcal{N} = \{\bn\} \cup \{\bn': \bn' \text{ is a descendant of } \bn\}$, and therefore extract the set of all nodes in the relaxed segment tree rooted at $\bn$.
\end{proof}
\noindent In the following content, we use $\cT(\bn)$ to denote the relaxed segment tree rooted at $\bn$.

\begin{proposition}
	\label{proposition:structiou-non-ancestor-disjoint}
	For a relaxed segment tree $\cT$ and $\bp, \bq \in N_\cT$,  $\bp$ is neither an ancestor nor a descendant of $\bq$ $\Leftrightarrow$ $I_\bp \cap I_\bq = \emptyset$.
\end{proposition}
\begin{proof}
	($\Rightarrow$) Let $\bz$ denote the least common ancestor of $\bp$ and $\bq$. There exists $\bp', \bq' \in C_\bz (\bp' \neq \bq')$ such that $I_{\bp'} \supseteq I_\bp$ and $I_{\bq'} \supseteq I_\bq$; therefore
	\begin{align*}
		I_p \cap I_q \subseteq I_{\bp'} \cap I_{\bq'}
		\overset{\text{Definition~\ref{def:structiou-segment-tree-node} (2)}}{=}\emptyset
		\Rightarrow I_\bp \cap I_\bq = \emptyset.
	\end{align*}
	\\
	($\Leftarrow$) If $I_{\bp} \cap I_{\bq} = \emptyset$, according to \cref{def:structiou-segment-tree-node} (3) and \cref{def:structiou-ancestor}, $\bp$ is not an ancestor of $\bq$ and vice versa.
\end{proof}

\section{The \structiou Metric}
\label{sec:structiou-structaiou}
We present the definition of the structured average \iou (\structiou) metric in \cref{sec:structiou-problem-formulation}, which measures the similarity between two relaxed segment trees.
We also introduce a polynomial-time algorithm to compute the \structiou metric in \cref{sec:structiou-solution}.
\subsection{Problem Formulation}
\label{sec:structiou-problem-formulation}

Given relaxed segment trees $\cT_1$ and $\cT_2$ with node sets $N_{\cT_1} = \{\bn_{1, i}\}_{i=1}^{|N_{\cT_1}|}$ and $N_{\cT_2} = \{\bn_{2, j}\}_{j=1}^{|N_{\cT_2}|}$, we can align the trees by matching their same-label nodes.
Let $\bn_{1,i}\leftrightarrow \bn_{2,j}$ denote the matching between the nodes $\bn_{1,i}$ and $\bn_{2,j}$.

\begin{definition}[conflicted node matchings; \cref{fig:example-conflicted-node-matching}]
	\label{def:structiou-conflicted-node-matching}
	The matchings $\bn_{1, i} \leftrightarrow \bn_{2, j}$ and $\bn_{1, k} \leftrightarrow \bn_{2, \ell}$ are \textit{conflicted} if any of the following conditions holds:
	\begin{enumerate}
		\item $\bn_{1,i}$ is an ancestor of $\bn_{1,k}$, and $\bn_{2,j}$ is not an ancestor of $\bn_{2,\ell}$;
		\item $\bn_{1,i}$ is not an ancestor of $\bn_{1,k}$, and $\bn_{2,j}$ is an ancestor of $\bn_{2,\ell}$;
		\item $\bn_{1,i}$ is a descendant of $\bn_{1,k}$, and $\bn_{2,j}$ is not a descendant of $\bn_{2,\ell}$;
		\item $\bn_{1,i}$ is not a descendant of $\bn_{1,k}$, and $\bn_{2,j}$ is a descendant of $\bn_{2,\ell}$.
	\end{enumerate}
\end{definition}
\noindent Intuitively, we would like the alignment to be consistent with the ancestor-descendant relationship between nodes.

\input{figures/502-structiou-examples.tex}

The optimal (i.e., maximally \iou-weighted) structured alignment between $\cT_1$ and $\cT_2$ is given by the solution to the following problem:
\begin{problem}[maximally \iou-weighted alignment]
\label{prob:structiou-max-iou-weighted-tree-alignment}
\begin{align*}
	 & \bA^* = \arg\max_{\bA}
	\quad \sum_{i=1}^{|N_{\cT_1}|}\sum_{j=1}^{|N_{\cT_2}|} a_{i,j}
	\iou\left(
	I_{\bn_{1, i}},
	I_{\bn_{2, j}}
	\right)
\end{align*}
\begin{align}
	\suchthat
	 & \sum_{j} a_{i,j} \leq 1 (\forall i \in \left[\left|N_{\cT_1}\right|\right]) \label{eq:max-iou-weighted-tree-alignment-constraint1},                                     \\
	 & \sum_{i} a_{i,j} \leq 1 (\forall j \in \left[\left|N_{\cT_2}\right|\right]) \label{eq:max-iou-weighted-tree-alignment-constraint2},                                     \\
	 & a_{i,j} + a_{k,\ell} \leq 1 \quad \text{if } \bn_{1, i}\leftrightarrow \bn_{2,j} \text{ and }  \bn_{1, k}\leftrightarrow \bn_{2,\ell} \text{ are conflicted.} \nonumber
\end{align}
\end{problem}
\noindent $\bA\in \{0, 1\}^{|N_{\cT_1}|\times|N_{\cT_2}|}$ denotes an alignment matrix: $a_{i, j} = 1$ indicates that the matching $\bn_{1, i} \leftrightarrow \bn_{2, j}$ is selected, otherwise $a_{i, j} = 0$.
The last constraint of \cref{prob:structiou-max-iou-weighted-tree-alignment} ensures that no conflicted matchings are selected.
\cref{eq:max-iou-weighted-tree-alignment-constraint1,eq:max-iou-weighted-tree-alignment-constraint2} imply one-to-one matching between nodes; that is, in a valid tree alignment, each node in $\cT_1$ can be matched with at most one node in $\cT_2$, and vice versa.
The solution to \cref{prob:structiou-max-iou-weighted-tree-alignment} gives the maximal possible sum of \iou over aligned node pairs.

\begin{definition}
	\label{def:structiou-structa-iou}
	The \textbf{structured average} \textbf{\iou} (\structiou) between $\cT_1$ and $\cT_2$ is given by
	\begin{align*}
		\overline{\iou}(\cT_1, \cT_2) = \frac{1}{|N_{\cT_1}| + |N_{\cT_2}|} \sum_{i=1}^{|N_{\cT_1}|}\sum_{j=1}^{|N_{\cT_2}|} a^*_{i,j} \mathiou\left(I_{\bn_{1, i}}, I_{\bn_{2, j}}\right),
	\end{align*}
	where $\bA^* = \{a^*_{i,j}\}$ is the solution to Problem~\ref{prob:structiou-max-iou-weighted-tree-alignment}.
\end{definition}

\subsection{Solution}
\label{sec:structiou-solution}

\input{algorithms/501-p-solution.tex}

We present a polynomial-time algorithm for the exact solution to \cref{prob:structiou-max-iou-weighted-tree-alignment} by breaking it down into structured subproblems and solving them recursively with dynamic programming.

We define the subproblem as follows: given relaxed segment trees $\cT_1$ and $\cT_2$, we would like to find the maximum \iou weighted alignment of $\cT_1$ and $\cT_2$, where the roots of $\cT_1$ and $\cT_2$ are aligned.
Without loss of generality, we assume that the root nodes of $\cT_1$ and $\cT_2$ are both indexed by $1$.
Formally,
\begin{problem}[maximum \iou weighted alignment, with root nodes aligned]
\label{prob:structiou-root-aligned-max-iou-weighted-tree-alignment}
\begin{align*}
	f_{\cT_1, \cT_2} = \max_{\bm{A}} \sum_{i=1}^{|N_{\cT_1}|}\sum_{j=1}^{|N_{\cT_2}|} a_{i,j}
	\mathiou\left(
	I_{\bn_{1, i}},
	I_{\bn_{2, j}}
	\right)
\end{align*}

\vspace{-20pt}

\begin{align*}
	\suchthat ~
	 & a_{1, 1} = 1;                                                                                                                                              \\
	 & \sum_{j} a_{i,j} \leq 1 (\forall i \in \left[\left|N_{\cT_1}\right|\right]),                                                                               \\
	 & \sum_{i} a_{i,j} \leq 1 (\forall j \in \left[\left|N_{\cT_2}\right|\right]),                                                                               \\
	 & a_{i,j} + a_{k,\ell} \leq 1 \quad \text{if } \bn_{1, i}\leftrightarrow \bn_{2,j} \text{ and} \bn_{1,k}\leftrightarrow \bn_{2,\ell} \text{ are conflicted},
\end{align*}
where $\bA\in \{0, 1\}^{|\cT_1|\times|\cT_2|}$ is the alignment matrix.
\end{problem}
While \cref{prob:structiou-max-iou-weighted-tree-alignment,prob:structiou-root-aligned-max-iou-weighted-tree-alignment} are not equivalent in principle, \cref{prob:structiou-max-iou-weighted-tree-alignment} can be reduced to \cref{prob:structiou-root-aligned-max-iou-weighted-tree-alignment} within $\mathcal{O}(1)$ time, by adding a dummy root node to each tree that associates with segments covering all the segments in both trees.
We now present a polynomial-time solution to \cref{prob:structiou-root-aligned-max-iou-weighted-tree-alignment}.

\begin{definition}
	\label{def:structiou-ordered-disjoint-descendant-sequence}
	Given a node $\bn$ of a relaxed segment tree, $\mathbf{D} = \left(\bn_1, \bn_2, \ldots, \bn_k\right)$ is an \textbf{ordered disjoint descendant sequence} of $\bn$ if
	\begin{enumerate}
		\item (ordered) for any $i, j \in [k]$ and $i < j$, $s_{\bn_i} < s_{\bn_j}$, where $s_{\bn_i}$ and $s_{\bn_j}$ are left endpoint of the associated intervals;
		\item (disjoint) for any $i, j \in [k]$ and $i\neq j$,
		      $I_{n_i} \cap I_{n_j} = \emptyset$;
		\item (descendant) for any $i \in [k]$, $\bn_i$ is a descendant of $\bn$.
	\end{enumerate}
\end{definition}
\begin{corollary}
	\label{cor:structiou-ordered-disjoint-descendant-sequence}
	In an ordered disjoint descendant sequence $\mathbf{D} = \left(\bn_1, \bn_2, \ldots, \bn_k\right)$ of $\bn$, $e_{\bn_i} \leq s_{\bn_{i+1}}$ for any $i \in [k-1]$.
\end{corollary}
\begin{proof}
	If there exists $i \in [k-1]$ such that $b_{\bn_i} > a_{\bn_{i+1}}$, then
	\begin{align*}
		  & I_{\bn_i}\cap I_{\bn_{i+1}}                                                                                                                                     \\
		= & (a_{\bn_i}, b_{\bn_i}) \cap (a_{\bn_{i+1}}, b_{\bn_{i+1}})                                                                                                      \\
		= & \left\{x: \max(a_{\bn_i}, a_{\bn_{i+1}}) < x < \min(b_{\bn_i}, b_{\bn_{i+1}})\right\}                                                                           \\
		= & \left\{x: a_{\bn_{i+1}} < x < \min(b_{\bn_i}, b_{\bn_{i+1}})\right\}  \quad \textit{(Definition~\ref{def:structiou-ordered-disjoint-descendant-sequence} (1)).}
	\end{align*}
	Since $b_{\bn_{i+1}} > a_{\bn_{i+1}}$ (definition of open intervals), \begin{align*}
		 & a_{\bn_{i+1}} < \min(b_{\bn_i}, b_{\bn_{i+1}})
		\Rightarrow I_{\bn_i}\cap I_{\bn_{i+1}} \neq \emptyset.
	\end{align*}
	This conflicts with Definition~\ref{def:structiou-ordered-disjoint-descendant-sequence} (2).
\end{proof}

The solution to \cref{prob:structiou-root-aligned-max-iou-weighted-tree-alignment} is given by the following recursion:
\begin{align}
	f_{\cT_1, \cT_2} = \iou\left(I_{\br_{\cT_1}}, I_{\br_{\cT_2}}\right) +
	\max_{|\mathbf{D}_1|=|\mathbf{D}_2|}
	\sum_{i=1}^{|\mathbf{D}_1|} f_{\cT(\bd_{1, i}), \cT(\bd_{2, i})} \label{eq:structiou-subspan-alignments},
\end{align}
where $\br_{\cT_1}$ and $\br_{\cT_2}$ denote the root nodes of $\cT_1$ and $\cT_2$ respectively; $|\cdot|$ denotes the length of a sequence; $\mathbf{D}_1 = (\bd_{1, 1}, \bd_{1, 2}, \ldots, \bd_{1, |\mathbf{D}_1|})$ and $\mathbf{D}_2 = (\bd_{2, 1}, \bd_{2, 2}, \ldots, \bd_{2, |\mathbf{D}_2|})$ are same-length ordered disjoint descendant sequences of $\br_{\cT_1}$ and $\br_{\cT_2}$ respectively.
\cref{eq:structiou-subspan-alignments} can be computed within polynomial time by solving a knapsack-style problem with dynamic programming. Specifically, let
\begin{align*}
	g[\cT_1, \cT_2, & e_1, e_2] = \max_{|\mathbf{D}_1^{e_1}|=|\mathbf{D}_2^{e_2}|}
	\sum_{j=1}^{|\mathbf{D}_1^{e_1}|} f_{\cT(\bd_{1, j}^{e_1}), \cT(\bd_{2, j}^{e_2})},
\end{align*}
where $e_1$ and $e_2$ are arbitrary scalars denoting the constraints of endpoints; \\
$\mathbf{D}_1^{e_1} = \left(\bd_{1, 1}^{e_1}, \ldots, \bd_{1, |\mathbf{D}_1^{e_1}|}^{e_1}\right)$ is an ordered disjoint descendant sequence of $\br_{\cT_1}$, where for any $j \in [|\mathbf{D}_1^{e_1}|]$, the right endpoint of the corresponding node $e_{\bd_{1, j}^{e_1}} \leq e_1$;
similarly, $\mathbf{D}_2^{e_2} = \left(\bd_{2, 1}^{e_2}, \bd_{2, 2}^{e_2}, \ldots, \bd_{2, |\mathbf{D}_2^{e_2}|}^{e_2}\right)$ is a disjoint descendant sequence of $\br_{\cT_2}$ of which the right endpoint of each node does not exceed $e_2$.
\cref{algo:structiou-p-solution} computes $g$ and \cref{eq:structiou-subspan-alignments} within polynomial time, and therefore leads to a polynomial-time solution to \cref{prob:structiou-root-aligned-max-iou-weighted-tree-alignment}.

\paragraph{Complexity analysis.}
Suppose $|\cT_1| = n$ and $|\cT_2| = m$.
To compute $f_{\cT_1, \cT_2}$, all we need to compute is $g[\cT_1', \cT_2', e_1', e_2']$ for all $\cT_1', \cT_2'$, $e_1'$ and $e_2'$.
Here, $\cT_1'$ and $\cT_2'$ enumerate over all subtrees of $\cT_1$ and $\cT_2$, respectively, and $e_1'$ and $e_2'$ enumerate over the endpoints of all nodes in both trees, respectively.
The update process requires $\mathcal{O}(1)$ time for each $\cT_1, \cT_2, e_1, e_2$.
The edge cases, i.e., $g$ values of terminal nodes, can be directly computed in $\mathcal{O}(1)$ time, and therefore, the overall time complexity to solve \cref{prob:structiou-root-aligned-max-iou-weighted-tree-alignment} is $\mathcal{O}(n^2m^2)$.

\section{Experiments}
\label{sec:structiou-experiments}
We present two example applications of \structiou: speech constituency parsing evaluation (\cref{sec:structiou-speech-constituency-parsing-evalution}) and text constituency parsing evaluation (\cref{sec:structiou-text-constituency-parsing-evaluation-english}), where the former is our main focus.
In each part, we show the connection between \structiou and existing metrics in appropriate settings and present the unique features of \structiou.

\subsection{Speech Constituency Parsing Evaluation}
\label{sec:structiou-speech-constituency-parsing-evalution}

\subsubsection{Datasets and Setups}
\label{sec:structiou-datasets-and-setups}
We use the NXT-Switchboard \citep[NXT-SWBD;][]{calhoun-etal-2010-nxt} dataset to train and evaluate models, where the parser can access the forced alignment word boundaries in both training and testing stages.
We train an off-the-shelf supervised constituency parsing model for speech transcriptions \citep{jamshid-lou-johnson-2020-improving} on the training set of NXT-SWBD, do early-stopping using \parseval $F_1$ on the development set, and perform all the analysis below on the development set.
The model achieves $F_1=85.4$ and \structiou (averaged across sentences)\footnote{Unless otherwise specified, all \structiou scores reported in this dissertation are computed by averaging across \structiou scores of individual sentences. We compare and discuss sentence-level and corpus-level \structiou in \cref{sec:structiou-sentence-vs-corpus-structaiou}}$=0.954$ on the standard development set.

\input{figures/503-structiou-speech-comparison.tex}

\subsubsection{Comparison to the \parseval $F_1$ score}
\label{subsec:structiou-comparison-to-parseval-f1-speech}
Since the forced alignment word boundaries are accessible by the models, the \parseval $F_1$ metric can be directly calculated between the predicted speech constituency parse tree and the ground truth.
We compare the values of \structiou and \parseval (\citet{sekine-collins-1997-evalb} implementation with default parameters) in the settings with forced-alignment word segmentation (\cref{fig:structiou-structa-iou-vs-parseval-speech}), and find a strong correlation between the two metrics.

\subsubsection{Analysis: \structiou with Perturbed Word Boundaries}
\label{subsec:structiou-perturbation-speech}
In textless speech parsing \citep{lai2023audio,tseng-etal-2023-cascading}, the word boundaries are unknown, and the parser-predicted word boundaries are usually imperfect.
As a controlled simulation to such settings, we perturb the forced alignment word boundaries of the predicted parse tree (\cref{fig:structiou-perturbation-example}), and calculate the \structiou score between the perturbed parse tree and the ground truth over the original forced alignment word boundaries.
Specifically, we suppose the word boundaries of a sentence with $n$ words are $\mathcal{B}=b_0, b_1, \dots, b_n$,\footnote{We assume no silence between spoken words; if any inter-word silence exists, we remove it.} and consider the following types of perturbation with a hyperparameter $\delta \in [0, 1]$ controlling the perturbation level:
\begin{itemize}
	\item \textbf{Noise}-$\delta$. We start with $\mathcal{B}^{(0)} = \mathcal{B}$, and update the boundaries iteratively as follows.
	      For each $i \in [n-1]$, we randomly draw a number $r_i$ from the uniform distribution $U(-\delta, \delta)$, and let $b_i^{(i)} = b_i^{(i-1)} + |r_i| * \left(b_{i+\text{sgn}(r_i)}^{(i-1)} - b_i^{(i-1)}\right)$, where $\text{sgn}(\cdot): \mathbb{R}\rightarrow \{1, -1\}$ denotes the sign function
	      \begin{align*}
		      \text{sgn}(x) = \begin{cases}
			                      1  & \text{if } x \geq 0; \\
			                      -1 & \text{if } x < 0.
		                      \end{cases}
	      \end{align*}
	      For all $j\neq i$ and $j\in [n]$, we let $b_j^{(i)}$ remain the same as $b_j^{(i-1)}$.
	      Finally, we take $\mathcal{B}^{(n-1)}$ as the perturbed word boundaries for the predicted tree.
	\item \textbf{Insert}-$\delta$. We randomly draw a number $r_i$ from the uniform distribution for each boundary index $i\in[n]$.
	      If $r_i < \delta$, we insert a word boundary at the position $b'_i$, randomly drawn from the uniform distribution $U(b_{i-1}, b_i)$, breaking the $i^{\textit{th}}$ spoken word into two (i.e., $[b_{i-1}, b'_i]$ and $[b'_i, b_i]$).
	\item \textbf{Delete}-$\delta$. Similarly to the insertion-based perturbation, we randomly draw a number $r_i$ from the uniform distribution $U(0, 1)$ for each boundary index $i \in [n-1]$, and delete the boundary $b_i$ if $r_i < \delta$.
	      Since such boundary deletion may break the predicted tree structure, we use the base model (\cref{sec:structiou-datasets-and-setups}) to re-predict the parse tree with the new word boundaries, where words concatenated by space are taken as the textual input \citep{jamshid-lou-johnson-2020-improving}.
\end{itemize}
A larger $\delta$ means a higher level of perturbation is applied, and we, therefore, expect a lower \structiou score; $\delta=0$ means no perturbation is applied, and the \structiou score is the same as that for the predicted parse trees with forced alignment word boundaries.
\input{figures/504-structiou-speech-perturbation-example.tex}

For each $\delta \in \{0.1, 0.2, \ldots, 1.0\}$, starting from the base model (for deletion-based perturbation) or its predicted parse trees (for noise and insertion-based perturbation), we run the perturbation five times and report both the mean and the standard deviation of the \structiou result after perturbation.

\paragraph{Results and discussion.}
We present how the \structiou value changes with respect to $\delta$ for different types of perturbation (\cref{fig:structiou-perturbation-speech}).
The figure's standard deviation is nearly invisible, showing that our metric is stable under a specific setting.
As desired, a larger $\delta$ leads to a lower \structiou score for all three types of perturbation.
Among the perturbation types, \structiou is the most sensitive to deletion and the least sensitive to noise-based perturbation.
Although the results are not comparable across perturbation types in the most rigorous sense, this reflects that \structiou, to some extent, is more sensitive to structural change of the trees than simple word boundary changes.

Although both word boundary insertion and deletion change the predicted tree structures, the former has less impact on the \structiou scores.
This also aligns with our expectation: boundary insertion only splits some of the spoken words into two and keeps the longer constituents; however, deletion may significantly change the tree structure, especially when it happens at the boundary of two long constituents.
\input{figures/505-structiou-speech-perturbation-results.tex}

\subsubsection{Corpus-Level vs. Sentence-Level Metric}
\label{sec:structiou-sentence-vs-corpus-structaiou}
Note that 39.7\% utterances in the NXT-SWBD development set contain only one spoken word, and the \structiou score of such sentences is always high---the metric degenerates to the \iou score between two intervals.
Averaging the \structiou scores across all sentence pairs in the dataset may therefore overly emphasize these short utterances.
To address this, we introduce the corpus-level \structiou score as an alternative, where \cref{def:structiou-structa-iou} is modified as follows:
\begin{definition}
	\label{def:structiou-structa-iou-corpus}
	The corpus-level \structiou between two sets of parsed trees $\mathcal{D}_1 = \{\cT_{1,k}\}$ and $\mathcal{D}_2 = \{\cT_{2,k}\}$ is given by
	\begin{align*}
		\overline{\iou}(\mathcal{D}_1, \mathcal{D}_2)
		= & \frac{\sum_{k=1}^{|\mathcal{D}_1|} \left(|\cT_{1,k}| + |\cT_{2,k}|\right)\overline{\iou}(\cT_{1,k}, \cT_{2,k})}{\sum_{k'=1}^{|\mathcal{D}_1|} |\cT_{1,k'}| + |\cT_{2,k'}|},
	\end{align*}
	where $|\mathcal{D}_1| = |\mathcal{D}_2|$, and a pair of $\cT_{1,k}$ and $\cT_{2,k}$ denotes the parse trees of the $k^{\textit{th}}$ sentence in the corpus respectively.
\end{definition}
\input{figures/506-structiou-speech-structaiou-with-length.tex}

We compare the corpus-level and sentence-level \structiou scores (\cref{fig:structiou-with-length}).
As desired, the corpus-level \structiou score has lower absolute values than the sentence-level one, and the difference is more significant when longer sentences are considered.
A similar phenomenon has also been found in text constituency parsing \citep{kim-etal-2019-compound} where corpus-level \parseval $F_1$ scores are lower than sentence-level ones.

\subsection{Text Constituency Parsing Evaluation: English}
\label{sec:structiou-text-constituency-parsing-evaluation-english}
We extend our experiment to the evaluation of text constituency parsing.
In this subsection, we suppose every written word corresponds to a segment of the same unit length---analogously, this can be considered speech parsing with evenly distributed word boundaries for both predicted and ground-truth trees.

\subsubsection{Correlation with \parseval $F_1$ Scores on the Penn Treebank}
We use the Penn Treebank \citep[PTB;][]{marcus-etal-1993-building} dataset to train and evaluate Benepar \citep{kitaev-klein-2018-constituency}, a state-of-the-art text constituency parsing model, doing early-stopping using labeled \parseval $F_1$ on the development set.
The base model achieves \parseval $F_1=94.4$ and \structiou (averaged across sentences) $=0.962$ on the standard development set.

\input{figures/507-structiou-text-comparison.tex}
We compare the \structiou scores with the \parseval $F_1$ scores on the development set (\cref{fig:structiou-vs-parseval-text}).
As in the speech parsing experiment, we find a strong correlation between the two metrics, showing that \structiou is consistent with the existing metric in the text parsing domain.

\subsubsection{\structiou vs. \parseval $F_1$ on Syntactically Ambiguous Sentences}
\label{sec:syntactically-ambiguous-sentences}
\input{figures/508-structiou-syntactic-ambiguity.tex}
We consider a particular setting of parsing syntactically ambiguous sentences, where the syntactically plausible parse tree of a sentence may not be unique (see examples in \cref{fig:structiou-example-ambiguous-sentences}).
We simplify the case shown in \cref{fig:structiou-example-ambiguous-sentences} and generate synthetic sentences with syntactic ambiguity with the template \texttt{N (P N)\{{\it n}\}}, where \texttt{P} denotes a preposition and \texttt{N} denotes a noun, and \texttt{{\it n}} determines how many times the \texttt{P N} pattern is repeated.
For \texttt{N (P N)\{2\}}, the two potential parse trees are shown in \cref{fig:structiou-example-ambiguous-sentences-synthetic}.

We compare the \parseval and \structiou in the following scenarios, choosing a random syntactically plausible parse tree as the ground truth:
\begin{itemize}
	\item \textbf{Ground truth vs. random parse trees}, where the random parse trees are constructed by recursively combining random consecutive words (or word groups) into a binary tree. We construct 100 random parse trees and report the average.
	\item \textbf{Ground truth vs. syntactically plausible parse trees}, where we report the lowest possible score between the ground truth and other syntactically plausible trees.
\end{itemize}
As shown in \cref{tab:structiou-syntactic-ambiguity-results}, the lowest possible \parseval $F_1$ score between the ground truth and another syntactically plausible tree is significantly lower than the score achieved by meaningless random trees; however, \structiou consistently assigns higher scores to the syntactically plausible parses, showing more tolerance to syntactic ambiguity.
\input{tables/501-structiou-syntactic-ambiguity-results.tex}

\subsection{Text Constituency Parsing Evaluation: Hebrew}
We extend our experiment to the evaluation of text constituency parsing on Hebrew, a morphologically rich language that allows different tokenizations of the same sentence.
In this subsection, we suppose every written character (instead of a word for English; \cref{sec:structiou-text-constituency-parsing-evaluation-english}) corresponds to a segment of the same unit length.
We evaluate \structiou by running a pre-trained Hebrew constituency parsing model \citep[\texttt{benepar\_he2};][]{kitaev-etal-2019-multilingual} on the SPMRL 2013 Hebrew development set \citep{seddah-etal-2013-overview}.
Similarly to the English text parsing evaluation result (\cref{sec:structiou-text-constituency-parsing-evaluation-english}), we obtain a high Spearman rank correlation coefficient of 0.823 (over 10-sentence buckets) between \structiou (0.959 averaged across sentences) and \parseval F1 scores measured by EVALB-SPMRL (93.3).

%
\begin{figure}[t]
	\centering
	\begin{minipage}{0.45\textwidth}
		\vspace{-40pt}
		\textbf{Ground-truth Tree} \\
		\begin{forest}for tree={inner sep=0pt,l=10pt,l sep=10pt}
			[FRAGQ
				[\textcolor{cyan}{NP}
					[\textcolor{orange}{SYN\_WDT}				[\textcolor{blue}{DTT} [\<'yzh>]]]
					[\textcolor{magenta}{NP} [\textcolor{green}{SYN\_NN} [\textcolor{pink}{NN} [\<p.sw`ym>]]]]
				]
				[\textcolor{brown}{SYN\_yyQM} [\textcolor{olive}{yyQM} [?]]]
			]
		\end{forest}
	\end{minipage}
	\begin{minipage}{0.45\textwidth}
		\textbf{Predicted Tree} \\
		\begin{forest}for tree={inner sep=0pt,l=10pt,l sep=10pt}
			[SQ
				[\textcolor{cyan}{NP}
					[\textcolor{orange}{SYN\_WDT}				[
							\textcolor{blue}{DTT} [{\<'yzh>}]]
					]
					[\textcolor{magenta}{NP}
						[SYN\_NNT
							[BNT
									[\<p.sw`>]
							]
						]
						[NP
								[
									\textcolor{green}{SYN\_NN} [\textcolor{pink}{NN} [\<ym>] ]
								]
						]
					]
				]
				[\textcolor{brown}{SYN\_yyQM} [\textcolor{olive}{yyQM} [?]]]
			]
		\end{forest}
	\end{minipage}
	\caption[Parse trees of the Hebrew sentence ?\<'yzh p.sw`ym> with two possible morphological analyses.]{
		Parse trees of the Hebrew sentence ?\<'yzh p.sw`ym> with two possible morphological analyses (\structiou = 0.635). The plural morpheme \<yM> appears as a separate token in the right-hand side tree.
		Best viewed in color: the aligned nodes on both sides are shown in the same color.
	}
	\label{fig:hebrew-par-example}
\end{figure}
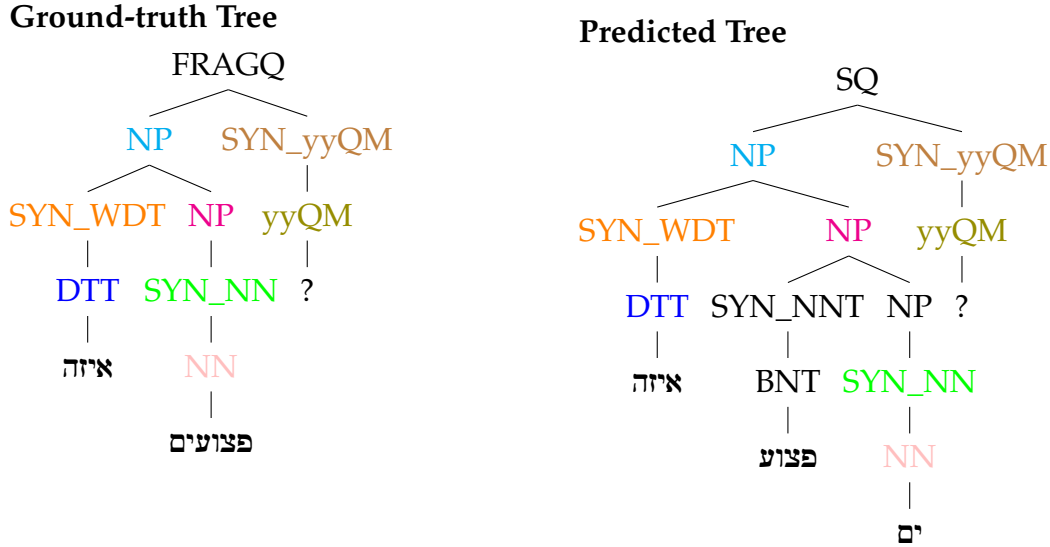

In addition, we demonstrate that \structiou naturally provides a metric that supports misaligned morphological analysis.
The default tokenization in the SPMRL dataset does not extract the plural morphemes \<wt> and \<yM>; therefore, simply extracting the plural morphemes forms another acceptable tokenization strategy (see \cref{fig:hebrew-par-example} for an example).
We break the nouns ending with these two morphemes and feed the new tokenization to the \texttt{benepar\_he2} model.\footnote{The \texttt{benepar\_he2} model is not trained on this tokenization; however, we expect the model still works well, since it uses XLM-R \citep{conneau-etal-2020-unsupervised} as the word embeddings, which provides syntactic information of the new tokenization. \label{footnote:benepar-he2-xlm-r}}
The prediction with our new tokenization receives a \structiou of 0.907 against the ground-truth---as desired, it is lower than 0.959 with the ground-truth tokenization.
However, the \structiou score remains high, reflecting the facts that (1) the manipulation introduces misalignment between parses, and (2) the Benepar model is fairly robust to such mismatch on tokenization (see \cref{footnote:benepar-he2-xlm-r}).
In contrast to \textsc{TedEval} \citep{tsarfaty-etal-2012-joint}, which aligns the tree structures by deleting and adding nodes \citep{bille-2005-survey} and takes into account these edits in the \parseval-based evaluation, \structiou provides an alternative way to evaluate the parsing quality under misaligned morphological analyses: instead of treating all the misaligned nodes as errors that result in the same penalization in the final metric, \structiou assigns partial credit to the aligned same-label nodes with $\iou > 0$.

\section{Conclusion and Discussion}
In this chapter, we present \structiou, the first metric that computes the similarity between two parse trees over continuous spoken word boundaries.
\structiou enables the evaluation of textless speech parsing \citep{lai2023audio,tseng-etal-2023-cascading}, where no text or speech recognizer is used or available to parse spoken utterances.

In the canonical text and speech parsing settings, \structiou complements the existing evaluation metrics \citep{black-etal-1991-procedure,roark-etal-2006-sparseval,tsarfaty-etal-2012-joint}.
Even for the evaluation of English constituency parsing, \structiou shows a higher tolerance to potential syntactic ambiguity under certain scenarios, providing an alternative interpretation of the parsing quality.

Faithful evaluation of parsing quality is crucial for developing both speech and text parsing models.
In supervised parsing, it has been common sense that higher evaluation metric scores (i.e.,  \parseval $F_1$) imply better models.
However, as we discussed in \citet{shi2020role}, the misalignment between linguistically annotated ground truths and model predictions, especially unsupervised parsing model predictions, does not necessarily indicate poor parsing quality of the models---instead, the models may have learned different but equally valid structures.
Conversely, annotations made by linguistic experts (such as the Penn Treebank) may exhibit discrepancies when compared to the responses of native speakers who lack formal linguistic training.
We suggest that future work investigate what properties of the parses are emphasized by each evaluation metric, and consider multi-dimensional evaluation metrics \interalia{kasai-etal-2022-bidimensional}.

%% file: figures/501-structiou-teaser.tex
\begin{figure}[t]
    \centering\small
    \begin{subfigure}[t]{\textwidth}
        \centering
        \begin{minipage}{0.23\textwidth}
            \centering
            \begin{forest} for tree={inner sep=0pt,l=10pt,l sep=10pt,s sep=5pt}
                [NP
                    [PRP [Your]]
                    [NN [turn]]
                ]
            \end{forest}
            \includegraphics[width=0.9\textwidth]{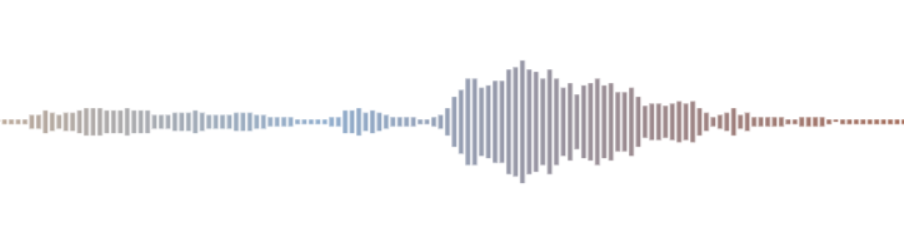}
        \end{minipage}
        \begin{minipage}{0.24\textwidth}
            \centering \normalsize
            {\small Forced \\[-2pt] Alignment}
            \vspace{-8pt}

            \flushleft
            \hspace{25pt}
            \begin{tikzpicture}
                \node[left] at (0,0) (Step 1) {};
                \draw [thick, ->] (Step 1) -- (1.2,0);
            \end{tikzpicture}
        \end{minipage}
        \begin{minipage}{0.23\textwidth}
            \centering
            \begin{forest} for tree={inner sep=0pt,l=10pt,l sep=6pt,s sep=5pt}
                [ \textcolor{magenta}{NP} \\ \textcolor{magenta}{(2.56--3.01)}
                    [ \textcolor{orange}{PRP} \\\textcolor{orange}{(2.56--2.72)}
                        [Your]
                    ]
                    [ \textcolor{cyan}{NN} \\\textcolor{cyan}{(2.72--3.01)}
                        [turn]
                    ]
                ]
            \end{forest}
        \end{minipage}
        \caption{
            \label{fig:structiou-example-speech-parse-tree-1}
            Ground-truth speech parse tree (right), obtained by forced alignment between the ground-truth text parse tree (left, top) and the spoken utterance (left, bottom).
        }
    \end{subfigure}
    \begin{subfigure}[t]{\textwidth}
        \centering \small
        \begin{minipage}{0.4\textwidth}
            \centering
            \begin{forest} for tree={inner sep=0pt,l=10pt,l sep=10pt,s sep=5pt}
                [ \textcolor{black}{VP} \\ (2.55--3.01)
                    [ VBP \\ (2.55--2.56) ]
                    [
                        \textcolor{magenta}{NP} \\ \textcolor{magenta}{(2.56-3.01)}
                        [ \textcolor{orange}{PRP} \\ \textcolor{orange}{(2.56--2.72)}]
                        [ \textcolor{cyan}{NN} \\ \textcolor{cyan}{(2.72--3.01)} ]
                    ]
                ]
            \end{forest}

            \vspace{5pt}
            (\structiou$=0.75$)
        \end{minipage}
        \begin{minipage}{0.4\textwidth}
            \centering
            \begin{forest} for tree={inner sep=0pt,l=10pt,l sep=10pt,s sep=5pt}
                [ \textcolor{magenta}{NP} \\ \textcolor{magenta}{(2.51--3.10)}
                    [ \textcolor{orange}{PRP}\\ \textcolor{orange}{(2.51--2.70)}]
                    [ \textcolor{cyan}{NN} \\ \textcolor{cyan}{(2.70--3.10)} ]
                ]
            \end{forest}

            \vspace{5pt}
            (\structiou$=0.81$)
        \end{minipage}
        \caption{
            \label{fig:structiou-example-speech-parse-tree-2}
            Predicted tree with good word boundaries and an errorful tree structure (left), or that with errorful word boundaries and a perfect tree structure (right).
        }
    \end{subfigure}

    \caption[Illustration of \structiou.]{
            \label{fig:structiou-example-speech-parse-tree}
            Illustration of how \structiou evaluates textless speech constituency parsing.
            Best viewed in color, where nodes with the same color are aligned.
            Numbers in parentheses are the starting and ending times of the corresponding spans (in seconds).
        }
\end{figure}

%% file: figures/502-structiou-examples.tex
\begin{figure*}[t]
    \centering
    \begin{minipage}[t]{0.15\textwidth}
        $\cT_1$:
        \begin{forest} for tree={inner sep=0pt,l=10pt,l sep=10pt,s sep=15pt}
            [A
                [B]
                [C
                        [D]
                        [E]
                ]
            ]
        \end{forest}
    \end{minipage}
    \begin{minipage}[t]{0.15\textwidth}
        $\cT_2$:
        \begin{forest} for tree={inner sep=0pt,l=10pt,l sep=10pt,s sep=15pt}
            [F
                [G]
                [H]
            ]
        \end{forest}
    \end{minipage}\hspace{0.05\textwidth}

    \begin{minipage}[t]{0.8\textwidth}
        \begin{align*}
            \text{A} \leftrightarrow \text{G} \text{ and } & \text{E} \leftrightarrow \text {H:} &  & \hspace{-23pt} \text{conflicted (violating rule 1)}       \\
            \text{C} \leftrightarrow \text{F} \text{ and } & \text{A} \leftrightarrow \text {H:} &  & \hspace{-23pt}\text{conflicted (violating rules 2 and 3)} \\
            \text{B} \leftrightarrow \text{H} \text{ and } & \text{C} \leftrightarrow \text {F:} &  & \hspace{-23pt}\text{conflicted (violating rule 4)}        \\
            \text{A} \leftrightarrow \text{F} \text{ and } & \text{C} \leftrightarrow \text {H:} &  & \hspace{-23pt}\text{not conflicted}
        \end{align*}
    \end{minipage}
    \caption[Examples of conflicted and non-conflicted node matchings.]{
        \label{fig:example-conflicted-node-matching}
        Examples of conflicted and non-conflicted node matchings (\cref{def:structiou-conflicted-node-matching}).
    }
\end{figure*}

%% file: algorithms/501-p-solution.tex
\begin{algorithm}[t!]
    \SetAlgoLined
    \SetKwInOut{Input}{Output}
    \KwIn{$\cT_1, \cT_2$}
    \KwOut{\cref{eq:structiou-subspan-alignments} $ = g[\cT_1, \cT_2, e_{\br_{\cT_1}}, e_{\br_{\cT_2}}]$}
    $g[\cT_1, \cT_2, x, y] \gets 0, \forall x, y$ \\
    $g'[\cT_1, \cT_2, x, y] := \max_{x' < x, y' < y} g[\cT_1, \cT_2, x', y'] $ \\
    $\bd_1 \gets$ sequence of descendants of $\br_{\cT_1}$, sorted in increasing order of right endpoint \\
    $\bd_2 \gets$ sequence of descendants of $\br_{\cT_2}$, sorted in increasing order of right endpoint \\
    \For{$i \gets 1 \ldots, |\bd_1|$}{
        \For{{$j \gets 1 \ldots, |\bd_2|$}}{
            $g[\cT_1, \cT_2, e_{\bd_{1, i}}, e_{\bd_{2, j}}] \gets \max\left(g[\cT_1, \cT_2, e_{\bd_{1, i}}, e_{\bd_{2, j}}], f_{\cT(\bd_{1, i}), \cT(\bd_{2, j})} + g'[\cT_1, \cT_2, s_{\bd_{1, i}}, s_{\bd_{2, j}}]\right)$ \\
            update $g'$ accordingly within $\mathcal{O}(1)$ time
        }
    }
    \KwRet{$g[\cT_1, \cT_2, e_{\br_{\cT_1}}, e_{\br_{\cT_2}}]$}
    \caption{Polynomial time solution to \cref{eq:structiou-subspan-alignments} \label{algo:structiou-p-solution}}
\end{algorithm}

%% file: figures/503-structiou-speech-comparison.tex
\begin{figure}[t]
    \centering
    \includegraphics[width=0.5\textwidth]{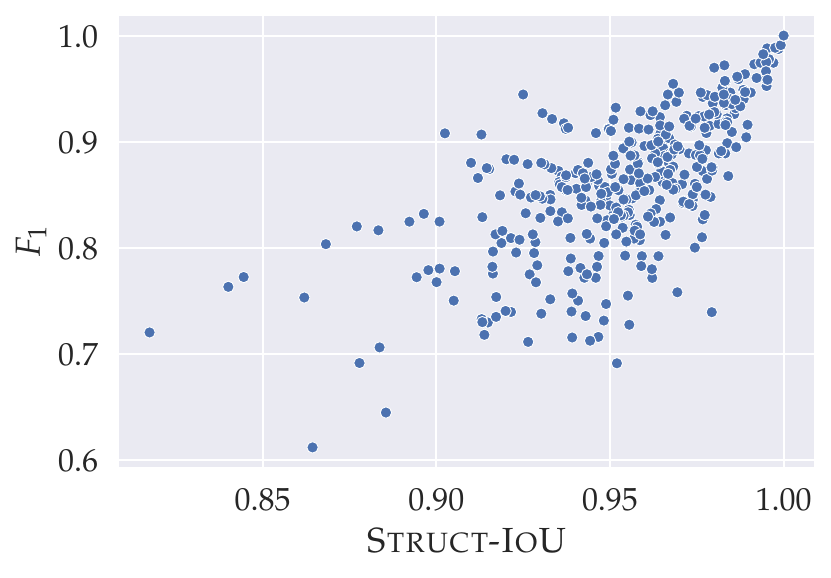}
    \caption[\structiou vs. \parseval $F_1$ on NXT-SWBD.]{\label{fig:structiou-structa-iou-vs-parseval-speech}
    \structiou vs. \parseval $F_1$ on NXT-SWBD (Spearman's correlation $\rho = 0.689$, p-value=$1.79\times 10^{-54}$). Each dot represents the results of the base model ($F_1$=85.4 on the full development set) on ten random examples from the development set.}
\end{figure}

%% file: figures/504-structiou-speech-perturbation-example.tex
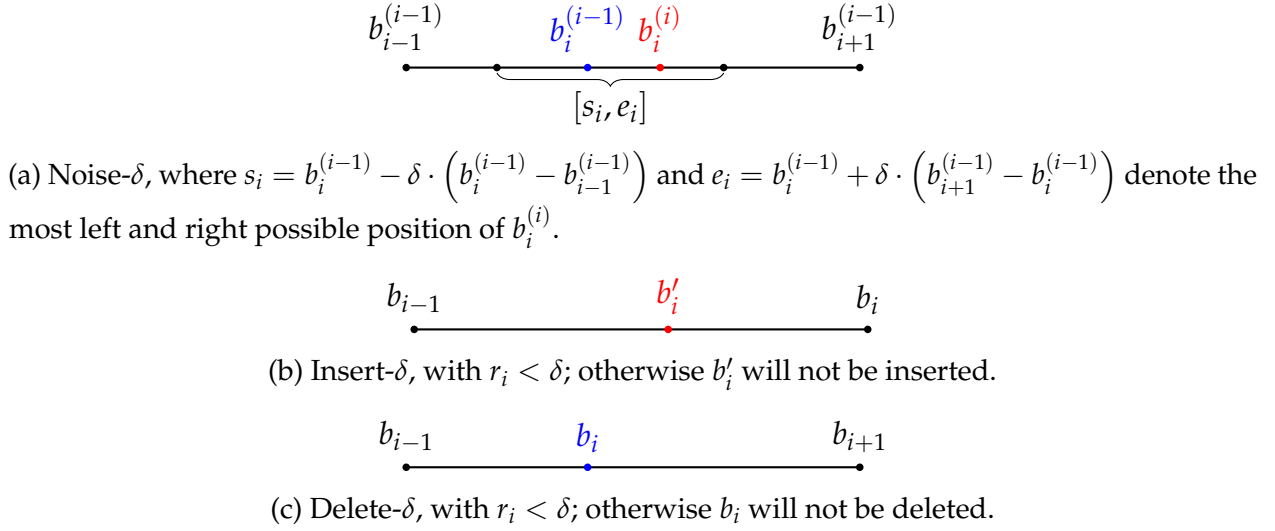
\begin{figure}[t]
    \centering
    \begin{subfigure}[t]{\textwidth}
        \centering
        \begin{tikzpicture}
            \draw[black,thick,] (0,0) -- (6,0)
            node[pos=0,structiounode,fill=black,label=above:{$b_{i-1}^{(i-1)}$}]{}
            node[pos=0.4,structiounode,fill=blue,label=above:{\textcolor{blue}{$b_i^{(i-1)}$}}]{}
            node[pos=1,structiounode,fill=black,label=above:{$b_{i+1}^{(i-1)}$}]{}
            node[pos=0.56,structiounode,fill=red,label=above:{\textcolor{red}{$b_{i}^{(i)}$}}]{}
            node[pos=0.7,structiounode,fill=black]{}
            node[pos=0.2,structiounode,fill=black]{}
            ;
            \coordinate (L) at (1.2, 0);
            \coordinate (R) at (4.2, 0);
            \draw[decorate,decoration={brace,amplitude=5pt,raise=2pt,mirror},yshift=0pt] (L) -- (R) node [midway, yshift=-15pt, xshift=0pt] {$[s_i, e_i]$};
        \end{tikzpicture}
        \caption{Noise-$\delta$, where $s_i = b_i^{(i-1)} - \delta \cdot \left(b_i^{(i-1)} - b_{i-1}^{(i-1)}\right)$ and $e_i = b_i^{(i-1)} + \delta \cdot \left(b_{i+1}^{(i-1)} - b_i^{(i-1)}\right)$ denote the most left and right possible position of $b_i^{(i)}$.}
        \label{subfig:structiou-noise-delta}
    \end{subfigure}
    \begin{subfigure}[t]{\textwidth}
        \centering \vspace{5pt}
        \begin{tikzpicture}
            \draw[black,thick,] (0,0) -- (6,0)
            node[pos=0,structiounode,fill=black,label=above:{$b_{i-1}$}]{}
            node[pos=1,structiounode,fill=black,label=above:{$b_i$}]{}
            node[pos=0.56,structiounode,fill=red,label=above:{\textcolor{red}{$b_i'$}}]{};
        \end{tikzpicture}
        \caption{Insert-$\delta$, with $r_i < \delta$; otherwise $b_i'$ will not be inserted.}
        \label{subfig:structiou-add-delta}
    \end{subfigure}
    \begin{subfigure}[t]{\textwidth}
        \centering \vspace{5pt}
        \begin{tikzpicture}
            \draw[black,thick,] (0,0) -- (6,0)
            node[pos=0,structiounode,fill=black,label=above:{$b_{i-1}$}]{}
            node[pos=1,structiounode,fill=black,label=above:{$b_{i+1}$}]{}
            node[pos=0.4,structiounode,fill=blue,label=above:{\textcolor{blue}{$b_i$}}]{};
        \end{tikzpicture}
        \caption{Delete-$\delta$, with $r_i < \delta$; otherwise $b_i$ will not be deleted.}
        \label{subfig:delete-delta}
    \end{subfigure}
    \caption[Examples of perturbation.]{
        \label{fig:structiou-perturbation-example}
        Examples of three types of perturbation: when applicable, the added boundaries are shown in red, and the deleted boundaries are shown in blue.
        Best viewed in color.
    }
\end{figure}

%% file: figures/505-structiou-speech-perturbation-results.tex
\begin{figure}[t]
    \centering
    \includegraphics[width=0.5\textwidth]{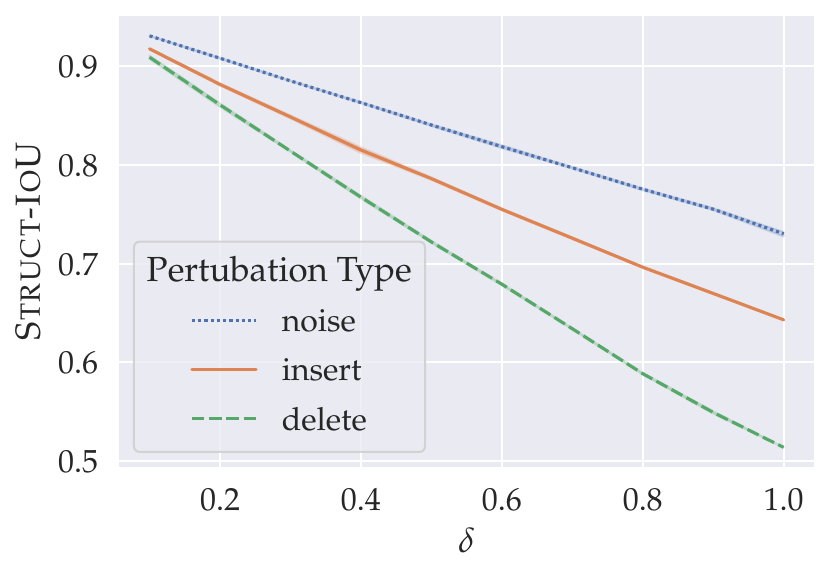}
    \caption{
        \label{fig:structiou-perturbation-speech} \structiou scores with respect to $\delta$ for different types of perturbations.
    }
\end{figure}

%% file: figures/506-structiou-speech-structaiou-with-length.tex
\begin{figure}[t]
    \centering
    \includegraphics[width=0.5\textwidth]{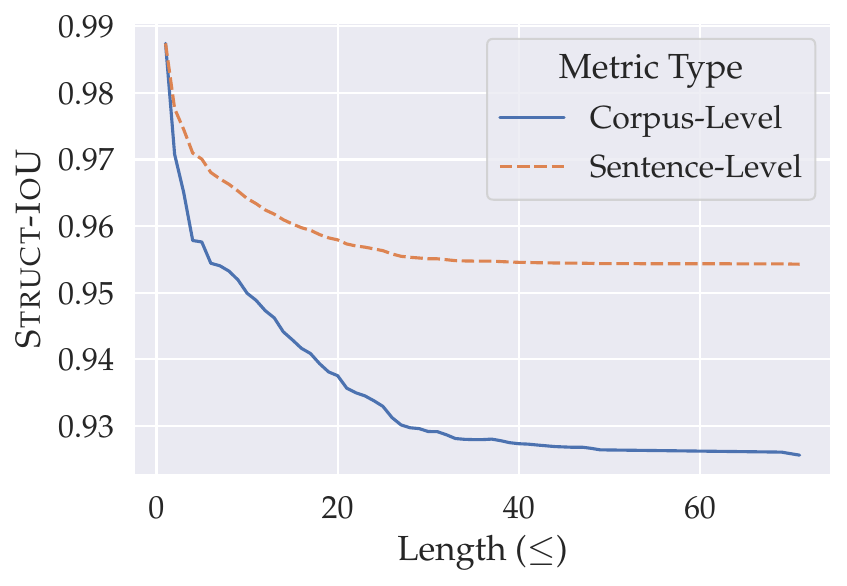}
    \caption[Corpus-level vs. sentence-level \structiou scores.]{
        \label{fig:structiou-with-length} Corpus-level and sentence-level \structiou scores of the predicted parse trees of the base model ($F_1 = 85.4$ on the development set), evaluated on development examples with less than or equal to a certain number of spoken words.
    }
\end{figure}

%% file: figures/507-structiou-text-comparison.tex
\begin{figure}[t]
    \centering
    \includegraphics[width=0.5\textwidth]{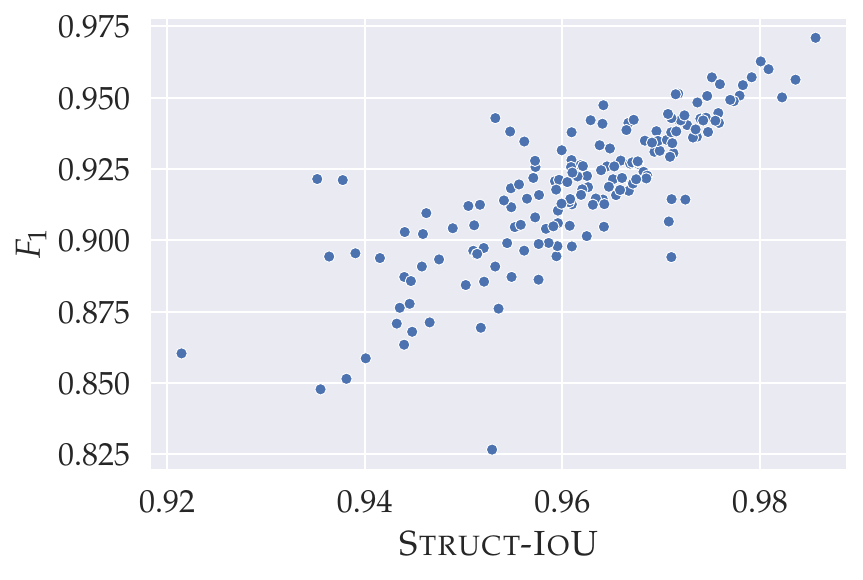}
    \caption[\structiou vs. \parseval $F_1$ on PTB.]{\label{fig:structiou-vs-parseval-text}
    Comparison of \structiou and \parseval $F_1$ (Spearman's rank correlation $\rho = 0.821$, p-value=$8.16\times 10^{-43}$). Each dot represents the results of the base model on 10 random examples from the PTB development set.}
\end{figure}

%% file: figures/508-structiou-syntactic-ambiguity.tex
\begin{figure}[t!]
    \centering
    \begin{forest} for tree={inner sep=0pt,l=12pt,l sep=10pt,s sep=12pt}
        [S
            [NP
                    [DT [The]]
                    [NN [girl]]
            ]
            [VP
                    [VBD [saw]]
                    [NP
                            [DT [a]]
                            [NN [cat]]
                            [PP
                                    [P [with]]
                                    [NP
                                            [DT [a]]
                                            [NN [telescope]]
                                    ]
                            ]
                    ]
            ]
        ]
    \end{forest}

    \vspace{-20pt}

    \begin{forest} for tree={inner sep=0pt,l=12pt,l sep=10pt,s sep=12pt}
        [S
            [NP
                    [DT [The]]
                    [NN [girl]]
            ]
            [VP
                    [VBD [saw]]
                    [NP
                            [DT [a]]
                            [NN [cat]]
                    ]
                    [PP
                            [P [with]]
                            [NP
                                    [DT [a]]
                                    [NN [telescope]]
                            ]
                    ]
            ]
        ]
    \end{forest}
    \caption[Example of syntactically ambiguous sentence (in English).]{\label{fig:structiou-example-ambiguous-sentences} An example syntactically ambiguous sentence: \textit{The girl saw a cat with a telescope}.
        Both parses are syntactically valid, but the first one implies that \textit{a cat} was holding the telescope, whereas the second implies \textit{the girl} was using the telescope.}
\end{figure}
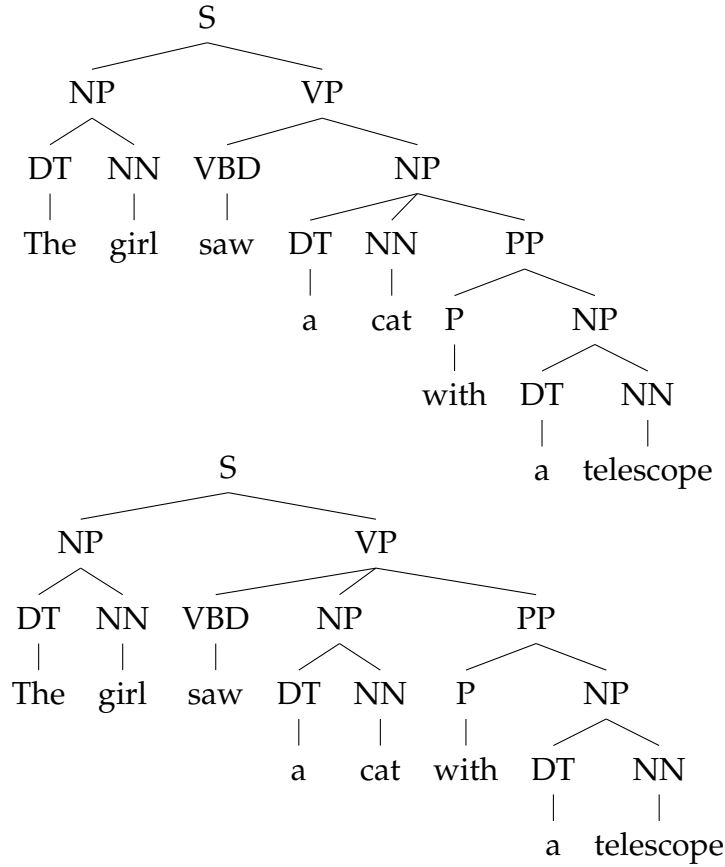

\begin{figure}[t]
    \centering
    \begin{forest} for tree={inner sep=0pt,l=10pt,l sep=10pt,s sep=15pt}
        [NP
            [NP [N]]
            [PP
                    [P]
                    [NP
                            [NP [N]]
                            [PP
                                    [P]
                                    [NP [N]]
                            ]
                    ]
            ]
        ]
    \end{forest}
    \hspace{10pt}
    \begin{forest} for tree={inner sep=0pt,l=10pt,l sep=10pt,s sep=15pt}
        [NP
            [NP
                    [NP [N]]
                    [PP
                            [P]
                            [NP
                                    [N]
                            ]
                    ]
            ]
            [PP
                    [P]
                    [NP [N]]
            ]
        ]
    \end{forest}
    \caption[Example of syntactically ambiguous sentence (synthetic).]{\label{fig:structiou-example-ambiguous-sentences-synthetic} Two syntactically plausible parses of the \texttt{N (P N)\{2\}}, where NP denotes a noun phrase, and PP denotes a prepositional phrase.}
\end{figure}
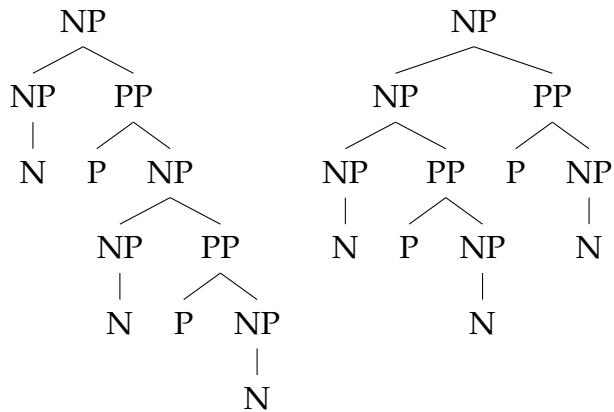

%% file: tables/501-structiou-syntactic-ambiguity-results.tex
\begin{table}[t!]
    \centering
    \begin{tabular}{lc}
        \toprule
        \bf Metric      & \bf Ground-Truth vs. Random, Average   \\
        \midrule
        \parseval $F_1$ & 27.3                                   \\
        \structiou      & 61.9                                   \\
        \midrule
                        & \bf Ground-Truth vs. Plausible, Lowest \\
        \midrule
        \parseval $F_1$ & 12.5                                   \\
        \structiou      & 63.6                                   \\
        \bottomrule
    \end{tabular}
    \caption[Comparison of \parseval $F_1$ and \structiou scores on syntactically plausible trees.]{
        \label{tab:structiou-syntactic-ambiguity-results}
        Average \parseval $F_1$ and \structiou scores between the ground truth and a random binary tree, and the lowest possible scores between the ground truth and another syntactically plausible tree.
        Experiments are done on the string ``\texttt{N \text{(}P N\text{)}}\{\texttt{8}\}''.
        For simplicity, we report the unlabeled scores, where all nonterminals are treated as having the same label.
    }
\end{table}

%% file: src/60-g2l2.tex
\chapter{Joint Syntax and Semantics Induction in Grounded Environments}
\label{chapter:g2l2}
\textit{Content in this chapter has been published as a conference paper at NeurIPS 2021 \citep{mao2021grammar}. Jiayuan Mao is the lead author of this work.}

We now start to consider program execution results as another source of grounding signals and consider using distant supervisions from both vision and program execution for grammar induction.
In this chapter, we present Grammar-Based Grounded Lexicon Learning (\gtlt), a lexicalist approach toward learning a compositional and grounded meaning representation of language from grounded data, such as paired images and texts.
At the core of \gtlt is a collection of lexicon entries, which map each word to a tuple of a syntactic type and a neuro-symbolic semantic program.
In \cref{fig:g2l2-teaser}, for example, the word {\it shiny} has a syntactic type of \emph{adjective}; its neuro-symbolic semantic program has the {\it symbolic} form $\lambda x.\textit{filter}(x, \textbf{SHINY})$, where the concept {\bf SHINY} is associated with a {\it neural network} embedding, which will be used to classify shiny objects.
Given an input sentence, \gtlt first looks up the lexicon entries associated with each token.
It then derives the meaning of the sentence as an executable neuro-symbolic program by composing lexical meanings based on syntax.
These programs can be executed on grounded inputs.
To facilitate learning in an exponentially growing compositional space, we introduce a joint parsing and expected execution algorithm (\ckyee), which does local marginalization over derivations to reduce the training time.
We evaluate \gtlt on two domains: visual reasoning and language-driven navigation.
Results show that \gtlt can generalize from small amounts of data to novel compositions of words.

\input{figures/601-g2l2-teaser.tex}

\section{Related Work}
\paragraph{Lexicalist theories.}
The lexicalist theories of syntax \citep{pollard-sag-1994-head,steedman-2000-syntactic,bresnan-etal-2016-lexical} argue that (1) the key syntactic principles by which words and phrases combine are extremely simple and general, and (2) nearly all of the complexity in syntax can be attributed to rich and detailed lexical entries for the words in the language.
For example, whereas the relationship between the active and passive voice, e.g., ``Kim saw a balloon'' versus ``A balloon was seen by Kim'', was treated in pre-lexicalist theories as a special syntactic rule converting between the sentences, in lexicalist theories, this relationship derives simply from the knowledge that the passive participle for the verb ``see'' is ``seen,'' which interacts with knowledge of other words to make both the active and passive forms of the sentence possible.
In lexicalist theories, the problem for the language learner thus becomes a problem of learning the words in the language, not a problem of learning numerous abstract rule schemas.
The combinatory categorial grammar \citep[CCG; ][]{steedman-2000-syntactic} framework we use is a well-established example of a lexicalist theory: there is a universal inventory of just three combinatory rules (\cref{fig:g2l2-teaser}a), but those rules can only be applied once richly specified lexical entries are learned for the words in a sentence.
We believe that this lexicalist-theory approach is a particularly good fit for the problem of grounded language learning: the visual context provides clues to the meaning of a word, and the grammatical behavior of the word is tied closely to this meaning.
This linking makes learning efficient.

\paragraph{Compositional generalization in NLP.}
Improving the compositional generalization of NLP systems has drawn great attention in recent years \citep{baroni-2020-linguistic}.
Most recent approaches towards this goal are built on deep learning-based models.
There are two representative approaches: building structured neural networks with explicit phrase-based structures or segments \citep{socher-etal-2013-recursive,zhu-etal-2015-long,tai-etal-2015-improved,shi2018tree,saqur-narasimhan-2020-multimodal}; and using data augmentation techniques \citep{andreas-2020-good,guo-etal-2020-sequence,akyurek-etal-2020-learning}.
However, these approaches either rely on additional annotation or pre-trained models for phrase structure inference or require domain-specific heuristics in data augmentation.
In contrast to both approaches, we propose to use combinatory grammar rules to constrain the learning of word meanings and how they are composed.

\paragraph{Neural latent trees.}
\ckyee is related to recent work using CKY-style modules for inducing latent trees.
However, our model is fundamentally different from work on unsupervised constituency parsing \interalia{kim-etal-2019-compound,shi-etal-2021-learning}, which use the CKY algorithm for inference over scalar span scores and those compute span representation vectors with CKY-style algorithms \interalia{maillard-clark-2018-latent,drozdov-etal-2019-unsupervised}.
Our key contribution is introducing the expected execution mechanism, where each span is associated with weighted, compressed programs.
Beyond enumerating all possible parsing trees as in \citep{maillard-clark-2018-latent}, G2L2 considers all possible programs associated with each span.
Our expected execution procedure works for different types (object set, integer, etc.) and even functor types, which makes our approximation exact for linear cases with polynomial complexity.

\paragraph{Grammar-based grounded language learning.}
There have also been work that learns grammatical structures from grounded texts \citep{artzi-zettlemoyer-2013-weakly,shi2019visually,zhao-titov-2020-visually,jin-schuler-2020-grounded}.
However, these approaches either rely on pre-defined lexicon entries \citep{artzi-zettlemoyer-2013-weakly} or only focus on inducing syntactic structures such as phrase-structure grammar \interalia{shi2019visually}.
Different from them, \gtlt jointly learns the syntactic types, semantic programs, and concept grounding, only based on a small set of combinatory grammar rules.

Grammar-based and grounded language learning has also been studied in linguistics, with related work to ours studying how humans use grammar as constraints in learning meaning \citep{steedman-2000-syntactic} and how learning syntactic rules and semantic meanings in language bootstrap each other \citep{abend-etal-2017-bootstrapping,taylor-gelman-1988-adjectives}.
However, most previous computational models have focused only on explaining small-scale lab experiments and do not address grounding in visual perception \citep{fazly2010probabilistic,gauthier2018word}.
In contrast, \gtlt is a neuro-symbolic model that integrates the combinatory categorial grammar formalism \citep{steedman-2000-syntactic} with joint perceptual learning and concept learning to learn meanings from images and texts directly.

\paragraph{Neuro-symbolic models for language grounding.}
Integrating symbolic structures such as programs and neural networks has shown success in modeling compositional queries in various domains, including image and video reasoning \citep{Hu2017Learning,Mascharka2018Transparency}, knowledge base query \citep{Andreas2016Learning}, and robotic planning \citep{andreas2017modular}.
This work uses symbolic domain-specific languages with neural network embeddings for visual reasoning in images and navigation sequence generation, following NS-CL \citep{mao2019neurosymbolic}.
However, in contrast to using a neural network--based semantic parser as in the aforementioned work, our model \gtlt focuses on learning grammar-based lexicon for compositional generalization in linguistic structures, such as novel word composition.

\section{Grammar-Based Grounded Lexicon Learning}
Our framework, \gtlt, learns grounded lexicons from cross-modal data, such as paired images and texts.
This section will use the visual reasoning task, specifically visual question answering (VQA), as the example.
However, the idea itself can be applied to other tasks and domains, such as image captioning and language-driven navigation.

\gtlt learns from a collection of VQA data tuples containing an image, a question, and an answer to the question.
In \gtlt, each word type $w$ is associated with one or multiple lexical entries comprising their syntactic types and semantic programs.
Given the input question, \gtlt first looks up the lexicon entries associated with each token in the sentence (\cref{fig:g2l2-model}I).
\gtlt then uses a chart parsing algorithm to derive the programmatic meaning representation of the entire sentence by recursively composing meanings based on syntax (\cref{fig:g2l2-model}II).
To answer the question, we execute the program on the image representation (\cref{fig:g2l2-model}III).
During training, we compare the answer derived from the model with the groundtruth answer to form the supervision for the entire system.
No additional supervision is needed, such as lexicon entries for certain words or concept labels.

\input{figures/602-g2l2-chart-parsing.tex}

\subsection{Grounded Lexicon}

\input{figures/603-g2l2-entry.tex}

At a high level, \gtlt follows the combinatory categorical grammar \citep[CCG; ][]{steedman-2000-syntactic} formalism to maintain lexicon entries and parse sentences.
Illustrated in \cref{fig:g2l2-entry}, each word $w$ (e.g., {\it shiny}) is associated with one or multiple entries.
Each entry $e_w^{(i)}$ is a tuple comprised of a syntax type $\textit{syn}_w^{(i)}$ (e.g., {\it objset/objset}), and a semantic meaning form $\textit{sem}_w^{(i)}$ (e.g., $\lambda x.\textit{filter}(x, \textbf{SHINY})$).
$\textit{sem}_w^{(i)}$ is a symbolic program represented in a typed domain-specific language (DSL) and can be executed on the input image.
Some programs contain concepts (in this case, \textbf{SHINY}) that can be visually grounded.

\paragraph{Typed domain-specific language.}
\gtlt uses a DSL to represent word meanings.
For the visual reasoning domain, we use the CLEVR DSL \citep{Johnson2017CLEVR}.
It contains object-level operations such as selecting all objects with a particular attribute (e.g., the shiny objects) or selecting all objects with a specific relationship with a certain object (e.g., the objects left of the cube).
It also supports functions that respond to user queries, such as counting the number of objects or querying a specific attribute (e.g., shape) of an object.
The language is typed: most functions take a set of objects or a single object as their inputs and produce another set of objects.
For example, the operation $\textit{filter}$ has the signature $\textit{filter}(\textit{objset}, \textit{concept}) \rightarrow \textit{objset}$ and returns all objects that have $\textit{concept}$ (e.g., all {\it shiny} objects) in the input set.

\paragraph{Syntactic types.} There are two types of syntactic types in \gtlt: primitive and complex.\footnote{In some domains we also use conjunctions ({\it CONJ}) in the coordination rule.} The primitive types are defined in the typed domain-specific language (e.g., {\it objset}, {\it int}).
A complex type, denoted as X/Y or X\textbackslash Y, is a functor type that takes an argument of type Y and returns an object of type X.
The direction of the slash indicates word order: for X/Y, the argument Y must appear on the right, whereas in X\textbackslash Y, it must appear on the left.
Note that X and Y can themselves be complex types, which allows us to define functor types with multiple arguments, such as (X\textbackslash Y)/Z, or even functors with functor arguments (e.g., (X\textbackslash Y)/(Z/Z)).

In \gtlt, the semantic type of a word (in the DSL), together with a set of directional and ordering settings for its arguments (that reflects how the word and its arguments should be linearized in text), uniquely determines the syntactic type of a word.
For example, the syntactic type for word {\it shiny} is {\it objset/objset}.
It first states that {\it shiny} acts as a function in meaning composition, which takes a subprogram that outputs a set of objects (e.g., $\textit{filter}(\textbf{CUBE})$) as its argument, and produces a new program whose output is also a set of objects, in this case,  $\textit{filter}(\textit{filter}(\textbf{CUBE}), \textbf{SHINY})$ Second, it states the direction of the argument, which should come from its right.

\paragraph{Neuro-symbolic programs.}
Some DSL functions involve concepts that are grounded in other modalities, such as the visual appearance and spatial relationships of objects.
Taking the function $\textit{filter}$ as an example: its secondary argument $\textit{concept}$ should be associated with the visual representation of objects.
In \gtlt, the meaning of each lexicon entry may involve one more constants (called ``concepts'') that are grounded on other modalities, possibly via deep neural embeddings.
In the case of {\it shiny}: $\lambda x.\textit{filter}(x, \textbf{SHINY})$.
The concept $\textbf{SHINY}$ is associated with a vector embedding in a joint visual-semantic embedding space, following \citet{kiros2014unifying}.
During program execution, we will be comparing the embedding of concept $\textbf{SHINY}$ with object embeddings extracted from the input image to filter out all {\it shiny} objects.

\paragraph{Lexicon learning.} \gtlt learns lexicon entries in the following three steps.
(i) First, we enumerate all possible semantic meaning programs derived from the DSL.
For example, in the visual reasoning domain, a candidate program is $\lambda x.\textit{filter}(x, \textbf{?})$, where $\textbf{?}$ denotes a concept argument.
When we try to associate this lexicon entry to the word {\it shiny}, the program is instantiated as $\lambda x.\textit{filter}(x, \textbf{SHINY})$, where \textbf{SHINY} is a new concept associated with a vector embedding.
Typically, we set a maximum number of arguments for each program and constrain its depth.
(ii) Next, for programs that have a primitive type, we use
its semantic type as the syntactic type (e.g., {\it objset}).
For programs that function with arguments, we enumerate possible arguments in the ordering of the arguments.
For example, the program $\lambda x.\textit{filter}(x, \textbf{SHINY})$ has two candidate syntactic types: {\it objset/objset} (the argument is on its right in language) and {\it objset\textbackslash objset} (the argument is on its left).
(iii) Finally, we associate each candidate lexicon entry with a learnable scalar weight $\tau(\cdot)$.
It is typical for a single word to have tens or hundreds of candidate entries, and we optimize these lexicon entry weights in the training process.
In practice, we assume no lexical ambiguity, i.e., {\it each word type has only one lexical entry}.
Thus, the ambiguity of parsing only comes from different syntactic derivation orders for the same lexical entries.
This also allows us to prune lexicon entries that do not lead to successful derivations during training.

\subsection{Program Execution}
Any fully grounded programs (i.e., programs without unbound arguments) can be executed based on the image representation.
We implement the Neuro-Symbolic Concept Learner \citep[NS-CL;][]{mao2019neurosymbolic} as our differentiable program executor, which consists of a collection of deterministic functional modules to realize the operations in the DSL.
NS-CL represents execution results in a ``soft'' manner: in the visual reasoning domain, a set of objects is represented as a vector mask $m$ of length $N$, where $N$ is the number of objects in the scene.
Each element, $m_i$ can be interpreted as the probability that object $i$ is in the set.
For example, the operation $\lambda x.\textit{filter}(x, \textbf{SHINY})$ receives an input mask $m$ and produces a mask $m'$ that selects all shiny objects in the input set.
The computation has two steps: (i) compare the vector embedding of concept $\textbf{SHINY}$ with all objects in the scene to obtain a mask $m^{(\textbf{SHINY})}$, denoting the probability of each object being {\it shiny}; (ii) compute the element-wise multiplication $m' = m \odot m^{\textbf{SHINY}}$, which can be further used as the input to other functions.
In NS-CL, the execution result of any program is fully differentiable with respect to the input image representation and concept embeddings (e.g., $\textbf{SHINY}$).

\input{figures/604-g2l2-ckyee.tex}
\input{algorithms/601-ckyee.tex}

\subsection{Joint Chart Parsing and Expected Execution}
\gtlt extends a standard dynamic programming algorithm for chart parsing (i.e., the CKY algorithm \citep{kasami1966efficient,younger1967recognition,cocke1969programming}) to compose sentence meaning from lexical meaning forms, based on syntax.
Denote $w_i$ as the input word sequence.
$e_{i}^{j}$ the $j$-th lexicon entry associated with word $w_i$, and $\tau(e_{i}^{j})$ the corresponding weight.
Consider all possible derivation of the question $\{\textit{derivation}_k\}$, $k=1,2,\dots$.
We define the following context-free probability distribution of derivations as:
\begin{center}
    \[ p(\textit{derivation}_k) \propto \exp \left( \sum_{e \in \textit{derivation}_k} \tau(e) \right). \]
\end{center}
The probability is exponentially proportional to the total weights $\tau(e)$ of all lexicon entries $e \in \textit{derivation}_k$ used by the specific derivation.

A straightforward implementation to support joint learning of lexicon weights $\tau$ and neural modules (e.g., $\textit{filter}(x, \textbf{SHINY})$), is to simply execute all possible derivations on the input image, and compare the answer with the groundtruth.
However, the number of possible derivations grows exponentially as the question length, making such computation intractable.
For example, in SCAN \citep{Lake2018Generalization}, each word has 178 candidate lexicons, and the number of lexicon combinations of a sentence with five words will be $178^5 \approx 10^{11}$.
To address this issue, we introduce the idea of expected execution, which essentially computes the ``expected'' execution result of all possible derivations.
We further accelerate this process by taking local marginalization.

Our \ckyee algorithm is illustrated in \cref{alg:g2l2-ckyee}.
It processes all spans $[\textit{left}, \textit{right})$ sequentially ordered by their length.
The composition for derivations of $[\textit{left}, \textit{right})$ has two stages.
First, it enumerates possible split point $k$ and tries to combine the derivation of $[\textit{left}, k)$ and $[k, \textit{right})$.
This step is identical to the standard CKY parsing algorithm.
Next, suppose there are two derivations $x$ and $y$ of span $[i, j)$, whose program structures are identical except for subtrees that can be partially evaluated (i.e., do not contain any unbounded arguments).
In that case, we will compress these two derivations into one, by marginalizing the execution result for that subtree.

See the example from \cref{fig:g2l2-subtree}.
Two programs have an identical structure, except for the second argument to the outer-most {\it relate} operation.
However, these sub-trees, highlighted in gray, can be partially evaluated on the input image, and both of them output a vector of scores indicating the objects being selected.
Denote $\tau_1$ and $\tau_2$ as the weight associated with two derivations, and $v_1$ and $v_2$ the partial evaluation results (vectors) for two subtrees.
We will replace these two candidate meaning forms with $z$:
\[ z \coloneqq \lambda x.\textit{relate}(x, v', \textbf{RIGHT}),\;\;\text{where~} v'\coloneqq \frac{\exp(\tau_1) v_1 + \exp(\tau_2) v_2}{\exp(\tau_1) + \exp(\tau_2)}, \tau(z) \coloneqq \tau_1 + \tau_2.  \]

\paragraph{Complexity.}
Intuitively, once we have determined the semantics of a constituent in the question, the actual concrete meaning form of the derivation does not matter for future program execution, if the meaning form can already be partially evaluated on the input image.
This joint parsing and expected execution procedure significantly reduces the exponential space of possible parsing to a polynomial space with respect to the number of possible program layouts that can not be partially evaluated, which, in practice, is small.
The complexity of CKY-E2 is polynomial with respect to the length $L$ of the sentence, and $M$ is the number of candidate lexicon entries.
More specifically, $O(L^3 M)$, where $O(L^3)$ comes from the chart parsing algorithm, and the number of derivations after the expected execution procedure is $O(M)$.
This result is obtained by viewing the maximum arity for functor types as a constant (e.g., 2).
Intuitively, for each span, all possible derivations associated with this span can be grouped into four categories: derivations of a primitive type, derivations of a 1-ary functor type, derivations of a 2-ary functor type, and derivations of a 2-ary functor type, with one argument binded.
All these numbers grow linearly with respect to $M$.

\paragraph{Correctness.}
One can theoretically prove that if all operations in the program layout are commutative with the expectation operator, i.e., if $\mathbb{E}\left([f\left(x\right)\right] = f\left(\mathbb{E}\left[x\right]\right)$, our \ckyee produces exact computation of the expected execution result.
These operations include tensor addition, multiplication (if tensors are independent), and concatenation, which cover most of the computation we will do in neuro-symbolic program execution.
For example, for {\it filter}, taking the expectation over different inputs before doing the filtering is the same as taking the expectation over the filter results of different inputs.
However, there are operations such as quantifiers whose semantics are not commutative with the expectation operator.
In practice, it is possible to still use the expected expectation framework to approximate.
We leave the application of other approximated inference techniques as future work.

\subsection{Learning}
Our model, \gtlt, can be trained end-to-end by looking at images and reading paired questions and answers.
We denote $\ell$ as a loss function that compares the output of a program execution (e.g., a probability distribution over possible answers) and the groundtruth.
More precisely, given all possible derivations $\textit{derivation}_k$, the image representation $I$, the answer $A$, and the executor $\mathcal{E}( \cdot, I )$, we optimize all parameters by minimizing the loss $\mathcal L$:
\begin{equation*}
    \mathcal L = \sum_{k} \left( p(\textit{derivation}_k) \cdot \ell \left( \mathcal{E}( \textit{derivation}_k, I ) , A \right) \right).
\end{equation*}
In practice, we use gradient-based optimization for both the neural network weights in concept grounding modules and the lexicon weights $\tau$.

\section{Experiment}
\label{sec:g2l2-expr}

We evaluate \gtlt on two domains: visual reasoning in CLEVR \citep{Johnson2017CLEVR} and language-driven navigation in SCAN \citep{Lake2018Generalization}.
Beyond the grounding accuracy, we also evaluate the compositional generalizability and data efficiency, comparing \gtlt with end-to-end neural models and modular neural networks.

\subsection{Visual Reasoning}
We first evaluate \gtlt on the visual reasoning tasks in the CLEVR domain \citep{Johnson2017CLEVR}, where the task is to reason and answer questions about images.
Our study uses a subset of the CLEVR dataset, which does not include sentences that involve coreference resolution and words with multiple meanings in different contexts.

\subsubsection{Domain-Specific Language}
Our DSL is based on the CLEVR DSL introduced in \citet{Johnson2017CLEVR}, and the neuro-symbolic realization of each functional module is extended from the Neuro-Symbolic Concept Learner~\citep[NS-CL; ][]{mao2019neurosymbolic}.
We refer readers to the original papers for a detailed introduction to the DSL and neuro-symbolic program execution.
Here, we only highlight the key aspects of our language and its neuro-symbolic realization, and discuss the difference between our implementation and the ones in the original papers.

Our visual reasoning DSL is a subset of CLEVR, containing six types and eight primitive operations.
\cref{tab:clevr-typesystem} illustrates all six types and how they are internally represented in neuro-symbolic execution.

\begin{table}[t!]
    \centering
    \begin{tabular}{lp{0.3\columnwidth}p{0.48\columnwidth}} \toprule
        \bf Type   & \bf Note                      & \bf Representation                                                                                                                                                                                                 \\ \midrule
        ObjConcept & Object-level concepts         & An embedding vector.                                                                                                                                                                                               \\ \midrule
        Attribute  & Object-level attributes       & A vector of length $K_{\textit{obj}}$, where $K_{\textit{obj}}$ is the number of                                                                                                                                   \\ \midrule
        RelConcept & Relational concepts           & An embedding vector.                                                                                                                                                                                               \\ \midrule
        ObjectSet  & A set of objects in the scene & A vector $\mathbf{m}$ of length $N$, where $N$ is the number of objects in the scene. Each entry $\mathbf{m}_i$ is a real value in $[0, 1]$, can be interpreted as the probability that object $i$ is in this set. \\ \midrule
        Integer    & An integer                    & A single non-negative real value, can be interpreted as the ``expected'' value of this integer.                                                                                                                    \\ \midrule
        Bool       & A Boolean value               & A single real value in $[0, 1]$, can be interpreted as the probability that this Boolean value is true.                                                                                                            \\ \bottomrule
    \end{tabular}
    \caption{The type system of the domain-specific language for visual reasoning.}
    \label{tab:clevr-typesystem}
\end{table}

\begin{table}[t!]
    \centering
    \setlength{\tabcolsep}{3pt}
    \begin{tabular}{p{0.55\columnwidth}p{0.40\columnwidth}} \toprule
        Signature                                                                                                   & Note                                                                                                                              \\ \midrule
        {\it scene}() $\longrightarrow$ ObjectSet                                                                   & Return all objects in the scene.                                                                                                  \\ \midrule
        {\it filter}($\mathbf{a}$: ObjectSet, $c$: ObjConcept) $\longrightarrow$ ObjectSet                          & Filter out a set of objects having the object-level concept (e.g., red) from the input object set.                                \\ \midrule
        {\it relate}($\mathbf{a}$: ObjectSet, $\mathbf{b}$: ObjectSet, $c$: RelConcept) $\longrightarrow$ ObjectSet & Filter out a set of objects in set $\mathbf{a}$ that have the relational concept (e.g., left) with the input object $\mathbf{b}$. \\ \midrule
        {\it intersection}($\mathbf{a}$: ObjectSet, $\mathbf{b}$: ObjectSet) $\longrightarrow$ ObjectSet            & Return the intersection of set $\mathbf{a}$ and set $\mathbf{b}$.                                                                 \\ \midrule
        {\it union}($\mathbf{a}$: ObjectSet, $\mathbf{b}$: ObjectSet) $\longrightarrow$ ObjectSet                   & Return the union of set $\mathbf{a}$ and set  $\mathbf{b}$.                                                                       \\ \midrule
        {\it query}($\mathbf{a}$: ObjectSet, $c$: Attribute) $\longrightarrow$ ObjConcept                           & Query the attribute (e.g., color) of the input object $\mathbf{a}$.                                                               \\ \midrule
        {\it exist}($\mathbf{a}$: ObjectSet) $\longrightarrow$ Bool                                                 & Check if the set is empty.                                                                                                        \\ \midrule
        {\it count}($\mathbf{a}$: ObjectSet) $\longrightarrow$ Integer                                              & Count the number of objects in the input set.                                                                                     \\ \bottomrule
    \end{tabular}
    \caption{All operations in the domain-specific language for visual reasoning.}
    \label{tab:clevr-dsl}
\end{table}
\cref{tab:clevr-dsl} further shows all operations in the DSL.
There are two main differences between the DSL used by \gtlt and the original CLEVR DSL.
First, we have removed the \textit{unique} operation, whose semantic meaning was to return the single object in a set of objects. For example, it can be used to represent the meaning of the word ``\textit{the}'' in ``\textit{\underline{the} red object}'', in which the semantic program of ``\textit{red object}'' yields a set of red objects and the semantic program of ``\textit{the}'' selects the unique object in that set. However, the meaning of ``\textit{the}'' may have a slightly different semantic type in different contexts, for example, ``\textit{what is \underline{the} color of ...}''. Since this has violated our assumption about each word having only one lexicon entry, we choose to remove this operation to simplify the learning problem.
Meanwhile, to handle the ``uniqueness'' of the object being referred to, in our realization of related operations, such as \textit{relate} and \textit{query}, we will implicitly choose the unique object being referred to, which we will detail in the following paragraphs.

\paragraph{Object-centric scene representation.}
In our visual reasoning domain, we have assumed access to a pre-trained object-detector that generates a list of bounding boxes of objects in the scene.
In our implementation, following \citet{mao2019neurosymbolic}, we use a pre-trained Mask R-CNN \cite{He2017Mask} to generate bounding boxes for each object proposal.
These bounding boxes, paired with the original image, are then sent to a ResNet-34 \citep{he2016deep} to extract a region-based representation (by RoI Align) and image-based representation, respectively.
We concatenate them to form a vector embedding for each object in the image.

\paragraph{Neuro-symbolic realization.}
The high-level idea for the program execution is to build a collection of functions that realize the semantics of each operation based on the vector embeddings of objects and concepts.
Taking the \textit{filter} operation as an example, denote $\mathbf{a}$ as a vector representation of the input set, $o_i$ the object embeddings, and $e_c$ the concept embedding.
We compute the vector representation $\mathbf{b}$ of the output set as:
\[ \mathbf{b}_i = \mathbf{a_i} \cdot \sigma\left(\left\langle o_i, e_c \right\rangle\right), \]
where $\sigma$ is the sigmoid function, and $\langle \cdot, \cdot, \rangle$ is the inner product of two vectors.
Intuitively, we first compute the inner product between the concept embedding $e_c$ and each object embedding, which gives as a vector of scores of whether object $i$ has concept $c$.
Next, we compute the element-wise multiplication between two vectors.

A key difference between our realization of these operations and the one in \citet{mao2019neurosymbolic} is that we use element-wise multiplication to simulate the intersection between two sets, and $1 - (1 - \mathbf{a})(1 - \mathbf{b})$ for union. In contrast, \citet{mao2019neurosymbolic} use element-wise min operation for intersection and max for union.
Both realizations can be motivated by real-valued logic: product logic vs. G\"odel logic.

\paragraph{Example.}
Here, we run a concrete example to illustrate the execution process of a program in the visual reasoning domain. Suppose we have an image containing three objects $o_1$, $o_2$ and $o_3$.
We have two additional vector embeddings for concepts \textbf{SHINY} and \textbf{CUBE}.
Furthermore, $\sigmadot{o_i}{SHINY} = [0.1, 0.8, 0.9]$, and $\sigmadot{o_i}{CUBE} = [0.8, 0.1, 0.9]$.

Consider the input sentence ``{\it How many shiny cubes are there}''. \cref{tab:clevr-execution} illustrates a step-by-step execution of the underlying program: $$\textit{count}(\textit{filter}(\textit{filter}(\textit{scene}(), \textbf{CUBE}), \textbf{SHINY})).$$

\begin{table}[t]
    \centering
    \begin{tabular}{lll} \toprule
        Program                                                                                             & Type               & Value                                     \\ \midrule
        $\textit{scene}()$                                                                                  & ObjectSet          & $[1, 1, 1]$                               \\ \midrule
        $\textit{filter}(\textit{scene}(), \textbf{CUBE})$                                                  & ObjectSet          & $[0.8, 0.1, 0.9]$                         \\ \midrule
        \mycell{$\textit{filter}(\textit{filter}(\textit{scene}(), \textbf{CUBE}), \textbf{SHINY})$}        & \mycell{ObjectSet} & \mycell{$[0.08, 0.08, 0.81]$              \\=$[0.8, 0.1, 0.9] \odot [0.1, 0.8, 0.9]$} \\ \midrule
        $\textit{count}(\textit{filter}(\textit{filter}(\textit{scene}(), \textbf{CUBE}), \textbf{SHINY}))$ & Integer            & $0.97 = \textit{sum}([0.08, 0.08, 0.81])$ \\ \bottomrule
    \end{tabular}
    \caption[Execution trace of the program $\textit{count}(\textit{filter}(\textit{filter}(\textit{scene}(), \textbf{CUBE}), \textbf{SHINY}))$.]{Execution trace of the program $\textit{count}(\textit{filter}(\textit{filter}(\textit{scene}(), \textbf{CUBE}), \textbf{SHINY}))$. $\textit{sum}$ denotes the ``reduced sum'' operation of a vector, which returns the summation of all entries in that vector. $\odot$ denotes element-wise multiplication for two vectors.}
    \label{tab:clevr-execution}
\end{table}

\paragraph{Expected execution.}
In the visual reasoning domain, we have only implemented the expected execution mechanism for subordinate program trees whose type is {\it objset}, although many other types such as {\it integer} and {\it bool} also naturally support expected execution.
This is because types such as {\it integer} and {\it bool} only appear at the sentence level, and thus, computing the ``expectation'' of such programs does not reduce the overall complexity.

Formally, the expected execution process compresses a list of semantic programs, denoted by $v_1, v_2, \cdots, v_K$, and their corresponding weights $\tau(v_i)$ into a single semantic program $v^*$ with weight $\tau(v^*)$. Suppose all $v_i$'s have type {\it objset}. We use $\bar{v}_i$ to denote the execution result of these programs. Each of them is a vector of length $N$, where $N$ is the number of objects in the scene. We compute $\bar{v}^*$ and $\tau(v^*)$ as the following:
\begin{eqnarray*}
    \bar{v}^* & = & \frac{1}{\sum_i \exp(\tau(v_i))} \sum_i \left( \exp(\tau(v_i)) \cdot \bar{v}_i \right),\\
    \tau(v^*) & = & \log \sum_i \exp( \tau(v_i) ).
\end{eqnarray*}
Intuitively, we normalize the weights using a softmax function to translate them into a distribution, and then compute the expectation of the vectors.

\paragraph{Candidate lexicons.} Recall that the process of lexicon learning has three stages. First, we generate an extensive collection of candidate semantic programs. Second, we generate candidate lexicon entries for each word by enumerating all possible candidate semantic programs generated in the first step and all possible ordering (linearization in a sentence) of its arguments. Third, we apply our \ckyee and gradient-based optimization to update the weights associated with each lexicon entry.

In our visual reasoning domain, we only consider the following candidate semantic programs and linearizations:
\begin{enumerate}
    \item  Syntactic type: $\textit{objset}$, semantic program: $\textit{scene}()$ (English noun).
    \item  Syntactic type: $\textit{objset}$, semantic program: $\textit{filter}(\textit{scene}(), \textbf{?})$ (English noun).
    \item  Syntactic type: $\textit{objset}/\textit{objset}$, semantic program: $\lambda x.\textit{filter}(x, \textbf{?})$ (English adjective).
    \item  Syntactic type: $\textit{objset}\backslash\textit{objset}/\textit{objset}$, semantic program: $\lambda x.\lambda y.\textit{relate}(x, y, \textbf{?})$ (English preposition I).
    \item  Syntactic type: $\textit{objset}\backslash\textit{objset}/\textit{objset}$, semantic program: $\lambda x.\lambda y.\textit{relate}(y, x, \textbf{?})$ (English preposition II)
    \item  Syntactic type: $\textit{bool}/\textit{objset}$, semantic program: $\lambda x.\textit{exist}(x)$.
    \item  Syntactic type: $\textit{integer}/\textit{objset}$, $\lambda x.\textit{count}(x)$.
    \item  Syntactic type: $\textit{word}/\textit{objset}$, $\lambda x.\textit{query}(x, \textbf{?})$.
    \item  Syntactic type: $\text{CONJ}_{\text{AND}}$, $\lambda f.\lambda g.(\lambda x.\textit{intersect}(f(x), g(x)))$ (generalized conjunction).
    \item  Syntactic type: $\text{CONJ}_{\text{OR}}$, $\lambda x.\lambda y.(\lambda z.\textit{intersect}(z, \textit{union}(x, y)))$ (generalized disjunction).
\end{enumerate}

As we will see later, when we compare the candidate lexicon entries for the visual reasoning domain and the language-driven navigation domain, the visual reasoning domain contains significantly fewer entries than the navigation domain. This is because much of the learning process in this domain is associated with learning the concept embeddings. In the following few paragraphs, we will explain how we instantiate concepts based on these lexicon entry templates and implement generalized conjunction and disjunction in our domain.

First, for each word (more precisely, word type), e.g., {\it shiny}, we will instantiate ten lexicon entries. For semantic programs that contain unbounded concept arguments (\textbf{?} marks), we will introduce a series word-type-specific concepts. Specifically in this domain, each word type will be associated with 3 concept representations: $\textbf{SHINY}_{\text{ObjConcept}}$, $\textbf{SHINY}_{\text{RelConcept}}$, and $\textbf{SHINY}_{\text{Attribute}}$. Based on \cref{tab:clevr-dsl}, the first two concepts will be represented as two embedding vectors, and the third concept will be represented as a vector, indicating which concepts belong to this attribute category. Next, we will instantiate these lexicon entries by filling in these concept representations. For example, one of the candidate lexicon entry for {\it shiny} is syntactic type: $\textit{objset}$, semantic program: $\textit{filter}(\textit{scene}(), \textbf{SHINY}_{\text{ObjConcept}})$.
During training, all these vector embeddings and the weights associated with each lexicon entry, will be optimized jointly.

Next, we discuss the implementation for two conjunctive lexicon entries. The grammar rule for $\text{CONJ}_{\text{AND}}$ is:
\[T~~\text{CONJ}_{\text{AND}}~~T \rightarrow T,\]
where $T$ is an arbitrary syntactic type (thus called generalized conjunction).
There are two typical use cases: {what is the shape of the \underline{red and shiny} object}, and {what is the shape of the object that is \underline{left of the cube and right of the sphere}.} In the first case, both arguments have syntactic type $\textit{objset}/\textit{objset}$. In the second case, both arguments have syntactic type $\textit{objset}\backslash\textit{objset}$. Note that CLEVR contains only the second case.

The grammar rule for $\text{CONJ}_{\text{OR}}$ is:
\[\textit{objset}~~\text{CONJ}_{\text{OR}}~~\textit{objset} \rightarrow \textit{objset}\backslash\textit{objset}.\]
It covers the case: {\it how many objects are \underline{blue cubes or red spheres}.} Our implementation is slightly different with human-defined lexicon entries for the word {\it or}, in particular, because the DSL we use is a small set of set-theoretic operations, which does not fully match the expressiveness of truth-conditional semantics. Thus, the current DSL does not support the representation of all words in the dataset (in particular, {\it or} and {\it are}). Thus, we have implemented this ad-hoc fix to handle disjunction.

Finally, we would like to emphasize again that since our DSL does not support representing all semantic programs of words, we allow certain words to be associated with an ``{\it empty}'' lexicon entry. This entry can be combined with any words or constituents next to it and does not participate in the composition of syntactic types and semantic programs. In \cref{tab:clevr-lexicon-example}, we show the lexicon entry associated with each word in the sentence ``{\it are there any shiny cubes?}'', learned by our model, \gtlt.

\begin{table}[t!]
    \centering
    \setlength{\tabcolsep}{12pt}
    \begin{tabular}{lll} \toprule
        Word Type & Syntactic Type                    & Semantic Program                                                       \\ \midrule
        are       & <EMPTY>                           & <EMPTY>                                                                \\
        there     & <EMPTY>                           & <EMPTY>                                                                \\
        any       & $ \textit{bool}/\textit{objset}$  & $\lambda x.\textit{exist}(x)$                                          \\
        shiny     & $\textit{objset}/\textit{objset}$ & $\lambda x.\textit{filter}(x, \textbf{SHINY}_{\text{ObjConcept}})$     \\
        cubes     & $\textit{objset}$                 & $\textit{filter}(\textit{scene}(), \textbf{CUBE}_{\text{ObjConcept}})$ \\
        \bottomrule
    \end{tabular}
    \caption[The learned lexicon entries associated with each word for a simple sentence: {\it are there any shiny cubes?}]{The learned lexicon entries associated with each word for a simple sentence: {\it are there any shiny cubes?} The derived semantic program for the full sentence is $\textit{exist}(\textit{filter}(\textit{filter}(\textit{scene}(), \textbf{CUBE}_{\text{ObjConcept}}), \textbf{SHINY}_{\text{ObjConcept}}))$}
    \label{tab:clevr-lexicon-example}
\end{table}

\paragraph{Setup.} Instead of using manually defined heuristics for curriculum learning or self-paced learning as in previous work \citep{mao2019neurosymbolic,Li2020Competence}, we employ a curriculum learning setup that is simply based on sentence length: we gradually add longer sentences into the training set.
This helps the model to learn basic words from very short sentences (6 words), and use the acquired lexicon to facilitate learning longer sentences (20 words).
Since CLEVR does not provide test set annotations, for all models, we hold out 10\% of the training data for model development and test them on the CLEVR validation split.

\subsubsection{Baselines}
We compare \gtlt with 4 baselines.
(1) MAC \citep{Hudson2018Compositional} is an end-to-end approach based on attention.
(2) TbD-Net \citep{Mascharka2018Transparency} uses a pre-trained semantic parser to parse the question into a symbolic program and executes the program with a neural module network \citep{andreas2016neural}.
(3) Similarly, NS-VQA \citep{Yi2018Neuro} also parses the question into a symbolic program.
It also extracts an abstract scene representation with pre-trained neural recognition models \citep{He2017Mask}.
It executes the program based on the abstract scene representation.
Both approaches require additional supervision for training the semantic parser, and NS-VQA requires additional annotation for training the visual recognition model.
(4) NS-CL \citep{mao2019neurosymbolic} jointly learns a neural semantic parser and concept embeddings by looking at images and reading paired questions and answers.
It requires the annotation for all concepts in the domain (e.g., colors and shapes).
In contrast, \gtlt can {\it automatically} discover visual concepts from texts.

\subsubsection{Results}
\input{tables/601-g2l2-clevr.tex}

\cref{tab:g2l2-clevr} summarizes the results.
We consider any model that performs in the 95--100 range to have more or less solved the task.
Small differences in numeric scores in this range, such as the fact that NS-CL outperforms our model on the ``purple'' generalization task by 0.2\%, are less important than the fact that our model far outperforms all competitors on ``count'' compositional generalization and the ``depth'' generalization task, both of which all competitor models are far from solving.

We first compare different models on the {\bf standard} training-testing split.
We train different models with either 10\% or 100\% of the training data and evaluate them on the validation set.
Our model achieves a comparable performance in terms of its accuracy and data efficiency.

Next, we systematically build three {\bf compositional generalization} test splits: {\it purple}, {\it right of}, and {\it count}.
Essentially, we remove about 90\% of the sentences containing the word {\it purple}, the phrase {\it right}, and {\it counting operations}, such as {\it how many ...?} and {\it what number of ...?}.
We only keep sentences up to a certain length (6 for purple, 11 for right, and 8 for count).
We ensure that each use case of these words appears in training questions.
After training, we test these models on the validation set with questions containing these words.
Overall, our model \gtlt outperforms all baselines on all three generalization splits.
In particular, it significantly outperforms other methods on the {\it count} split.
The {\it count} split is hard for the baseline methods because this split requires models to generalize to sentences with deeper structures, for example, from ``{\it how many red objects are there?}'' to ``{\it how many red objects are right of the cube?}''
Note that, during training, all models have seen example uses of similar structures such as ``{\it what's the shape of the red object}'' and ``{\it what's the shape of the red object right of the cube?}''

Finally, we test generalization to sentences with deeper structures ({\bf depth}).
Specifically, we define the ``hop number'' of a question as the number of intermediate objects being referred to, in order to locate the target object.
For example, the ``hop number'' of the question ``{{\it how many red objects are right of the cube?}}'' is 1.
We train different models on 0-hop and 1-hop questions and test them on 2-hop questions.
Our model strongly outperforms all baselines.

The results on the {\bf compositional generalization} and {\bf depth} splits yield two conclusions.
First, disentangling grounded concept learning (associating words with visual appearances) and reasoning (e.g., filtering or counting subsets of objects in a given scene) improves data efficiency and generalization.
On CLEVR, neuro-symbolic approaches that separately identify concepts and perform explicit reasoning (NS-VQA, NS-CL, and \gtlt) consistently generalize better than approaches that do not (MAC, TbD).
The comparison between TbD and NS-VQA is informative: TbD fails on the “right of” task even in the case where the semantic parser is providing correct programs, while NS-VQA, which uses the same parser but explicitly represents compositional symbolic concepts for reasoning, succeeds in this task.
Crucially, of the three neuro-symbolic methods, \gtlt achieves strong performance with less domain-specific knowledge than other methods: NS-VQA needs groundtruth programs; NS-CL needs the concept vocabulary; \gtlt requires neither.
Second, our model is the only one to perform well on the hardest ``out-of-sample'' generalization tests: holding out ``count'' and generalizing to deeper embeddings.
The other, easier generalization tests all have close neighbors in the training set, differing by just one word.
In contrast, the length, depth, and ``count'' tests require generalizing to sentences that differ in multiple words from any training example.
They appear to require---or at least benefit especially well from---\gtlt lexical-grammatical approach to capturing the meaning of complex utterances with explicit constituent-level (as opposed to simply word-level) composition.

\subsection{Language-Driven Navigation}
\label{sec:g2l2-scan}
The second domain we consider is language-driven navigation.
We evaluate models on the SCAN dataset \citep{Lake2018Generalization}: a collection of sentence and navigational action sequence pairs.
There are six primitive actions: {\it jump}, {\it look}, {\it walk}, {\it run}, {\it lturn}, and {\it rturn}, where an instruction {\it turn left twice and run} will be translated to {\it lturn lturn run}.
All instructions are generated from a finite context-free grammar, so that we can systematically construct train-test splits for different types of compositional generalizations.

\subsubsection{Domain-Specific Language}
Our DSL for the language-driven navigation domain is a simple string manipulation language that supports creating new strings, concatenating two strings, and repeating a string multiple times. Our DSL contains only two primitive types: action sequence, abbreviated as ActSeq, and integer.
Formally, we summarize the list of operations in our language-driven navigation domain in \cref{tab:scan-dsl}.

\begin{table}[t!]
    \centering
    \setlength{\tabcolsep}{3pt}
    \begin{tabular}{p{0.45\columnwidth}p{0.5\columnwidth}} \toprule
        Signature                                                                          & Note                                                                                            \\ \midrule
        {\it empty}() $\longrightarrow$ ActSeq                                             & Create an empty string (of length 0).                                                           \\ \midrule
        {\it newprim}() $\longrightarrow$ ActSeq                                           & Create a string containing only one primitive action. In SCAN, there are in total 6 primitives. \\ \midrule
        {\it newint}() $\longrightarrow$ Integer                                           & Create a single integer. In SCAN, we only support integers \{2, 3, 4\}.                         \\ \midrule
        {\it concat}($\mathbf{a}$: ActSeq, $c$: ActSeq) $\longrightarrow$ ActSeq           & Concatenate two input strings.                                                                  \\ \midrule
        {\it repeat}($\mathbf{a}$: ActSeq, $\mathbf{b}$: Integer) $\longrightarrow$ ActSeq & Repeat the input string for multiple times.                                                     \\\bottomrule
    \end{tabular}
    \caption{All operations in the domain-specific language for language-driven navigation.}
    \label{tab:scan-dsl}
\end{table}

\paragraph{Probabilistic string representation.}
We represent each string in a ``probabilistic'' manner. In particular, each string $s$ is represented as a tuple $\langle L^s, C^s\rangle$. $L^s$ is a categorical distribution of the length. $C^s$ is a three-dimensional tensor, indexed by $\ell, k, c$, where $C^s_{\ell, k, c} = p(s[k] = c | \textit{length}(s) = \ell)$. Thus, $C$ has the shape $[L+1, L, |V|]$, where $L$ is the max length of a string and $V$ is the action vocabulary. For simplicity, we constrain that $C^s_{\ell, k, c} \equiv 0$ for all $k > \ell$.

It is straightforward to represent empty strings: $L_0 = 1$, or strings with a single action primitive $a$: $L_1 = 1$ and $C_{1, 0, a} = 1$. Now we explain our implementation of the $\textit{concat}$ and the $\textit{repeat}$ operation.

For $z = \textit{concat}(x, y)$:
\begin{eqnarray*}
    L^z_\ell &=& \sum_{0 \le i \le \ell} \left( L^x_i \cdot L^y_{(\ell - i)} \right);\\
    C^z_{\ell, k, c} &=& \frac{1}{L^z_\ell} \sum_{0 \le i \le \ell} \left( L^x_i \cdot L^y_{(\ell - i)} \cdot (C^x_{i, k, c} + C^y_{\ell - i, k - i, c}) \right).
\end{eqnarray*}
The high-level idea is to enumerate the possible length of both strings.

Similarly, for $z = \textit{repeat}(x, m)$,
\begin{eqnarray*}
    L^z_\ell &=& \begin{cases}
        L^x_{\ell / m} & \text{if $\ell \text{~mod~} m = 0$} \\
        0              & \text{otherwise}
    \end{cases}\\
    C^z_{\ell, k, c} &=& \begin{cases}
        L^x_{\ell / m, k \text{~mod~} (\ell / m), c} & \text{if $\ell \mod m = 0$ and $k < \ell$} \\
        0                                            & \text{otherwise}
    \end{cases}.
\end{eqnarray*}

\paragraph{Expected execution.}
In this domain, we only perform expected execution for semantic programs of type ActSeq, whose execution results can be represented using the probabilistic string representation.
Denote $\bar{s}_i$ as the execution results for $K$ programs, and $\tau(s_i)$ the corresponding weights. We define $p(s_i) = \textit{softmax}\left( \left\{\tau(s_i)\right\} \right)_i = \frac{\exp \tau(s_i)}{ \sum_j \exp \tau(s)j) }$. We compute the expected string $\bar{s}$ and its weight $\tau(s)$ as:
\begin{equation*}
    L^s_\ell = \sum_i p(s_i) L^{s_i}_\ell; \quad C^s_{\ell, k, c}  = \frac{ \sum_i \left( p(s_i) L^{s_i}_\ell \cdot C^{s_i}_{\ell, k, c} \right) }{L^s_{\ell}}.
\end{equation*}

\paragraph{Candidate lexicons.}
We use a simple enumerate algorithm to generate candidate lexicon entries for our language-driven navigation DSL. Specifically, we first enumerate candidate semantic programs for each lexicon entry that satisfy the following constraints:
\begin{enumerate}
    \item There are at most three operations.
    \item There are at most two arguments.
    \item There is at most one argument whose type is a functor.
    \item There is no argument of type \textit{Integer}.
\end{enumerate}

\cref{tab:scan-program-sample} lists a couple of programs generated by the algorithm and their corresponding types.

\begin{table}[t]
    \centering\small
    \setlength{\tabcolsep}{3pt}
    \begin{tabular}{lp{0.55\linewidth}} \toprule
        Type                                                           & Program (Note)                                                          \\ \midrule
        \mycell{ActSeq}                                                & \mycell{$\textit{walk}()$                                               \\
        The simplest program that constructs a string with                                                                                       \\
        a single action primitive: {\bf WALK}.}                                                                                                  \\ \midrule
        \mycell{(ActSeq) $\longrightarrow$ ActSeq}                     & \mycell{$\lambda x.\textit{concat}(\textit{look}(), x)$                 \\ Prepend a {\bf LOOK} action to an input string.}\\ \midrule
        \mycell{(ActSeq, ActSeq) $\longrightarrow$ ActSeq}             & \mycell{$\lambda x.\lambda y.\textit{concat}(x, y)$                     \\ Concatenate two strings.}\\ \midrule
        \mycell{(ActSeq, ActSeq) $\longrightarrow$ ActSeq}             & \mycell{$\lambda x.\lambda y.\textit{concat}(\textit{repeat}(x, 2), y)$ \\ Repeat the first string twice and concatenate with the \\
        second string.}                                                                                                                          \\ \midrule
        \mycell{((ActSeq) -> ActSeq, ActSeq) $\longrightarrow$ ActSeq} & \mycell{$\lambda x.\lambda y.\textit{concat}(y, x(\textit{walk}()))$    \\ The first argument ($x$) is a function which maps a \\
        ActSeq to another ActSeq. The second argument $y$ is                                                                                     \\
        an ActSeq. The function invokes $x$ with a simple string                                                                                 \\
        {\bf WALK}, and                                                                                                                          \\ concatenate the result with $y$.}\\
        \bottomrule
    \end{tabular}
    \caption{Sample semantic programs generated by the enumeration process based on our language-driven navigation DSL.}
    \label{tab:scan-program-sample}
\end{table}

Based on the candidate semantic types, we first instantiate candidate lexicon entries by enumerating possible ordering (linearization) of the arguments. For example, the simple program $\lambda x.\textit{concat}(\textit{look}(), x)$ has two possible linearizations: {\it ActSeq}/{\it ActSeq} and {\it ActSeq}\textbackslash{\it ActSeq}. As discussed in the main paper, in order to handle parsing ambiguities, we further introduce two finer-grained syntactic types for the {\it ActSeq} type: {\it S} and {\it V}. In practice, we only allow the following set of syntactic types: {\it V}, {\it V/V}, {\it V\textbackslash V}, {\it V\textbackslash V/V}, {\it V\textbackslash V/(V\textbackslash V)}, and {\it S\textbackslash V/V}. In total, we have 178 candidate lexicon entries for each word.

\paragraph{Setup.} We use a string-editing domain-specific language (DSL) for modeling the meaning of words in the SCAN dataset.
At a high level, the model supports three primitive operations: constructing a new constant string (consisting of primitive operations), concatenating two strings, and repeating the input string a number of times.

For \gtlt, we generate candidate lexicons by enumerating functions in the string-editing DSL with up to 2 arguments, and the function body has a maximum depth of 3.
We also allow at most one of the arguments to be functor-typed, for example, {\it V\textbackslash V/(V\textbackslash V)}.
To handle parsing ambiguities, we use two primitive syntax types $S$ and $V$, while both of them are associated with the semantic type of {\it string}.
In total, we have 178 candidate lexicon entries for each word.

\subsubsection{Baselines}
We compare \gtlt to seven baselines.
(1) Seq2seq \citep{Sutskever2014Sequence} trains an LSTM-based encoder-decoder model.
We follow the hyperparameter setups of \citep{Lake2018Generalization}.
(2) Transformer \citep{vaswani2017attention} is a 4-head Transformer-based autoregressive seq2seq model.
We tuned the hidden size (i.e., the dimension of intermediate token representations) within \{100, 200, 400\}, as well as the number of layers (for both the encoder and the decoder) from \{2, 4, 8\}.
Other methods are based on different data augmentation schemes for training an LSTM seq2seq model.
Specifically, (3) GECA augments the original training splits using heuristic span recombination rules; (4) WordDrop \citep{guo-etal-2020-sequence} performs random dropout for input sequence (while keeping the same label); (5) similarly, SwitchOut \citep{wang2018switchout} randomly replaces an input token with a random token from the vocabulary; (6) SeqMix \citep{guo-etal-2020-sequence} uses soft augmentation techniques following \citep{zhang2017mixup}, which composes an ``weighted average'' of different input sequences; (7) recomb-2 \citep{akyurek-etal-2020-learning} learns recombination and resampling rules for augmentation.

\subsubsection{Results}
\input{tables/602-g2l2-scan.tex}
We compare models on three train-test splits (\cref{tab:g2l2-scan}).
In {\bf Simple}, the training and test instructions are drawn from the same distribution.
We compare the data efficiency of various models by using either 10\% or 100\% of the training data and test them on the same test split.
While all models can achieve a nearly-perfect accuracy with 100\% training data, our model \gtlt shows an advantage with only a small amount of data.
Next, in {\bf Compositional}, we have held out the sentences containing certain phrases, such as {\it jump} and {\it around right}.
For these held-out phrases, only valid non-contextual examples containing them (i.e., \textit{jump} in isolation and no example for \textit{around right}) are available during training.
During the test, algorithms need to make systematical generalizations of these phrases in novel contexts.
Finally, in {\bf Length}, all training examples have an action length less than or equal to 22, while that of a test example is up to 48.
Our model consistently reaches perfect performance in all considered settings, even on the cross-length generalization task where GECA does not help improve performance.
These results are consistent with the conclusions we derived from the CLEVR dataset.
Specifically, data-augmentation techniques for SCAN can solve simple generalization tests (e.g., {\it jump}, where tests all have close neighbors in the training set, differing by just one word) but not the hard ones (e.g., {\it length}, where test sentences can different in multiple words from any training examples).

\subsubsection{Case Study}
\gtlt is expressive enough to achieve perfect accuracy on the SCAN dataset: a set of lexicon entries matches the groundtruth in SCAN.
However, our learning algorithm does not always converge on the correct lexicon, but when it fails, the failure can be identified based on training-set accuracy.
So, we perform model selection based on the training accuracy for \gtlt: after a sufficient number of epochs, if the model has not reached perfect accuracy (100\%), we re-initialize the weights and train the model again.
Our results show that, among 100 times of training, the model reaches 100\% accuracy 74\% of the time
For runs that do not achieve 100\% accuracy, the average performance is 0.94.

Since \gtlt directly learns human-interpretable lexicon entries associated with each individual word, we can further inspect the failure cases made by it when the training accuracy does not converge to 0.
We find that the most significant failure mode is the word {\it and} (e.g., {\it jump and run}) and {\it after} (e.g., {\it jump after run}).
Both of them are treated as connectives in SCAN.
Sometimes \gtlt fails to pick the syntax type {\it S\textbackslash V/V} over the type {\it V\textbackslash V/V}.
The entry {\it V\textbackslash V/V} will succeed in parsing most cases (e.g., {\it jump and run}), except that it will introduce ambiguous parsing for sentences such as ``{\it jump and run twice}'': {\it jump and \underline{run twice}} {\it vs.} {\it \underline{jump and run} twice}.
Based on the definition of the SCAN, only the first derivation is valid.
In contrast, using {\it S\textbackslash V/V} resolves this ambiguity.
Depending on the weight initialization and the example presentation order, \gtlt sometimes gets stuck at the local optima of {\it V\textbackslash V/V}.
However, we can quickly identify this by the training accuracies---\gtlt can reach perfect performance on all considered splits by simply retraining with another random seed; therefore, we only select those with 100\% training accuracy as valid models.

\section{Conclusion and Discussion}
\label{sec:g2l2-conclusion}

In this chapter, we have presented \gtlt, a lexicalist approach towards learning the compositional and grounded meaning of words.
\gtlt builds in a compact but potentially universal set of combinatory grammar rules and learns grounded lexicon entries from a collection of sentences and their grounded meaning without human annotated lexicon entries.
The lexicon entries represent the semantic type of the word, the ordering settings for its arguments, and the grounding of concepts in its semantic program.
To facilitate lexicon entry induction in an exponentially growing space, we introduced \ckyee for joint chart parsing and {\it expected execution}.

Through systematical evaluation on both visual reasoning and language-driven navigation domains, we demonstrate the data efficiency and compositional generalization capability \gtlt, and its general applicability in different domains.
The design of \gtlt suggests several research directions.
First, in \gtlt, we have made strong assumptions on the context-independence of the lexicon entry and the application of grammar rules; handling linguistic ambiguities and pragmatics needs further exploration \citep{Frank2012Predicting}.
Second, meta-learning models that can leverage learned words to bootstrap the learning of novel words, such as syntactic bootstrapping \citep{gauthier2018word}, is a meaningful direction.
Finally, future work may consider integrating \gtlt with program-synthesis algorithms \citep{ellis2023dreamcoder} for learning more generic and complex semantic programs.

%% file: figures/601-g2l2-teaser.tex
\begin{figure}[t!]
    \centering
    \includegraphics[width=\textwidth]{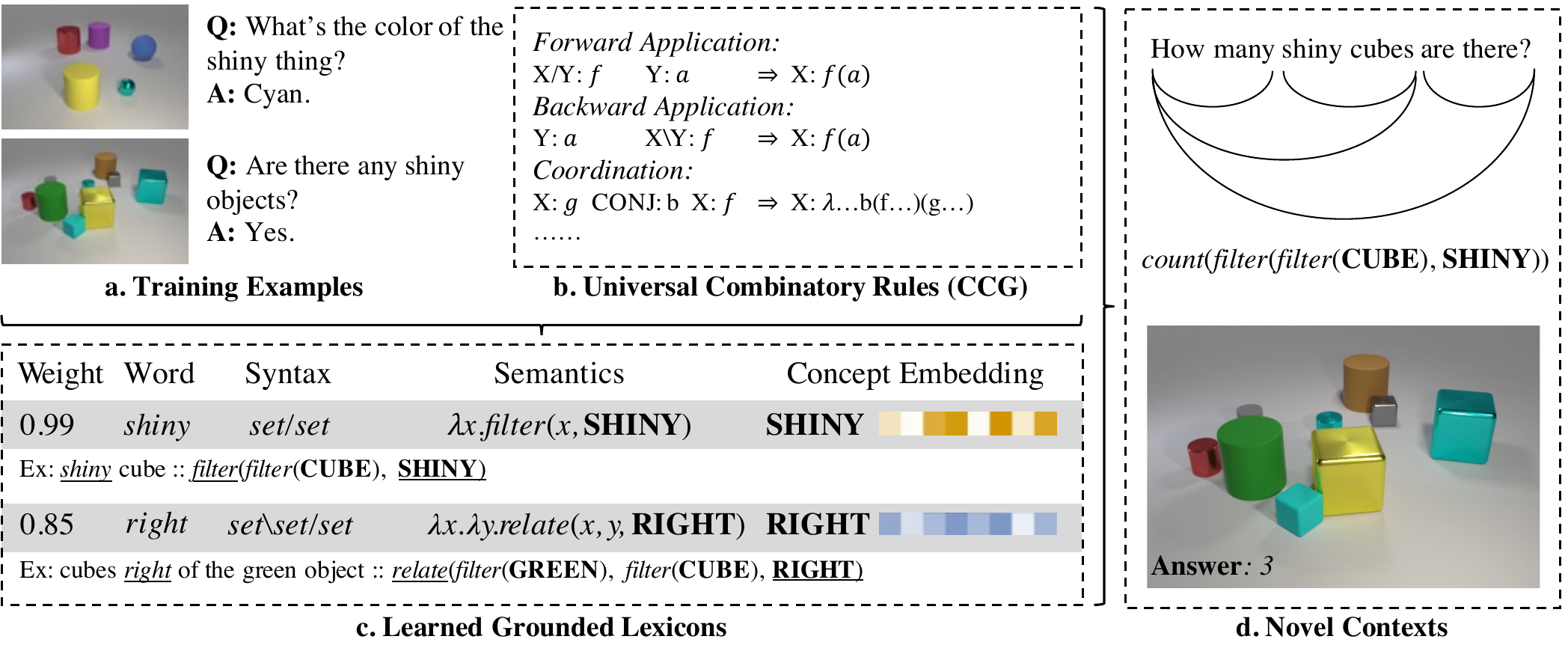}
    \caption[Illustruation of idea of grounded lexicon learning.]{
        Illustruation of idea of grounded lexicon learning.
        The model learns from grounded language data, for example, by looking at images and reading parallel question-answer pairs.
        It learns a collection of grounded lexicon entries comprised of weights, syntax types, semantics forms, and, optionally, grounded embeddings associated with semantic concepts.
        These lexicon entries can be used to parse questions into programs.
    }
    \label{fig:g2l2-teaser}
\end{figure}

%% file: figures/602-g2l2-chart-parsing.tex
\begin{figure}[t!]
    \centering
    \includegraphics[width=\textwidth]{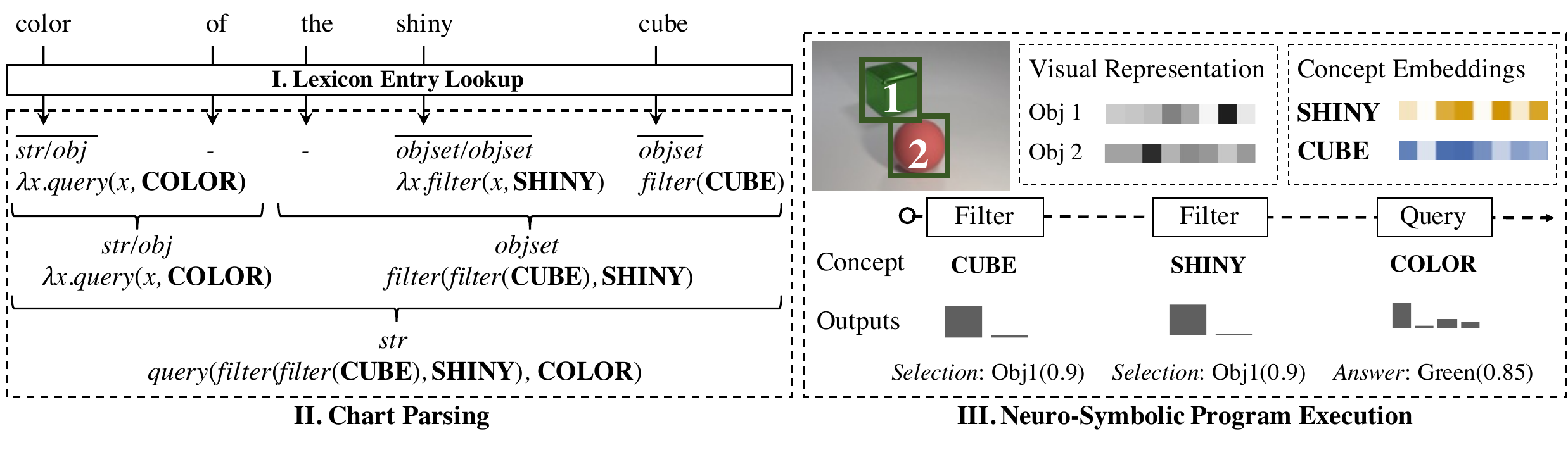}
    \caption[Illustruation of the \gtlt model.]{\gtlt parses the input sentence into an executable neuro-symbolic program by first (I) lookup the lexicon entry associated with each word, followed by (II) computes the most probable parsing tree and the corresponding tree with a chart parsing algorithm. The derived program can be grounded and executed on an image with a neuro-symbolic reasoning process \citep{mao2019neurosymbolic} (III).}
    \label{fig:g2l2-model}
\end{figure}

%% file: figures/603-g2l2-entry.tex
\begin{figure}[t]
    \centering
    \includegraphics[width=0.45\textwidth]{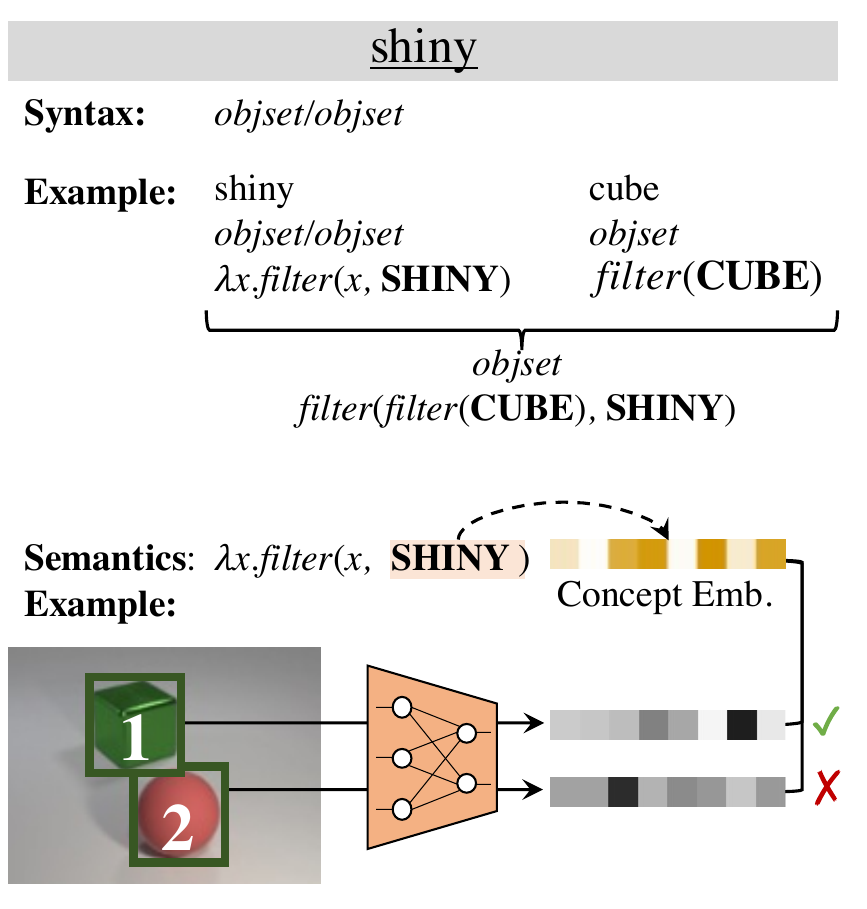}
    \caption[Illustration of a lexicon entry in \gtlt.]{
        Illustration of a lexicon entry in \gtlt.Each word is associated with a grounded lexicon, comprised of its syntactic type and a neuro-symbolic semantic program.
    }
    \label{fig:g2l2-entry}
\end{figure}

%% file: figures/604-g2l2-ckyee.tex
\begin{figure}[t!]
    \centering
    \includegraphics[width=0.5\textwidth]{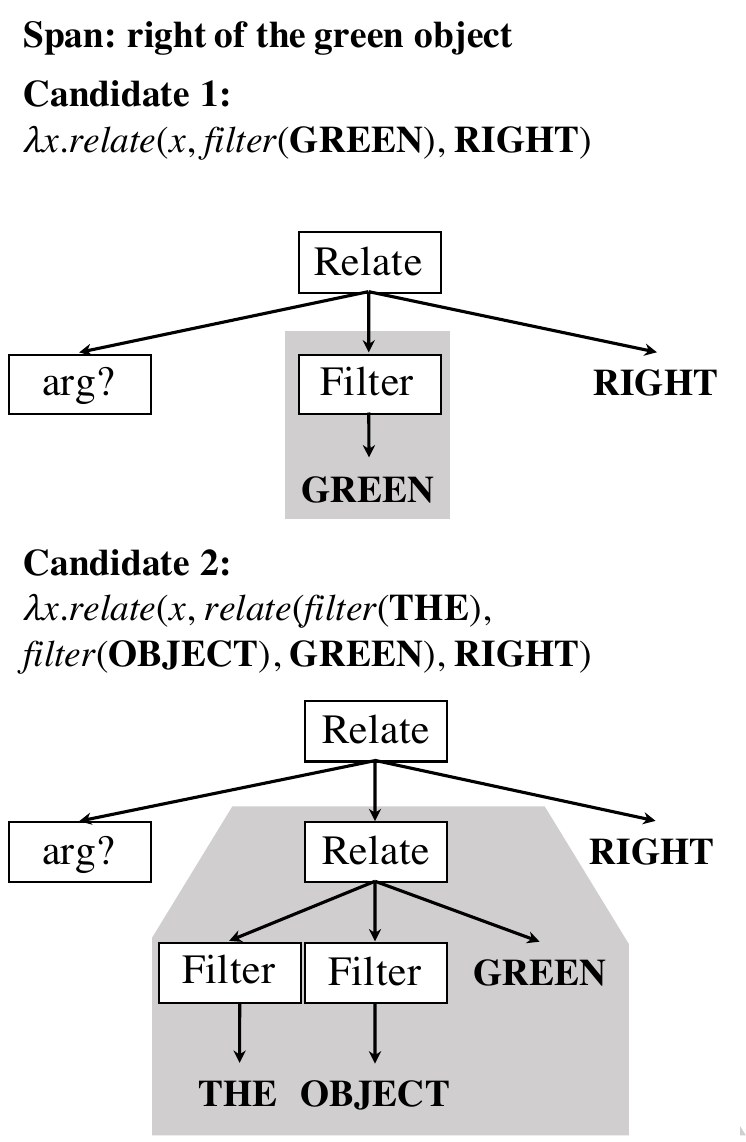}
    \caption[Illustration of two mergeable semantic programs.]{
        An illustrative example of two semantic programs that can be merged by computing the expected execution results of two subtrees (highlighted in gray).
        Both subtrees output a vector of scores that indicates the selected objects.
    }
    \label{fig:g2l2-subtree}
\end{figure}

%% file: algorithms/601-ckyee.tex
\begin{algorithm}[t!]
  \SetAlgoLined
  \SetKwInOut{Input}{Output}
  \SetKwFunction{ckyeefunc}{\textsc{ExpectedExecution}}
  \KwIn{\\
  $w_i$: the input sentence; \\
  $L$: sentence length; \\
  $e_i^j$: the $j$-th lexicon entry associated with word $w_i$; \\
  $\tau(e_i^j)$: lexicon weights.}
  \KwOut{$\mathit{exe}_k$ the execution result of the all possible derivations and their weights $\tau(\mathit{exe}_k)$.}
  \For{$\mathit{i} \gets 0$ to $L - 1$}{
    Initialize $\mathit{dp}[\mathit{i}, \mathit{i+1}]$ with lexicon entries $e_i^{*}$ and weights $\tau(e_i^*)$
  }
  \For{$\mathit{length} \gets 1$ to $L$}{
    \For{$\mathit{left} \gets 0$ to $L - \mathit{length}$}{
      $\mathit{right} \gets \mathit{left} + \mathit{length}$ \\
      $\mathit{dp}[\mathit{left}, \mathit{right}] \gets $ empty list \\
      \For{$\mathit{k} \gets \mathit{left} + 1$ to $\mathit{right} - 1$}{
        Try to combine nodes in $\mathit{dp}[\mathit{left}, \mathit{k}]$ and $\mathit{dp}[\mathit{k}, \mathit{right}]$ \\
        Append successful combination to $\mathit{dp}[\mathit{left}, \mathit{right}]$
      }
      $\textsc{ExpectedExecution}(\mathit{dp}[\mathit{left}, \mathit{right}])$
    }
  } \vspace{5pt}
  \textbf{Function} \ckyeefunc{$a$: a list of derivations} \\
    \While{$\exists x, y \in a$ are identical except for subtrees of the same type}{
      Create $z$ from $x$ and $y$ by computing the expected execution results for non-identical subtrees \\
      $\tau(z) \gets \tau(x) + \tau(y)$ \\
      Replace $x$ and $y$ in $a$ with $z$
    }
  \caption{The CKY-E$^2$ algorithm.}
  \label{alg:g2l2-ckyee}
\end{algorithm}

%% file: tables/601-g2l2-clevr.tex
\begin{table}[t]
    \centering\small
    \setlength{\tabcolsep}{3pt}
    \begin{tabular}{l cc c cc c ccc c}
    \toprule
         \bf \multirow{2}{*}{\vspace{-5pt}Model} & \bf \multirow{2}{*}{\vspace{-5pt}Prog?} & \bf \multirow{2}{*}{\vspace{-5pt}Concept?} &&  \multicolumn{2}{c}{\bf Standard} && \multicolumn{3}{c}{\bf Compositional Generalization} & \bf Depth \\
         \cmidrule{5-6}  \cmidrule {8-10}
                                                   &&&& 10\%   & 100\% && \it ~~~purple~~~ & \it ~~~right of~~~ & \it ~~~count~~~ &  \\
    \midrule
         MAC        & N/N & N/N && 85.39 & 98.61  && 97.14 & 90.85 & 54.87  & 77.40  \\ \midrule
         TbD-Net  & Y/Y & N/N && 44.52 & 98.04  && 89.57 & 49.92 & 63.37  & 53.13\\
         NS-VQA                 & Y/Y & Y/Y && \textbf{98.57} & 98.57  && 95.52 & \textbf{99.80} & 81.81  & 50.45 \\
         NS-CL         & Y/N & Y/N && 98.51 & \textbf{98.91}  && \textbf{98.02} & 99.01 & 18.88  & 81.60 \\
         \midrule
         \gtlt (ours)                             & Y/N & Y/N && 98.11 & 98.25  && 97.82 & 98.59 & \textbf{96.76}  & \textbf{98.49} \\
    \bottomrule \\
    \end{tabular}
    \caption[Accuracy on the CLEVR dataset.]{Accuracy on the CLEVR dataset. Our model achieves a comparable results with state-of-the-art approaches on the standard training-testing split. It significantly outperforms all baselines on generalization to novel word compositions and to sentences with deeper structures. The best number in each column is bolded. The second column indicates whether the model uses program-based representation of question meaning and whether it needs program annotation for training questions. The third column indicates whether the model explicitly models individual concepts and whether it needs concept annotation for objects during training.}
    \label{tab:g2l2-clevr}
\end{table}

%% file: tables/602-g2l2-scan.tex
\begin{table}[t]
    \centering\small
    \setlength{\tabcolsep}{6pt}
    \begin{tabular}{l cc c     cc c}
    \toprule
         \bf \multirow{2}{*}{\vspace{-5pt}Model} &  \multicolumn{2}{c}{\bf Simple} && \multicolumn{2}{c}{\bf Compositional Generalization} & \bf Length \\
         \cmidrule{2-3}  \cmidrule {5-6}
                   & 10\%  & 100\% &&  \it jump & \it around right \\
    \midrule
         seq2seq     & 0.93\stderr{0.05} & 0.99\stderr{0.01}          %
         && 0.00\stderr{0.00}$^\dagger$ & 0.00\stderr{0.00}$^\dagger$ & 0.15\stderr{0.02} \\
         Transformer   & 0.71\stderr{0.24} & 0.78\stderr{0.11}%
         && 0.00\stderr{0.00}~~ & 0.10\stderr{0.08}~~ & 0.02\stderr{0.01} \\
         GECA               & 0.99\stderr{0.00} &   0.98\stderr{0.01}                  %
         && 0.87\stderr{0.05}$^\dagger$ & 0.82\stderr{0.11}$^\dagger$ & 0.15\stderr{0.02} \\
         WordDrop $^*$     & 0.56\stderr{0.02}                & 0.62\stderr{0.02}       %
         && 0.52\stderr{0.02}~~    & 0.70\stderr{0.06}~~ & 0.18\stderr{0.01} \\
         SwitchOut $^*$    & 0.99\stderr{0.01}                & 0.99\stderr{0.01}                         %
         && 0.98\stderr{0.02}~~    & 0.97\stderr{0.02}~~ & 0.17\stderr{0.02} \\
         SeqMix $^*$       & --                & --                         %
         && 0.98$^\ddagger$~~~~~~~~~    & 0.89$^\ddagger$~~~~~~~~~ & -- \\
         recomb-2 & --                & --                         %
         && 0.88\stderr{0.07}$^\dagger$ & 0.82\stderr{0.08}$^\dagger$ & -- \\
         \gtlt (ours)                               & \textbf{1.00}\stderr{0.00}  & \textbf{1.00}\stderr{0.00} & & \textbf{1.00}\stderr{0.00}~~~ & \textbf{1.00}\stderr{0.00}~~~& \textbf{1.00}\stderr{0.00}\\
    \bottomrule \\
    \end{tabular}
    \caption[Accuracy on the SCAN dataset.]{
     Accuracy on the SCAN dataset, averaged across 10 valid runs when applicable, $\pm$ denotes standard deviation. The best number in each column is bolded. $\dagger$: results taken from \citet{akyurek-etal-2020-learning}; $\ddagger$: results taken from \citet{guo-etal-2020-sequence}.
    Both papers have only presented results on the compositional generalization split.
    $*$: applied after GECA.
    The results for GECA are based on the released implementation by the authors.
    All the models are selected with respect to the accuracy on the training set. }
    \label{tab:g2l2-scan}
\end{table}

%% file: src/70-mbrexec.tex
\chapter{Learning Semantic Parses through Program Execution Consistency}
\label{chapter:mbrexec}
\textit{Content in this chapter has been published as a conference paper at EMNLP 2022 \citep{shi2022natural}.}

\input{figures/701-mbrexec-teaser.tex}

We now focus on learning semantic parses in realistic settings, such as translating natural language into executable code using generative models \interalia{chen-etal-2021-evaluating,austin-etal-2021-program,li-etal-2022-competition}.
While these models do not explicitly incorporate program semantics (i.e., execution results) during training, they can generate correct solutions for many problems.
However, choosing a \emph{single} correct program from a generated set for each problem remains challenging.

In this work, we introduce execution result-based minimum Bayes risk decoding (\mbrexec; \cref{fig:mbrexec-overview}) for program selection and show that it improves the few-shot performance of pre-trained code models on natural-language-to-code tasks.
We select output programs from a generated candidate set by marginalizing over program implementations that share the same semantics.
Because exact equivalence is intractable, we execute each program on a small number of test inputs to approximate semantic equivalence.
Across datasets, execution or simulated execution significantly outperforms the methods that do not involve program semantics.
We find that \mbrexec consistently improves over all investigated execution-unaware selection methods, suggesting it is an effective approach for natural language-to-code translation.

\section{Related Work}
\label{sec:mbrexec-related}

\paragraph{Language to code translation with neural networks.}
With the progress of neural network--based language modeling and conditioned text generation, there has been much work exploring natural language to code generation with end-to-end neural model architectures \interalia{xiao-etal-2016-sequence,ling-etal-2016-latent,rabinovich-etal-2017-abstract,dong-lapata-2018-coarse,suhr-etal-2018-learning,xu-etal-2020-incorporating,lachaux-etal-2021-dobf}.
Recently, large Transformer-based~\citep{vaswani2017attention} pre-trained code models have shown surprisingly strong generation performance across programming languages \interalia{chen-etal-2021-evaluating,austin-etal-2021-program,li-etal-2022-competition}.
In this work, we explore selection (i.e., inference) methods to apply to these pre-trained models, showing that selecting programs using their execution results can greatly improve program generation.

Multiple benchmarks have been proposed to evaluate code model performance \interalia{yin-etal-2018-learning,miceli-barone-sennrich-2017-parallel,lu-etal-2021-codexglue}.
In this work, we evaluate on three text-to-code datasets: MBPP \citep[Python;][]{austin-etal-2021-program}, Spider \citep[SQL;][]{yu-etal-2018-spider} and NL2Bash \citep[Bash;][]{lin-etal-2018-nl2bash}, covering a range of programming languages.

\paragraph{Prompting pre-trained language models.}
The GPT-2 \citep{radford-etal-2019-language} and GPT-3 \citep{brown-etal-2020-language} models have shown strong prompting performance: after conditioning on a task-related prompt, the language models are often able to make accurate output predictions for unseen inputs. These results lead to prompt-based approaches for few-shot or zero-shot text classification \interalia{shin-etal-2020-autoprompt,gao-etal-2021-making,min-etal-2022-noisy}, question answering \citep{khashabi-etal-2020-unifiedqa}, machine translation \citep{radford-etal-2019-language}, and evaluation of generated text \citep{yuan-etal-2021-bartscore}, where no more than a few examples are used to construct the prompts. Few-shot examples are usually formatted into natural language prompts, and continuations generated by the models for these prompts are then converted to task-specific predictions. The prompt formatting can be either manually designed \citep{jiang-etal-2020-know} or automatically learned \citep{li-liang-2021-prefix,lester-etal-2021-power}.
We refer the readers to \citet{liu-etal-2023-pre} for a more comprehensive survey.

In this work, we prompt a pre-trained code model \citep[Codex;][]{chen-etal-2021-evaluating} in a few-shot setting (\cref{sec:mbrexec-sample-collection}) and perform execution-based selection over the samples. We also find that the Codex model performs well with fairly programming-language-agnostic prompt formatting (\cref{tab:mbrexec-prompt-format}).
\input{tables/701-mbrexec-prompt-format.tex}

\paragraph{Minimum Bayes risk decoding.}
In structured prediction, minimum Bayes risk (MBR) decoding \citep{bickel-doksum-1977-mathematical} selects a structured output that minimizes the expected errors in the structure by introducing an explicit loss function to the decision criterion.
This method has outperformed the maximum a posteriori (MAP) method on many tasks, including syntactic parsing \citep{titov2006bayes,shi2019visually,zhang-etal-2020-efficient}, statistical machine translation \citep{kumar-byrne-2004-minimum,zhang-gildea-2008-efficient}, and neural machine translation \citep{eikema-aziz-2020-map,eikema-aziz-2022-sampling}.

In machine translation, MBR decoding is usually implemented by reranking candidates \interalia{goel-byrne-2000-minimum, kumar-byrne-2004-minimum, tromble-etal-2008-lattice}. Let $F$ denote the input and $E$ denote the corresponding ground-truth translation. Given a loss function $\ell(\cdot, \cdot)$ between translations and a probability model $P(E\mid F)$, MBR decoding can be formulated as
\begin{align}
    \hat{E} = \arg\min_{E'\in \mathcal{E}_h} \sum_{E\in\mathcal{E}_e} \ell(E, E') P(E\mid F),
    \label{eq:mbr-decoding}
\end{align}
where $\mathcal{E}_h$ is the \textit{hypothesis space}, and $\mathcal{E}_e$ is the \textit{evidence space}: both are sets of possible translations.

We define execution-based MBR loss functions and show they are crucial in the selection processes for natural language to code with a pre-trained large language model.

\section{Proposed Approach: \mbrexec}
Our execution-based framework consists of two parts: (1) collecting samples from a pre-trained code model (\cref{sec:mbrexec-sample-collection}) and (2) selecting the best candidate using minimum Bayes risk decoding (\cref{sec:mbrexec-mbr-exec}).
\subsection{Sample Collection}
\label{sec:mbrexec-sample-collection}
To obtain the corresponding code, we query the pre-trained code model with few-shot prompts followed by the text description, using a unified mark-up style few-shot prompting template (\cref{tab:mbrexec-prompt-format}).\footnote{While existing work on prompting language models usually requires a task-specific design of prompts \interalia{shin-etal-2020-autoprompt,zhong-etal-2021-factual,gao-etal-2021-making}, we find that a fairly general pattern (\cref{tab:mbrexec-prompt-format}), which does not involve any programming language-specific information, works well across programming languages on Codex.}
In addition to the generated programs themselves, most existing models also allow us to have the associated probability of generating each generated token $w_i$ conditioned on the prompt tokens $C=\langle c_1, \ldots, c_n \rangle$ and all the previously generated tokens $w_1, \ldots, w_{i-1}$, denoted by $P(w_i \mid C, w_1, \ldots w_{i-1})$.

\subsection{Execution-Based MBR Decoding}
\label{sec:mbrexec-mbr-exec}
Given a problem in its natural language description $C$, we sample a set of programs $\mathcal{P}=\{p_i\}_{i=1}^N$ using the method in \cref{sec:mbrexec-sample-collection}. We formulate the execution-based MBR (\mbrexec) decoding by selecting
\begin{align}
    \hat{p} & = \arg\min_{p \in \mathcal{P}} \mathcal{L}_\textit{MBR}(p; \mathcal{P}) \nonumber                                       \\
            & =\arg\min_{p\in \mathcal{P}} \sum_{p_\textit{ref} \in \mathcal{P}} \ell(p, p_\textit{ref}) \label{eq:mbr-exec-decoding}
\end{align}
as the best candidate, where $\mathcal{L}_\textit{MBR}(\cdot; \cdot)$ denotes the MBR loss of a program conditioned on a set of references, and $\ell$ is a predefined, execution-based loss function that examines the discrepancy between two programs. Intuitively, this finds a consensus candidate which has a low loss relative to all other candidates. The above implementation is an unbiased estimation of Eq~\eqref{eq:mbr-decoding}.

We introduce the following execution result-based loss function:
\begin{align*}
    \ell (p_i, p_j) & = \max_{t \in \mathcal{T}} \mathbbm{1}\left[p_i(t) \neq p_j(t)\right],
\end{align*}
where $\mathcal{T}$ is the set of available test inputs,\footnote{Our \mbrexec decoding process does not involve any ground-truth test case output, nor the ground-truth programs. This is compatible with many real scenarios, e.g., in a programming competition, where valid test inputs are easier to access than ground-truth output.} and $p_i(t)$ denotes the execution result of program $p_i$ when having $t$ as the input. When a program fails to execute on a test case, it is considered not equivalent to any other programs, even if they fail to execute as well. Intuitively, $\ell$ assigns equivalence ($0$ loss) if and only if two programs have the same output on all considered test cases.

There may be multiple programs receiving the same MBR loss $\mathcal{L}_\textit{MBR}(\cdot; \mathcal{P})$, which are all minima. We break any ties by selecting the program with the largest likelihood among them.

\section{Experiments}
\label{sec:mbrexec-expr}
We evaluate (\cref{sec:mbrexec-expr-results-main}) and analyze (\cref{sec:mbrexec-expr-analysis}) the performance of \mbrexec, starting with introducing the datasets and evaluation metrics (\cref{sec:mbrexec-expr-datasets}), as well as non-execution-based baselines (\cref{sec:mbrexec-expr-baselines}) for \mbrexec. Finally, we show and discuss oracle performances on the considered tasks (\cref{sec:mbrexec-expr-oracle}).

\subsection{Datasets and Evaluation Metrics}
\label{sec:mbrexec-expr-datasets}
We consider three datasets that cover a range of programming languages: MBPP \citep[Python;][]{austin-etal-2021-program}, Spider \citep[SQL;][]{yu-etal-2018-spider}, and NL2Bash \citep[Bash;][]{lin-etal-2018-nl2bash}.

\input{tables/704-mbrexec-detailed-prompt-mbpp.tex}

\paragraph{MBPP.}
The MBPP dataset \citep{austin-etal-2021-program}\footnote{
    \url{https://github.com/google-research/google-research/tree/master/mbpp}
} consists of 974 basic Python programming problems, with 500 used for testing and the rest for training or few-shot prompting.
There are ground-truth programs and three assertions (i.e., test cases with input and ground-truth output) associated with the description of each problem. When collecting the samples, we use one assertion as the extra information (\texttt{[INFO]}; \cref{tab:mbrexec-prompt-format}).\footnote{The main goal of \texttt{[INFO]} in MBPP is to inform Codex about the desired function name for more straightforward evaluation -- while the assertions are not a necessary part of the prompt, we use them as \texttt{[INFO]} for simplicity and compatibility with past work \citep{austin-etal-2021-program}. } Programs are evaluated with execution accuracy, where a program is considered as passing if all three test cases are correct.

\input{tables/705-mbrexec-detailed-prompt-spider.tex}

\paragraph{Spider.}
The Spider dataset \citep{yu-etal-2018-spider}\footnote{\url{https://yale-lily.github.io/spider}} is a text-to-SQL dataset, which requires a model to translate text descriptions into SQL commands. There are 7,000 examples for training and 1,034 for development. When prompting models to produce candidate commands, we concatenate the corresponding SQL table and column names as the \texttt{[INFO]}. Commands are evaluated with execution accuracy, where a command is considered as passing if it returns the same result as the ground-truth command when executed on the same database.

\input{tables/706-mbrexec-detailed-prompt-nl2bash.tex}

\paragraph{NL2Bash.}
The NL2Bash dataset \citep{lin-etal-2018-nl2bash} aims to translate natural language to bash commands. We omit \texttt{[INFO]} in the sample collection process. Because it is challenging to execute bash commands in a sandbox, we split a bash command with \texttt{bashlex},\footnote{\url{https://pypi.org/project/bashlex/}} a rule-based bash parser, and use the token-level BLEU-4 score between commands as the estimation of execution result similarity. We consider a command not executable when \texttt{bashlex} fails to parse it. Following \citet{lin-etal-2018-nl2bash}, commands are evaluated with character-level BLEU-4 score.

Across datasets, we use 15 examples from the training set for few-shot prompting.
Detailed examples showing prompt formatting can be found in \cref{tab:mbrexec-real-prompt-mbpp,tab:mbrexec-real-prompt-spider,tab:mbrexec-real-prompt-nl2bash}.
Unless otherwise specified, we collect samples by querying Codex with five different prompts, each containing three examples, using temperature 0.3. We combine the candidates sampled across the five prompts to get a set of candidate samples to use in our selection methods.
For execution on MBPP and Spider, we apply a memory limit of 128GB and a time limit of 10 seconds on a single Intel(R) Xeon(R) CPU E5-2698 v4 @ 2.20GHz CPU, and consider the programs that exceed these limits as inexecutable; unless otherwise specified, we only execute each program on the first test input provided for the example and use the output for calculating the Bayes risk in the inference process.

\subsection{Baselines}
\label{sec:mbrexec-expr-baselines}
We compare the most basic baselines with no selection, prompting Codex with three examples in \cref{tab:mbrexec-prompt-format} format:\footnote{We use the \texttt{code-davinci-001} engine throughout this work.}
\begin{itemize}[leftmargin=*]\setlength{\itemsep}{0pt}
    \item \textbf{Greedy decoding.} We perform token-by-token greedy decoding to generate the output.
    \item \textbf{Sampling.} We sample the output token by token with a fixed temperature, where we set the temperature as 0.3 in all of our experiments.
\end{itemize}
In addition, we consider the following baseline sample selection methods:
\begin{itemize}[leftmargin=*]\setlength{\itemsep}{0pt}
    \item \textbf{Maximizing likelihood} (\maxlikelihood). Given a set of sampled candidate programs, we select the one with the largest log-likelihood. Formally, we select
          \begin{align*}
              \hat{p} = \arg\max_{p \in \mathcal{P}} \prod_{i=1}^{n_p} P(w_{p, i} \mid C, w_{p, 1}, \ldots, w_{p, i-1}),
          \end{align*}
          where $n_p$ denotes the number of tokens in a generated program $p$, and $w_{p, i}$ denotes its $i$-th token.
    \item \textbf{Maximizing average log likelihood} (\maxavglikelihood) across tokens. In order to address the practical issue that ML typically favors shorter sequences, we follow \citet{chen-etal-2021-evaluating} and propose another baseline that uses the average log-likelihood across tokens as the selection criterion, where we select
          \begin{align*}
              \hat{p} = & \arg\max_{p \in \mathcal{P}} \frac{1}{n_p} \sum_{i=1}^{n_p} \log P(w_{p, i} \mid C, w_{p, 1}, \ldots, w_{p, i-1}).
          \end{align*}
    \item \textbf{BLEU score based MBR} (\mbrbleu). To study the effect of execution-based MBR in sample selection, we consider BLEU-based MBR, where the Bayes risk is calculated using the following risk function:
          \begin{align*}
              \ell_\textit{BLEU}(p_i, p_j) = -\text{BLEU}(p_i, p_j),
          \end{align*}
          where $\text{BLEU}(p_i, p_j)$ is the BLEU score of the two programs. We use character-level (MBR-char\textsc{bleu}) or token-level (MBR-token\textsc{bleu}) BLEU-4 in all of our experiments.
\end{itemize}

\input{figures/702-mbrexec-main-results.tex}
\input{tables/702-mbrexec-main-comparisons.tex}
\subsection{Primary Results}
\label{sec:mbrexec-expr-results-main}
We evaluate \mbrexec on the three datasets (\cref{sec:mbrexec-expr-datasets}) with dataset-specific metrics, using one test case for each problem.
\mbrexec outperforms all baselines without a selection process by a significant margin (\cref{tab:mbrexec-main-results-basic-baseline}).
In addition, we find that \mbrexec outperforms all baseline selection methods (\cref{fig:mbrexec-main-results}) and is especially effective on the two datasets (MBPP and Spider) that use execution-based evaluation.
In addition, the \mbrbleu metrics are also solid and robust across datasets, suggesting the effectiveness of finding a consensus candidate with generally low discrepancy with other samples.

While more samples lead to better performance for most methods, \maxavglikelihood consistently performs worse with larger sample size, as we find that \maxavglikelihood generally favors programs with unnecessary repetitions,\footnote{This issue has been found in existing open-ended text generation models, while methods such as unlikelihood training \citep{welleck-etal-2020-neural} may help reduce degeneration (i.e., the generation of unnecessarily repetitive output).} and a larger sample size generally leads to a more significant chance to have such a sample.

\subsection{Analysis}
\label{sec:mbrexec-expr-analysis}
We analyze the performance of \mbrexec from the following perspectives: the effectiveness across different sample collection temperatures (\cref{sec:mbrexec-expr-temperature}), the effectiveness of using groups of 3-shot prompts (\cref{sec:mbrexec-expr-15shot}), and the contribution of using execution results instead of simply checking the executability of programs (\cref{sec:mbrexec-expr-executability}).

\subsubsection{Effect of Sample Temperature}
\label{sec:mbrexec-expr-temperature}
\input{figures/703-mbrexec-temperature.tex}
\input{tables/703-mbrexec-0.3-vs-greedy.tex}
We first compare sampling with temperature 0.3 to greedy decoding (i.e., temperature $\tau=0$) from the Codex model (\cref{tab:mbrexec-0.3-vs-greedy}). When having the same number of examples, \mbrexec on sampled candidates with temperature 0.3 consistently reaches competitive or better performance than that on greedy decoded candidates.

We plot the performance of \mbrexec for various sampling temperatures (\cref{fig:mbrexec-temperature}). Across datasets, we find that \mbrexec with a decoding temperature lower than 0.5 usually leads to reasonably good performance. When the temperature approaches 1.0, the results rapidly drop for all considered selection methods on MBPP and Spider; however, \maxavglikelihood generally achieves higher performance on NL2bash with a higher temperature.

According to the evidence discussed above, we recommend sampling with a low temperature (specifically, lower than 0.5) for candidate sample collection and performing \mbrexec for final program selection for better results.

\input{figures/704-mbrexec-15-shot.tex}
\subsubsection{Effect of Different 3-shot Prompts}
\label{sec:mbrexec-expr-15shot}
We analyze the necessity of choosing multiple groups of 3-shot instead of simply concatenating the available 15 examples as the prompt (\cref{fig:mbrexec-15-shot}).\footnote{We only include MBPP and NL2Bash results here as concatenating 15 Spider examples usually results in exceeding the token number limit of the pre-trained models.}
We allow different orders of the 15 examples when collecting samples.
On both MBPP and NL2Bash datasets, we find that using different groups of 3-shot prompts outperforms concatenating all 15 examples, suggesting that different groups of fewer-shot prompts followed by post-hoc decoding may be more effective than using all available examples for all time.

\input{figures/705-mbrexec-executability.tex}
\subsubsection{Executability vs. Execution Results}
\label{sec:mbrexec-expr-executability}
We perform an ablation study to identify the contribution of execution results vs. program executability (\cref{fig:mbrexec-ablation-executability}) on the MBPP and Spider datasets.\footnote{We did not include NL2bash since \mbrexec does not execute the commands. However, the comparison between \mbrexec and MBR-token\textsc{bleu} in \cref{fig:mbrexec-temperature}(c) shows that using an external bash parser as an executability estimator leads to more consistent and generally better performance.}
We try to execute all candidates on the test cases and perform baseline candidate methods only on the candidates that successfully execute within the time limit. On both datasets, we find that simply involving executability checking significantly helps improve the performance of all non-semantic feature-based selection methods; on Spider, applying \maxlikelihood over executable commands even outperforms \mbrexec across sample sizes.

\subsubsection{Soft Loss as the Bayes Risk Function}
\input{figures/706-mbrexec-soft-hard.tex}
While all the above evaluations are based on executing one test case per problem, more test cases can lead to more accurate judgments of semantic equivalence between programs \citep{zhong-etal-2020-semantic}.
Therefore, we introduce more test cases and compare $\ell$ (\cref{sec:mbrexec-mbr-exec}) with $\ell_\textit{soft}$, a soft version of the loss function, as the Bayes risk function in \mbrexec. We define $\ell_\textit{soft}$ as follows:
\begin{align*}
    \ell_\textit{soft} (p_i, p_j) & = \frac{1}{|\mathcal{T}|}\sum_{t \in \mathcal{T}} \mathbbm{1}\left[p_i(t) \neq p_j(t)\right],
\end{align*}
which assesses equivalence based on the number of test cases that receive the same output.
If only one test case is available, $\ell$ and $\ell_\textit{soft}$ are equivalent.

We experiment with the MBPP dataset (\cref{fig:mbrexec-mbpp-hard-soft-exec}) as it provides three test cases per problem. While multiple test cases clearly outperform \mbrexec with one test case across sample sizes, we did not find a significant difference between $\ell_\textit{hard}$ and $\ell_\textit{soft}$, nor between using two or three test cases.

\subsection{Oracle Performance}
\label{sec:mbrexec-expr-oracle}
\input{figures/707-mbrexec-oracle.tex}
We report the upper bound performance of all inference methods (\cref{fig:mbrexec-oracle}). Here, we define the expected Pass@K on one problem $q$ by
\begin{align*}
     & \textit{ExPass@K}(q) = \mathbb{E}_{|\mathcal{P}| = K} \left[
        \max_{p\in \mathcal{P}}
        \min_{t \in \mathcal{T}_q}
        \mathbbm{1}\left[
            p(t) = G(t)
            \right]
        \right],
\end{align*}
where $G(t)$ denotes the ground-truth output for test case input $t$.
Intuitively, to calculate the performance upper bound, a problem $q$ is considered to be solved if one program in the candidate sample set $P$ passes all associated test cases $\mathcal{T}_q$. The dataset-level expected Pass@K is defined as the average expected Pass@K across all problems.

In addition, we report the supervised performance on these datasets, where all available training data are used for model training or finetuning: for MBPP, the results are from \citet{austin-etal-2021-program}, where they use all 374 training examples to finetune their pre-trained code model; for Spider, we compare to the current state-of-the-art result  \citep{scholak-etal-2021-picard}; for NL2Bash, we finetune GPT-2 \citep{radford-etal-2019-language} with all training examples with the same prompting set up as \cref{tab:mbrexec-prompt-format}.

However, it is worth noting that the upper bounds outperform the state-of-the-art supervised performances on all datasets by a significant margin when a reasonable sample is given.
This further demonstrates the effectiveness of the pre-trained code models and points out a potential next step in the direction: while such models can generate correct programs, designing an effective inference algorithm may be a promising way to translate natural language commands to code in real-world applications.

\section{Conclusion and Discussion}
\label{sec:mbrexec-discussion}
In this chapter, we present and systematically analyze \mbrexec, an execution--based inference algorithm for pre-trained language to code models, on datasets that cover three representative programming languages. Our results showed that doing execution, even with access only to inputs (not outputs) for test cases, or with only access to an executability checker, substantially helps improve the quality of generated programs, especially in the settings that use execution accuracy as the evaluation metric (MBPP and Spider). Given the consistently strong performance, we suggest future work on program synthesis with large pre-trained models consider \mbrexec as an effective selection algorithm. When we are not able to execute programs, or there are no test inputs available, our results suggest considering an alternative MBR metric (e.g., \mbrbleu) as the selection algorithm.

The method proposed in this chapter connects deeply with \citet{holtzman-etal-2021-surface}.
In addition to the potentially insufficient training of models, \citet{holtzman-etal-2021-surface} offers a potential interpretation of why the maximum-probability program does not always give the correct execution results---they may compete with each other in terms of the surface form probability.
In contrast to the domain-specific pointwise mutual information solution for question answering proposed by \citet{holtzman-etal-2021-surface}, we directly consider the external execution results, which are arguably more interpretable and reliable in noisy real-world settings.

Our results also align with the concurrent work by \cite{wang-etal-2023-self}, where they propose using consistency-based selection for mathematical reasoning.
More concretely, they sample multiple intermediate reasoning paths from the pre-trained large language models, and have them perform majority voting to select the final answer.
Indeed, there are three potential views of both work presented in this chapter and \cite{wang-etal-2023-self}:
\begin{itemize}
    \item \textbf{Majority voting.}
          Each of the sampled programs (this work) or reasoning paths \citep{wang-etal-2023-self} is considered as a vote, and the final answer is selected by majority voting.
    \item \textbf{Minimum Bayes risk decoding.}
          Unifying in the theoretical framework proposed in this work, it can be viewed that \citet{wang-etal-2023-self} (1) define a Bayes risk function that considers the discrepancy between reasoning paths by $\mathbbm{1}\left[x\neq y\right]$, where $x$ and $y$ are two reasoning paths of the same problem, and (2) select the final answer by minimizing the expected discrepancy.
    \item \textbf{Sample-based marginal distribution estimation.} Arguably more naturally, both methods can be viewed as estimating the marginal probability of execution results (this work) or final answers to a math problem \citep{wang-etal-2023-self} by sampling.
          Here, the models sample multiple examples and marginalize over all possible program forms (this work) or reasoning paths \citep{wang-etal-2023-self} that share the same execution results (this work) or final answers \citep{wang-etal-2023-self} to select the final program or answer.
\end{itemize}
The Bayes risk minimization framework provides a more flexible way to design new variants of the underlying probability distribution, compared to the other two views, through flexible modification of the Bayes risk function.
While this work estimates the semantics of programs in an arguably more reliable way, we also acknowledge that \citet{wang-etal-2023-self} have offered a more general approach that can only rely on pre-trained models without any external sources.

This work and \citet{wang-etal-2023-self} also share a common limitation: both methods are computationally expensive due to the need for multiple examples from the pre-trained models.
We suggest that future work explore more efficient ways to estimate the marginal distribution of programs or reasoning paths, which may lead to more practical applications of these methods in real-world scenarios.

%% file: figures/701-mbrexec-teaser.tex
\begin{figure}[t!]
    \centering
    \includegraphics[width=0.8\textwidth]{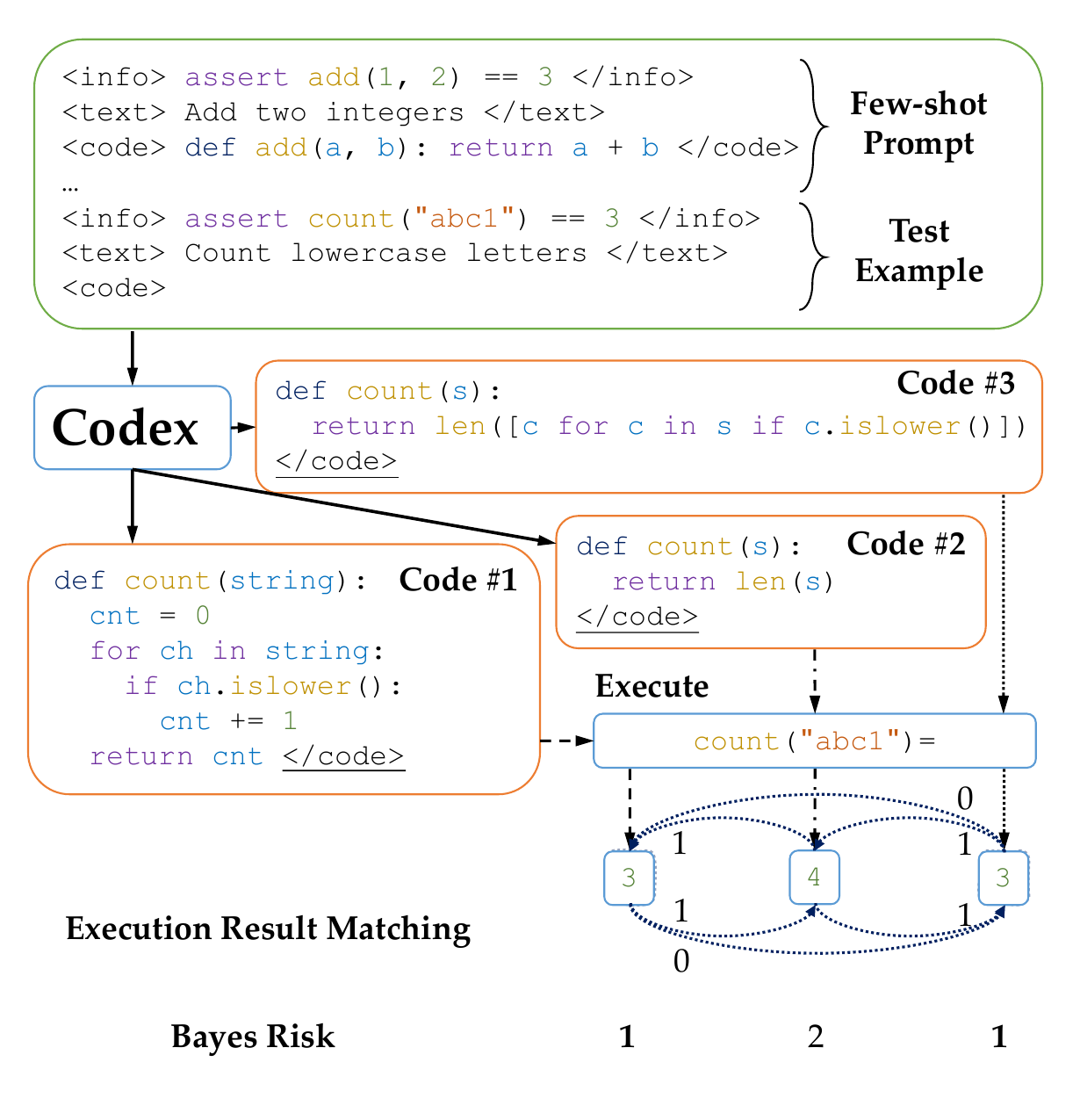}
    \caption[Illustration of \mbrexec.]{Illustration of \mbrexec on translating natural language to Python code: we (1) sample programs from a pre-trained language-to-code model such as Codex \citep{chen-etal-2021-evaluating}, (2) execute each program on one test case, and (3) select the example with the minimal execution result--based Bayes risk.
    Numbers around dotted lines denote the 0/1 matching loss between execution results, while the Bayes risk of a program is defined by the sum of the loss between itself and other examples.
    In the figure, either Code \#1 or Code \#3 can be selected. Ground-truth program output is not needed for selection.}
    \label{fig:mbrexec-overview}
\end{figure}

%% file: tables/701-mbrexec-prompt-format.tex
\begin{table}[t!]
    \centering \small
    \begin{tabular}{l}
        \toprule
        \textbf{\textit{General Template}}\\
            \texttt{{<info>[INFO]</info>}} \qquad \textit{(optional)} \\
            \texttt{{<text>[TEXT]</text>}} \\
            \texttt{{<code>}[CODE]</code>} \\
        \midrule \midrule
        \textbf{\textit{Instantiation 1: Python, One-Shot Example}} \\
        \midrule
        \textbf{\textit{Exemplar}} \\
            \texttt{<info>assert add(1, 2) == 3</info>}  \\
            \texttt{<text>Write a function that adds 2 integers</text>} \\
            \texttt{<code>def add(a, b): return a + b</code>} \\
        \midrule
        \textbf{\textit{Query}} \\
            \texttt{<info>assert cat() == "cat"</info>}  \\
            \texttt{<text>Write a function that outputs the string "cat"</text>} \\
            \texttt{<code>} \\
        \midrule \midrule
        \textbf{\textit{Instantiation 2: Bash, One-Shot Example}} \\
        \midrule
        \textbf{\textit{Exemplar}} \\
            \texttt{<text>show the files in the current directory</text>} \\
            \texttt{<code>ls</code>} \\
        \midrule
        \textbf{\textit{Query}} \\
            \texttt{<text>show the first 5 lines of a.txt</text>}  \\
            \texttt{<code>} \\
        \bottomrule
    \end{tabular}
    \caption[Prompt formatting for \mbrexec sample collection.]{\label{tab:mbrexec-prompt-format} Prompt formatting template for queries to pre-trained code models. For instantiation, we substitute \texttt{[TEXT]} and \texttt{[CODE]} with natural language descriptions and corresponding code snippets, respectively. We also provide compatibility for an optional \texttt{[INFO]} section to provide the model extra information (e.g., the desired function identifier and example function calls) that helps code generation. In general, we expect the pre-trained code models to generate a \texttt{</code>} token at the end of each code snippet given its pattern following ability \citep{brown-etal-2020-language,chen-etal-2021-evaluating}. Otherwise, we will truncate the generated code to 1024 tokens. }
\end{table}

%% file: tables/704-mbrexec-detailed-prompt-mbpp.tex
\begin{table}[t!]
    \centering \scriptsize
    \begin{tabular}{p{0.96\textwidth}}
    \toprule
    \textbf{MBPP: Prompt} \\
    \midrule ~\\[-25pt]
    \begin{lstlisting}[breaklines=true]
<info>assert camel_to_snake('GoogleAssistant') == 'google_assistant'</info>
<text>Write a function to convert camel case string to snake case string by using regex.</text>
<code>import re
def camel_to_snake(text):
  str1 = re.sub('(.)([A-Z][a-z]+)', r'\\1_\\2', text)
  return re.sub('([a-z0-9])([A-Z])', r'\\1_\\2', str1).lower()</code>
<info>assert sort_dict_item({(5, 6) : 3, (2, 3) : 9, (8, 4): 10, (6, 4): 12} ) == {(2, 3): 9, (6, 4): 12, (5, 6): 3, (8, 4): 10}</info>
<text>Write a function to sort dictionary items by tuple product of keys for the given dictionary with tuple keys.</text>
<code>def sort_dict_item(test_dict):
  res = {key: test_dict[key] for key in sorted(test_dict.keys(), key = lambda ele: ele[1] * ele[0])}
  return  (res)
</code>
<info>assert reverse_list_lists([[1, 2, 3, 4], [5, 6, 7, 8], [9, 10, 11, 12], [13, 14, 15, 16]])==[[4, 3, 2, 1], [8, 7, 6, 5], [12, 11, 10, 9], [16, 15, 14, 13]]</info>
<text>Write a function to reverse each list in a given list of lists.</text>
<code>def reverse_list_lists(lists):
    for l in lists:
        l.sort(reverse = True)
    return lists </code>
<info>assert remove_Occ(\"hello\",\"l\") == \"heo\"</info>
<text>Write a python function to remove first and last occurrence of a given character from the string.</text>
<code>
    \end{lstlisting} \\[-5pt]
    \midrule
    \midrule
    \textbf{MBPP: Response} \\
    \midrule ~\\[-25pt]
    \begin{lstlisting}[breaklines=true]
def remove_Occ(str1,ch):
    return str1[:str1.index(ch)] + str1[str1.rindex(ch)+1:]</code>
    \end{lstlisting} \\[-5pt]
    \bottomrule
    \end{tabular}
    \caption[MBPP example prompt and response from Codex.]{\label{tab:mbrexec-real-prompt-mbpp} MBPP example prompt and response from Codex: we use the first assertion in the dataset as the extra information (i.e., \texttt{[INFO]} in Table~\ref{tab:mbrexec-prompt-format}). The content in the last \texttt{<info>...</info>} and \texttt{<text>...</text>} marks in the prompt corresponds to the test problem.}
\end{table}

%% file: tables/705-mbrexec-detailed-prompt-spider.tex
\begin{table}[ht!]
    \centering \scriptsize
    \begin{tabular}{p{0.96\textwidth}}
    \toprule
    \textbf{Spider: Prompt} \\
    \midrule ~\\[-25pt]
    \begin{lstlisting}[breaklines=true]
<info>e_learning | * | Course_Authors_and_Tutors : author_id , author_tutor_ATB , login_name , password , personal_name , middle_name , family_name , gender_mf , address_line_1 | Students : student_id , date_of_registration , date_of_latest_logon , login_name , password , personal_name , middle_name , family_name | Subjects : subject_id , subject_name | Courses : course_id , author_id , subject_id , course_name , course_description | Student_Course_Enrolment : registration_id , student_id , course_id , date_of_enrolment , date_of_completion | Student_Tests_Taken : registration_id , date_test_taken , test_result</info>
<text>Which course authors teach two or more courses? Give me their addresses and author IDs.</text>
<code>SELECT T1.address_line_1 ,  T2.author_id FROM Course_Authors_and_Tutors AS T1 JOIN Courses AS T2 ON T1.author_id  =  T2.author_id GROUP BY T2.author_id HAVING Count(*)  >=  2</code>
<info>flight_1 | * | flight : flno , origin , destination , distance , departure_date , arrival_date , price , aid | aircraft : aid , name , distance | employee : eid , name , salary | certificate : eid , aid</info>
<text>Show origin and destination for flights with price higher than 300.</text>
<code>SELECT origin ,  destination FROM Flight WHERE price  >  300</code>
<info>driving_school | * | Addresses : address_id , line_1_number_building , city , zip_postcode , state_province_county , country | Staff : staff_id , staff_address_id , nickname , first_name , middle_name , last_name , date_of_birth , date_joined_staff , date_left_staff | Vehicles : vehicle_id , vehicle_details | Customers : customer_id , customer_address_id , customer_status_code , date_became_customer , date_of_birth , first_name , last_name , amount_outstanding , email_address , phone_number , cell_mobile_phone_number | Customer_Payments : customer_id , datetime_payment , payment_method_code , amount_payment | Lessons : lesson_id , customer_id , lesson_status_code , staff_id , vehicle_id , lesson_date , lesson_time , price</info>
<text>When did the staff member with first name as Janessa and last name as Sawayn leave the company?</text>
<code>SELECT date_left_staff FROM Staff WHERE first_name = \"Janessa\" AND last_name = \"Sawayn\";</code>
<info>concert_singer | * | stadium : Stadium_ID , Location , Name , Capacity , Highest , Lowest , Average | singer : Singer_ID , Name , Country , Song_Name , Song_release_year , Age , Is_male | concert : concert_ID , concert_Name , Theme , Stadium_ID , Year | singer_in_concert : concert_ID , Singer_ID</info>
<text>How many singers do we have?</text>
<code>
    \end{lstlisting}\\[-5pt]
    \midrule
    \midrule
    \textbf{Spider: Response} \\
    \midrule ~\\[-25pt]
    \begin{lstlisting}[breaklines=true]
SELECT COUNT(*) FROM singer;</code>
    \end{lstlisting} \\[-5pt]
    \bottomrule
    \end{tabular}
    \caption[Spider example prompt and response from Codex.]{\label{tab:mbrexec-real-prompt-spider} Spider example prompt and response from Codex: following \citet{xie-etal-2022-unifiedskg}, we use the concatenation of corresponding table and column names in the dataset as the extra information (i.e., \texttt{[INFO]} in Table~\ref{tab:mbrexec-prompt-format}). The content in the last \texttt{<info>...</info>} and \texttt{<text>...</text>} marks in the prompt corresponds to the test problem.}
\end{table}

%% file: tables/706-mbrexec-detailed-prompt-nl2bash.tex
\begin{table}[t!]
    \centering \scriptsize
    \begin{tabular}{p{0.96\textwidth}}
    \toprule
    \textbf{NL2Bash: Prompt} \\
    \midrule ~\\[-25pt]
    \begin{lstlisting}[breaklines=true]
<text>Print file information of command \"bash\"</text>
<code>echo $(ls -l $(which bash))</code>
<text>Recursively change the owner and group of all files in \"/your/directory/to/fuel/\" to \"nginx\"</text>
<code>chown nginx:nginx /your/directory/to/fuel/ -R</code>
<text>Copy \"src/prog.js\" and \"images/icon.jpg\" to \"/tmp/package\" keeping relative path names</text>
<code>rsync -R src/prog.js images/icon.jpg /tmp/package</code>
<text>Adds execution permissions on a script ./etc/bash_completion within Homebrew home folder path.</text>
<code>
    \end{lstlisting} \\[-5pt]
    \midrule
    \midrule
    \textbf{NL2Bash: Response} \\
    \midrule ~\\[-25pt]
    \begin{lstlisting}
chmod +x /usr/local/etc/bash_completion</code>
    \end{lstlisting}[breaklines=true] \\[-0pt]
    \bottomrule
    \end{tabular}
    \caption[NL2Bash example prompt and response from Codex.]{\label{tab:mbrexec-real-prompt-nl2bash} NL2Bash example prompt and response from Codex: we did not use any extra information. The content in the last \texttt{<text>...</text>} marks in the prompt corresponds to the test problem.}
\end{table}

%% file: figures/702-mbrexec-main-results.tex
\begin{figure}[t!]
    \centering
    \begin{subfigure}[t]{0.48\textwidth}
        \includegraphics[width=\textwidth]{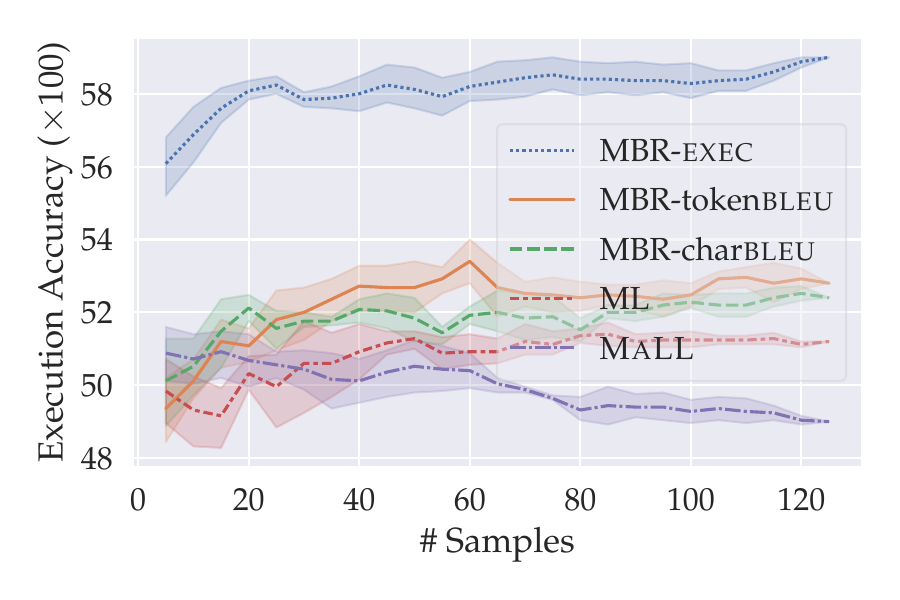}
        \caption{MBPP}
    \end{subfigure}
    \begin{subfigure}[t]{0.48\textwidth}
        \includegraphics[width=\textwidth]{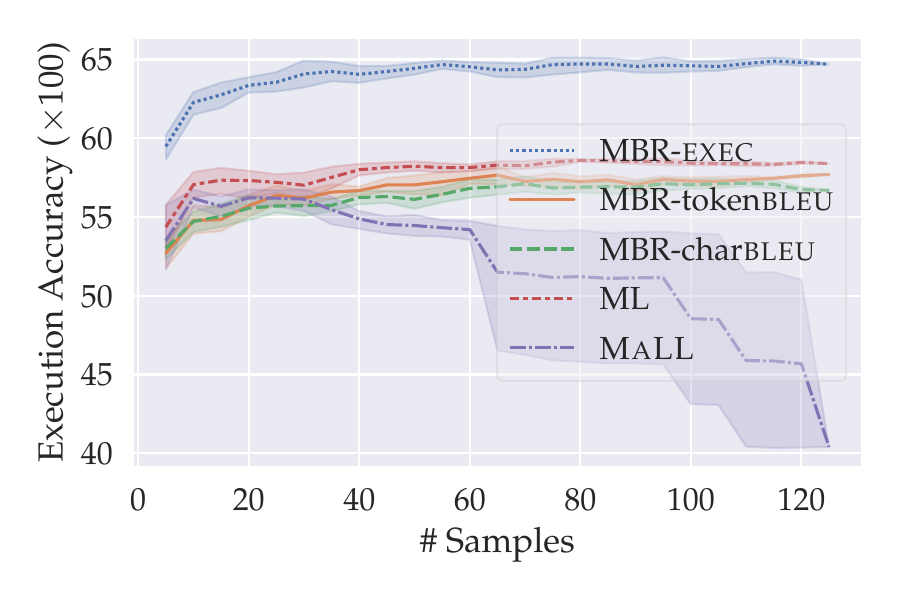}
        \caption{Spider}
    \end{subfigure}
    \begin{subfigure}[t]{0.48\textwidth}
        \includegraphics[width=\textwidth]{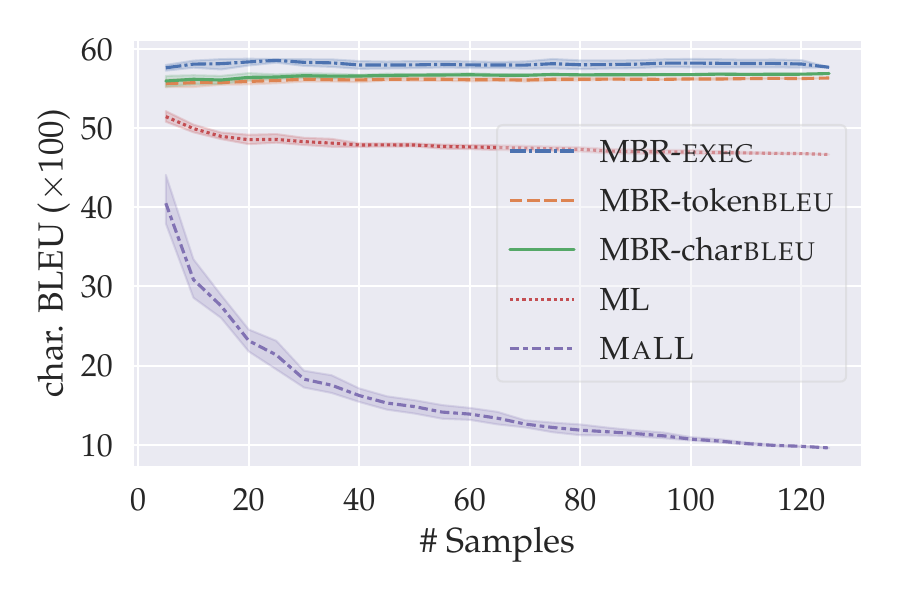}
        \caption{NL2Bash}
    \end{subfigure}
    \caption[Primary evaluation results of \mbrexec.]{Primary evaluation results of \mbrexec: performance of the evaluated selection criteria (best viewed in color). For each sample size, we evaluate the methods on five different groups of samples and report the average performance (lines) and the standard deviations (shaded regions). All samples are collected from Codex with a temperature of 0.3.}
    \label{fig:mbrexec-main-results}
\end{figure}

%% file: tables/702-mbrexec-main-comparisons.tex
\begin{table}[t!]
    \centering \small
    \begin{tabular}{lccc}
        \toprule
        \textbf{Method} & \textbf{MBPP} & \textbf{Spider} & \textbf{NL2Bash} \\
        \midrule
        Greedy (3-shot) &  $47.3\pm2.5$ \hspace{-6pt} & $50.8\pm2.6$ \hspace{-6pt}   & $52.8\pm2.9$ \\
        Sample (3-shot) &  $47.7\pm1.5$ \hspace{-6pt} & $48.5\pm2.6$ \hspace{-6pt}  &  $53.0\pm2.9$ \\
        \midrule
        \mbrexec &  $\textbf{58.2}\pm0.3$ \hspace{-6pt} &  $\textbf{63.6}\pm 0.8$  \hspace{-6pt} &  $\textbf{58.5}\pm 0.3$ \\
        \bottomrule
    \end{tabular}
    \caption[Comparison between \mbrexec and baselines without selection process.]{Comparison between \mbrexec and baselines without a selection process. For both \mbrexec and Sample (3-shot), we collected samples with a temperature of 0.3. All numbers involve the same set of 125 samples for each case: for greedy and sample baselines, we report the average performance of them all; for \mbrexec, we report the result with 25 examples, averaged across five experiments. }
    \label{tab:mbrexec-main-results-basic-baseline}
\end{table}

%% file: figures/703-mbrexec-temperature.tex
\begin{figure}[t!]
    \centering
    \begin{subfigure}[t]{0.48\textwidth}
        \includegraphics[width=\textwidth]{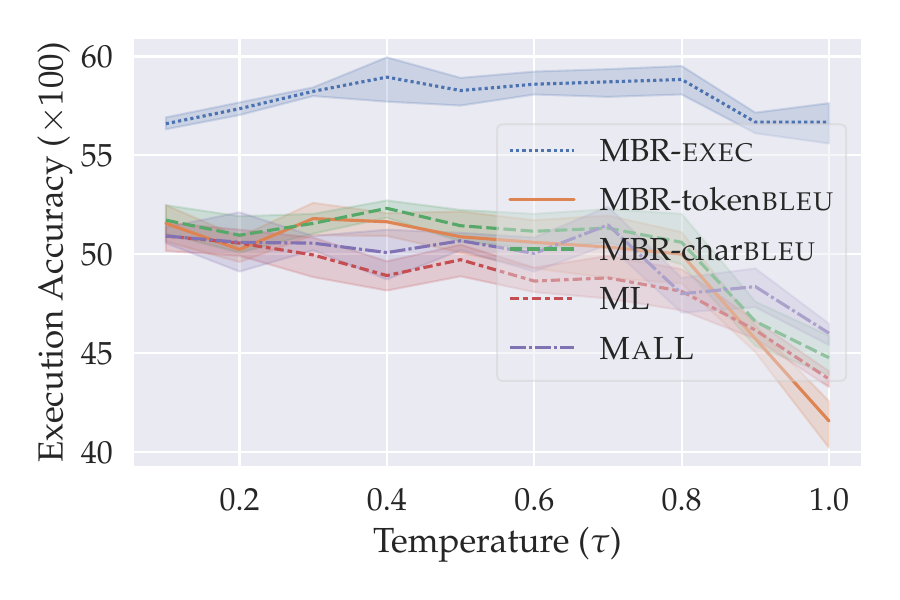}
        \caption{MBPP}
    \end{subfigure}
    \begin{subfigure}[t]{0.48\textwidth}
        \includegraphics[width=\textwidth]{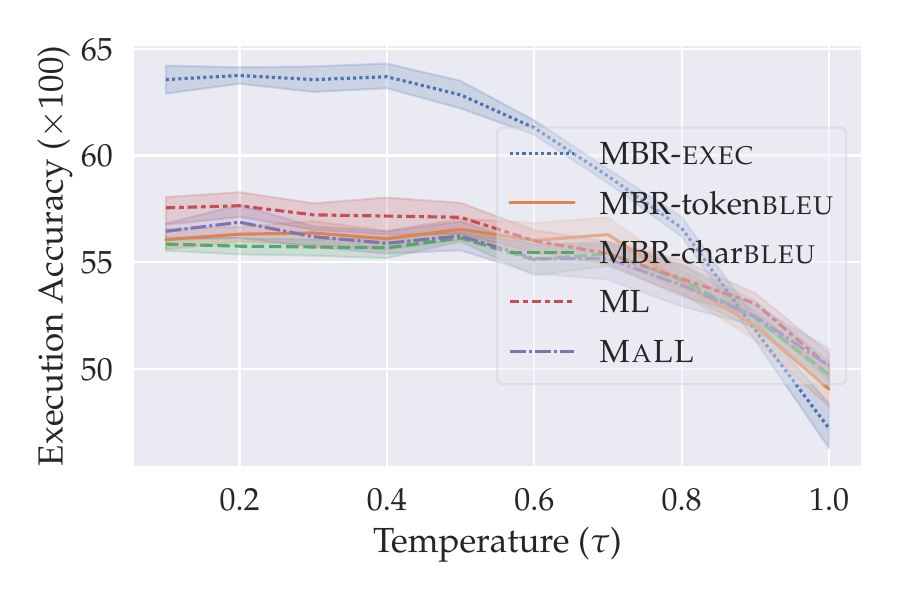}
        \caption{Spider}
    \end{subfigure}
    \begin{subfigure}[t]{0.48\textwidth}
        \includegraphics[width=\textwidth]{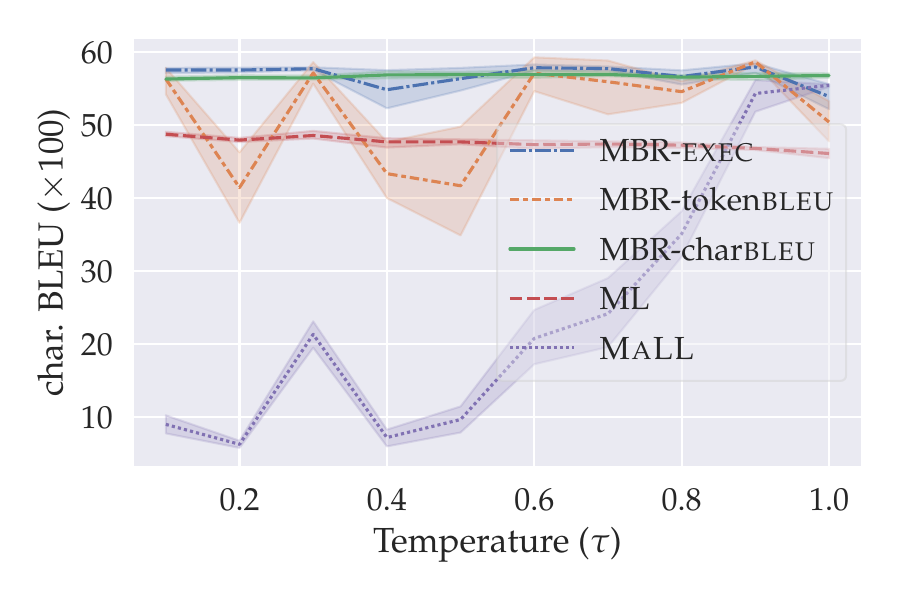}
        \caption{NL2Bash}
    \end{subfigure}
    \caption[Performance of the evaluated selection criteria across temperatures.]{Performance of the evaluated selection criteria across temperatures (best viewed in color). We perform the methods for each temperature on five groups of 25 examples and report the average performance (lines) and the standard deviations (shaded regions).}
    \label{fig:mbrexec-temperature}
\end{figure}

%% file: tables/703-mbrexec-0.3-vs-greedy.tex
\begin{table}[t!]
    \centering
    \small
    \begin{tabular}{lcc}
        \toprule
        \textbf{Dataset} & \textbf{Greedy ($\tau=0$)} & \textbf{Sample ($\tau=0.3$)} \\
        \midrule
        MBPP    & $56.0$    &  $\textbf{58.2}\pm0.3$\\
        Spider  & $62.1$    &  $\textbf{63.6}\pm0.8$\\
        NL2Bash & $58.4$    &  $\textbf{58.5}\pm0.3$\\
        \bottomrule
    \end{tabular}
    \caption[\mbrexec performance on greedily decoded and sampled programs.]{
        \label{tab:mbrexec-0.3-vs-greedy} \mbrexec performance on greedily decoded and sampled programs: for each problem, we use 25 groups of 3-shot prompts, decode or sample one program with each prompt, and use \mbrexec to select the best program.
        For sampling with temperature 0.3, we repeat the process five times and report the average performance and standard deviations.
        The dataset-specific metric can be found at \cref{sec:mbrexec-expr-datasets}.
        The best number in each row is in boldface.
        Note that the greedy performances are different from those reported in \cref{tab:mbrexec-main-results-basic-baseline}, as we perform \mbrexec here over greedy decoding outputs while reporting the average performance in \cref{tab:mbrexec-main-results-basic-baseline}.
    }
\end{table}

%% file: figures/704-mbrexec-15-shot.tex
\begin{figure}[t!]
    \centering
    \begin{subfigure}[t]{0.48\textwidth}
        \includegraphics[width=\textwidth]{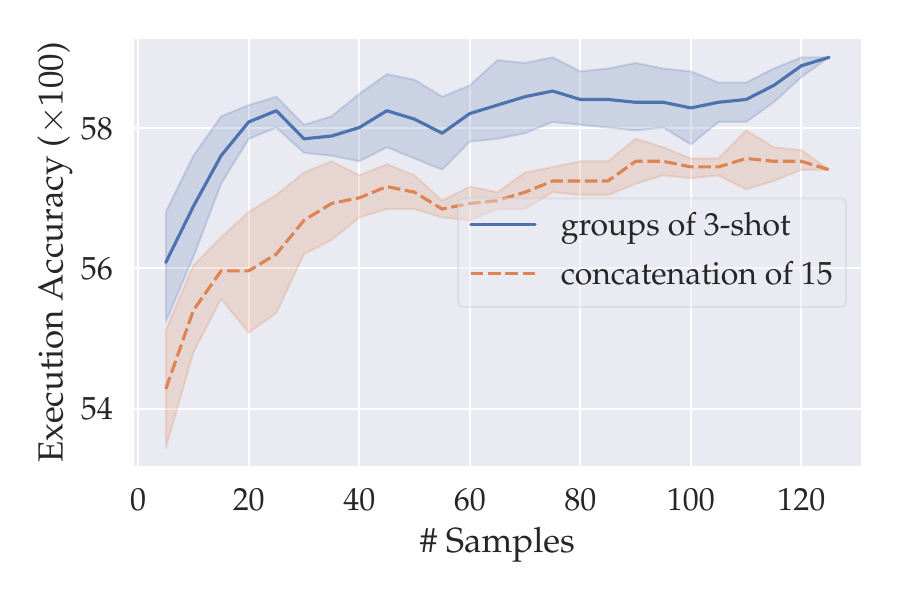}
        \caption{MBPP}
    \end{subfigure}
    \begin{subfigure}[t]{0.48\textwidth}
        \includegraphics[width=\textwidth]{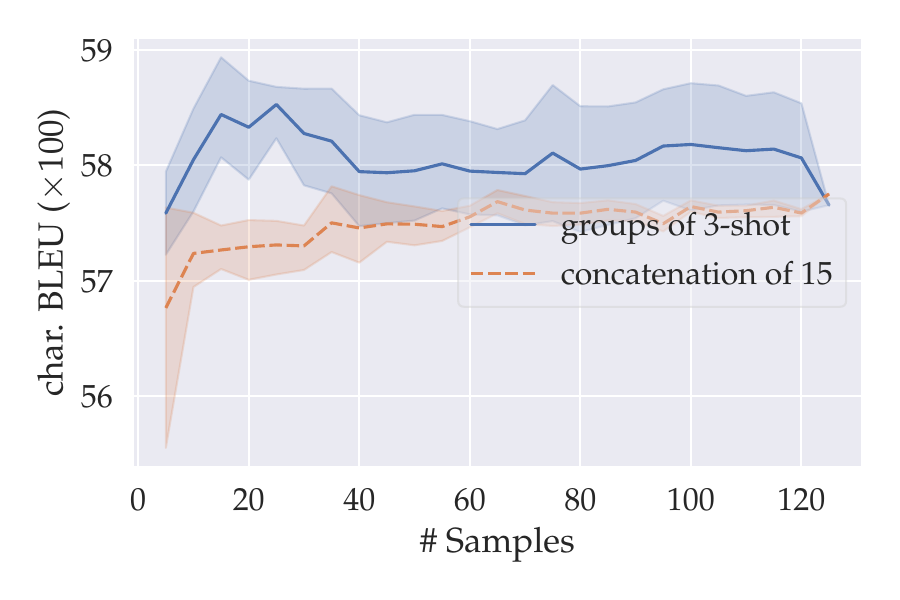}
        \caption{NL2Bash}
    \end{subfigure}
    \caption[Performance with groups of 3-shot prompt vs. concatenation of 15 prompts.]{Performance with different types of prompts, where \textit{groups of 3-shot} denotes the prompt formatting in \cref{tab:mbrexec-prompt-format}, while \textit{concatenation of 15} denotes concatenating all available 15 examples as prompts for data collection. }
    \label{fig:mbrexec-15-shot}
\end{figure}

%% file: figures/705-mbrexec-executability.tex
\begin{figure}[ht!]
    \centering
    \begin{subfigure}[t]{0.45\textwidth}
        \includegraphics[width=\textwidth]{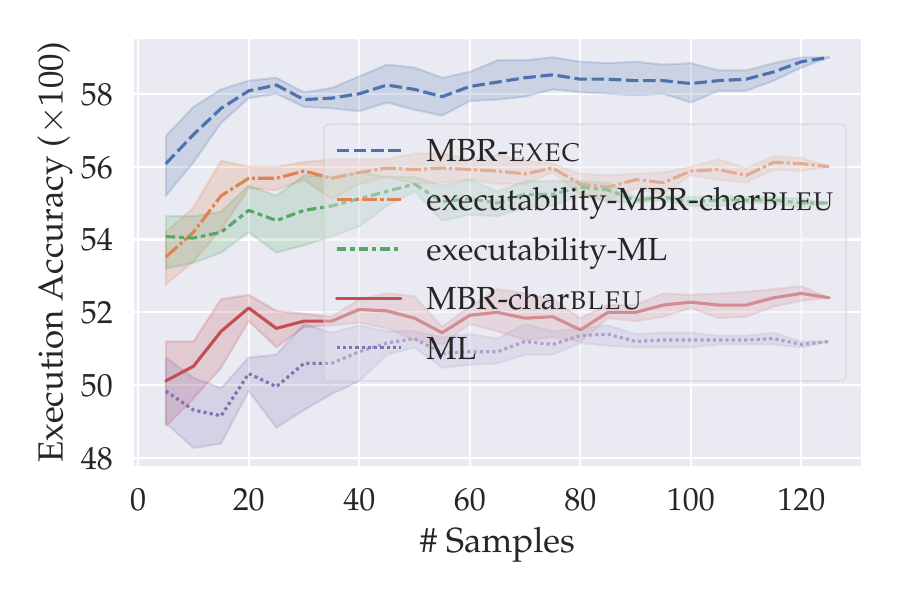}

        \vspace{-10pt}
        \caption{MBPP}
    \end{subfigure}
    \hspace{-10pt}
    \begin{subfigure}[t]{0.45\textwidth}
        \includegraphics[width=\textwidth]{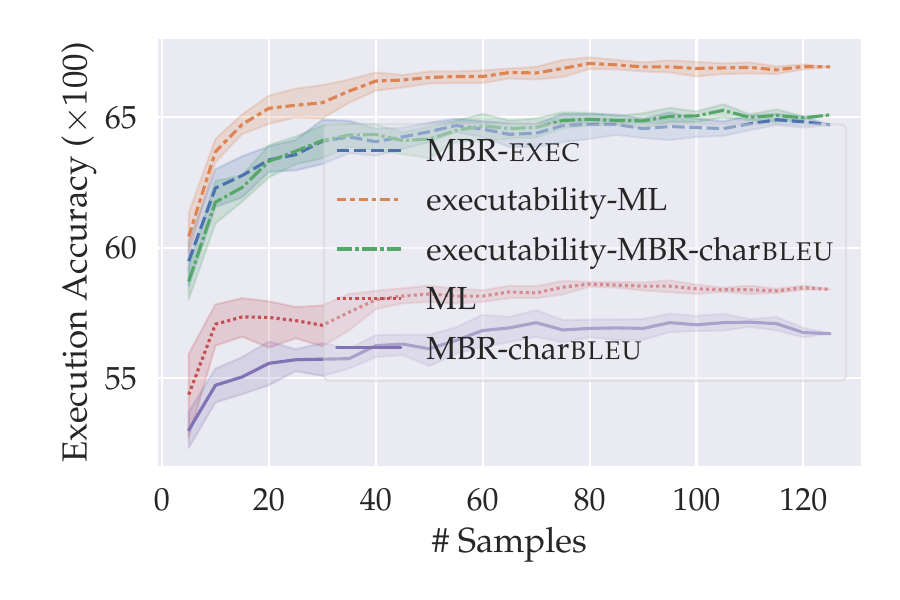}

        \vspace{-10pt}
        \caption{Spider}
    \end{subfigure}
    \hspace{-10pt}
    \begin{subfigure}[t]{0.45\textwidth}
        \includegraphics[width=\textwidth]{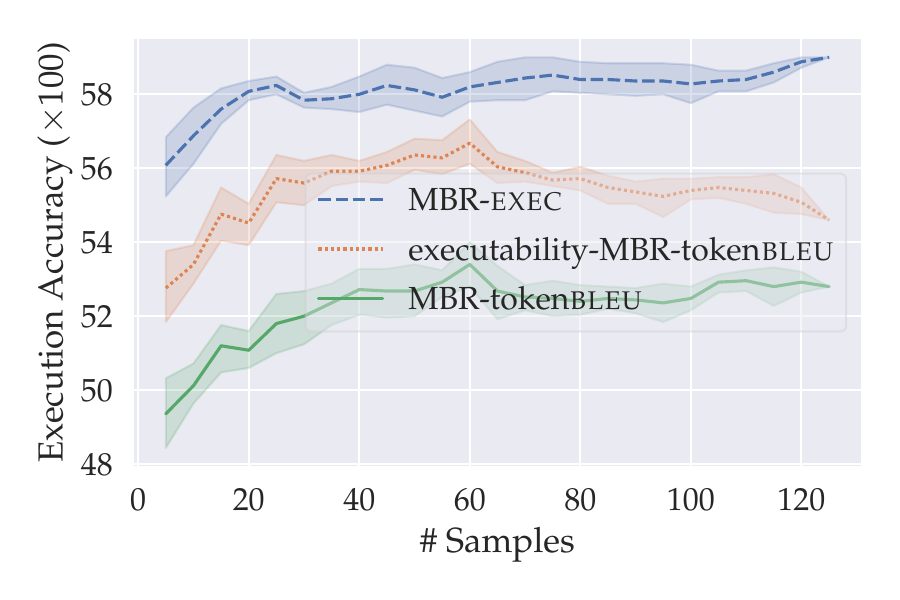}

        \vspace{-10pt}
        \caption{MBPP (MBR-token\textsc{BLEU})}
    \end{subfigure}
    \hspace{-10pt}
    \begin{subfigure}[t]{0.45\textwidth}
        \includegraphics[width=\textwidth]{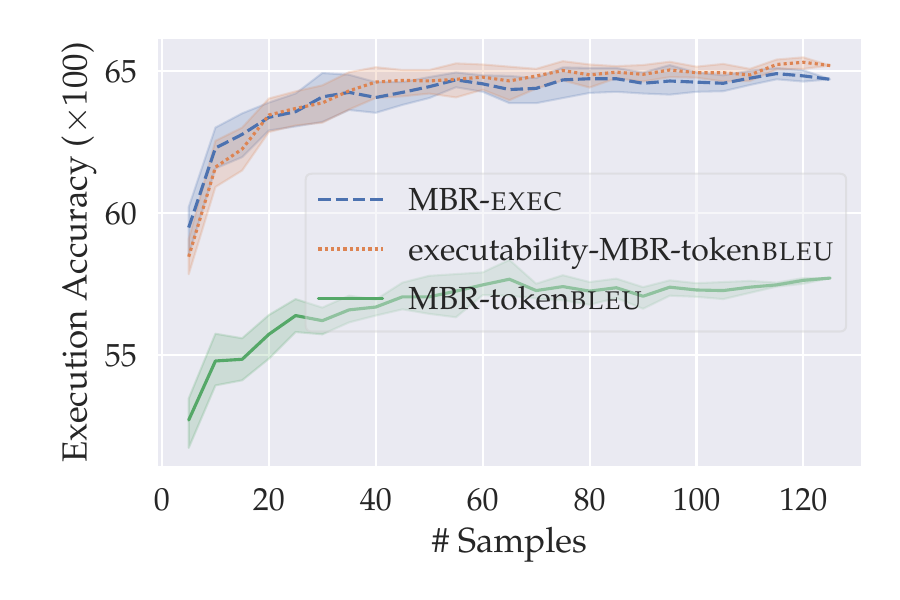}

        \vspace{-10pt}
        \caption{Spider (MBR-token\textsc{BLEU})}
    \end{subfigure}
    \hspace{-10pt}
    \begin{subfigure}[t]{0.45\textwidth}
        \includegraphics[width=\textwidth]{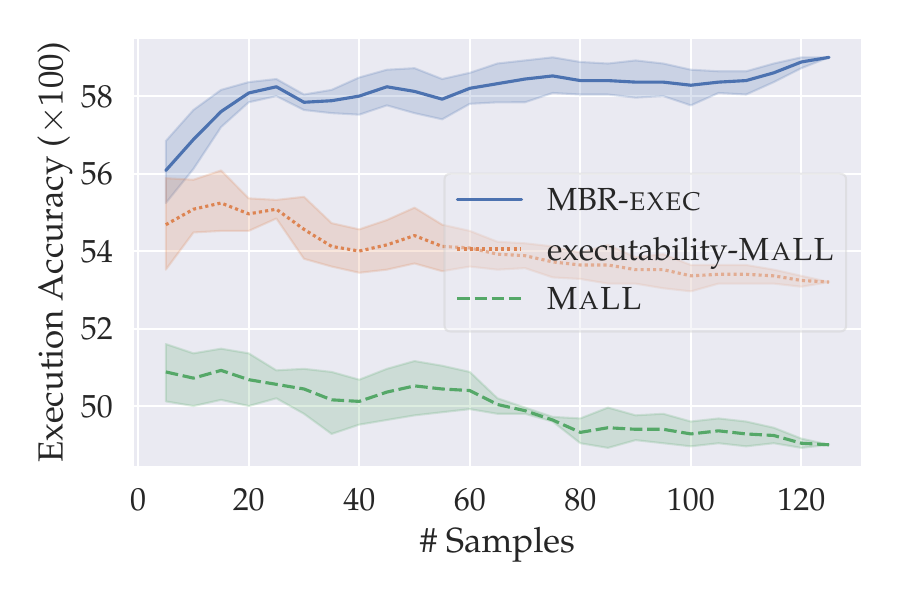}

        \vspace{-10pt}
        \caption{MBPP (\maxavglikelihood)}
    \end{subfigure}
    \hspace{-10pt}
    \begin{subfigure}[t]{0.45\textwidth}
        \includegraphics[width=\textwidth]{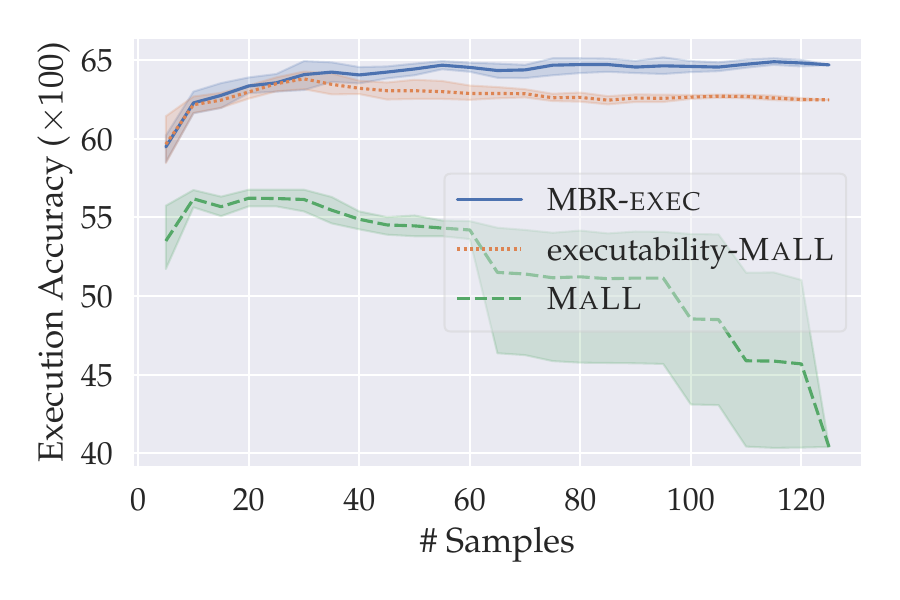}

        \vspace{-10pt}
        \caption{Spider (\maxavglikelihood)}
    \end{subfigure}
    \caption[Comparison between applying methods to all possible candidates vs. applying methods to only executable candidates.]{Comparison between applying methods to all possible candidates vs. applying methods to only executable candidates (best viewed in color), where executability-$X$ denotes applying selection criteria $X$ on executable candidates only. We present MBR-token\textsc{BLEU} and \maxavglikelihood and their combination with executability check separately for clarity.}
    \label{fig:mbrexec-ablation-executability}
\end{figure}

%% file: figures/706-mbrexec-soft-hard.tex
\begin{figure}[t!]
    \centering
    \includegraphics[width=0.48\textwidth]{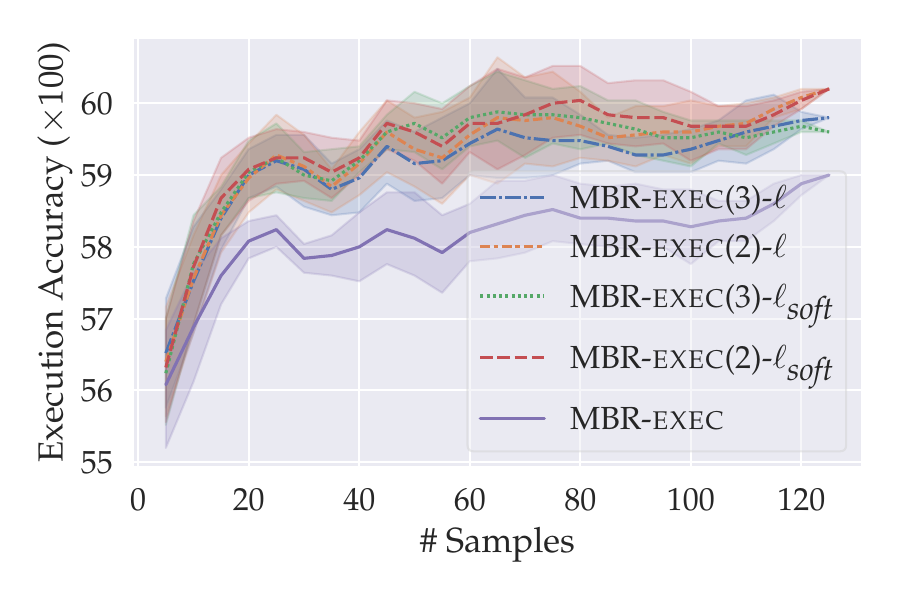}
    \caption[Comparison between soft and hard \mbrexec.]{Execution accuracies with respect to sample size on the MBPP dataset, where the number in the parentheses denotes the number of test cases per problem used for \mbrexec. Best viewed in color.}
    \label{fig:mbrexec-mbpp-hard-soft-exec}
\end{figure}

%% file: figures/707-mbrexec-oracle.tex
\begin{figure}[t!]
    \centering
    \begin{subfigure}[t]{0.48\textwidth}
        \includegraphics[width=\textwidth]{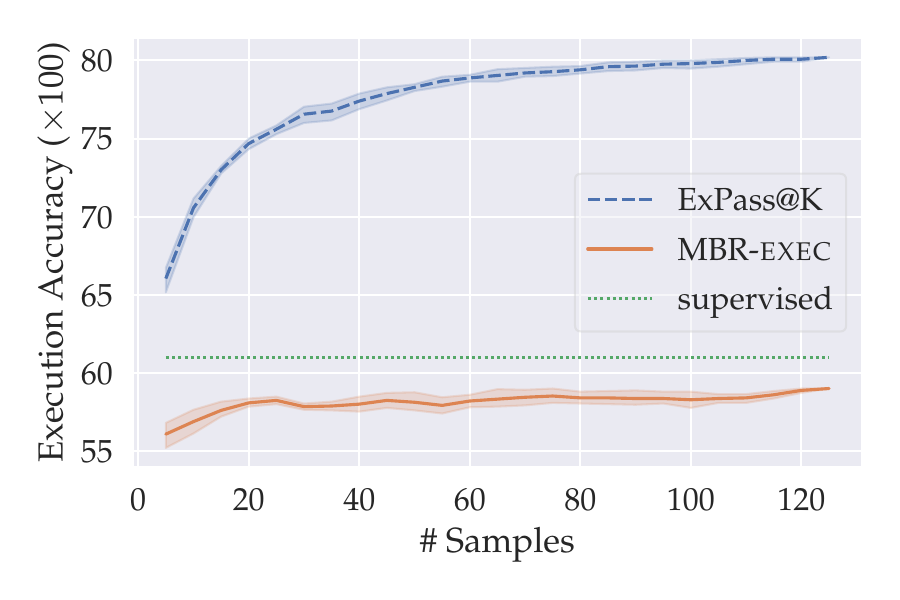}
        \caption{MBPP}
    \end{subfigure}
    \hspace{-10pt}
    \begin{subfigure}[t]{0.48\textwidth}
        \includegraphics[width=\textwidth]{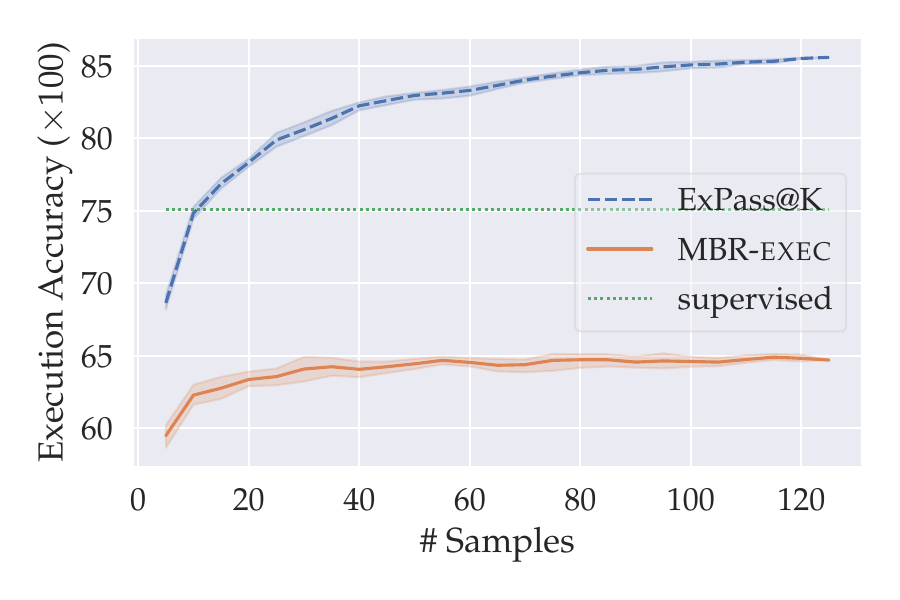}
        \caption{Spider}
    \end{subfigure}
    \hspace{-10pt}
    \begin{subfigure}[t]{0.48\textwidth}
        \includegraphics[width=\textwidth]{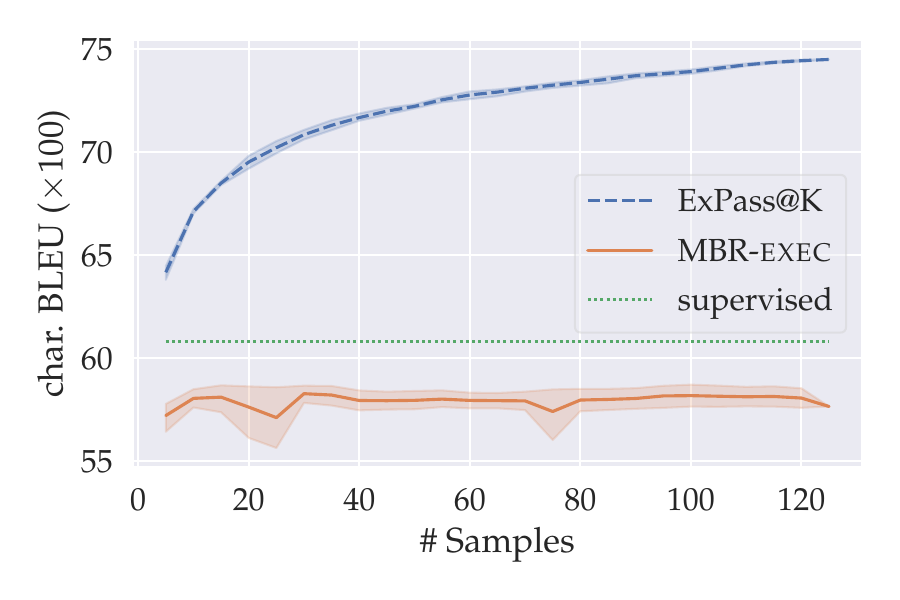}
        \caption{NL2Bash}
    \end{subfigure}
    \caption[Sample size--oracle performance curves on the considered datasets.]{Sample size--oracle performance curves on the considered datasets. We calculate each expected Pass@K with five different sets of candidates for each sample size while using the same sets to perform \mbrexec for fair comparison. }
    \label{fig:mbrexec-oracle}
\end{figure}

%% file: src/80-mlpalign.tex
\chapter{From Pre-Trained Contextualized Representations to Unsupervised Cross-Lingual Word Alignment}
\label{chapter:mlpalign}
\textit{Content in this has been published as part of a conference paper at ACL 2021 \citep{shi2021bilingual}, of which the main research question is on bilingual lexicon induction.}

This chapter presents a method for learning cross-lingual word alignment from contextualized representations.
The task of cross-lingual word alignment aims to find the word correspondences between two mutually translatable sentences in different languages (\cref{fig:mlpalign-example}).
Prior work, SimAlign \citep{sabet-etal-2020-simalign}, has proposed to directly adapt pre-trained contextualized multilingual language models, calculate the similarity between the representations of words in different languages, and then use different inference strategies to obtain the final alignment.
Following SimAlign, our method, which is initially designed for bilingual lexicon induction \citep{shi2021bilingual}, fuses multiple inference strategies to improve the alignment quality by learning a multi-layer perceptron (MLP) that takes statistical features, such as word frequency, cooccurring word pairs, and the similarity between the representations of words in different languages, as the input, and predicts the alignment probability.
Results on cross-lingual word alignment benchmarks show that our method outperforms SimAlign and builds a new state of the art for the task.

\input{figures/801-mlpalign-example.tex}

\section{Related Work}
\paragraph{Word alignment.}
Word alignment is a fundamental problem in statistical machine translation, of which the goal is to align words that are translations of each within parallel sentences \citep{brown-etal-1993-mathematics}. Most methods assume parallel sentences for training data \interalia{och-ney-2003-systematic,dyer-etal-2013-simple,peter-etal-2017-generating}.
Following the approach of IBM models \citep{brown-etal-1993-mathematics}, existing work has achieved decent performance with statistical methods \citep{och-ney-2003-systematic,dyer-etal-2013-simple}.
Recent work has demonstrated the potential of neural networks on the task \interalia{peter-etal-2017-generating,garg-etal-2019-jointly,zenkel-etal-2019-adding}.
In contrast, \citet{sabet-etal-2020-simalign} propose SimAlign, which does not train on parallel sentences but instead aligns words that have the most similar pre-trained multilingual representations \citep{devlin-etal-2019-bert,conneau-etal-2020-unsupervised}. SimAlign achieves competitive or superior performance compared to conventional alignment methods despite not using parallel sentences, and it provides one of the baseline components for our work.
This work presents a simple yet effective method to improve performance over SimAlign (\cref{sec:mlpalign-word-alignment}).

\paragraph{Bitext mining/parallel corpus mining.}
Bitext mining has been a long studied task \interalia{resnik-1999-mining,shi-etal-2006-dom,abdul-rauf-schwenk-2009-use}.
Most methods train neural multilingual encoders on bitext, which are then used with efficient nearest neighbor search to expand the training set \interalia{espana2017empirical,schwenk-2018-filtering,guo-etal-2018-effective,artetxe-schwenk-2019-margin}.
Recent work has also shown that unsupervised mining is possible~\citep{tran-etal-2020-cross,keung-etal-2020-unsupervised}. We use CRISS \citep{tran-etal-2020-cross}\footnote{\url{https://github.com/pytorch/fairseq/tree/master/examples/criss}} as one of our component models.

\section{Method}
\label{sec:mlpalign-word-alignment}

\subsection{Unsupervised Bitext Construction}
We follow \citet{tran-etal-2020-cross} to perform bitext construction through encoding-based retrieval: taking the average across the contextualized embeddings of tokens as sentence representation, we perform nearest neighbor search, and mine bitext using the margin-based max-score method \citep{artetxe-schwenk-2019-margin}.
We use CRISS \citep{tran-etal-2020-cross} as the base model in this work.

The margin score between sentence representations $\mathbf{s}$ and $\mathbf{t}$ is defined by
\begin{align}
    \label{eq:mlpalign-margin-score}
    \textit{score}(\mathbf{s}, \mathbf{t}) = \frac{\cos\left(\mathbf{s}, \mathbf{t}\right)}{\sum_{\mathbf{t}'\in \textit{NN}_k(\mathbf{t})}\frac{\cos(\mathbf{s}, \mathbf{t}')}{2k} + \sum_{\mathbf{s}'\in \textit{NN}_k(\mathbf{s})}\frac{\cos(\mathbf{s}', \mathbf{t})}{2k}},
\end{align}
where $\textit{NN}_k(\cdot)$ denotes the set of $k$ nearest neighbors of a vector in the corresponding space.
In this work, we keep the top 20\% of the sentence pairs with scores larger than one as the constructed bitext.

\subsection{Feature-Based Word Aligner}
Let $\mathcal{B} = \{\langle \mathbf{s}_i, \mathbf{t}_i \rangle \}_{i=1}^N$ denote the constructed bitext, where $N$ denotes the number of sentence pairs, and $\mathbf{s}_i$ and $\mathbf{t}_i$ denote a pair of sentences in the source and target language respectively.
In a pair of bitext $\langle \mathbf{s}, \mathbf{t}\rangle$, $\mathbf{s} = \langle s_1, \ldots, s_{\ell_s} \rangle$ and $\mathbf{t} = \langle t_1, \ldots, t_{\ell_t} \rangle$ denote sentences consisting of $\ell_s$ and $\ell_t$ word tokens respectively.

For a pair of bitext, \simalign with a specified inference algorithm produces word alignment $\mathcal{A} = \{\langle a_i, b_i \rangle\}_i$, denoting that the word tokens $s_{a_i}$ and $t_{b_i}$ are aligned.
\citet{sabet-etal-2020-simalign} has proposed different algorithms to induce alignment from the same similarity matrix, and the best method varies across language pairs.
In this work, we consider the relatively conservative (i.e., having higher precision) \textit{argmax} and the higher recall \textit{itermax} algorithm, and denote the alignments by $\mathcal{A}_\textit{argmax}$ and $\mathcal{A}_\textit{itermax}$ respectively.
Concretely, let $\bm{S} \in \mathbb{R}^{\ell_s\times \ell_t}$ denote the similarity matrix between two sentences, the \textit{argmax} algorithm aligns any word pair $\langle s_i, t_j \rangle$ that satisfies
\begin{align*}
    i &= \argmax_{i'} \bm{S}_{i',j}, \\
    j &= \argmax_{j'} \bm{S}_{i,j'}.
\end{align*}
A word pair $\langle s_i, t_j \rangle$ will be aligned by the \textit{itermax} algorithm if their similarity is both row-wise and column-wise maximum in the similarity matrix.
The \textit{itermax} algorithm applies \textit{argmax} iteratively---at each step, it appends all word pairs selected by \textit{argmax} to the alignment set and mask-out those selected positions,\footnote{Empirically, this can be done by setting the corresponding row and column of the similarity matrix to a sufficiently small value.} and repeats the process until no word pair can be selected.
It is obvious that $\mathcal{A}_\textit{argmax} \subseteq \mathcal{A}_\textit{itermax}$, since the first step of the latter produces the same result as the former.

We consider the following features for each word token pair $\langle s_i, t_j \rangle$:
\begin{itemize}
    \setlength\itemsep{0em}
    \item Count of alignment: we consider both one-to-one alignment and many-to-one alignment of $s$ and $t$ separately as two features since the task of lexicon induction is arguably biased toward one-to-one alignment.
    \item Count of co-occurrence in the mined bitext.
    \item The count of $s$ in the source language and $t$ in the target language.
    \item Contextualized word similarity: we feed the words in contexts into CRISS, use the average pooling of the output subword embeddings, and consider both cosine similarity and dot-product similarity as features.
\end{itemize}

We formulate word alignment as the task of ternary classification. For a pair of word tokens $\langle s_i, t_j \rangle$, the gold label $y_{\langle s_i, t_j\rangle}$ is defined as
\newcommand\1{\mathbbm{1}}
\begin{align*}
\1 [\langle i,j \rangle \in \mathcal{A}_\textit{argmax} ] +\1 [\langle i,j \rangle \in \mathcal{A}_\textit{itermax} ].
\end{align*}
Intuitively, the labels $0$ and $2$ represent confident alignment or non-alignment by both methods, while the label $1$ models the potential alignment.

The MLP takes the features $\mathbf{x}_{\langle s_i,t_j \rangle} \in \mathbb{R}^7$ of the word token pair and computes the probability of each label $y$ by
\begin{align*}
    \bm{\hat{h}} &= \mathrm{ReLU}\left(\mathbf{W}_1\mathbf{x}_{\langle s_i,t_j\rangle} + \mathbf{b}_1\right) \\
    \bm{g} &= \mathbf{W}_2 \cdot \bm{\hat{h}} + \mathbf{b}_2  \\
    P_\Phi(y \mid  s_i, t_j, \mathcal{S}, \mathcal{T}) &= \frac{\exp\left(g_y\right)}{\sum_{y'}\exp\left(g_{y'}\right)},
\end{align*}
where $\Phi = \{\mathbf{W}_1 \mathbf{W}_2, \mathbf{b}_1, \mathbf{b}_2\}$.
On the training stage, we maximize the log-likelihood of ground-truth labels:
\begin{align*}
    \Phi^* & = \arg\max_{\Phi} \sum_{\langle\mathcal{S}, \mathcal{T}\rangle \in \mathcal{B}}\sum_{s_i \in \mathcal{S}}\sum_{t_j \in \mathcal{T}} \log P_\Phi(y_{\langle s_i,t_j \rangle} \mid s_i,t_j, \mathcal{S}, \mathcal{T}).
\end{align*}
On the inference stage, we keep all word token pairs $\langle s_i, t_j\rangle$ that have
\begin{align*}
    \mathbb{E}_P[y] := \sum_y y\cdot P(y \mid s_i,t_j,\mathcal{S},\mathcal{T}) > 1
\end{align*}
as the prediction.

\section{Experiments}
\input{tables/801-mlpalign-results.tex}
We evaluate different word alignment methods (\cref{tab:word-alignment}) on existing word alignment datasets,\footnote{\url{http://www-i6.informatik.rwth-aachen.de/goldAlignment}~(de-en); \\
\url{https://web.eecs.umich.edu/~mihalcea/wpt} (en-fr and ro-en); \\
\url{https://web.eecs.umich.edu/~mihalcea/wpt05} (en-hi)}  following \citet{sabet-etal-2020-simalign}.
We investigate four language pairs: German--English (de-en), English--French (en-fr), English--Hindi (en-hi), and Romanian--English (ro-en).
While \citet{sabet-etal-2020-simalign} initially implement SimAlign with an XLM-R backbone, we also consider mBART \citep{liu2020multilingual} and CRISS \citep{tran-etal-2020-cross} as the backbone model.
Indeed, we find that, in general, backbone models that achieve better performance on general-purpose multilingual benchmarks also perform better on word alignment.
The CRISS-based \simalign already achieves competitive performance with the state-of-the-art method \citep{garg-etal-2019-jointly}, which requires supervised bitext for training.
By ensembling the \emph{argmax} and \emph{itermax} CRISS-based \simalign results, we set the new state of the art of word alignment without using any bitext supervision.

\section{Conclusion and Discussion}
This chapter discusses a lightweight and effective method to extract word alignment relations from pre-trained contextualized multilingual language models.
In the next chapter, we will discuss how such cross-lingual word alignment can bridge languages and enable grounding the syntax of one language to another.

Since SimAlign was proposed, it has become common wisdom that the cross-lingual word alignment information is encoded in the representations of pre-trained multilingual language models.
By calculating the similarity between the representations of words in different languages, SimAlign has achieved competitive performance without using parallel sentences, significantly saving the cost of obtaining bitext.
Following this line, our work demonstrates that fully unsupervised word alignment can even outperform methods requiring supervised bitexts.
While the feature-based MLP method is one of many possible ways to improve the alignment quality, it takes advantage of simplicity and interpretability.

While the method was initially designed for bilingual lexicon induction \citep{shi2021bilingual}, where we used statistical features from pre-trained contextualized multilingual models to induce mutually translatable word pairs, it is easily adapted to the word alignment task.
The improvement over single-model SimAlign suggests that pre-trained language models may be generally more potent than directly using their representations.
The method can be further improved by incorporating more sophisticated features and models.

%% file: figures/801-mlpalign-example.tex
\begin{figure}[t]
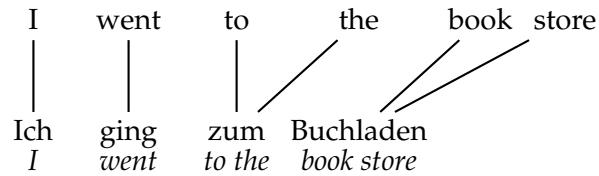

    \small
    \centering
    \begin{dependency}[arc edge, arc angle=70, text only label, label style={above}]
    \begin{deptext}[column sep=.1cm]
    I \& ~~~went~~~ \& to \& the \& book \&store \\[30pt]
    Ich \& ging \& zum \& Buchladen \\
    \small \textit{I} \& \small \textit{went} \& \small \textit{to the} \& \small \textit{book store}\\
    \end{deptext}
    \draw [-, thick, solid] (\wordref{1}{1}) -- (\wordref{2}{1});
    \draw [-, thick, solid] (\wordref{1}{2}) -- (\wordref{2}{2});
    \draw [-, thick, solid] (\wordref{1}{3}) -- (\wordref{2}{3});
    \draw [-, thick, solid] (\wordref{1}{4}) -- (\wordref{2}{3});
    \draw [-, thick, solid] (\wordref{1}{5}) -- (\wordref{2}{4});
    \draw [-, thick, solid] (\wordref{1}{6}) -- (\wordref{2}{4});
    \end{dependency}
    \caption[Example of cross-lingual word alignment in English and German.]{
        Example of cross-lingual word alignment in English and German.
        The lines indicate the word correspondences between the two sentences.
        The word alignment relations include both one-to-one (e.g., I and Ich) and many-to-one (e.g., book store and Buchladen) correspondences.
    }
    \label{fig:mlpalign-example}
\end{figure}

%% file: tables/801-mlpalign-results.tex
\begin{table}[t!]
    \centering \small
    \begin{tabular}{lcccc}
    \toprule
         \bf Model &  de-en & en-fr & en-hi & ro-en \\
    \midrule
        GIZA++ \citep{och-ney-2003-systematic}$^{\dagger}$ & 0.22 & 0.09 & 0.52 & 0.32 \\
        Fast Align \citep{dyer-etal-2013-simple}$^{\dagger}$ & 0.30 & 0.16 & 0.62 & 0.32\\
        \citet{garg-etal-2019-jointly} & 0.16 & 0.05 &  N/A & 0.23\\
        \citet{zenkel-etal-2019-adding} & 0.21 & 0.10 & N/A & 0.28 \\
    \midrule
        \multicolumn{5}{l}{\simalign\ \citep{sabet-etal-2020-simalign}} \\
        \quad XLM-R \citep{conneau-etal-2020-unsupervised}-\emph{argmax}$^{\dagger}$ & 0.19 & 0.07 & 0.39 & 0.29 \\
        \quad mBART \citep{liu2020multilingual}-\emph{argmax} & 0.20 & 0.09 & 0.45 & 0.29 \\
        \quad CRISS \citep{tran-etal-2020-cross}-\emph{argmax}$^{*}$ & 0.17 & 0.05 & 0.32 & 0.25 \\
        \quad CRISS \citep{tran-etal-2020-cross}-\emph{itermax}$^{*}$ & 0.18 & 0.08 & 0.30 & 0.23 \\
        MLP (ours)$^{*}$ & \bf 0.15 & \bf 0.04 & \bf 0.28 & \bf 0.22 \\
    \bottomrule
    \end{tabular}
    \caption[Average error rate (AER) for word alignment.]{Average error rate (AER) for word alignment (lower is better). The best numbers in each column are bolded. Models in the top section require ground-truth bitext, while those in the bottom section do not. $*$: models that involve unsupervised bitext construction. $\dagger$: results copied from \citet{sabet-etal-2020-simalign}. }
    \label{tab:word-alignment}
\end{table}

%% file: src/90-subdp.tex
\chapter{Zero-Shot Cross-Lingual Dependency Parsing with Substructure Distribution Projection}
\label{chapter:subdp}
\textit{Content in \cref{chapter:subdp} has been published as a conference paper at ACL 2022 \citep{shi2022substructure}.}

Zero-shot cross-lingual dependency parsing requires the prediction of dependency parses without seeing any parsing example in the target language; instead, the model may use annotated parses in other languages.
This chapter discusses how high-quality zero-shot cross-lingual dependency parsing can be achieved with unsupervised word alignment \interalia{sabet-etal-2020-simalign,shi2021bilingual} as the bridge between languages.

\input{figures/901-subdp-intro}

We present substructure distribution projection (\subdp), a technique that projects a distribution over structures in one domain to another by projecting substructure distributions separately.
Models for the target domain can then be trained using the projected distributions as soft silver labels.
Our motivation is to address the limitations of existing annotation projection methods, which typically project ``hard'' structures and usually have a low recall for the target structures.\footnote{Throughout this chapter, we refer to dependency parse trees with 0/1 arc and label probabilities, i.e., conventional dependency trees, as \textit{hard trees}; in contrast, we refer to collections of per-word head distributions and per-arc label distributions with continuous probabilities as \textit{soft trees}.}
As illustrated in \cref{fig:subdp-intro-hard}, most annotation projection methods typically output partial hard dependency trees, where there either is or is not an arc between any pair of words.
In addition, most bitext-based work has relied on one-to-one word alignment between bitext pairs \interalia{ma-xia-2014-unsupervised,lacroix-etal-2016-frustratingly,rasooli-etal-2021-wikily}, discarding information in many-to-one alignments.
In contrast, \subdp (\cref{fig:subdp-intro-soft}) projects substructure distributions, i.e., the conditional probability distribution of the corresponding head given a word.\footnote{Projection of the distribution over whole parse trees has been considered by \citet{ma-xia-2014-unsupervised}, while \subdp has a much lower time complexity---see \cref{sec:subdp-related} for more discussion. }
When the source parse is a hard tree, \subdp has the same behavior as prior work \citep[e.g.,][]{lacroix-etal-2016-frustratingly} for arcs that are only involved in one-to-one alignments; for many-to-one alignments, \subdp projects the corresponding arcs into \emph{soft} arc distributions in the target language.
Therefore, in \subdp, a target language word may have multiple heads in the projected trees, where their probabilities sum to one.
More generally, \subdp may take dependency arc or label distributions (i.e., soft trees) in the source language(s) instead of hard trees as the input.
As in annotation projection approaches, the projected soft trees are then used to train a target language parser.

We evaluate \subdp on zero-shot cross-lingual dependency parsing, taking dependency arcs as substructures: we project the predicted dependency arc distributions in the source language(s) to target language(s) and train a target language parser on the resulting distributions.
Given an English treebank as the only source of human supervision,
\subdp achieves higher unlabeled attachment scores than all prior work on the Universal Dependencies v2.2 \citep{nivre-etal-2020-universal} test set across eight diverse target languages, as well as the best-labeled attachment score on six languages.
In addition, \subdp improves zero-shot cross-lingual dependency parsing with very few (e.g., 50) supervised bitext pairs across a broader range of target languages.

\section{Related Work}
\label{sec:subdp-related}
\paragraph{Zero-shot cross-lingual dependency parsing.}\footnote{Also referred to as zero-shot dependency parsing in recent literature \citep{schuster-etal-2019-cross,wang-etal-2019-cross}.}
Existing approaches can be classified into the following categories:
\begin{enumerate}
    \item \textbf{Delexicalized training} \interalia{zeman-resnik-2008-cross,mcdonald-etal-2011-multi,cohen-etal-2011-unsupervised,durrett-etal-2012-syntactic,rosa-zabokrtsky-2015-klcpos3}, which only considers delexicalized features (e.g., part-of-speech tags) in training.
    However, in most recent work, the availability of pre-defined delexicalized features is not assumed.
    \item \textbf{Transfer with cross-lingual embeddings} \interalia{tackstrom-etal-2012-cross,guo-etal-2015-cross,schuster-etal-2019-cross}, which assumes that cross-lingual word representations, including word clusters \citep{tackstrom-etal-2012-cross,ammar-etal-2016-many}, word type embeddings \citep{guo-etal-2015-cross,guo-etal-2016-representation,duong-etal-2015-cross,ammar-etal-2016-many,wick-etal-2016-minimally}, or contextualized cross-lingual word embeddings \citep{schuster-etal-2019-cross,wang-etal-2019-cross,he-etal-2019-cross,ahmad-etal-2019-difficulties,ahmad-etal-2019-cross}, provide shared features for words with similar syntactic roles.
    \item \textbf{Treebank translation}, which translates treebanks in the source language(s) into the target language(s) \citep{tiedemann-etal-2014-treebank,tiedemann-2015-improving,tiedemann-agic-2016-synthetic} or a code-switching mode \citep{zhang-etal-2019-cross}, and uses them to train target language parsers.
    \item \textbf{Annotation projection},\footnote{We use \emph{annotation projection} to denote the projection of predicted parses following \citet{rasooli-collins-2019-low} and \citet{zhang-etal-2019-cross}, and \textit{treebank translation} for the projection of human-annotated trees following \citet{tiedemann-etal-2014-treebank}.} which trains a parser in the source language(s) and projects the predicted source language parse trees to the target language(s) using bitext \citep{hwa-etal-2005-bootstrapping,ma-xia-2014-unsupervised,agic-etal-2016-multilingual}.
    Additional strategies are usually used to improve the projection quality, such as keeping confident edges only \citep{li-etal-2014-soft,lacroix-etal-2016-frustratingly}, projection from multiple source languages \citep{tackstrom-etal-2013-target,agic-etal-2016-multilingual,rasooli-collins-2017-cross}, density based iterative filtering \citep{rasooli-collins-2015-density,rasooli-collins-2017-cross,rasooli-collins-2019-low}, and noisy self-training \citep{kurniawan-etal-2021-ppt}.
\end{enumerate}

These approaches make different assumptions on annotation availability, such as gold part-of-speech tags \interalia{zeman-resnik-2008-cross,cohen-etal-2011-unsupervised,durrett-etal-2012-syntactic}, a reasonably good translator, which uses extra annotated data in the training process \citep{tiedemann-etal-2014-treebank,tiedemann-2015-improving,zhang-etal-2019-cross},
high-quality bilingual lexicons \interalia{durrett-etal-2012-syntactic,guo-etal-2015-cross,guo-etal-2016-representation}, or language-specific constraints \citep{meng-etal-2019-target}.
Most bitext-based work assumes annotated bitext \interalia{ma-xia-2014-unsupervised,li-etal-2014-soft,lacroix-etal-2016-frustratingly} or bitext constructed from extra signals \citep[e.g., Wikipedia; ][]{rasooli-etal-2021-wikily}.
However,  \citet{he-etal-2019-cross}, \citet{schuster-etal-2019-cross}, \citet{ahmad-etal-2019-difficulties,ahmad-etal-2019-cross}, and \citet{kurniawan-etal-2021-ppt} only require minimal annotations (i.e., source language treebanks and unlimited raw text in relevant languages).
We are mainly interested in the minimal annotation setting and will compare it to this line of work.

Our proposed method, \subdp, falls into the category of annotation projection.
Some of the benefits of \subdp relative to prior work are that it works well with minimal annotations, allows soft word alignment (\cref{sec:subdp-preliminary}), supports both labeled and unlabeled parsing, and has a low time complexity $\mathcal{O}(n^2)$ for non-projective parsing.\footnote{
    In contrast, \citet{ma-xia-2014-unsupervised} require $\mathcal{O}(n^4)$ time for non-projective unlabeled dependency parsing.
}
\subdp can be easily extended to other tasks, such as sequence labeling, where we can define substructures \citep{shi2021substructure} and substructure distributions.

\paragraph{Multilingual contextualized representations.}
Recent contextualized models pre-trained on multilingual text \interalia{devlin-etal-2019-bert,conneau-etal-2020-unsupervised,tran-etal-2020-cross} are effective across a wide range of cross-lingual NLP tasks, including bitext retrieval \citep{tran-etal-2020-cross}, bilingual lexicon induction \citep{shi2021bilingual}, cross-lingual named entity recognition \citep{pires-etal-2019-multilingual,mulcaire-etal-2019-polyglot}, and cross-lingual dependency parsing \citep{schuster-etal-2019-cross,wang-etal-2019-cross}.
In this work, we apply two of the contextualized pre-trained models, XLM-R \citep{conneau-etal-2020-unsupervised} and CRISS \citep{tran-etal-2020-cross}, to generate unsupervised bitext.

\paragraph{Soft-label methods.}
Calculating the cross entropy loss between model output and a soft distribution (instead of one-hot labels) has been applied to knowledge distillation \interalia{hinton-etal-2015-distilling,you-etal-2017-learning,sanh-etal-2019-distillbert}, cross-lingual named entity recognition \citep{wu-etal-2020-single}, and
for handling annotation discrepancy \citep{fornaciari-etal-2021-beyond}.
Our approach is a soft-label method with additional post-processing to the output of the original models.

\section{Method}
Our pipeline for zero-shot cross-lingual dependency parsing consists of three steps:
(1) train a bi-affine dependency parser $\mathcal{P}_1$ in the source language $L_1$,
(2) project annotations on $L_1$ sentences to their parallel sentences in the target language $L_2$ (\cref{sec:subdp-ddp}), and
(3) train another bi-affine dependency parser $\mathcal{P}_2$ for $L_2$ (\cref{sec:subdp-optimization}).
We first present some background (\cref{sec:subdp-background}) and preliminaries (\cref{sec:subdp-preliminary}).

\subsection{Background}
\label{sec:subdp-background}
\paragraph{Bi-affine dependency parser.}
For a sentence with $n$ words $\langle w_1, \ldots, w_n\rangle$,\footnote{For convenience, we assume that $w_1$ is an added dummy word that has one dependent -- the root word of the sentence.} we denote the word features when acting as heads and dependents by $\bm{H}\in \mathbb{R}^{n \times d_h}$ and $\bm{D} \in \mathbb{R}^{n \times d_d}$ respectively, where $d_h$ and $d_d$ denote the dimensionality of the corresponding features.
The probability of word $w_i$ having head $w_j$ can be formulated as an $n$-way classification problem:
\begin{align}
    \bm{S}^\textit{(arc)} &= \bm{D}\bm{W}^{\textit{(arc)}}\bm{H}^\intercal \label{eq:bi-affine} \\
    P(w_j\!\mid\!w_i) &= \frac{\exp\left(\bm S^\textit{(arc)}_{i, j}\right)}{\sum_{k=1}^{n} \exp\left(\bm S^\textit{(arc)}_{i, k}\right)}, \label{eq:arc-prob}
\end{align}
where $\bm W^{\textit{(arc)}} \in \mathbb{R}^{d_d\times d_h}$ is the parameters of the bi-affine module.\footnote{While \cref{eq:bi-affine} is in a bi-linear form, in practice, we can always append a constant feature column to both $\bm H$ and $\bm D$, resulting in a bi-affine model.}
Given $\log P(w_j \mid w_i)$ for every pair of $i$ and $j$, the dependency trees can be inferred by finding the spanning arborescence of maximum weight using the Chu--Liu--Edmonds algorithm \citep{chu1965shortest,edmonds1968optimum}.
We use the algorithm proposed by \citet{tarjan1977finding}, which has an $\mathcal{O}(n^2)$ time complexity for each sentence.

We denote the candidate dependency label set by $L$.
Parameterized by $\bm{W}^\textit{(label)} \in \mathbb{R}^{d_d\times d_h \times |L|}$, we define the probability that the arc from head $w_j$ to dependent $w_i$ has the label $\ell$ by
\begin{align}
    \bm{S}^\textit{(label)}_{i, j, \ell} &= \sum_{p} \sum_{q} \bm{D}_{i,p}\bm{W}^{\textit{(label)}}_{p, q, \ell} \bm{H}_{j,q} \nonumber \\
    P(\ell\!\mid\!w_j\!\rightarrow\!w_i) &= \frac{\exp\left(\bm S^\textit{(label)}_{i, j, \ell}\right)}{\sum_{k=1}^{|L|} \exp\left(\bm S^\textit{(label)}_{i, j, k}\right)}, \label{eq:subdp-label-prob}
\end{align}

Given the probability definitions above, we train the model to maximize the log-likelihood of the training data.
More details can be found in \citet{dozat2016deep}.

We use bi-affine dependency parsers as the backbone for all parsers in this work, though it is worth noting that \subdp works for any parser that produces a set of arc and label distributions.

\paragraph{CRISS.}
CRISS \citep{tran-etal-2020-cross} is an unsupervised machine translation model trained with monolingual corpora, starting from mBART \citep{liu-etal-2020-mbart}, a multilingual pre-trained sequence-to-sequence model with a mask-filling denoising objective.
During the training process, CRISS iteratively (1) encodes sentences in the monolingual corpora with its encoder, (2) mines bitext based on encoding similarity, and (3) uses the mined bitext to fine-tune the model with a machine translation objective.
In this work, we use CRISS to generate an unsupervised translation of English sentences to construct bitext and apply its encoder to extract word features for an ablation study.

\paragraph{SimAlign.} SimAlign \citep{sabet-etal-2020-simalign} is a similarity-based word aligner: given a pair of source and target sentence $\langle s, t\rangle$, SimAlign computes a contextualized representation for each token in both $s$ and $t$ using multilingual pre-trained models \citep{devlin-etal-2019-bert,conneau-etal-2020-unsupervised}, and calculates the similarity matrix $S$, where $S_{i,j}$ represents the cosine similarity between tokens $s_i$ and $t_j$.
The \texttt{argmax} inference algorithm selects position pairs $\langle i, j\rangle$, where $S_{i,j}$ is both horizontal and vertical maximum and outputs the word pairs corresponding to such position pairs as the word alignment.
This work uses XLM-R \citep{conneau-etal-2020-unsupervised} based SimAlign with the \texttt{argmax} algorithm to extract word alignment for \subdp.
Notably, pre-trained multilingual models usually use subwords, a more fine-grained level than words, for tokenization.
The \texttt{argmax} algorithm may therefore generate many-to-one alignments.
More details can be found in \citet{sabet-etal-2020-simalign}.

Unlike bitext-based word alignment \citep{och-ney-2003-systematic,dyer-etal-2013-simple}, such as GIZA++ \citep{och-ney-2003-systematic} and \texttt{fast\_align} \citep{dyer-etal-2013-simple}, SimAlign does not require any bitext to produce high-quality alignments, and therefore better fits the low-resource scenario with very few bitext pairs available.

\subsection{Preliminaries}
\label{sec:subdp-preliminary}

\paragraph{Dependency annotations in $L_1$.}
As in the most common data settings for supervised dependency parsing, we assume access to sentences with dependency annotations:
for a sentence $\langle w_1, \ldots, w_n \rangle$, there is a dummy word $w_1$, whose unique dependent is the root word; every other word $w_i$ is labeled with $h_i$ and $r_i$, denoting that the head of $w_i$ is $w_{h_i}$, with the dependency relation $r_i$.
We use these annotations to train an $L_1$ bi-affine dependency parser $\mathcal{P}_1$, following the procedure described in \cref{sec:subdp-background}.

\paragraph{Bitext.} We denote the available $m$ pairs of bitext by $\mathcal{B} = \{\langle s^{(k)}, t^{(k)}\rangle\}_{k=1}^m$, where $\{s^{(k)}\}$ and $\{t^{(k)}\}$ are sentences in $L_1$ and $L_2$ respectively.
\paragraph{Word alignment.}
For a bitext pair $\langle s, t\rangle$, we generate the word alignment matrix $\bm\tilde{A} \in \{0, 1\}^{|s|\times|t|}$ with SimAlign, where $\bm{\tilde{A}}_{i,j}=1$ denotes that there exists an alignment between $s_i$ and $t_j$.

We would like the word alignment matrices to be right stochastic, i.e., (1) each element is non-negative and (2) each row sums to one, to ensure that the results after projection remain distributions.
To handle words that have zero or more than one aligned words in the other language, we introduce the following two matrix operators.

\textbf{The \textit{add-dummy-position} operator $\Delta(\cdot)$}:
\begin{align*}
    \Delta: \mathbb{R}^{r\times c} &\rightarrow \mathbb{R}^{(r+1)\times(c+1)} (\forall r, c \in \mathbb{N}_+) \\
    \Delta(\bm{M})_{i,j} &= \bm{M}_{i,j}  (1 \leq i \leq r, 1\leq j \leq c); \\
    \Delta(\bm{M})_{i,c+1} &= \boldsymbol{0}[\bm{M}_{i,1},\ldots,\bm{M}_{i,c}] (1\leq i\leq r);\\
    \Delta(\bm{M})_{r+1,j} &= 0 (1 \leq j \leq c); \\
    \Delta(\bm{M})_{r+1,c+1} &= 1,
\end{align*}
where $\boldsymbol{0}[\cdot]=1$ when all input values are zero and otherwise $0$.

\textbf{The \textit{row normalization} operator $\mathcal{N}^{\mathcal{R}}(\cdot)$}:
\begin{align*}
    \mathcal{N}^{\mathcal{R}}:& \mathbb{R}^{r\times c} \rightarrow \mathbb{R}^{r\times c} (\forall r, c \in \mathbb{N}_+) \\
    \mathcal{N}^{\mathcal{R}}(\bm{M})_{i, j} &= \frac{\bm{M}_{i, j}}{\sum_{\ell} \bm{M}_{i, \ell}}.
\end{align*}
Intuitively, the added dummy positions correspond to \emph{null} words in the word alignment literature \interalia{dyer-etal-2013-simple,schulz-etal-2016-word,sabet-etal-2020-simalign}.
We denote the source-to-target alignment matrix by $\bm{A}^{s\rightarrow t} = \mathcal{N}^\mathcal{R}\left(\Delta(\tilde{\bm A})\right)$, and the target-to-source alignment matrix by $\bm{A}^{t\rightarrow s} = \mathcal{N}^\mathcal{R}\left(\Delta(\tilde{\bm A}^\intercal)\right)$.
Both are right stochastic matrices by definition.

\subsection{Dependency Distribution Projection}
\label{sec:subdp-ddp}
\paragraph{Arc distribution projection. } We consider a pair of bitext $\langle s, t\rangle$.
Let $P_1(s_j \mid s_i)$ denote the arc probability produced by the parser $\mathcal{P}_1$.
Like the dummy position notation, we specify a dummy $(|s|+1)^{th}$ word whose head is itself, that is,
\begin{align*}
    P_1(s_i \mid s_{|s|+1}) = 0, ~P_1(s_{|s|+1} \mid s_{|s|+1}) = 1.
\end{align*}
We project $P_1(\cdot \mid \cdot)$ to $\hat{P}_2(t_q\mid t_p)$, the arc probability distributions in the parallel $L_2$ example $t$, 
\begin{align}
    \hat{P}_2(t_q\!\mid\!t_p)\!=\!\sum_{i=1}^{|s|+1} \sum_{j=1}^{|s|+1} \bm{A}^{t\rightarrow s}_{p,i} P_1(s_j\!\mid\!s_i) \bm{A}^{s\rightarrow t}_{j,q}. \label{eq:3-matrices-prob}
\end{align}
It is guaranteed that $\hat{P}_2(\cdot \mid t_p)$ is a distribution for any $t_p$.
\begin{proposition}
    Suppose that $P_1(\cdot\mid s_i)$ is a probability distribution for any $s_i$, and that $\bm{A}^{t\rightarrow s}$ and $\bm{A}^{s\rightarrow t}$ are right-stochastic matrices (i.e., each row of the matrices defines a probability distribution). Let $P_2(t_p\mid t_q) = \sum_{i=1}^{|s|+1}\sum_{j=1}^{|s|+1} \bm{A}^{t\rightarrow s}_{p,i}P_1(s_j\mid s_i)\bm{A}^{s\rightarrow t}_{j,q}$. We have that $P_2(\cdot\mid t_p)$ is a distribution for any $t_p$.
    \label{proposition:distribution-arc}
\end{proposition}
\begin{proof}
    First, for any combination of $i, j, p, q$, we have that $\bm{A}^{t\rightarrow s}_{p, i} \geq 0$, $P_1(s_j\mid s_i) \geq 0$, $\bm{A}^{s\rightarrow t}_{j,q}\geq 0$, therefore,
    \begin{align*}
        P_2(t_q\mid t_p) = \sum_{i=1}^{|s|+1}\sum_{j=1}^{|s|+1} \bm{A}^{t\rightarrow s}_{p,i}P_1(s_j\mid s_i)\bm{A}^{s\rightarrow t}_{j,q} \geq 0
    \end{align*}
    On the other hand,
    \begin{align*}
        &\sum_{q=1}^{|t|+1} P_2(t_q\mid t_p) \\
        =& \sum_{q=1}^{|t|+1}\sum_{i=1}^{|s|+1}\sum_{j=1}^{|s|+1} \bm{A}^{t\rightarrow s}_{p,i}P_1(s_j\mid s_i)\bm{A}^{s\rightarrow t}_{j,q} \\
        =& \sum_{i=1}^{|s|+1}\sum_{j=1}^{|s|+1} \bm{A}^{t\rightarrow s}_{p,i} P_1(s_j\mid s_i) \left(\sum_{q=1}^{|t|+1}\bm{A}^{s\rightarrow t}_{j,q}\right) \\
        =& \sum_{i=1}^{|s|+1}\sum_{j=1}^{|s|+1} \bm{A}^{t\rightarrow s}_{p,i} P_1(s_j\mid s_i) \\
        =& \sum_{i=1}^{|s|+1} \bm{A}^{t\rightarrow s}_{p,i} \left(\sum_{j=1}^{|s|+1}P_1(s_j\mid s_i)\right) \\    =& \sum_{i=1}^{|s|+1} \bm{A}^{t\rightarrow s}_{p,i} \\
        =& 1.
    \end{align*}
\end{proof}

Note that if we adopt matrix notations, where we denote $\hat{P}_2(t_q\mid t_p)$ by $\bm{\hat{P}}^{(2)}_{p, q}$ and denote $P_1(s_j\mid s_i)$ by $\bm{P}^{(1)}_{i, j}$, \cref{eq:3-matrices-prob} is equivalent to
\begin{align*}
    \bm{\hat{P}}^{(2)} = \bm{A}^{t\rightarrow s}\bm{P}^{(1)}\bm{A}^{s\rightarrow t}.
\end{align*}

\noindent\textbf{Label distribution projection. } Let $P_1(\ell\!\mid\!s_j\!\rightarrow\!s_i)$ denote the label probability produced by $\mathcal{P}_1$.
For dummy positions, we simply add a uniform distribution, that is,
\begin{align*}
    P_1(\ell\!\mid s_j\!\rightarrow\!s_i) = \frac{1}{L} && \text{if $i$ or $j=|s|+1$.}
\end{align*}
We project $P_1(\cdot\!\mid\!\cdot\!\rightarrow\!\cdot)$ to $\hat{P}_2(\ell\!\mid\!t_q\!\rightarrow\!t_p)$, the label distributions in the parallel $L_2$ example $t$, by
\begin{align*}
    \hat{P}_2(\ell\!\mid\!t_q\!\rightarrow\!t_p)\!=\!\!\sum_{i=1}^{|s|+1}\!\sum_{j=1}^{|s|+1}\! \bm{A}^{t\rightarrow s}_{p,i}P_1(\ell\!\mid\!s_j\!\rightarrow\!s_i)\bm{A}^{t\rightarrow s}_{q,j}
\end{align*}
$\hat{P}_2(\cdot\!\mid\!t_q\!\rightarrow\!t_p)$ is provably a distribution for any pair of $t_p$ and $t_q$.
\begin{proposition}
    Suppose that $P_1(\cdot\mid s_j\rightarrow s_i)$ is a probability distribution for any combination of $s_i$ and $s_j$, and that $\bm{A}^{t\rightarrow s}$ is a right-stochastic matrix.
    Let $P_2(\ell\mid t_q \rightarrow t_p) = \sum_{i=1}^{|s|+1}\sum_{j=1}^{|s|+1}\bm{A}^{t\rightarrow s}_{p,i}P_1(\ell\mid s_j\rightarrow s_i)\bm{A}^{t\rightarrow s}_{q,j}$.
    We have that $P_2(\cdot \mid t_q \mid t_p)$ is a probability distribution for any $t_p$ and $t_q$.
\end{proposition}
\begin{proof}
    Similarly to the proof of \cref{proposition:distribution-arc}, it is easy to show that for any $\ell, t_p, t_q$,
\begin{align*}
    P_2(\ell\mid t_q\rightarrow t_p) \geq 0.
\end{align*}

We next consider the sum over $\ell$ for a specific pair of $t_p$ and $t_q$, where we have
\begin{align*}
    &\sum_{\ell=1}^{|L|}P_2(\ell\mid t_q\rightarrow t_p) \\
    =& \sum_{\ell=1}^{|L|}\sum_{i=1}^{|s|+1}\sum_{j=1}^{|s|+1}\bm{A}^{t\rightarrow s}_{p,i}P_1(\ell\mid s_j\rightarrow s_i)\bm{A}^{t\rightarrow s}_{q,j} \\
    =& \sum_{i=1}^{|s|+1}\sum_{j=1}^{|s|+1}\bm{A}^{t\rightarrow s}_{p,i}\bm{A}^{t\rightarrow s}_{q,j} \left( \sum_{\ell=1}^{|L|}P_1(\ell\mid s_j\rightarrow s_i)\right) \\
    =& \sum_{i=1}^{|s|+1}\sum_{j=1}^{|s|+1}\bm{A}^{t\rightarrow s}_{p,i}\bm{A}^{t\rightarrow s}_{q,j} \\
    =& \sum_{i=1}^{|s|+1}\bm{A}^{t\rightarrow s}_{p,i}\left(\sum_{j=1}^{|s|+1}\bm{A}^{t\rightarrow s}_{q,j}\right) \\
    =& \sum_{i=1}^{|s|+1}\bm{A}^{t\rightarrow s}_{p,i} \\
    =& 1.
\end{align*}
\end{proof}

\subsection{Optimization}
\label{sec:subdp-optimization}
We train another bi-affine dependency parser $\mathcal{P}_2$ on language $L_2$, by minimizing the cross entropy between its produced probability $P_2$ and the soft silver labels $\hat{P}_2$.
Note that the added dummy word denoting the null alignment is not eventually used in the final dependency inference process and may introduce extra noise to the model, so we instead calculate the \textit{partial} cross-entropy loss, which does not consider elements involving dummy words.
Concretely, we compute the partial arc cross-entropy loss for one example $t$ as follows: %
\begin{align*}
    \mathcal{L}_\textit{arc}^{(t)}(P_2, \hat{P}_2) \!
    = \!-\!\sum_{p=1}^{|t|}\!\sum_{q=1}^{|t|}\!\hat{P}_2(t_q\!\mid\!t_p)\log  P_2(t_q\!\mid\!t_p)
\end{align*}
Similarly, the partial label cross-entropy loss can be computed as follows:
\begin{align*}
    \mathcal{L}_\textit{label}^{(t)}(P_2, \hat{P}_2)
    & = -\sum_{\ell=1}^{|L|}\sum_{p=1}^{|t|}\sum_{q=1}^{|t|}\\
    & \hat{P}_2(\ell\mid t_q\rightarrow t_p)\log P_2(\ell \mid t_q\rightarrow t_p)
\end{align*}
Finally, we train the parameters of $\mathcal{P}_2$ to minimize
\begin{align}
    \sum_{\langle s,t \rangle \in \mathcal{B}} \mathcal{L}_\textit{arc}^{(t)}(P_2, \hat{P}_2) + \mathcal{L}_\textit{label}^{(t)}(P_2, \hat{P}_2). \label{eq:subdp-objective}
\end{align}

\section{Experiments}
\input{tables/901-fully-unsup.tex}
Throughout all experiments, the subword representation is a weighted sum of layer-wise representation from a frozen pre-trained model, where each layer has a scalar weight optimized together with other network parameters to minimize \cref{eq:subdp-objective}.
We convert subword features to word features by endpoint concatenation, following \citet{toshniwal2020cross}.
We use the Adam optimizer \citep{kingma2015adam} to train all models, where the source language parser is trained for 100 epochs with initial learning rate $2\times 10^{-3}$ following the baseline implementation by \citet{zhang-etal-2020-efficient}, and the target language parser is trained for 30 epochs with initial learning rate $5\times 10^{-4}$.\footnote{We do not observe further training loss decrease when training for more epochs. The learning rate for \subdp is tuned to optimize the development loss for German, where the German gold trees remain unused.}
We use the loss against silver projected distributions on the development set for \subdp and the development LAS against projected trees for baselines for early stopping.\footnote{\subdp does not provide a set of hard silver trees for LAS and UAS calculation.}
For evaluation, we ignore all punctuation following the most common convention \interalia{ma-xia-2014-unsupervised,rasooli-collins-2015-density,kurniawan-etal-2021-ppt}.
If not specified,
\begin{itemize}[leftmargin=*,itemsep=-2mm]
    \item All models in target languages are initialized with the trained source language parser.
    \item All word alignments are obtained by XLM-R based SimAlign \citep{sabet-etal-2020-simalign}, using BPE tokenization and the \texttt{argmax} algorithm.
    \item XLM-R is used as the feature extractor.
\end{itemize}
We report results on the standard development sets to avoid tuning on the test sets for analysis purposes.

\subsection{Results: Fully Unsupervised Transfer}

We compare \subdp to prior work in the minimal annotation setting (\cref{tab:subdp-ud2.2-fully-unsup}), where an English dependency treebank is the only annotation that involves human effort.
We select target languages from the overlap between those considered by \citet{kurniawan-etal-2021-ppt}, those covered by XLM-R \citep{conneau-etal-2020-unsupervised} training corpora, and those supported by CRISS \citep{tran-etal-2020-cross}, resulting in eight languages: Arabic (ar), Hindi (hi), Korean (ko), Turkish (tr), German (de), Spanish (es), French (fr), and Italian (it).

We translate English sentences using the unsupervised model CRISS to construct the required bitext.\footnote{In experiments, we translate English treebank sentences; in more general cases, any source language sentence can be taken for bitext construction.}
To ensure the quality of the unsupervised bitext, we discard (1) translations where at least 80\% of words appear in the corresponding source sentences, which are likely to be copied, (2) those containing a CRISS language token other than the target language, which are likely to be false translation into another language, and (3) those with 80\% or more words appearing in the translated sentence more than once, which are likely to be repetitions.

Transferring from an English parser, \subdp achieves the best UAS across all eight target languages and the best LAS on six languages out of eight.
In addition, we find that \subdp is consistent across random seeds, with a standard deviation less than $0.8$ for every number in \cref{tab:subdp-ud2.2-fully-unsup}.

\subsection{Ablation Study}
\label{sec:subdp-ablation-subsection}
\input{figures/902-subdp-ablation.tex}
\input{figures/903-subdp-ablation.tex}
We introduce the following baselines with the same annotated data availability for an  ablation study:

\begin{enumerate}
    \item \textbf{Direct transfer of English models (DT).}
    We train a bi-affine dependency parser on English treebanks and test the model on other languages.
    This approach is expected to outperform a random baseline as it has a pre-trained cross-lingual language model-based feature extractor, which may implicitly enable cross-lingual transfer.
    For this baseline, we test both XLM-R and CRISS encoders, as \subdp benefits from both models.
    \item \textbf{Self-training (ST). } Following \citet{kurniawan-etal-2021-ppt}, we apply an XLM-R DT parser to the target language,\footnote{We only consider XLM-R as the feature extractor for ST as it achieves better average DT results.} and train another parser on the predicted hard trees.
    \item \textbf{Hard projection (Hard).} It is intuitive to compare \subdp against the hard tree projection baseline \citep{lacroix-etal-2016-frustratingly}, where we use the same set of bitext and alignments to project trees to the target languages, keeping only the edges with both sides aligned in a one-to-one alignment.
    We use the projected trees to train a parser in the target language.
    \item \textbf{Random target parser initialization (RandI). } Instead of using the trained English model as the initialization of target parsers, we randomly initialize the weights in this baseline.
    This approach matches with \subdp in every component except the target parser initialization.
\end{enumerate}
All baselines use bi-affine dependency parsers, with pre-trained cross-lingual language models (XLM-R or CRISS) as feature extractors.

We compare the LAS and UAS between \subdp and the baselines above (\cref{fig:subdp-ablation-las,fig:subdp-ablation-uas}), and find that
\begin{itemize}
    \item Across all languages, \subdp significantly outperforms DT with either XLM-R or CRISS word feature extractor.
    ST consistently improves over DT but is much less competitive than \subdp, indicating that the gain of \subdp over prior work is not simply from more powerful word features.
    \item While hard treebank projection using the method proposed by \citet{lacroix-etal-2016-frustratingly} is quite competitive, \subdp consistently produces competitive (Arabic, German, Spanish) or better (Hindi, Korean, Turkish, French, Italian) results.
    \item Comparing \subdp to RandI, we find that initializing the target language parser with a trained source language (English in this work) parser helps improve performance across the board; therefore, source parser initialization should be considered as a general step in future work on zero-shot cross-lingual dependency parsing.
\end{itemize}

\subsection{Analysis: Effect of Alignment Methods}
\input{tables/902-subdp-alignments.tex}
Since most existing work has used only one-to-one alignment for annotation projection
\interalia{ma-xia-2014-unsupervised,lacroix-etal-2016-frustratingly,rasooli-etal-2021-wikily}, we would like to analyze the effect of introducing many-to-one alignment edges in \subdp.
We filter SimAlign BPE \texttt{argmax} to obtain a more conservative version, dropping all many-to-one edges (i.e., those that have a word linked to multiple edges),\footnote{~This approach is different from Hard as it takes soft source trees as the input, yielding soft target trees as silver labels to train target language parsers.} and compare it to the BPE \texttt{argmax} algorithm (\cref{tab:subdp-alignments}).

While the confident one-to-one alignment achieves further improvement on Arabic and all four nearby languages, we find that the many-to-one BPE \texttt{argmax} alignment is important to the superior transfer performance on Hindi, Korean, and Turkish.
Given the fact that the scores are quite similar for Arabic, the results generally suggest using the many-to-one SimAlign BPE \texttt{argmax} alignments for transferring from English to distant languages while using the more confident one-to-one alignments for nearby languages.

\subsection{Results: Multiple Source Languages}
\input{tables/903-subdp-multiple-source.tex}
Following \citet{schuster-etal-2019-cross}, we use Universal Dependencies v2.0 \citep{mcdonald-etal-2013-universal} to evaluate zero-shot cross-lingual transfer from multiple source languages (\cref{tab:subdp-multiple-source}).\footnote{We do not report performance for Portuguese and Swedish as CRISS does not cover them; however, the annotated treebanks in these languages are used as source treebanks when applicable.}
For each language among German (de), Spanish (es), French (fr), Italian (it), Portuguese (pt), and Swedish (sv), annotated treebanks from all other languages and English can be used for training and development purposes.
For \subdp, we generate bitext from all applicable source languages with CRISS.

\subdp outperforms the previous state-of-the-art on German by 13.5 LAS, but under-performs the DT baseline on the other three languages.
However, suppose we start with a trained \subdp parser for a target language and use the standard training data (i.e., treebanks in other languages) to train further a bi-affine dependency parser (DT w/ \subdp init.).
In that case, we can achieve better results than DT across the board, obtaining competitive or even better LAS than methods that use extra annotations other than source treebanks \citep{zhang-barzilay-2015-hierarchical,guo-etal-2016-representation}.

\subsection{Results: Transfer with Supervised Bitext}
We further evaluate \subdp in another scenario where a few bitext pairs are available.
We consider a larger set of eighteen target languages, including Arabic (ar), Czech (cs), German (de), Spanish (es), Finnish (fi), French (fr), Hindi (hi), Hugarian (hu), Italian (it), Japanese (ja), Korean (ko), Norwegian (no), Portuguese (pt), Russian (ru), Tamil (ta), Telugu (te), Vietnamese (vi), and Chinese (zh).
We transfer from English to each target language with Wikimatrix bitext \citep{schwenk-etal-2021-wikimatrix}, where the examples are mined with an encoding similarity-based bitext miner trained with annotated bitext.
We vary the number of
Wikimatrix bitext pairs, selecting the number of pairs within the geometric sequence $\{50\times 2^{k}\}_{k=0}^9$,
leaving 10\% of the examples for development.
\input{figures/904-subdp-data-efficiency.tex}

On average and for nearby languages (\cref{fig:subdp-data-efficiency}), we find that the performance of \subdp with 50 pairs of bitext is quite close to that with 25K pairs of bitext.
Although some distant languages generally require more bitext for further improvement, \subdp outperforms the direct transfer baseline by a nontrivial margin with a small amount (e.g., 800-1.6K pairs) of bitext.

\section{Conclusion and Discussion}
Our work is in line with recent work \citep{rasooli-etal-2021-wikily}, which shows that cross-lingual transfer can be done effectively with weak supervision, such as Wikipedia links. Our results go further and study the setting of zero additional supervision beyond the source language treebank, demonstrating the potential of zero-shot cross-lingual dependency parsing with zero additional supervision, even between distant languages that do not share vocabulary or subwords. Our work suggests a new protocol for dependency annotation of low-resource languages: (1) train a pre-trained multilingual model following existing work such as XLM-R \citep{conneau-etal-2020-unsupervised} and CRISS \citep{tran-etal-2020-cross}, (2) annotate a small number of bitext pairs or generate bitext with trained unsupervised translation models, and (3) train a zero-shot cross-lingual dependency parser using \subdp.

Our contribution to zero-shot cross-lingual dependency parsing is arguably orthogonal to contextualized representation alignment \citep{schuster-etal-2019-cross,wang-etal-2019-cross}, where pre-trained multilingual language models are finetuned for better transfer.
In contrast, we use the frozen pre-trained models to extract features.
In addition, projection quality controls by heuristics-based filtering \citep{rasooli-collins-2015-density} may also be combined with \subdp to improve the performance.

Our results, on the other hand, demonstrate that multilingual pre-trained models may have more applications beyond representation-based direct transfer---information extracted from these models without further supervision (e.g., word alignment in this work) may further benefit downstream tasks (e.g., zero-shot cross-lingual dependency parsing in this work) with appropriate usage.

While this work depends on pre-trained multilingual models such as CRISS \citep{tran-etal-2020-cross}, which require extensive computational resources to train from scratch, \subdp may be applied whenever bitext alignment and cross-lingual word embeddings are available.
In addition, the required pre-trained cross-lingual models are useful for general purposes and can be applied to other downstream NLP tasks.

We suggest that \subdp can be extended to other scenarios wherever relevant parallel signals are available, such as cross-lingual named entity recognition, cross-lingual constituency parsing, or zero-shot scene graph parsing for images using only the dependency supervision in text. We leave the further exploration of \subdp on other tasks and a more comprehensive cross-lingual parsing quality analysis for future work.

%% file: figures/901-subdp-intro.tex
\begin{figure}[t!]
    \centering
    \begin{subfigure}[t]{\textwidth}
        \centering
        \includegraphics[width=0.62\textwidth]{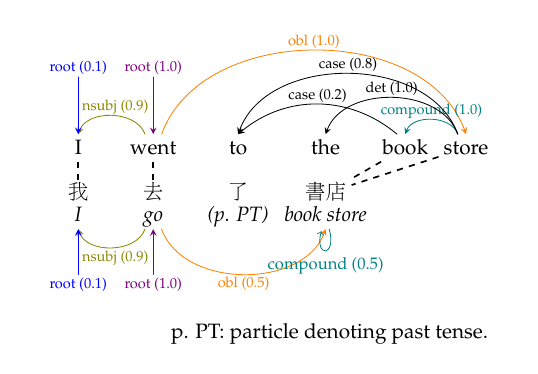}
        \caption{Soft trees projected by \subdp. The self-loops are discarded when training models.}
        \label{fig:subdp-intro-soft}
    \end{subfigure}
    \vspace{10pt}

    \begin{subfigure}[t]{\textwidth}
        \centering
        \hspace{10pt}
        \includegraphics[width=0.52\textwidth]{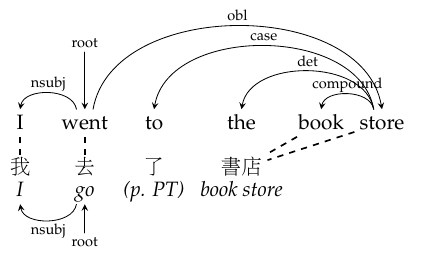}
        \caption{Hard trees projected with only one-to-one alignments \citep{lacroix-etal-2016-frustratingly}.}
        \label{fig:subdp-intro-hard}
    \end{subfigure}

    \caption[Illustration of \subdp vs. \citet{lacroix-etal-2016-frustratingly}.]{Illustration of \subdp vs. \citet{lacroix-etal-2016-frustratingly}.
    The English sentence (up) is projected to German (bottom) using the dependency arcs as substructures.
    Same-colored solid arcs denote corresponding arcs in the source and target trees, while dashed arcs exist in the ground truth but are not projected.
    The arcs are projected separately, and the target parser is trained on the projected distributions.
    Best viewed in color.
    }
    \label{fig:subdp-intro}
\end{figure}

%% file: tables/901-fully-unsup.tex
\newcommand{\emptyresult}{---}
\begin{table}[t!]
    \centering \small \addtolength{\tabcolsep}{0pt}
    \begin{tabular}{lccccccccccccccccc}
        \toprule
        \bf Method && \multicolumn{7}{c}{\it distant languages} && \multicolumn{7}{c}{\it nearby languages} \\
        \cmidrule{3-9} \cmidrule{11-17}
        \textbf{LAS} && ar && hi && ko && tr && de && es && fr && it & \\
        \midrule
        \citeauthor{meng-etal-2019-target} && \emptyresult && \emptyresult && \emptyresult && \emptyresult && \emptyresult && \emptyresult && \emptyresult && \emptyresult & \\
        \citeauthor{he-etal-2019-cross}   && \emptyresult && \emptyresult && \emptyresult && \emptyresult && \emptyresult && \emptyresult && \emptyresult && \emptyresult & \\
        \citeauthor{ahmad-etal-2019-cross} && 27.9 && 28.0 && 16.1 && \emptyresult && 61.8 && 65.8 && 73.3 && 75.6 & \\
        \citeauthor{kurniawan-etal-2021-ppt} && 38.5 && 28.3 && 16.1 && 20.6 && 63.5 && 69.2 && \textbf{74.5} && \textbf{77.7} & \\
        \subdp (ours) && \textbf{41.3} && \textbf{38.9} && \textbf{31.2} && \textbf{33.5} && \textbf{71.7} && \textbf{70.4} && 71.0 && 75.0 & \\
        \midrule
        \textbf{UAS} \\
        \cmidrule{1-1}
        \citeauthor{meng-etal-2019-target} && 47.3 && 52.4 && 37.1 && 35.2 && 70.8 && 75.8 && 79.1 && 82.0 \\
        \citeauthor{he-etal-2019-cross} && 55.4 && 33.2 && 37.0 && 36.1 && 69.5 && 64.3 && 67.7 && 70.7 \\
        \citeauthor{ahmad-etal-2019-cross} && 27.9 && 28.0 && 16.1 && \emptyresult && 61.8 && 65.8 && 73.3 && 75.6 \\
        \citeauthor{kurniawan-etal-2021-ppt} && 48.3 && 36.4 && 34.6 && 38.4 && 74.1 && 78.3 && 80.6 && 83.7 \\
        \subdp (ours) && \textbf{63.8} && \textbf{58.3} && \textbf{54.3} && \textbf{56.9} && \textbf{82.8} && \textbf{83.9} && \textbf{84.8} && \textbf{88.2}\\
        \bottomrule
    \end{tabular}
    \caption[LAS and UAS on Universal Dependencies v2.2.]{
        Labeled attachment scores (LAS) and unlabeled attachment scores (UAS) on the Universal Dependencies v2.2 \citep{nivre-etal-2020-universal} standard test set, transferring from English. Following \citet{kurniawan-etal-2021-ppt}, our results are averaged across five runs with different random seeds; the best number in each column is in boldface.
    }
    \label{tab:subdp-ud2.2-fully-unsup}
\end{table}

%% file: figures/902-subdp-ablation.tex
\begin{figure}[t!]
    \includegraphics[width=0.28\textwidth]{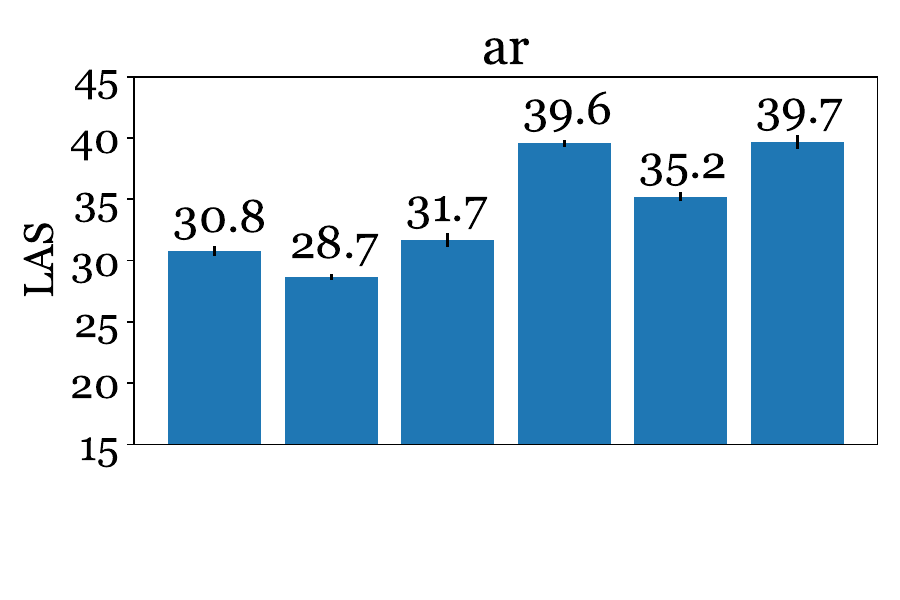} \hspace{-25pt}
    \includegraphics[width=0.28\textwidth]{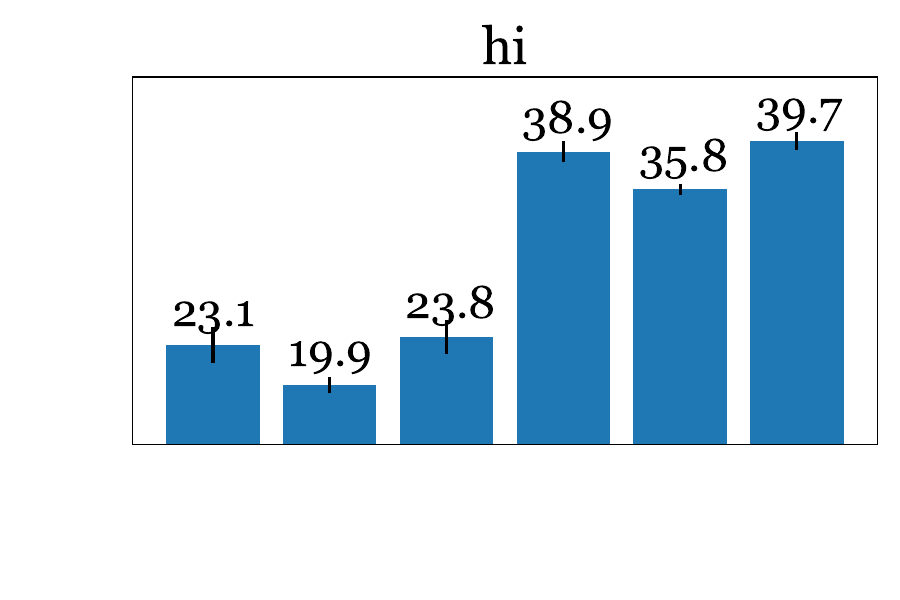} \hspace{-25pt}
    \includegraphics[width=0.28\textwidth]{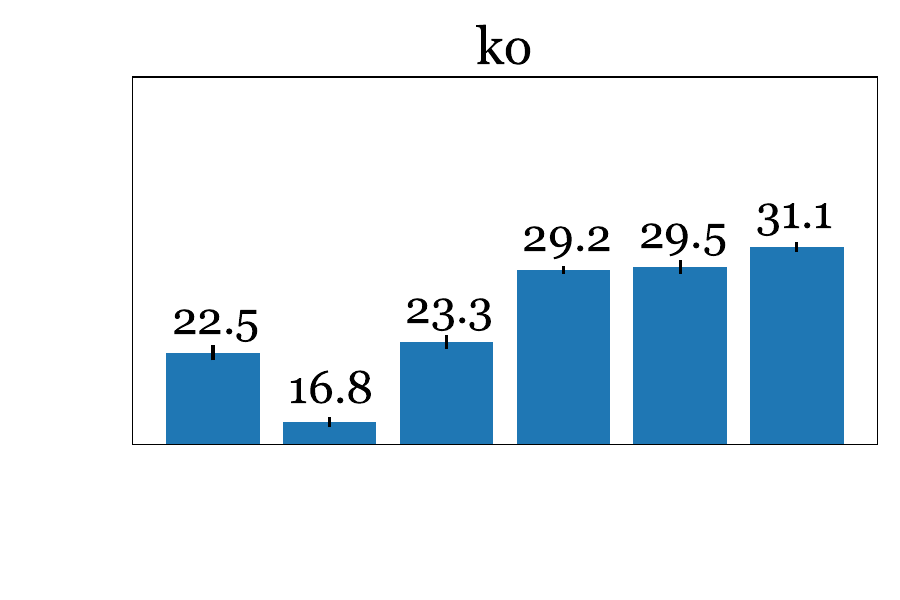} \hspace{-25pt}
    \includegraphics[width=0.28\textwidth]{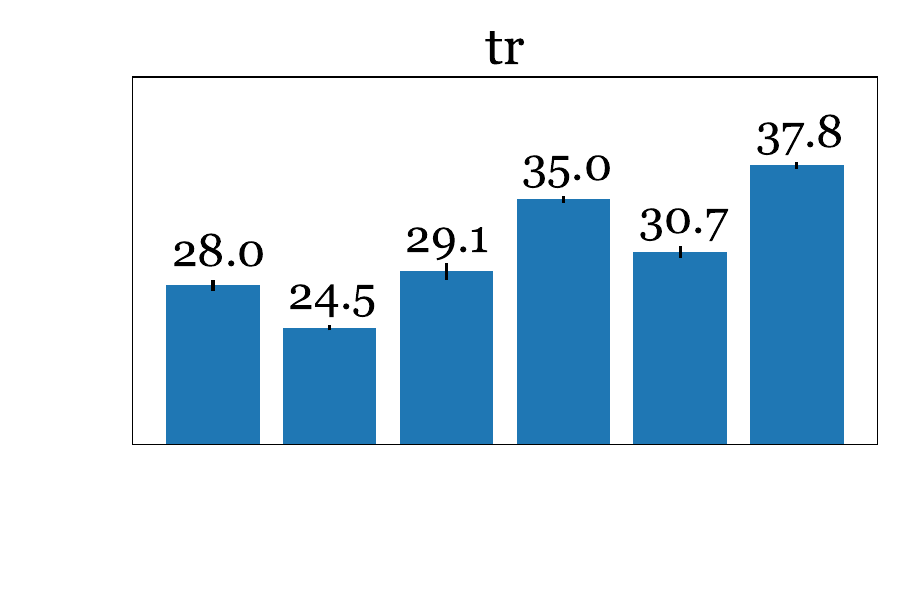} \\[-22pt]
    \includegraphics[width=0.28\textwidth]{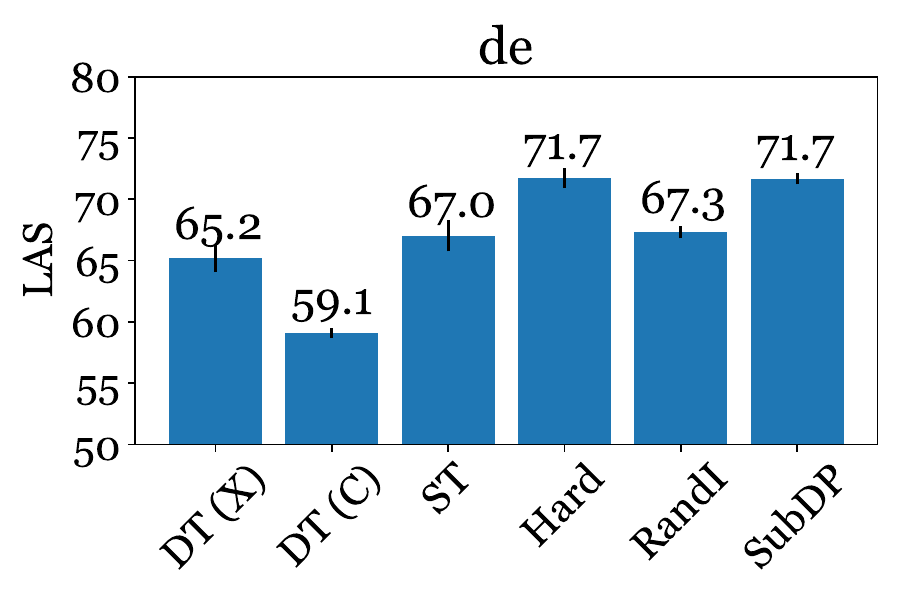} \hspace{-25pt}
    \includegraphics[width=0.28\textwidth]{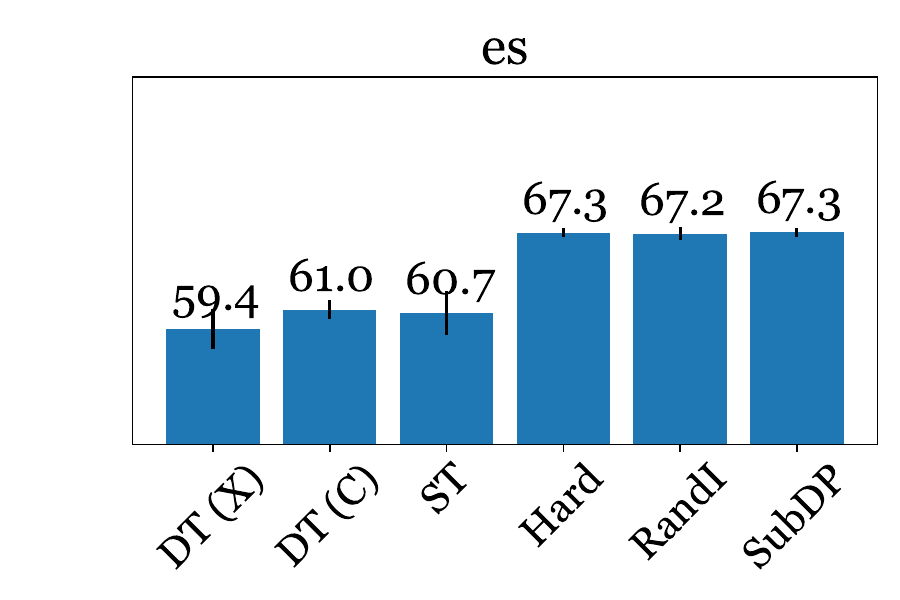} \hspace{-25pt}
    \includegraphics[width=0.28\textwidth]{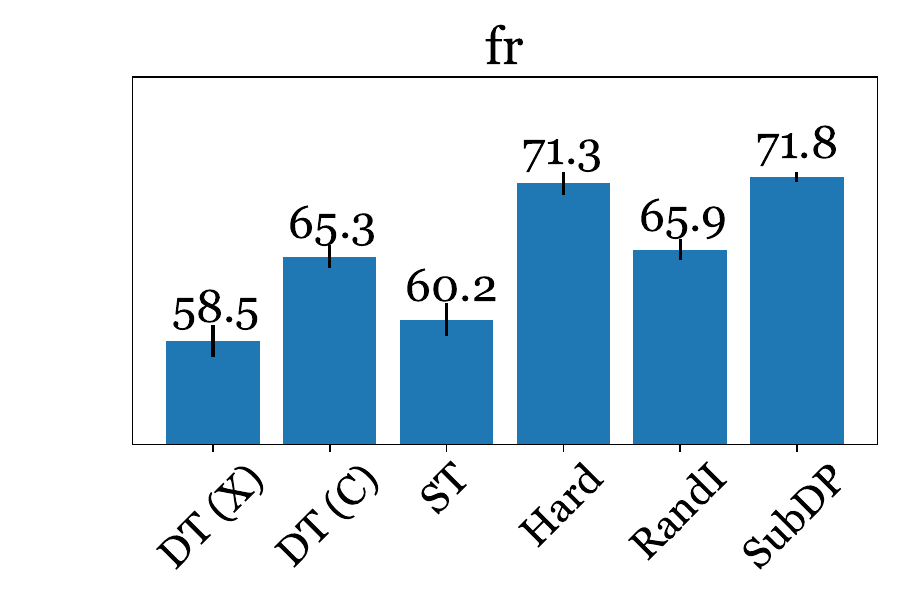} \hspace{-25pt}
    \includegraphics[width=0.28\textwidth]{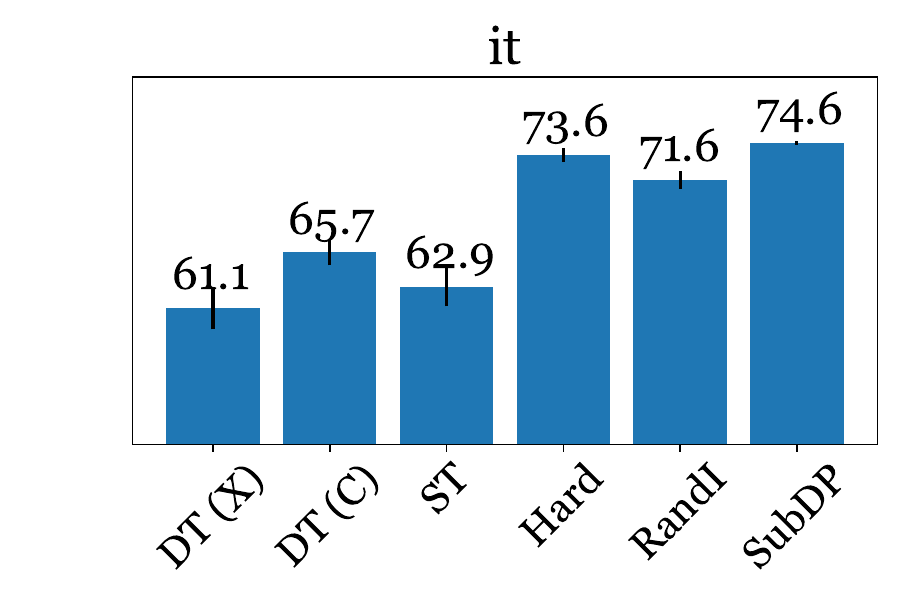}
    \caption[LAS on the Universal Dependencies v2.2 standard development set.]{
        LAS on the Universal Dependencies v2.2 standard development set.
        The standard deviations are denoted by black lines at the top of the bars.
        All numbers are averaged across five runs. DT(X): direct transfer by XLM-R representations; DT (C): direct transfer by CRISS representations.
    }
    \label{fig:subdp-ablation-las}
\end{figure}

%% file: figures/903-subdp-ablation.tex
\begin{figure}[t!]
    \includegraphics[width=0.28\textwidth]{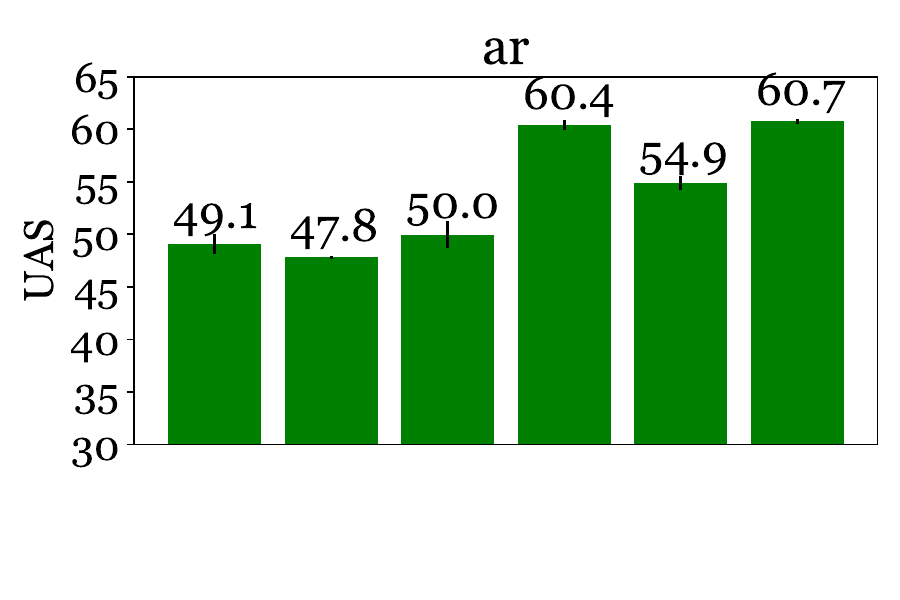} \hspace{-25pt}
    \includegraphics[width=0.28\textwidth]{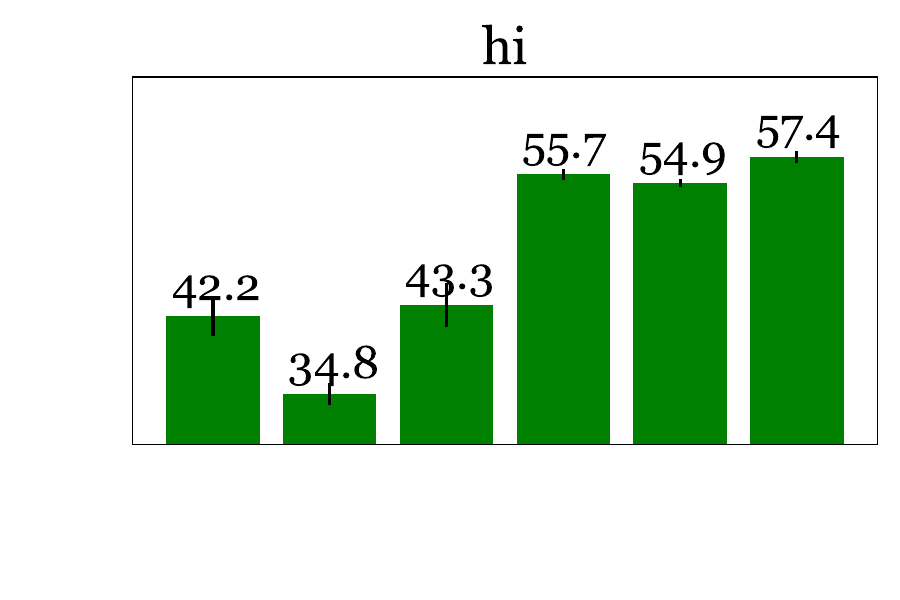} \hspace{-25pt}
    \includegraphics[width=0.28\textwidth]{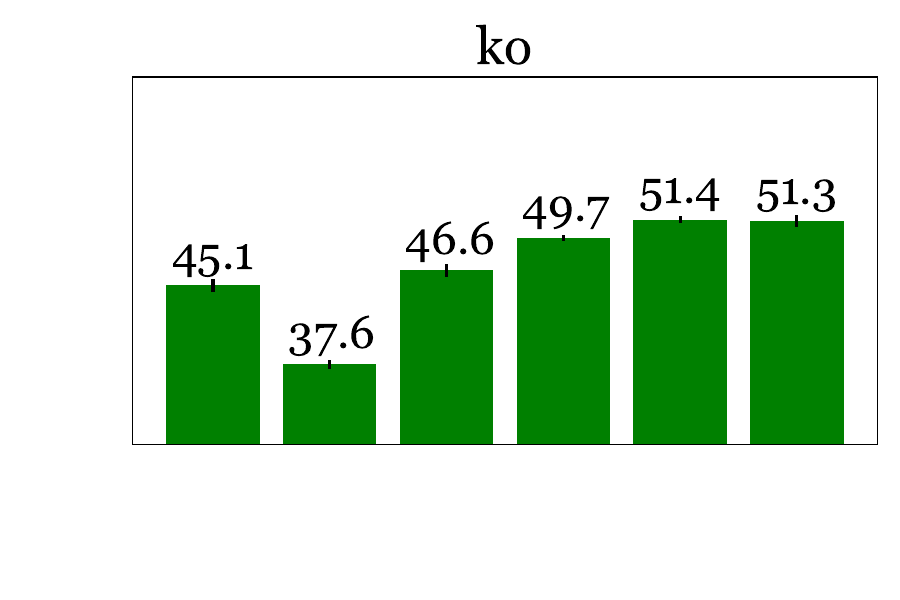} \hspace{-25pt}
    \includegraphics[width=0.28\textwidth]{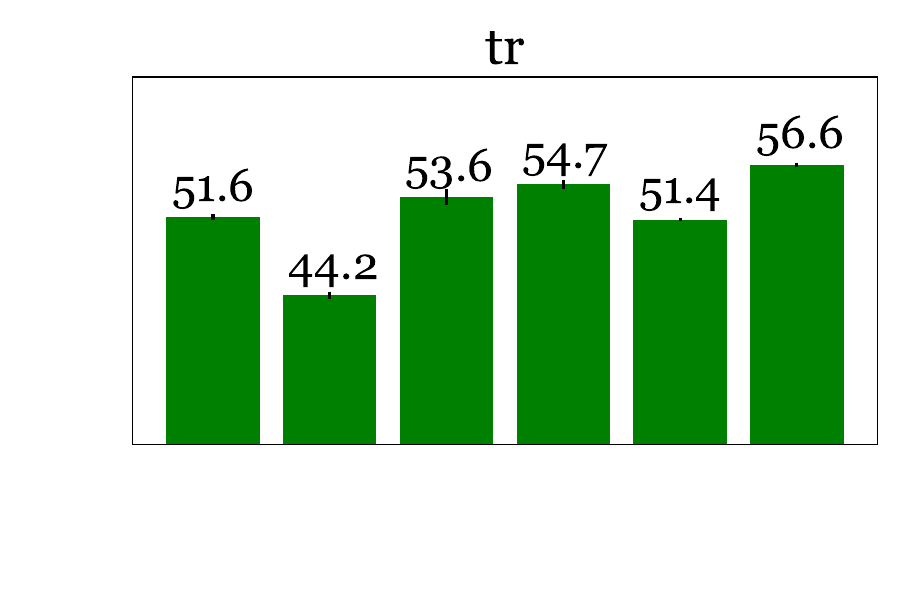} \\[-22pt]
    \includegraphics[width=0.28\textwidth]{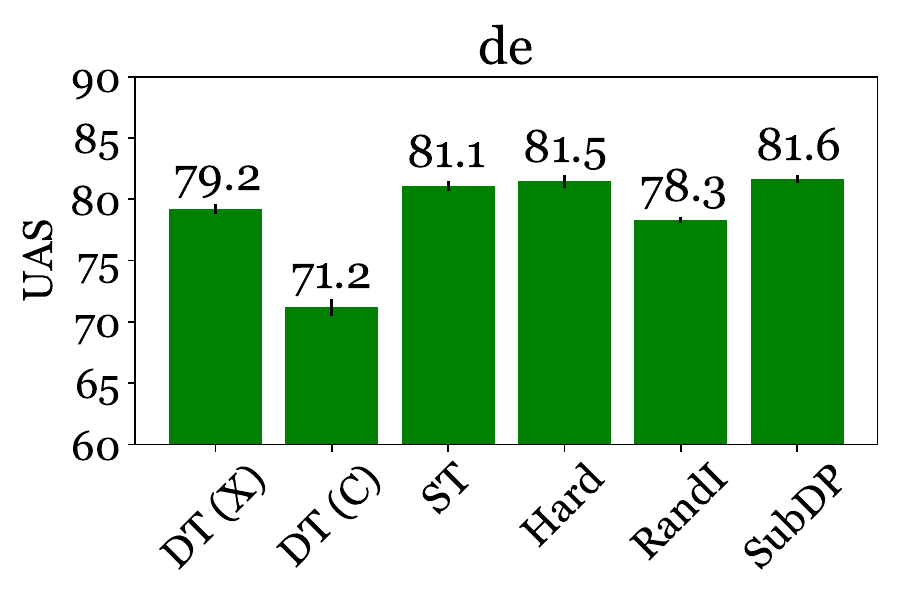} \hspace{-25pt}
    \includegraphics[width=0.28\textwidth]{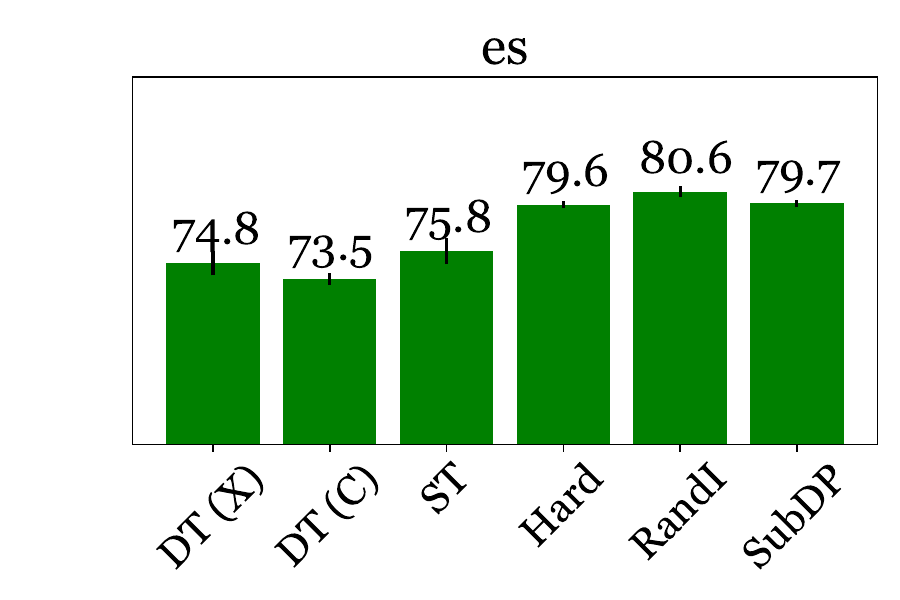} \hspace{-25pt}
    \includegraphics[width=0.28\textwidth]{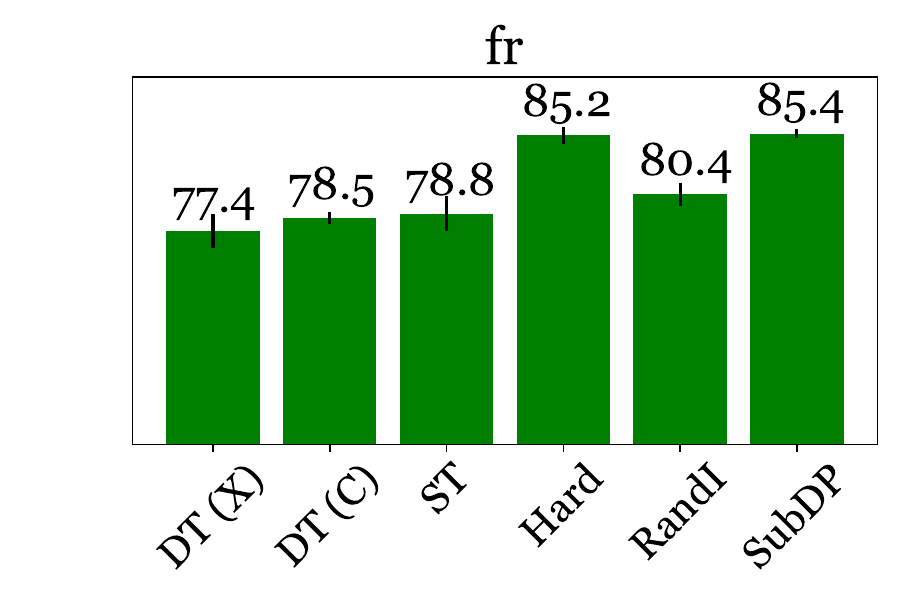} \hspace{-25pt}
    \includegraphics[width=0.28\textwidth]{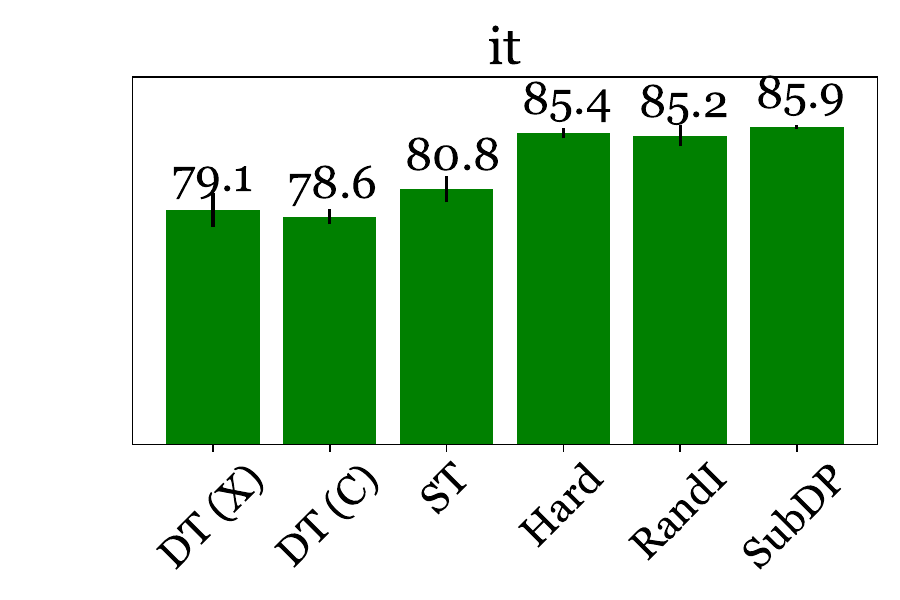}
    \caption[UAS on the Universal Dependencies v2.2 standard development set.]{
        UAS on the Universal Dependencies v2.2 standard development set.
        The standard deviations are denoted by black lines at the top of the bars.
        All numbers are averaged across five runs. DT(X): direct transfer by XLM-R representations; DT (C): direct transfer by CRISS representations.
    }
    \label{fig:subdp-ablation-uas}
\end{figure}

%% file: tables/902-subdp-alignments.tex
\begin{table}[t!]
    \centering
    \small
    \begin{tabular}{lccccc}
        \toprule
        \bf Lang. & \multicolumn{2}{c}{\tt BPE argmax} & \multicolumn{2}{c}{\tt 1:1 only}\\
        & LAS & UAS & LAS & UAS \\
        \midrule
        ar &  39.7 & 60.7 & \bf 40.2 & \bf 61.1 \\
        hi & \bf 39.7 & \bf 57.4 & 38.7 & 56.5 \\
        ko & \bf 31.1 & \bf 51.3 & 27.3 & 49.6\\
        tr & \bf 37.8 & \bf 56.7 & 33.3 & 55.8 \\
        \midrule
        avg. \textit{distant} & \bf 37.1 & \bf 56.5 & 34.8 & 55.8 \\
        \midrule
        \midrule
        de & 71.7 & 81.6 & \bf 72.6 & \bf 83.8 \\
        es & 67.3 & 79.7 & \bf 70.4 & \bf 84.2 \\
        fr & 71.8 & 85.3 & \bf 72.6 & \bf 87.7 \\
        it & 74.6 & 85.9 & \bf 76.0 & \bf 88.8 \\
        \midrule
        avg. \textit{nearby} & 71.4 & 83.1 & \bf 72.9 & \bf 86.1 \\
        \bottomrule
    \end{tabular}
    \caption[LAS and UAS on the Universal Dependencies v2.2 development set, using different alignment methods.]{
        LAS and UAS on the Universal Dependencies v2.2 \citep{nivre-etal-2020-universal} standard development set, averaged across five runs with different random seeds. 1:1 only denotes the filtered one-to-one alignments.
        The best LAS and UAS for each language are in boldface.
    }
    \label{tab:subdp-alignments}
\end{table}

%% file: tables/903-subdp-multiple-source.tex
\begin{table}[t!]
    \centering \small
    \begin{tabular}{lcccc}
        \toprule
         \bf Method & de & es & fr & it \\
         \midrule
         \citet{zhang-barzilay-2015-hierarchical} & 62.5 & 78.0 & \bf 78.9 & 79.3 \\
         \citet{guo-etal-2016-representation} & 65.0 & 79.0 & 77.7 & 78.5 \\
         \citet{schuster-etal-2019-cross}$^\ddagger$ & 61.7 & 76.6 & 76.3 & 77.1\\
         DT (XLM-R)$^{\ddagger,*}$ & 73.1 & 82.2 & 75.5 & 79.5 \\
         \subdp (XLM-R)$^{\ddagger,*}$ & \bf 78.5 & 72.1 & 73.1 & 74.3 \\
         DT w/ \subdp init.$^{\ddagger,*}$ & 76.1 & \bf 82.6 & 77.7 & \bf 81.9\\
         \bottomrule
    \end{tabular}
    \caption[LAS on Universal Dependencies v2.0 with multiple source languages.]{
        LAS on Universal Dependencies v2.0 \citep{mcdonald-etal-2013-universal} standard test set.
        $\ddagger$: methods with minimal annotation. $*$: results from our experiments; other results are taken from \citet{schuster-etal-2019-cross}.
        The best number for each target language is in boldface.
    }
    \label{tab:subdp-multiple-source}
\end{table}

%% file: figures/904-subdp-data-efficiency.tex
\begin{figure}[t!]
    \centering
    \includegraphics[width=0.4\textwidth]{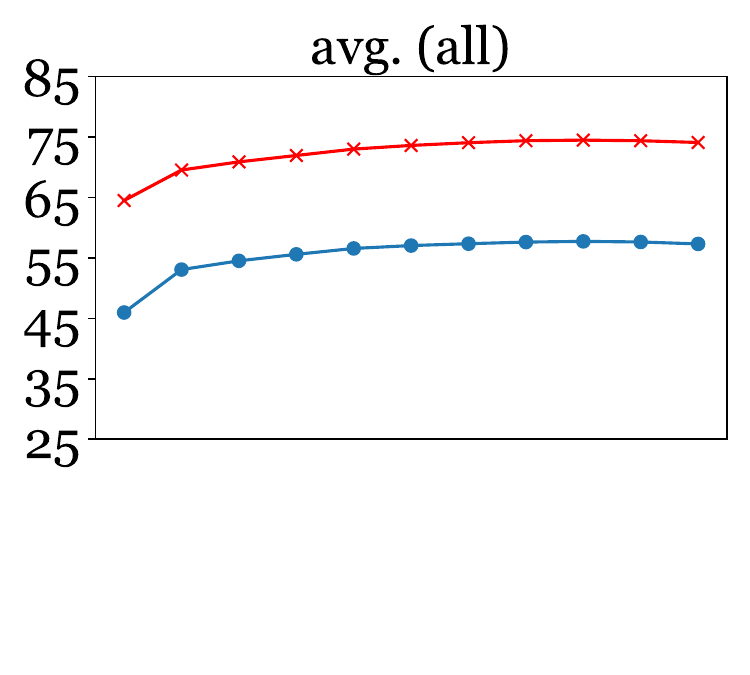} \\[-50pt]
    \includegraphics[width=0.4\textwidth]{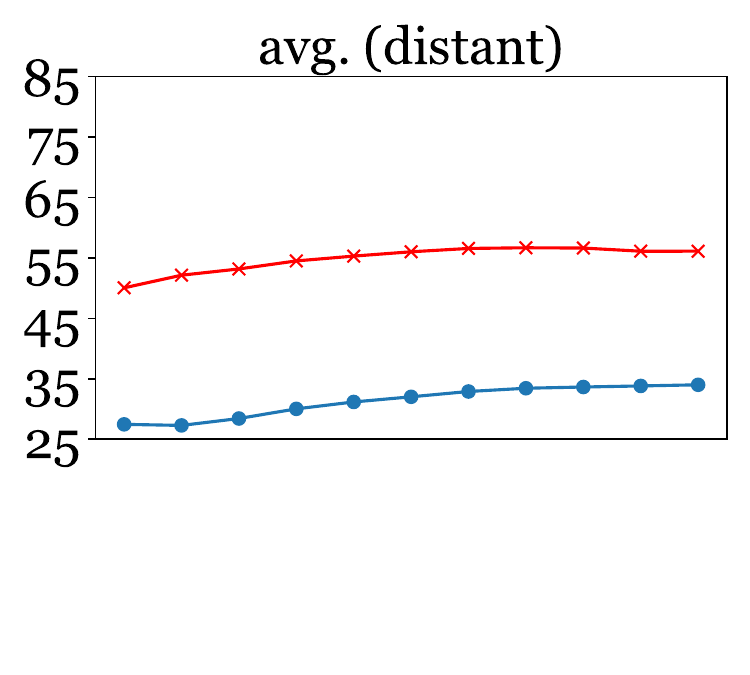} \\[-50pt]
    \includegraphics[width=0.4\textwidth]{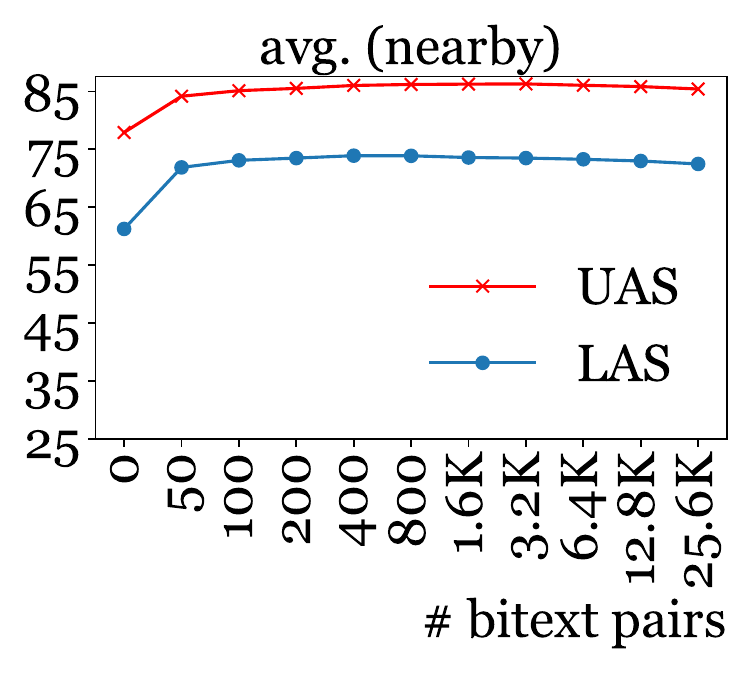}
    \caption[Data efficiency analysis for \subdp.]{
        Averaged LAS and UAS on the Universal Dependencies v2.2 standard development set with respect to the number of bitext pairs.
        For each language, we run five times with different random seeds.
        The $x$-axis is on a log scale.
        Using zero bitext pairs corresponds to the direct transfer (DT; \cref{sec:subdp-ablation-subsection}) baseline.
        All European languages are considered nearby, while the remaining are treated as distant languages.
    }
    \label{fig:subdp-data-efficiency}
\end{figure}

%% file: src/discussion.tex
\chapter{Conclusion and Discussion}
\label{chapter:discussion}
In this dissertation, we have explored the problem of learning language structures through grounding.
Instead of following the supervision paradigm where the model is trained with explicit annotations of language structures, our key contribution is to propose a paradigm that learns these structures through arguably distant grounding signals.
The grounding signals take various forms, including visual signals, acoustic signals, execution results of programs, and information from another language.
These grounded settings offer advantages over pure text-based methods through two perspectives: (1) the cross-modal annotations are much easier to collect than explicit annotations required by supervised learning, as many of them exist naturally in the world, such as image captions, videos, and question-answer pairs, and (2) grounding signals naturally serve as bridges that connect language with the real world, offering the potential to learn more interpretable language structures.

The contents and findings presented in this dissertation are connected to multiple subareas in natural language processing, computational linguistics, and machine learning.
We summarize the contributions of this dissertation in the following aspects:
\begin{itemize}
    \item \cref{chapter:vgnsl,chapter:avnsl,chapter:g2l2} propose novel settings and models for grammar induction from visually grounded text and speech.
    \item \cref{chapter:structiou} proposes an evaluation metric, \structiou, for measuring the quality of induced speech constituency parse trees, which can also be applied to text constituency parsing evaluation.
    \item \cref{chapter:g2l2} propose a model that learns joint syntactic and semantic structures from visual grounding signals and program execution results, enabling nearly perfect compositional generalization.
    \item \cref{chapter:mbrexec} propose a decoding method that uses program execution results to guide the generation of code conditioned on natural language descriptions. For the first time, we show that few-shot natural language-to-code translation can achieve comparable performance to supervised methods, with awareness of program execution results.
    \item \cref{chapter:mlpalign,chapter:subdp} propose to consider cross-lingual transfer within the paradigm of learning language structures through cross-lingual grounding, and introduce models and methods for cross-lingual word alignment and dependency parsing.
          Our systems achieve state-of-the-art performance on both tasks, respectively.
\end{itemize}

Most work in this dissertation is built based on a hypothesis that language structures are, to some extent, learnable through grounding.
While the empirical results in this dissertation support this hypothesis, demonstrating that various forms of grounding signals improve performance over pure text-based methods on corresponding tasks, many open questions and challenges still need to be addressed in future work.

First, while there are shared features between text and grounding modalities, such as the shared semantics between image regions and words in image captioning, the discrepancy between these modalities is also significant---it is still arguable whether the grounding signals can provide sufficient information for learning language structures.
For example, it has been clearly shown that visual grounding signals in \vgnsl are insufficient to retrieve all constituents, especially verb phrases, even in visually grounded settings (\cref{table:vgnsl-main-result}).
In addition, the recent trend of large language models has shown that most of the grammatical rules of natural language can be learned implicitly from a large amount of text data.
Related to this line of discussions, whether grounding signals can provide additional benefits over large-scale text data in learning language structures remains unclear.
It would be interesting to design and implement methods that disentangle the contribution of each type of grounding signal, as well as the intrinsic information from text, and explore how to combine them effectively.

Second, in the cross-lingual grounding settings, due to the limitation of resources, we have not attributed the performance in a detailed way to the quality of the corpora, the quality of the cross-lingual alignment, or the difference between languages.
Extending the experiment to a more diverse set of languages, especially the underrepresented and low-resource ones, may reveal more insights into the effectiveness of cross-lingual grounding signals.
This line also connects to historical language processing, especially those that lack a consensus on interpretations, where grounding in various forms may facilitate the understanding.
It is also worth noting that different languages may carry cultural and historical information that is not directly translatable, and the cross-modal grounding signals may help discover these differences and similarities.

Third, many of the methods proposed in this dissertation are limited in terms of computational efficiency and scalability.
For example, the time complexity for computing the \structiou metric is bivariate quadratic, and \gtlt is not scalable to large-scale datasets due to the high computational cost of the program search while being polynomial.
It would be interesting to explore more efficient methods to implement these algorithms through parallel computing, approximate search with conventional techniques or neural networks, or finding more efficient analytical solutions by exploiting the sparsity of the search space.

Finally, we have shown that grounded language learning can be meaningfully connected with syntactic and semantic structures; however, natural language is not limited to these two types of structures---this dissertation does not involve content related to discourse, pragmatics, and other types of linguistic structures.
Additionally, although structure is everywhere with language, modeling linguistic phenomena does not necessarily imply being built based on structures---for example, it is unlikely that the grammatical genders in gendered languages or the classifiers in Mandarin Chinese have a direct correspondence with complex structures.
We envision a future where grounding will benefit a more comprehensive set of linguistic tasks and facilitate the fundamental understanding of language.
